\documentclass[sigconf,screen]{acmart}

\usepackage[symbol]{footmisc}
\usepackage{times}

\PassOptionsToPackage{hidelinks}{hyperref}
\usepackage{amsmath}

\usepackage{graphicx}
\usepackage{subcaption}

\usepackage{booktabs}
\usepackage{multirow}
\usepackage{longtable}
\usepackage{tabularx}
\usepackage{array}

\usepackage{algorithm}
\usepackage{algorithmic}
\usepackage{stfloats}

\usepackage[most]{tcolorbox}
\usepackage{xcolor}
\usepackage{colortbl}

\definecolor{darkseagreen}{RGB}{180,212,192}
\definecolor{darkseablue}{RGB}{136,194,219}
\definecolor{darkseared}{RGB}{234,167,162}
\usepackage{pifont}
\usepackage{caption}
\usepackage[switch]{lineno}
\usepackage{tcolorbox}
\definecolor{myblue}{RGB}{150,150,230}
\DeclareMathOperator*{\ours}{\displaystyle \textbf{EmoBench-M}}

\newcommand\logo{\raisebox{-4pt}{\includegraphics[width=1.2em]{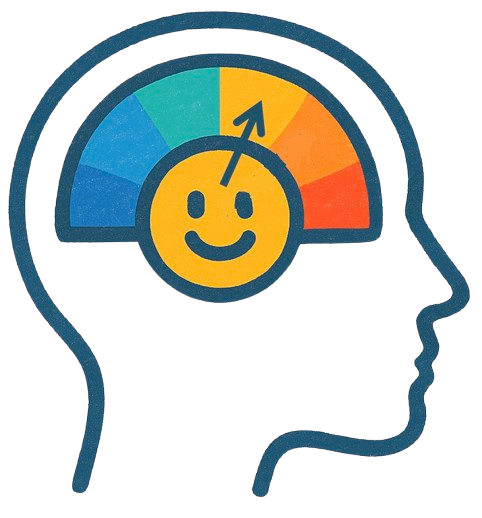}}}

\newcommand{\tightcolorbox}[2]{%
  \begingroup
  \setlength{\fboxsep}{0pt}%
  \colorbox{#1}{%
    \raisebox{0pt}[\ht\strutbox][\dp\strutbox]{%
      \hspace{0.25ex}#2\hspace{0.25ex}}}%
  \endgroup
}

\newcommand{\best}[1]{\tightcolorbox{red!30}{\bfseries #1}}
\newcommand{\second}[1]{\tightcolorbox{blue!30}{#1}}

\AtBeginDocument{%
  }

\setcopyright{acmlicensed}
\copyrightyear{2018}
\acmYear{2018}
\acmDOI{XXXXXXX.XXXXXXX}
\acmConference[Conference acronym 'XX]{Make sure to enter the correct
  conference title from your rights confirmation email}{June 03--05,
  2018}{Woodstock, NY}
\acmISBN{978-1-4503-XXXX-X/2018/06}

\begin{document}

\title[EmoBench-M]{\logo ~EmoBench-M: Benchmarking Emotional Intelligence \\ for Multimodal Large Language Models}

\author{He Hu}
\email{huhe@gml.ac.cn}
\affiliation{%
  \institution{Shenzhen University}
  \city{Shenzhen}
  \country{China}}

\author{Lianzhong You}
\affiliation{%
  \institution{Guangdong Laboratory of Artificial Intelligence and Digital Economy (SZ)}
  \city{Shenzhen}
  \country{China}}

\author{Hongbo Xu}
\affiliation{%
  \institution{Guangdong Laboratory of Artificial Intelligence and Digital Economy (SZ)}
  \city{Shenzhen}
  \country{China}}

\author{Qianning Wang}
\affiliation{%
  \institution{Auckland University of Technology}
  \city{Auckland}
  \country{New Zealand}}

\author{Fei Richard Yu}
\affiliation{%
  \institution{Shenzhen University}
  \city{Shenzhen}
  \country{China}}

\author{Fei Ma}
\affiliation{%
  \institution{Guangdong Laboratory of Artificial Intelligence and Digital Economy (SZ)}
  \city{Shenzhen}
  \country{China}}

\author{Zebang Cheng}
\affiliation{%
  \institution{ Shenzhen University}
  \city{Shenzhen}
  \country{China}}

\author{Zheng Lian}
\affiliation{%
  \institution{Institute of Automation, Chinese Academy of Sciences}
  \city{Beijing}
  \country{China}}

\author{Yucheng Zhou}
\authornotemark[2]
\affiliation{%
  \institution{SKL-IOTSC, CIS, University of Macau}
   \city{Macau}
  \country{China}}

\author{Laizhong Cui}
\email{cuilz@szu.edu.cn}
\authornotemark[2]
\affiliation{%
  \institution{Shenzhen University}
  \city{Shenzhen}
  \country{China}}

\renewcommand{\shortauthors}{Trovato et al.}

\begin{abstract}
With the integration of multimodal large language models (MLLMs) into robotic systems and AI applications, embedding emotional intelligence (EI) capabilities is essential for enabling these models to perceive, interpret, and respond to human emotions effectively in real-world scenarios. Existing static, text-based, or text-image benchmarks overlook the multimodal complexities of real interactions and fail to capture the dynamic, context-dependent nature of emotional expressions, rendering them inadequate for evaluating MLLMs’ EI capabilities. To address these limitations, we introduce EmoBench-M, a systematic benchmark grounded in established psychological theories, designed to evaluate MLLMs across 13 evaluation scenarios spanning three hierarchical dimensions: foundational emotion recognition (FER), conversational emotion understanding (CEU), and socially complex emotion analysis (SCEA). Evaluation was conducted on 27 state-of-the-art MLLMs, using both objective task-specific metrics and LLM-based evaluation, revealing a substantial performance gap relative to human-level competence. Even the best performing models, Gemini-3.0-Pro and GPT-5.2, achieve the highest scores on EmoBench-M, 70.5 and 66.5 points. Specialized models such as AffectGPT exhibit uneven performance across EmoBench-M, demonstrating strengths in certain scenarios but generally lacking comprehensive emotional intelligence. By providing a comprehensive, multimodal evaluation framework, EmoBench-M captures both the strengths and weaknesses of current MLLMs across diverse emotional contexts. All benchmark resources, including datasets and code, are publicly available at \textbf{\href{https://emo-gml.github.io/}{\textcolor{myblue}{Emobench-M}}}, facilitating further research and advancement in MLLM emotional intelligence.
\end{abstract}



\begin{teaserfigure}
  \includegraphics[width=\textwidth]{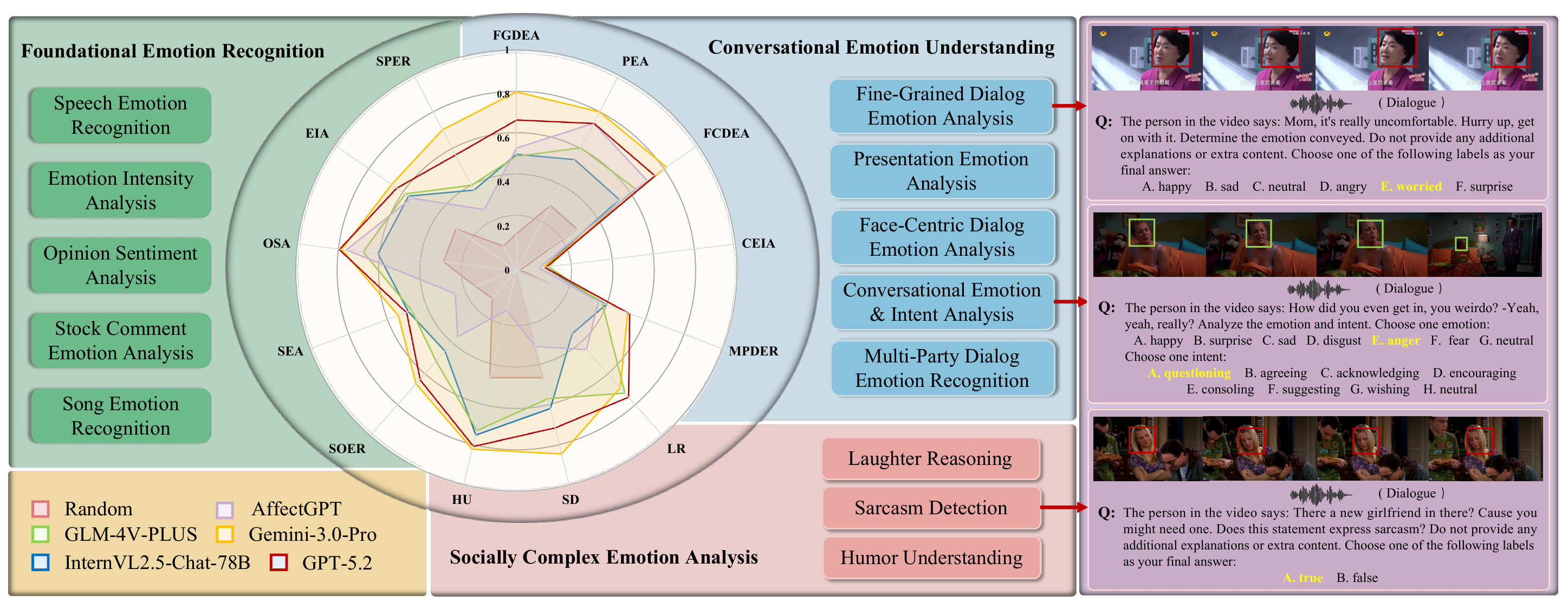}
  \caption{\small Taxonomy for Evaluating Emotion Intelligence (EI) Capabilities of Multimodal Large Language Models (MLLMs): The diagram outlines the categories of ``Foundational Emotion Recognition'', ``Conversational Emotion Understanding'', and ``Socially Complex Emotion Analysis'' along with their respective evaluation scenarios. It also presents a performance comparison of different methods on the proposed dataset $\ours$. The ``Random'' baseline refers to a heuristic approach that randomly selects labels from the available candidates.}
  \Description{Enjoying the baseball game from the third-base
  seats. Ichiro Suzuki is preparing to bat.}
  \label{fig:intro}
\end{teaserfigure}


\maketitle

\section{Introduction}

\begin{table*}[!t]
\caption{\small Comparison of benchmarks related to emotional intelligence. ``Psych-based?'' refers to whether the benchmark is grounded in psychological theories. ``GEN.'' stands for Generation task. }
\centering
% \resizebox{\linewidth}{!}{
\setlength{\tabcolsep}{1mm}
\begin{tabularx}{\textwidth}{
l
c
c
>{\centering\arraybackslash}X
>{\centering\arraybackslash}X
>{\centering\arraybackslash}X
}
\toprule
\textbf{Benchmark} & \textbf{Psych-based?} & \textbf{Task} & \textbf{Multimodality} & \textbf{Answer Type} & \textbf{Evaluator} \\
\midrule
% Text Annotation \cite{Text_Annotation}  & \textcolor{blue}{\ding{51}}  & 1  & Text                  & Multi-choice  & Metric  \\
EmotionBench \cite{Emotionally}    & \textcolor{blue}{\ding{51}}  & 1  & Text                  & Multi-choice  & Metric  \\
SOUL \cite{SOUL}              & \textcolor{red}{\ding{55}}  & 2  & Text                  & Multi-choice \& GEN. & Metric \& LLM  \\
EmoBench \cite{EmoBench}        & \textcolor{blue}{\ding{51}}  & 8  & Text                  & Multi-choice  & Metric  \\
ChatGPT2AC \cite{ChatGPT2ERG}        & \textcolor{red}{\ding{55}}  & 11 & Text                  & Multi-choice  & Metric  \\\midrule
MOSABench \cite{MOSABench}      & \textcolor{red}{\ding{55}}  & 1  & Text \& Image          & Multi-choice  & Metric  \\
SarcasmBench \cite{SarcasmBench}      & \textcolor{red}{\ding{55}}  & 1  & Text \& Image          & Multi-choice  & Metric  \\
MM-InstructEval \cite{MM-InstructEval}   & \textcolor{red}{\ding{55}}  & 6  & Text \& Image          & Multi-choice  & Metric  \\\midrule
MC-EIU \cite{MC-EIU}            & \textcolor{red}{\ding{55}}  & 1  & Video \& Audio \& Text & Multi-choice  & Metric  \\
OV-MER \cite{Open-vocabulary}           & \textcolor{red}{\ding{55}}  & 1  & Video \& Audio \& Text & Multi-choice  & Metric  \\
MER-UniBench \cite{AffectGPT}           & \textcolor{red}{\ding{55}}  & 3  & Video \& Audio \& Text & Multi-choice  & Metric  \\
\rowcolor{blue!10} % 设置这一行的颜色（例如浅蓝）
\textbf{EmoBench-M (Ours)} & \textcolor{blue}{\ding{51}} & \textbf{13} & Video \& Audio \& Text & Multi-choice \& GEN. & Metric \& LLM \\
\bottomrule
\end{tabularx}
% }
\vspace{2mm}
\label{tab:intro}
\vspace{-2mm}
\end{table*}

% \begin{figure*}[t]
%     \centering
%     \includegraphics[width=\linewidth]{figs/main-pip.png}
%     \caption{\small Taxonomy for Evaluating Emotion Intelligence (EI) Capabilities of Multimodal Large Language Models (MLLMs): The diagram outlines the categories of ``Foundational Emotion Recognition'', ``Conversational Emotion Understanding'', and ``Socially Complex Emotion Analysis'' along with their respective evaluation scenarios. It also presents a performance comparison of different methods on the proposed dataset $\ours$. The ``Random'' baseline refers to a heuristic approach that randomly selects labels from the available candidates.}
%     \label{fig:radar}
%     \label{fig:intro}
% \end{figure*}

Emotional Intelligence (EI), initially conceptualized by Salovey and Mayer \cite{EmotionalIntelligence}, emphasizes the ability to perceive, understand, regulate, and apply emotions in oneself and others. 
Recent advancements in multimodal large language models (MLLMs) have significantly improved human-computer interaction and natural language understanding, and integrating MLLMs into robotic control systems has become increasingly prevalent \cite{RobotSurvey, Emotion-LLaMA}. 
Incorporating EI capabilities within MLLMs is essential for improving robotic performance in real-world environments. It will enable robots to address human emotional needs better and ensure more effective interactions.

However, there is currently no universal benchmark to evaluate the EI capabilities of MLLMs comprehensively. Table~\ref{tab:intro} lists existing benchmarks for evaluating EI, demonstrating that most are designed for text-only or text-image EI tasks \cite{EmoBench, MM-InstructEval, Emotionally}, and most are not grounded in established psychological theories.
Real-world MLLM-driven human-robot interactions typically occur in dynamic, multimodal environments. 
Unlike static text and images, videos with audio provide richer and more complex multimodal information, including dynamic facial expressions, body language, and vocal tone, which more authentically convey the flow of emotions and the interactive process. 
Evaluating MLLMs in multimodal environments is crucial because it allows for a more comprehensive understanding of their ability to interpret and respond to diverse emotional cues in real-world scenarios. 

Building on established psychological theories of EI \cite{EmotionalIntelligence, EmotionBench}, we explore the EI capabilities of MLLMs across three primary dimensions: 
\noindent
\textbf{Level I: Foundational Emotion Recognition.} This dimension focuses on accurately identifying emotional states through explicit signals such as facial expressions, vocal tone, and body language \cite{ekman1992there}. It also emphasizes the extraction of emotional information from multimodal signals \cite{poria2019emotion}.  
\noindent
\textbf{Level II: Conversational Emotion Understanding.} Extending beyond foundational recognition, this dimension requires the ability to track emotional dynamics within conversations and to comprehend the contextual and situational meanings of emotions \cite{gross2002emotion,MELD,MC-EIU}.  
\noindent
\textbf{Level III: Socially Complex Emotion Understanding.} Representing an advanced level of EI, this dimension involves understanding emotions influenced not only by internal affective states but also by external social and cultural contexts. It requires AI systems to exhibit mentalizing capabilities, the ability to infer others' emotions and intentions based on environmental cues \cite{frith2006neural, SarcasmBench}.  
Building on these three dimensions, we propose a novel multimodal EI benchmark, Emobench-M, for MLLMs. 
As shown in Figure~\ref{fig:intro}, our benchmark includes 13 scenarios covering diverse contexts such as music and presentation, multi-party dialogues, and social conversation. 
By utilizing multimodal data, i.e., video with audio, Emobench-M enables a more comprehensive evaluation of the EI of MLLMs. 
Moreover, controversial samples were excluded after a thorough human review to ensure the quality of the benchmark. 

To the best of our knowledge, Emobench-M is the first comprehensive benchmark to evaluate EI at the multimodality level. 
We evaluate various open-source MLLMs (e.g., Video-LLaMA2 \cite{VideoLLaMA_2} and InternVL2.5 \cite{InternVL}) and closed-source MLLMs (e.g., GLM-4V \cite{GLM-4} and Gemini \cite{Gemini}) on Emobench-M. 
Our findings indicate that the EI capability of MLLMs in multimodal and realistic environments remains substantially below human performance in many scenarios. 
Moreover, we conduct an extensive evaluation of MLLMs across varying model sizes and reasoning levels. 
We will release our code and data to encourage further research in MLLM's EI.

\section{Related Work}\label{app:related}
\subsection{Multimodal Large Language Models}
With the success of LLMs in various natural language processing (NLP) tasks, such as reasoning \cite{ThoT} and euphemism detection \cite{Euphemism}, numerous efforts have been made to extend LLMs to multimodal areas, i.e., MLLMs, enabling them to process additional types of information, including images, videos, and audio \cite{VideoLLaMA_2, MiniCPM-V}. MLLMs excel in multimodal perception and reasoning and handle more diverse tasks with inputs from different multimodality \cite{MiniGPT4, InternVideo2}. 
For instance, Qwen2-Audio \cite{Qwen2-Audio} specializes in integrating audio and text, demonstrating strong performance in auditory perception tasks. 
The MiniCPM-V \cite{MiniCPM-V}, LongVA \cite{LongVA}, GLM \cite{GLM-4}, InternVL \cite{InternVL}, and InternVideo2 \cite{InternVideo2} have made significant strides in vision understanding and multimodal dialogue generation.
Video-LLaMA2 \cite{VideoLLaMA_2} not only focuses on vision understanding but also enhances audio-video understanding capabilities. Additionally, the Gemini \cite{Gemini1, Gemini, Gemini2}, an LLM natively supporting multimodal capabilities, can seamlessly understand, manipulate, and integrate information from different modalities. 
Moreover, some works further improve vision reasoning ability in MLLMs by visual dependency \cite{VisualDependency}, in-context learning \cite{VICL}. However, despite these advances, existing MLLMs still struggle with fine-grained, temporally dynamic, and socially contextualized emotional understanding, particularly in complex real-world multimodal interactions.

\subsection{Evaluation of Emotional Intelligence}
Given that EI is essential for understanding and responding to human emotions, many studies have focused on evaluating the EI capabilities of LLMs. MERBench \cite{MERBench} standardizes evaluation for multimodal emotion recognition by addressing inconsistencies in feature extractors and offering a unified framework. It introduces MER2023 \cite{mer2023}, a dataset focused on the Chinese language, emphasizing multi-label learning and robustness analysis.
Moreover, MC-EIU \cite{MC-EIU} offers a joint evaluation of emotion and intent in multimodal conversations.
MOSABench \cite{MOSABench} introduces a novel method for multi-object sentiment analysis, emphasizing the challenges MLLMs face in handling spatial complexities.
\citet{GPT-4V} evaluates GPT-4's visual capabilities in emotion recognition tasks but reveals limitations in recognizing micro-expressions and leveraging temporal data effectively. EmotionBench \cite{EmotionBench} employs emotional appraisal theory to evaluate LLMs, exposing misalignments between LLM responses and human emotional behaviors. To deep dive into EI of LLM,  EIBench \cite{Both_Matter} and EmoBench \cite{EmoBench} are based on established psychological theories to evaluate LLM with various EI tasks, and they expose significant gaps between current LLMs and human-like emotional intelligence. In addition, EQ-Bench \cite{EQ-Bench} and SOUL \cite{SOUL} focus on nuanced EI aspects, including emotion intensity prediction and justification generation, revealing performance disparities between small and large models. 

\begin{table*}[!t]\small
\caption{\small Emotion Recognition Tasks and Metrics. ``n-CLS'' denotes an n-class classification task, and ``GEN'' represents a generation task. ``Num'' indicates the number of samples. ``ACC'' denotes accuracy. For the ``Laughter Reasoning'', we employ an LLM as the evaluator, and evaluation prompts are shown in Appendix~\ref{app:evalprompt}. To ensure fair and consistent comparisons in future research, we adopt the open-source Qwen2.5-72B \protect\cite{yang2024qwen2}. Details of the specific categories in the classification tasks are provided in the Appendix~\ref{app:dataset}. } 
\centering
\resizebox{\linewidth}{!}{
\setlength{\tabcolsep}{4mm}
\begin{tabular}{lcccc}
\toprule
\textbf{Evaluation Scenario} & \textbf{Data Source} & \textbf{Task} & \textbf{Num} & \textbf{Metric} \\
\hline\midrule
\rowcolor{darkseagreen}\multicolumn{5}{c}{\textbf{Level I: Foundational Emotion Recognition}} \\\midrule
Song Emotion Recognition (SOER) & RAVDESS(song) \cite{ryerson} & 6-CLS & 500 & ACC \\
Speech Emotion Recognition (SPER) & RAVDESS(speech) \cite{ryerson} & 8-CLS & 500 & ACC \\
Opinion Sentiment Analysis (OSA) & CMU-MOSI \cite{MOSI} & 3-CLS & 500 & ACC \\
Emotion Intensity Analysis (EIA) & CMU-MOSEI \cite{CMU-MOSEI} & 3-CLS & 500 & ACC \\
Stock Comment Emotion Analysis (SEA) & FMSA-SC \cite{FMSA-SC} & 5-CLS & 250 & ACC \\
\hline\midrule
\rowcolor{darkseablue}\multicolumn{5}{c}{\textbf{Level II: Conversational Emotion Understanding}} \\\midrule
Fine-Grained Dialog Emotion Analysis (FGDEA) & MER2023 \cite{mer2023} & 6-CLS & 411 & ACC \\
Presentation Emotion Analysis (PEA) & CH-SIMSv2 \cite{CH-SIMSv2} & 3-CLS & 500 & ACC \\
Face-Centric Dialog Emotion Analysis (FCDEA) & CH-SIMS \cite{CH-SIMS} & 3-CLS & 457 & ACC \\
Conversational Emotion \& Intent Analysis (CEIA) & MC-EIU \cite{MC-EIU} & 7-\&8-CLS & 500 & ACC \\
Multi-Party Dialog Emotion Recognition (MPDER) & MELD \cite{MELD} & 7-CLS & 500 & ACC \\
\hline\midrule
\rowcolor{darkseared}\multicolumn{5}{c}{\textbf{Level III: Socially Complex Emotion Analysis}} \\\midrule
Humor Understanding (HU) & UR-FUNNY \cite{UR-FUNNY} & 2-CLS & 448 & ACC \\
Sarcasm Detection (SD) & MUStARD \cite{MUStARD} & 2-CLS & 500 & ACC \\
Laughter Reasoning (LR) & SMILE \cite{SMILE} & GEN & 80 & LLM \\
\bottomrule
\end{tabular}
}
\vspace{3mm}
\label{tab:emotion_tasks}
\vspace{-4mm}
\end{table*}

\subsection{Multimodal Benchmarks for LLMs}
The rapid development of Multimodal Large Language Models (MLLMs) has necessitated diverse benchmarks to systematically evaluate their capabilities across perception, reasoning, and application domains. MME \cite{MME} and MMT-Bench \cite{MMT-Bench} serve as comprehensive benchmarks for foundational multimodal tasks and general-purpose intelligence, providing standardized datasets and evaluation protocols for consistent comparison across models. MultiTrust \cite{MultiTrust} evaluates trustworthiness across truthfulness, safety, robustness, fairness, and privacy risks, emphasizing responsible deployment considerations. HumanVBench \cite{humanvbench} and MVBench \cite{MVBench} focus on human-centric and temporal understanding in video content, exposing gaps in cross-modal and temporal alignment and highlighting limitations in dynamic scenario reasoning. Specialized benchmarks address domain-specific challenges: MathScape \cite{MathScape} targets multimodal mathematical reasoning, M3SciQA \cite{M3SciQA} focuses on scientific question answering, BenchLMM \cite{BenchLMM} evaluates robustness under style shifts, and BLINK \cite{BLINK} targets core visual perception tasks. \citet{MedLVLM} assesses medical diagnostic capabilities and generalization of LVLMs, while SEED-Bench-2-Plus \cite{SEED-Bench-2-Plus} evaluates text-rich visual reasoning, such as interpreting charts and maps, highlighting ongoing challenges in multimodal integration. Collectively, these benchmarks provide a comprehensive evaluation landscape, enabling systematic analysis of both the strengths and limitations of current MLLMs across diverse tasks and real-world scenarios.

\section{EmoBench-M}
\subsection{Evaluation Taxonomy}
To systematically evaluate MLLM EI capabilities, the evaluation focuses on three dimensions based on established psychological theories of EI \cite{EmotionalIntelligence}: ``Foundational Emotion Recognition'', ``Conversational Emotion Understanding'', and ``Socially Complex Emotion Analysis''.  Table~\ref{tab:emotion_tasks} details evaluation scenarios in each dimension.

\subsubsection{Foundational Emotion Recognition}
Foundational emotion recognition, a core aspect of Emotional Intelligence (EI), focuses on identifying basic emotions such as anger, happiness, and sadness \cite{ekman1992there,scherer2005emotions}. This dimension evaluates a MLLMs' ability to extract and integrate emotional information from multimodal signals (video, audio, and text) to recognize these fundamental emotions, a crucial capability for higher-level EI.  The MLLMs' proficiency in discerning emotions conveyed through speech, music, and video is assessed.
Song and Speech Emotion Recognition uses data sourced from \cite{ryerson}, which provides video clips with audio-visual emotional cues. Opinion Sentiment Analysis utilizes data sourced from \cite{MOSI}, focusing on speech and facial expressions in opinion videos.  Emotion Intensity Analysis goes beyond simple polarity; data sourced from the CMU-MOSEI dataset \cite{CMU-MOSEI} is used to assess both the emotional state and its intensity from audio and video. This requires the model to identify specific emotion categories (e.g., happiness, sadness, anger) and quantify their intensity levels across diverse video content. Stock Comment Emotion Analysis employs data sourced from \cite{FMSA-SC}, analyzing emotions expressed in stock-related video comments.

\subsubsection{Conversational Emotion Understanding}
Conversational emotion understanding requires MLLMs to track emotional dynamics and interpret their contextual significance \cite{poria2019emotion,hazarika2018conversational}.  This involves identifying emotional shifts in multi-party conversations, leveraging semantic and tonal cues, and adapting to dynamic contexts, including inter-participant emotional interplay.  Several scenarios in this dimension:
Fine-grained Dialog Emotion Analysis (data source from \cite{mer2023}) captures subtle emotional shifts.  Face-centric Dialog Emotion Analysis (data source from CH-SIMS \cite{CH-SIMS}) focuses on facial expressions and verbal/visual cues in interactive, conversational settings.  Presentation Emotion Analysis (data source from CH-SIMSv2 \cite{CH-SIMSv2}) examines emotions in formal presentations. Crucially, CH-SIMSv2 extends CH-SIMS by encompassing broader presentation styles, multi-speaker scenarios, and more diverse non-verbal cues.  Conversational Emotion and Intent Analysis (data source from \cite{MC-EIU}) detects emotions and infers intentions. Multi-party Dialog Emotion Recognition (data source from \cite{MELD}) analyzes multi-party conversations, classifying seven emotions based on speech and facial cues.

\begin{table*}[!t]\scriptsize
\caption{\small Performance comparison of different methods on $\ours$ (\textbf{F}oundational \textbf{E}motion \textbf{R}ecognition). \best{Red} and \second{blue} indicate the best and the second best results among all models.}
\centering
\setlength{\tabcolsep}{1.5mm}
\renewcommand{\arraystretch}{1.05}
\begin{tabular}{lcccc@{\hspace{5mm}}cccccccccccc}
\toprule
\multirow{2}{*}{\textbf{Method}} &
\multirow{2}{*}{\textbf{Params}} &
\multirow{2}{*}{\textbf{A}} &
\multirow{2}{*}{\textbf{V}} &
\multirow{2}{*}{\textbf{T}} &
\multicolumn{2}{c}{\textbf{SOER}} &
\multicolumn{2}{c}{\textbf{SPER}} &
\multicolumn{2}{c}{\textbf{OSA}} &
\multicolumn{2}{c}{\textbf{EIA}} &
\multicolumn{2}{c}{\textbf{SEA}} &
\multicolumn{2}{c}{\textbf{Avg.}} \\
\cmidrule(lr){6-7} \cmidrule(lr){8-9} \cmidrule(lr){10-11} 
\cmidrule(lr){12-13} \cmidrule(lr){14-15} \cmidrule(lr){16-17}
 & & & & & ACC & WAF & ACC & WAF & ACC & WAF & ACC & WAF & ACC & WAF & ACC & WAF \\
\midrule
\rowcolor{gray!25}\multicolumn{17}{c}{\textit{Open-Source Model}} \\
\midrule
InternVL2.5 \cite{InternVL} & 4B &  & \checkmark & \checkmark & 50.5 & 47.6 & 41.2 & 35.5 & 71.8 & 75.5 & 60.1 & 60.6 & 48.8 & 46.2 & 54.5 & 53.1 \\
Video-LLaMA2 \cite{VideoLLaMA_2} & 7B &  & \checkmark & \checkmark & 52.4 & 47.7 & 42.4 & 35.0 & 31.0 & 39.8 & 50.2 & 46.8 & 50.8 & 38.6 & 45.4 & 41.6 \\
% Video-LLaMA2-16F \cite{VideoLLaMA_2} & 7B &  & \checkmark & \checkmark & 45.0 & 41.6 & 46.0 & 41.2 & 64.0 & 68.2 & 56.5 & 54.9 & 45.5 & 39.1 & 51.4 & 49.0 \\
Qwen2-Audio \cite{Qwen2-Audio} & 7B & \checkmark &  & \checkmark & \second{65.8} & 59.8 & \second{71.7} & \second{65.1} & 66.2 & 72.9 & 59.6 & 58.9 & 36.4 & 36.1 & 59.9 & 58.6 \\
Video-LLaMA2.1-16F \cite{VideoLLaMA_2} & 7B &  & \checkmark & \checkmark & 41.2 & 35.2 & 31.4 & 24.7 & 
75.4 & 77.7 & 61.6 & 60.9 & 44.8 & 43.2 & 50.9 & 48.3 \\
Video-LLaMA2.1-AV \cite{VideoLLaMA_2} & 7B & \checkmark & \checkmark & \checkmark & 50.4 & 42.2 & 37.7 & 30.5 & 73.0 & 76.4 & 57.6 & 58.2 & 33.2 & 33.6 & 50.4 & 48.2 \\  
LongVA-DPO \cite{LongVA} & 7B &  & \checkmark & \checkmark & 50.2 & 45.4 & 44.2 & 40.3 & 33.8 & 42.4 & 45.7 & 39.4 & 54.8 & 46.7 & 45.7 & 42.8 \\
Qwen2.5-VL \cite{Qwen-VL} & 7B &  & \checkmark & \checkmark & 47.8 & 43.3 & 36.4 & 28.9 & 62.8 & 70.7 & 60.0 & 59.4 & 51.6 & 48.1 & 51.7  & 50.1  \\
InternVideo2-Chat \cite{InternVideo2} & 8B &  & \checkmark & \checkmark & 55.2 & 50.5 & 44.0 & 35.1 & 45.4 & 55.1 & 56.0 & 55.3 & 52.4 & 42.0 & 50.6 & 47.6 \\
MiniCPM-V-2.6 \cite{MiniCPM-V} & 8B &  & \checkmark & \checkmark & 26.6 & 20.5 & 21.8 & 16.2 & 56.5 & 65.3 & 50.5 & 48.4 & 44.5 & 37.0 & 40.0 & 37.5 \\
InternVL2.5 \cite{InternVL} & 8B &  & \checkmark & \checkmark & 40.3 & 36.2 & 40.8 & 36.0 & 67.8 & 74.2 & 62.0 & 62.6 & 45.0 & 40.3 & 51.2 & 49.9 \\
Qwen3-VL\cite{Qwen3} & 8B &  & \checkmark & \checkmark & 56.0 & 53.2 & 43.2 & 38.0 & 77.6 & 79.9 & 66.4 & 66.7 & 46.0 & 44.8 & 57.8  & 56.5 \\
Qwen3-VL-Thinking \cite{Qwen3} & 8B &  & \checkmark & \checkmark & 31.6 & 25.3 & 21.6 & 15.8 & \best{81.0} & 81.6 & 61.8 & 60.5 & 37.5 & 40.2 & 46.7  & 44.7 \\
Qwen2.5-Omni \cite{Qwen2.5-Omni} & 7B & \checkmark & \checkmark & \checkmark & 50.8 & 46.4 & 40.8 & 32.8 & 64.6 & 71.8 & 39.6 & 54.3 & 52.8 & 39.5 & 49.7 & 49.0 \\
Emotion-LLaMA \cite{Emotion-LLaMA} & 7B & \checkmark & \checkmark  & \checkmark & 44.8 & 36.9 & 33.4 & 23.7 & 23.0 & 29.5 & 41.1 & 34.1 & 42.0 & 36.2 & 36.9 & 32.1 \\
AffectGPT \cite{AffectGPT} & 7B & \checkmark & \checkmark & \checkmark & 40.2  & 34.3 & 31.2 & 25.0 & 77.6 & 78.9 & 57.8 & 73.0 & 29.6 & 30.3 & 47.3  & 48.3 \\
\midrule
Qwen2.5-VL \cite{Qwen-VL} & 32B &  & \checkmark & \checkmark & 44.2 & 41.0 & 34.0 & 28.1 & 66.8 & 73.9 & 64.2 & 63.8 & 47.6 & 47.6 & 51.4 & 50.9 \\
InternVL2.5 \cite{InternVL} & 38B &  & \checkmark & \checkmark & 53.6 & 51.0 & 44.2 & 39.0 & 70.4 & 76.6 & \second{66.8} & \second{67.4} & 52.8 & 43.3 & 57.6 & 55.5 \\
Video-LLaMA2 \cite{VideoLLaMA_2} & 72B &  & \checkmark & \checkmark & 56.0 & 54.8 & 44.2 & 41.3 & 49.2 & 60.1 & 50.3 & 46.1 & 53.6 & 45.7 & 50.7 & 49.6 \\
InternVL2.5 \cite{InternVL} & 78B &  & \checkmark & \checkmark & 48.8 & 47.7 & 41.2 & 37.8 & 63.2 & 71.2 & 59.4 & 59.0 & 52.4 & 38.9 & 53.0 & 50.9 \\
Qwen2.5-VL \cite{Qwen-VL} & 72B &  & \checkmark & \checkmark & 44.8 & 42.1 & 35.6 & 29.6 & 72.4 & 78.3 & 62.7 & 62.1 & \second{58.4} & \second{53.9} & 54.8 & 53.2 \\
\midrule
\rowcolor{gray!25}\multicolumn{17}{c}{\textit{Closed-Source Model (API)}} \\
\midrule
GLM-4V-PLUS \cite{GLM-4} & API & & \checkmark & \checkmark & 54.9 & 48.8 & 43.7 & 34.8 & 70.0 & 75.0 & 61.2 & 59.9 & 50.8 & 45.0 & 56.1 & 52.7 \\
Gemini-1.5-Flash \cite{Gemini} & API & \checkmark & \checkmark & \checkmark & 62.0 & 60.1 & 52.0 & 50.1 & 75.0 & 78.6 & 65.0 & 64.8 & 44.4 & 46.5 & 59.7 & 60.0 \\
Gemini-2.0-Flash \cite{Gemini2} & API & \checkmark & \checkmark & \checkmark & 63.3 & 62.2 & 55.8 & 53.5 & 68.8 & 75.8 & 63.5 & 63.0 & 55.6 & 48.6 & 61.4 & 60.6 \\
Gemini-2.0-Flash-Thinking \cite{Gemini2} & API & \checkmark & \checkmark & \checkmark & 53.4 & 51.0 & 53.0 & 49.4 & 79.4 & 81.9 & 66.5 & 66.8 & 36.0 & 38.1 & 57.7 & 57.4 \\
Gemini-3.0-Flash \cite{gemini3} & API & \checkmark & \checkmark & \checkmark & 58.6 & 56.6 & 59.6 & 60.4 & 69.6 & 75.8 & 63.5 & 63.9 & \best{58.8} & \best{55.9} & 62.0 & 62.5  \\
Gemini-3.0-Pro \cite{gemini3} & API & \checkmark & \checkmark & \checkmark & \best{68.6} & \best{67.3} & \best{72.1} & \best{71.5} & 79.8 & \best{83.0} & \best{68.7} & \best{68.7} & 57.3 & \second{53.9} & \best{69.3} & \best{68.9}  \\
% GPT-4o \cite{gpt4o} & API &  & \checkmark & \checkmark & 51.3 & 51.1 & 45.7 & 44.1 & 73.2 & 78.4 & 64.7 & 65.0 & 42.4 & 41.6 & 55.5 & 56.0 \\
GPT-5.2 \cite{GPT-5} & API &  & \checkmark & \checkmark & \second{65.8} & \second{63.0} & 59.4 & 58.0 & \second{80.8} & \second{82.7} & 66.0 & 66.0 & 53.2  & 52.1 & \second{65.0} & \second{64.4} \\

\bottomrule
\end{tabular}
\vspace{3mm}
\vspace{-2mm}
\label{tab:performance_fer}
\end{table*}

\subsubsection{Socially Complex Emotion Analysis}
Emotional expression is influenced by internal drives and external social/cultural contexts \cite{frith2006neural}. Socially complex emotion understanding is an advanced EI dimension, encompassing the ability to identify, comprehend, and respond to nuanced emotions and social intentions in intricate social scenarios. This dimension primarily evaluates emotions arising in complex social contexts, requiring deeper inference of emotions like humor, sarcasm, and latent feelings based on social interactions and norms. Humor Understanding utilizes data sourced from~\cite{UR-FUNNY}. Sarcasm Detection uses data sourced from \cite{MUStARD}. Laughter Reasoning employs data sourced from \cite{SMILE} to analyze the complex emotions in various social situations.

%如果三位专家 完全不一致（1:1:1）,则进入 讨论仲裁,三人需就该案例进行讨论,达成一致,确定最终标签。

\subsection{Data Collection and Processing}

\begin{figure}[!t]
    \centering
    \includegraphics[width=1\linewidth]{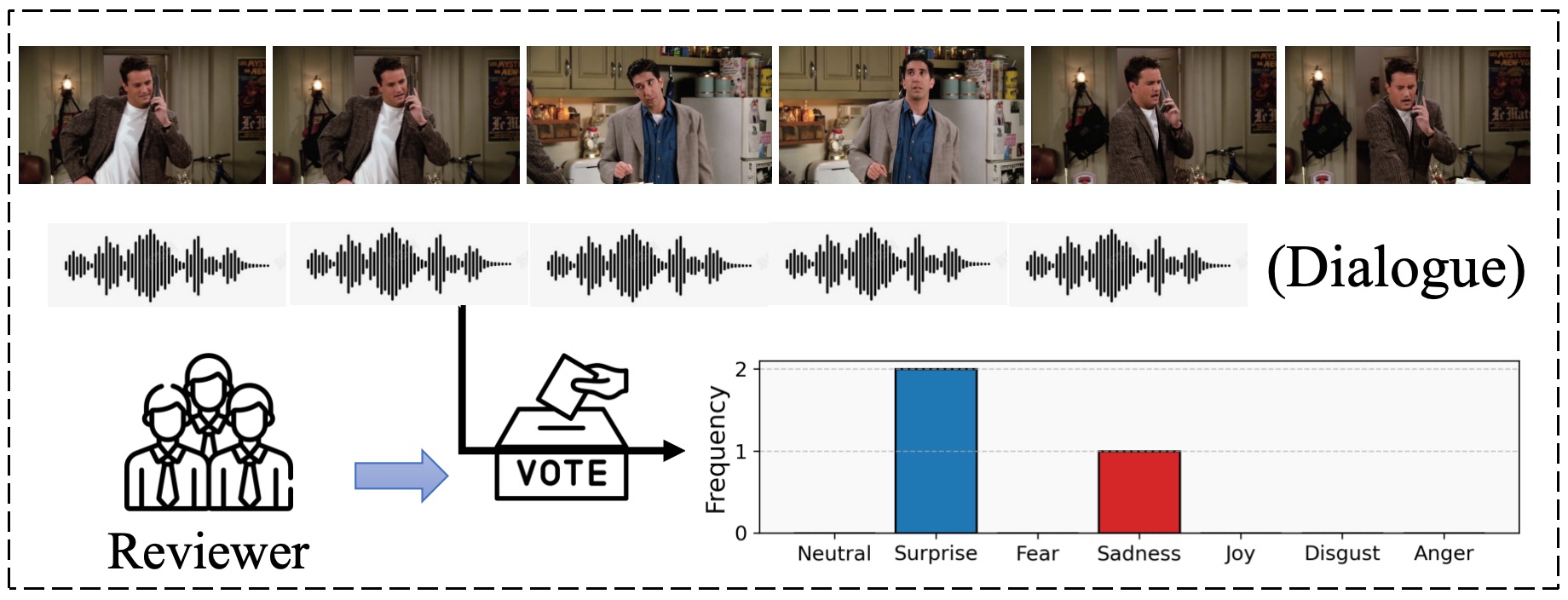}
    \vspace{-6mm}
    \caption{\small Data Filtering and Label Verification Process. Bar charts show original dataset label (\textcolor{red}{red}) and label from our reviewers (\textcolor{blue}{blue}).}
    \label{fig:filter}
    \vspace{-3mm}
\end{figure}

The EmoBench-M benchmark was meticulously curated to evaluate the EI capabilities of MLLMs across a diverse range of tasks. As illustrated in Figure~\ref{fig:filter}, the data collection and processing pipeline involved a rigorous, formalized procedure for data filtering and class balancing to ensure high quality and fairness.

\paragraph{Filtering and Quality Assurance.}
To eliminate ambiguous or mislabeled samples from the initial datasets, we implemented a multi-reviewer verification process. Each sample underwent manual review by a panel of $N=3$ graduate students with research experience in affective computing, following a unified annotation guideline. Let $s$ be a sample from an initial dataset $D_{\text{initial}}$, and let $y_{\text{orig}}(s)$ be its original label. Each reviewer $i$ assigned an emotion label $v_i(s)$ from the set of possible labels $L$.

We formalized the reviewers' consensus using a majority vote. For any given label $l \in L$, the vote count from the reviewers for sample $s$ is given by:
\begin{align}
C(l, s) = \sum_{i=1}^{N} \mathbb{I}(v_i(s) = l)
\end{align}
where $\mathbb{I}(\cdot)$ is the indicator function, which is 1 if the condition is true and 0 otherwise. The consensus label from the reviewers, $y_{\text{rev}}(s)$, is the label that receives the maximum number of votes:
\begin{align}
y_{\text{rev}}(s) = \underset{l \in L}{\arg\max} \, C(l, s)
\end{align}
A sample $s$ was retained only if the original label $y_{\text{orig}}(s)$ matched the reviewers' consensus label $y_{\text{rev}}(s)$. While emotional perception can be inherently ambiguous, this strict filtering strategy prioritizes label reliability over ambiguity to ensure fair and consistent benchmarking. This filtering criterion ensures that only samples with high inter-annotator agreement and consistency with the original labels are included. Consequently, the final filtered dataset, $D_{\text{filtered}}$, is defined as:
\begin{align}
D_{\text{filtered}} = \{s \in D_{\text{initial}} \mid y_{\text{orig}}(s) = y_{\text{rev}}(s) \}
\end{align}
This robust procedure guarantees that EmoBench-M is built upon high-quality, unambiguous data, thereby enhancing the benchmark's reliability and validity.

\paragraph{Class Imbalance Correction.}
To ensure a fair and unbiased evaluation, we addressed potential class imbalances within the filtered data. For each task, we capped the maximum number of test samples at 500. If the size of a filtered dataset for a task, $|D_{\text{filtered}}|$, exceeded this threshold, we performed a targeted down-sampling procedure.

This procedure iteratively removes samples from the majority class until the dataset size is reduced to 500. Specifically, in each step of the process, we identify the majority class $c_{\text{maj}}$ with the highest number of instances:
\begin{align}
c_{\text{maj}} = \underset{c \in L}{\arg\max} \, |D_{c}|
\end{align}
where $|D_{c}|$ is the number of samples with label $c$ in the current dataset. A single sample is then randomly removed from this majority class. This process is repeated until $|D_{\text{filtered}}| = 500$. This method preserves the presence of minority classes while creating a more balanced class distribution, which prevents the benchmark from being skewed towards dominant emotions and promotes better model generalization across the full spectrum of emotions.

\paragraph{Dataset Statistics.} 
EmoBench-M encompasses tasks with varying sample sizes, ranging from 80 to 500 samples per task, as detailed in Table~\ref{tab:emotion_tasks}. Each task is designed to evaluate different facets of emotional intelligence, spanning from basic emotion recognition to understanding more complex social emotions. The performance metric is accuracy (ACC), and LLM-based evaluation is employed for generation tasks.

\begin{table*}[!t]\scriptsize
\caption{\small Performance comparison of different methods on $\ours$ (\textbf{S}ocially \textbf{C}omplex \textbf{E}motion \textbf{A}nalysis). ``B-4'', ``R-L'', and ``BS'' denote BLEU-4 \protect\cite{Bleu}, ROUGE-L \protect\cite{Rouge}, and BERTScore \protect\cite{BERTScore}. MM and Logic refer to the LLM-based evaluation scores for multimodal content association and logical judgment.}
\centering
\setlength{\tabcolsep}{1.5mm}
\renewcommand{\arraystretch}{1.05}
\begin{tabular}{lcccc@{\hspace{5mm}}ccccccccccccc}
\toprule
\multirow{2}{*}{\textbf{Method}} &
\multirow{2}{*}{\textbf{Params}} &
\multirow{2}{*}{\textbf{A}} &
\multirow{2}{*}{\textbf{V}} &
\multirow{2}{*}{\textbf{T}} &
\multicolumn{2}{c}{\textbf{HU}} &
\multicolumn{2}{c}{\textbf{SD}} &
\multicolumn{6}{c}{\textbf{LR}} &
\multicolumn{2}{c}{\textbf{Avg.}} \\
\cmidrule(lr){6-7}\cmidrule(lr){8-9}\cmidrule(lr){10-15}\cmidrule(lr){16-17}
 &  &  &  &  & ACC & WAF & ACC & WAF & B-4 & R-L & BS & Logic & MM & Total & ACC & WAF \\
\midrule
\rowcolor{gray!25}\multicolumn{17}{c}{\textit{Open-Source Model}} \\\midrule
InternVL2.5 \cite{InternVL} & 4B &  & \checkmark & \checkmark & 56.6 & 56.6 & 52.7 & 52.7 & 0.0 & 13.0 & 15.2 & 18.0 & 19.8 & 37.8 & 54.7 & 54.7 \\
Video-LLaMA2 \cite{VideoLLaMA_2} & 7B &  & \checkmark & \checkmark & 60.7 & 59.0 & 55.8 & 54.8 & 4.6 & 29.8 & 36.3 & 33.9 & 33.5 & 67.4 & 58.3 & 56.9 \\
% Video-LLaMA2-16F \cite{VideoLLaMA_2} & 7B &  & \checkmark & \checkmark & 67.9 & 67.7 & 59.8 & 58.9 & 4.0 & 28.6 & 34.4 & 33.3 & 32.6 & 65.9 & 63.9 & 63.3 \\
Qwen2-Audio \cite{Qwen2-Audio} & 7B & \checkmark &  & \checkmark & 52.5 & 37.4 & 53.3 & 41.4 & 4.0 & 26.4 & 32.8 & 30.8 & 30.4 & 61.2 & 52.9 & 39.4 \\
Video-LLaMA2.1-16F \cite{VideoLLaMA_2} & 7B &  & \checkmark & \checkmark & 67.0 & 65.4 & 53.8 & 43.8 & 5.2 & 31.9 & 42.6 & 25.9 & 25.8 & 51.7 & 60.4 & 54.6 \\
Video-LLaMA2.1-AV \cite{VideoLLaMA_2} & 7B & \checkmark & \checkmark & \checkmark & 54.7 & 50.3 & 53.4 & 45.0 & 11.5 & 33.5 & 89.7 & 20.0 & 20.5 & 40.5 & 54.1 & 47.7 \\
LongVA-DPO \cite{LongVA} & 7B &  & \checkmark & \checkmark & 63.6 & 63.3 & 51.6 & 40.0 & 0.0 & 12.4 & 8.2 & 21.9 & 23.3 & 45.2 & 57.6 & 51.7 \\
Qwen2.5-VL \cite{Qwen-VL} & 7B &  & \checkmark & \checkmark & 65.2 & 64.2 & 55.4 & 55.2 & 15.6 & 36.3 & 91.1 & 34.8  & 34.1 & 68.9 & 60.3 & 59.7  \\
InternVideo2-Chat \cite{InternVideo2} & 8B &  & \checkmark & \checkmark & 68.1 & 67.8 & 61.2 & 61.0 & \best{19.5} & \best{45.8} & \best{92.5} & 31.9 & 29.6 & 61.5 & 64.7 & 64.4 \\
MiniCPM-V-2.6 \cite{MiniCPM-V} & 8B &  & \checkmark & \checkmark & 55.1 & 53.9 & 49.6 & 38.6 & 2.3 & 27.5 & 39.1 & 32.9 & 32.0 & 64.9 & 52.4 & 46.3 \\
InternVL2.5 \cite{InternVL} & 8B &  & \checkmark & \checkmark & 66.5 & 66.5 & 59.6 & 59.0 & 0.0 & 12.8 & 16.1 & 17.1 & 19.3 & 36.4 & 63.1 & 62.8 \\
Qwen3-VL \cite{Qwen3} & 8B &  & \checkmark & \checkmark & 68.1 & 66.5 & 61.2 & 61.0 & 15.9 & 35.6 & 90.9 & 36.7  & 35.8  & 72.5 & 64.7  & 63.8 \\
Qwen3-VL-Thinking \cite{Qwen3} & 8B &  & \checkmark & \checkmark & 67.2 & 63.5 & 59.9 & 53.0 &  4.1 & 22.7 & 85.7 & 38.1  & 36.9  & 75.1  & 63.6 & 58.3 \\
Qwen2.5-Omni \cite{Qwen2.5-Omni} & 7B & \checkmark & \checkmark & \checkmark & 64.5 & 63.7 & 62.8 & 62.8 & \second{18.5} & \second{43.6} & \second{91.1} & 34.1 & 33.3 &67.4  & 63.7  & 63.3 \\
Emotion-LLaMA \cite{Emotion-LLaMA} & 7B & \checkmark & \checkmark & \checkmark & 58.0 & 56.8 & 53.0 & 50.7 & 1.5 & 21.2 & 87.6 & 25.4 & 25.9 & 51.3 & 55.5 & 53.8 \\
AffectGPT \cite{AffectGPT} & 7B & \checkmark & \checkmark  & \checkmark & 18.5 & 26.5 & 35.4 & 41.1 & 4.9  & 19.8 & 86.3 & 22.8 & 25.1 & 47.9  & 27.0 & 33.8  \\
\midrule

Qwen2.5-VL \cite{Qwen-VL} & 32B &  & \checkmark & \checkmark & 74.9 & 74.8 & 62.6 & 60.9 & 14.4  & 34.5 & 90.9 & 35.4 & 34.4 & 69.9 & 68.8 & 67.9 \\
InternVL2.5 \cite{InternVL} & 38B &  & \checkmark & \checkmark & 73.0 & 72.7 & 61.2 & 61.2 & 0.3  & 13.3 & 17.5 & 17.1 & 18.6 & 35.7 & 67.1 & 67.0 \\
Video-LLaMA2 \cite{VideoLLaMA_2} & 72B &  & \checkmark & \checkmark & 67.9 & 64.5 & 51.0 & 35.5 & 7.4 & 35.4 & 48.0 & 34.0 & 32.6 & 66.6 & 59.5 & 50.0 \\
InternVL2.5 \cite{InternVL} & 78B &  & \checkmark & \checkmark & 76.8 & 76.8 & 64.4 & 63.6 & 0.0 & 13.0 & 18.2 & 18.2 & 20.0 & 38.2 & 70.6 & 70.2 \\
Qwen2.5-VL \cite{Qwen-VL} & 72B &  & \checkmark & \checkmark & 77.7 & 77.7 & 65.6 & 65.6 & 15.7 & 35.3 & 91.0 & 37.9 & 36.4 & 74.3 & 71.7 & 71.7 \\
\midrule
\rowcolor{gray!25}\multicolumn{17}{c}{\textit{Closed-Source Model (API)}} \\\midrule
GLM-4V-PLUS \cite{GLM-4} & API &  & \checkmark & \checkmark & 74.7 & 74.6 & 59.8 & 58.7 & 14.8 & 33.9 & 90.2 & 37.6 & 36.6 & 74.2 & 67.3 & 66.7 \\
Gemini-1.5-Flash \cite{Gemini} & API & \checkmark & \checkmark & \checkmark & 72.8 & 71.0 & 58.6 & 52.2 & 13.8 & 33.5 & 90.2 & 37.9 & 36.4 & 74.3 & 65.7 & 61.6 \\
Gemini-2.0-Flash \cite{Gemini2} & API & \checkmark & \checkmark & \checkmark & 79.2 & 79.2 & 64.8 & 62.3 & 15.3 & 33.6 & 90.4 & 36.5 & 35.4 & 71.9 & 72.0 &  70.8 \\
Gemini-2.0-Flash-Thinking \cite{Gemini2} & API & \checkmark & \checkmark & \checkmark & 75.6 & 74.7 & 60.4 & 55.8 & 15.8 & 35.5 & 66.4 & 37.5 & 36.6 & 74.1 & 68.0 & 65.3 \\
Gemini-3.0-Flash \cite{gemini3} & API & \checkmark & \checkmark & \checkmark & 79.0 & 78.3 & \second{80.0} & \second{79.9} & 12.6 & 31.2 & 90.1 & \best{38.4} & \second{37.7} & \second{76.1} & \second{79.5}  & \second{79.1} \\
Gemini-3.0-Pro \cite{gemini3} & API & \checkmark & \checkmark & \checkmark &\best{83.4}  &\best{83.3}  & \best{85.6}  & \best{85.6} & 13.4 & 32.2 & 90.0 & 35.9  & 35.1 & 71.0 & \best{84.5}  & \best{84.5} \\
% GPT-4o \cite{gpt4o} & API &  & \checkmark & \checkm}ark & 76.1 & 75.6 & 59.9 & 59.0 & 16.1 & 37.0 & \second{91.3} & \best{38.2} & \best{37.7} & \best{75.9} & 68.0  & 67.3 \\
GPT-5.2 \cite{GPT-5} & API &  & \checkmark & \checkmark & \second{81.7} & \second{81.7} & 73.0 & 73.0 & 14.3 & 34.0 & 90.5 & \second{38.3} & \best{37.9} & \best{76.2} &77.4  &77.4  \\
\bottomrule
\end{tabular}
\vspace{3mm}
\vspace{-4mm}
\label{tab:performance_scea}
\end{table*}

\section{Experiments}

\subsection{Experimental Settings}

\paragraph{Task Formulation.}
We evaluate all MLLMs in a zero-shot setting on EmoBench-M to assess their inherent multimodal emotional intelligence capabilities. For classification tasks, models are prompted to predict emotion categories directly from multimodal inputs, including audio, video, and text. For generative tasks, models are instructed to provide detailed explanations or inferences about the emotional context. To ensure a fair comparison across models, we apply a unified prompting strategy wherever feasible, introducing only minor adaptations when required by input modality. Specifically, for audio-only models, references to “video” in prompts are replaced with “audio”, and for vision-language models that accept multiple images but not full videos, “video” is replaced with “video frames”. These adjustments preserve the intended task while accommodating model-specific input constraints. Consistent prompt templates and evaluation settings are applied across all experiments (see Appendix~\ref{app:prompt} for details).  

\paragraph{Models.}
As shown in Table~\ref{tab:config}, we evaluate a diverse set of LLMs, including both open-source and closed-source systems, to provide a comprehensive analysis of their multimodal emotional intelligence capabilities. The open-source models include InternVL series \cite{InternVL}, InternVideo2 \cite{InternVideo2}, MiniCPM-V \cite{MiniCPM-V}, Video-LLaMA2 \cite{VideoLLaMA_2}, LongVA \cite{LongVA}, Qwen-VL series \cite{Qwen-VL}, Qwen2-Audio \cite{Qwen2-Audio}, and the emotion-specific MLLMs Emotion-LLaMA \cite{Emotion-LLaMA} and AffectGPT \cite{AffectGPT}, covering parameter scales ranging from 4B to 78B. All open-source models are executed using their officially released codebases to ensure reproducibility. Closed-source models evaluated include the GLM-4V-PLUS \cite{GLM-4}, Gemini series \cite{Gemini}, and GPT-5.2 \cite{GPT-5}, which are accessed exclusively via official APIs. More details are provided in Appendix~\ref{app:modelconfig}.

\begin{table*}[!t]\scriptsize
\caption{\small Performance comparison of different methods on $\ours$ (\textbf{C}onversational \textbf{E}motion \textbf{U}nderstanding).}
\centering
\setlength{\tabcolsep}{1.5mm}
\renewcommand{\arraystretch}{1.05}
\begin{tabular}{lcccc@{\hspace{5mm}}cccccccccccc}
\toprule
\multirow{2}{*}{\textbf{Method}} &
\multirow{2}{*}{\textbf{Params}} &
\multirow{2}{*}{\textbf{A}} &
\multirow{2}{*}{\textbf{V}} &
\multirow{2}{*}{\textbf{T}} &
\multicolumn{2}{c}{\textbf{FGDEA}} &
\multicolumn{2}{c}{\textbf{PEA}} &
\multicolumn{2}{c}{\textbf{FCDEA}} &
\multicolumn{2}{c}{\textbf{CEIA}} &
\multicolumn{2}{c}{\textbf{MPDER}} &
\multicolumn{2}{c}{\textbf{Avg.}} \\
\cmidrule(lr){6-7} \cmidrule(lr){8-9} \cmidrule(lr){10-11} 
\cmidrule(lr){12-13} \cmidrule(lr){14-15} \cmidrule(lr){16-17}
 & & & & & ACC & WAF & ACC & WAF & ACC & WAF & ACC & WAF & ACC & WAF & ACC & WAF \\
\midrule
\rowcolor{gray!25}\multicolumn{17}{c}{\textit{Open-Source Model}} \\\midrule
InternVL2.5 \cite{InternVL}            & 4B  &  & \checkmark & \checkmark & 56.9 & 57.9 & 66.8 & 68.6 & 67.5 & 69.0 & 14.0 & 10.7 & 41.2 & 39.9 & 49.3 & 49.2 \\
Video-LLaMA2 \cite{VideoLLaMA_2}       & 7B  &  & \checkmark & \checkmark & 45.7 & 46.4 & 45.2 & 50.2 & 42.7 & 45.2 &  8.1 &  5.0 & 30.7 & 27.9 & 34.5 & 34.9 \\
% Video-LLaMA2-16F \cite{VideoLLaMA_2}   & 7B  &  & \checkmark & \checkmark & 52.9 & 53.6 & 31.6 & 35.4 & 63.0 & 66.6 &  8.3 &  3.3 & 29.6 & 27.5 & 37.1 & 37.3 \\
Qwen2-Audio \cite{Qwen2-Audio}& 7B  & \checkmark &  & \checkmark & 51.6 & 50.1 & 59.0 & 63.7 & 55.6 & 59.1 &  7.6 &  6.0 & 42.7 & 40.8 & 43.3 & 43.9 \\
Video-LLaMA2.1-16F \cite{VideoLLaMA_2} & 7B  &  & \checkmark & \checkmark & 52.8 & 51.8 & 65.6 & 66.1 & 68.5 & 68.4 &  7.9 &  6.8 & 35.5 & 35.3 & 46.1 & 45.7 \\
Video-LLaMA2.1-AV \cite{VideoLLaMA_2}  & 7B  & \checkmark & \checkmark & \checkmark & 51.5 & 48.5 & 68.2 & 69.0 & 67.6 & 67.7 &  6.5 &  4.4 & 36.6 & 34.4 & 46.1 & 44.8 \\
LongVA-DPO \cite{LongVA}               & 7B  &  & \checkmark & \checkmark & 51.1 & 51.0 & 33.2 & 37.2 & 33.3 & 35.2 &  6.1 &  5.4 & 37.0 & 34.6 & 32.1 & 32.7 \\
Qwen2.5-VL  \cite{Qwen-VL}               & 7B  &  & \checkmark & \checkmark & 48.9 & 49.0 & 58.6 & 64.3 & 57.3 & 61.5 & 7.6  & 8.0  & 46.0 & 45.5 & 43.7  & 45.7   \\
InternVideo2-Chat \cite{InternVideo2}  & 8B  &  & \checkmark & \checkmark & 58.0 & 55.4 & 50.8 & 56.2 & 49.2 & 53.2 &  8.9 &  6.2 & 34.2 & 31.6 & 40.2 & 40.5 \\
MiniCPM-V-2.6 \cite{MiniCPM-V}         & 8B  &  & \checkmark & \checkmark & 48.9 & 49.0 & 58.6 & 63.7 & 57.1 & 61.3 & 11.7 &  9.0 & 39.2 & 37.5 & 43.1 & 44.1 \\
InternVL2.5 \cite{InternVL}            & 8B  &  & \checkmark & \checkmark & 48.9 & 48.5 & 61.0 & 65.2 & 62.5 & 65.7 & 12.4 & 13.2 & 43.8 & 42.3 & 45.7 & 47.0 \\
Qwen3-VL \cite{Qwen3}            & 8B  &  & \checkmark & \checkmark & 64.5 & 64.7 & 73.6 & 74.8 & 72.9 & 73.6 & \best{15.7} & 13.1 & 49.5 & 48.8 & 55.2 & 55.0 \\
Qwen3-VL-Thinking \cite{Qwen3}            & 8B  &  & \checkmark & \checkmark & 52.8 & 51.7 & 73.0 & 73.7 & 67.0 & 67.7 & 6.5 & 6.3 & 39.0 & 38.3 & 47.7 & 47.5 \\
Qwen2.5-Omni \cite{Qwen2.5-Omni}            & 7B  & \checkmark & \checkmark & \checkmark & 57.4 & 57.0 & 62.8 & 67.8 & 63.5 & 66.8 & 6.28 & 11.8 & 43.0 & 41.5 & 46.6 & 49.0 \\
Emotion-LLaMA \cite{Emotion-LLaMA}        & 7B  & \checkmark & \checkmark & \checkmark & 62.0 & 59.2 & 24.6 & 23.8 & 25.2 & 20.4 &  2.90&  3.80& 38.9 & 34.1 & 30.7 & 28.3 \\
AffectGPT \cite{AffectGPT} & 7B  & \checkmark & \checkmark & \checkmark & 55.5 & 56.8 & \second{75.4} & 75.3 & 72.0 & 71.1 &  10.3&  \best{18.7}& 39.8 & 40.7 & 50.6  & 52.5  \\
\midrule
Qwen2.5-VL \cite{Qwen-VL} & 32B &  & \checkmark & \checkmark & 57.6 & 57.5 & 64.8 & 69.1 & 62.8 & 66.5 &12.0  &11.7  & 48.1 & 48.5 & 49.1 & 50.7 \\
InternVL2.5 \cite{InternVL}           & 38B &  & \checkmark & \checkmark & 56.1 & 57.2 & 66.2 & 70.4 & 65.2 & 68.2 & 13.5 & 13.3 & 43.5 & 42.5 & 48.9 & 50.3 \\
Video-LLaMA2 \cite{VideoLLaMA_2}      & 72B &  & \checkmark & \checkmark & 43.1 & 41.2 & 42.8 & 49.5 & 41.4 & 45.4 & 11.6 & 12.5 & 47.6 & 48.1 & 37.3 & 39.3 \\
InternVL2.5 \cite{InternVL}           & 78B &  & \checkmark & \checkmark & 52.7 & 53.1 & 56.8 & 62.9 & 56.7 & 61.1 & 12.6 & 11.7 & 43.5 & 41.8 & 44.5 & 46.1 \\
Qwen2.5-VL \cite{Qwen-VL}    & 72B &  & \checkmark & \checkmark & 51.6 & 51.5 & 64.2 & 68.1 & 64.3 & 67.6 & 11.4 & 10.7 & 47.8 & 48.4 & 47.9 & 49.3 \\
\midrule
\rowcolor{gray!25}\multicolumn{17}{c}{\textit{Closed-Source Model (API)}} \\\midrule
GLM-4V-PLUS \cite{GLM-4}                  &  API  &  & \checkmark & \checkmark & 51.8 & 53.1 & 62.8 & 67.1 & 65.4 & 67.1 & 14.7 & 13.0 & 41.6 & 40.5 & 47.3 & 48.2 \\
Gemini-1.5-Flash \cite{Gemini}            &  API  & \checkmark & \checkmark & \checkmark & 67.2 & 67.2 & 72.3 & 74.3 & 73.2 & 74.0 & \second{15.6} & 15.8 & 49.5 & 49.2 & 55.6 & 56.1 \\
Gemini-2.0-Flash \cite{Gemini2}           &  API  & \checkmark & \checkmark & \checkmark & 64.2 & 64.8 & 70.9 & 73.6 & 71.9 & 73.8 & 11.1 & 12.7 & 48.7 & 48.5 & 53.4 & 54.7 \\
Gemini-2.0-Flash-Thinking \cite{Gemini2}  &  API  & \checkmark & \checkmark & \checkmark & 64.5 & 65.4 & 71.2 & 72.8 & 71.6 & 72.6 & 12.0 & 13.7 & 51.5 & 51.8 & 54.2 & 55.3 \\
Gemini-3.0-Flash \cite{gemini3}            &  API  & \checkmark & \checkmark & \checkmark & \second{72.5} & \second{73.1} & 73.0 & 76.4 & 73.7 & \second{76.5} & 14.6 & \second{16.0} & 47.0 & 46.0 & 56.2 & \second{57.6} \\
Gemini-3.0-Pro \cite{gemini3}            &  API  & \checkmark & \checkmark & \checkmark & \best{81.0} & \best{81.1} & \best{81.2} & \best{81.9} & \best{82.6} & \best{83.4} & 12.6  & 13.9 & \second{54.0} & \second{53.5} &\best{62.3}  &\best{62.8}  \\
GPT-5.2 \cite{GPT-5} & API &  & \checkmark & \checkmark & 68.4 & 68.3 & \second{75.4} & \second{76.5} & \second{75.9} & 76.4 & 13.1 & 11.8 & \best{54.6} & \best{54.5} & \second{57.5} &  57.5\\
\bottomrule
\end{tabular}
\vspace{3mm}
\vspace{-4mm}
\label{tab:performance_ceu}
\end{table*}

\begin{table}[!t]\scriptsize
\caption{\small Performance comparison (\%) on $\ours$. Mean ACC was used per dimension; LR
total score was used in SCEA.}
\centering
\setlength{\tabcolsep}{3mm}
% \resizebox{\linewidth}{!}{
\begin{tabular}{lcccc}
\toprule
\bf Method & \bf FER & \bf CEU & \bf SCEA & \bf Avg. \\
\midrule
\rowcolor{gray!25}\multicolumn{5}{c}{\textit{Open-Source Model}} \\\midrule
Random &  23.1 &  19.8 &  33.3 &  25.4 \\
InternVL2.5-4B & 54.5 & 49.3 & 49.0 & 50.9 \\
Video-LLaMA2-7B & 45.4 & 34.5 & 61.3 & 47.1 \\
% Video-LLaMA2-7B-16F & 51.4 & 37.1 & 64.5 & 51.0 \\
Qwen2-Audio-7B & 59.9 & 43.3 & 55.7 & 53.0 \\
Video-LLaMA2.1-7B-16F & 50.9 & 46.1 & 57.5 & 51.5 \\
Video-LLaMA2.1-7B-AV & 50.4 & 46.1 & 49.5 & 48.7 \\
LongVA-DPO-7B & 45.7 & 32.1 & 53.5 & 43.8 \\
Qwen2.5-VL-7B & 51.7  & 43.7 & 63.2 & 52.9 \\
InternVideo2-Chat-8B & 50.6 & 40.2 & 63.6 & 51.5 \\
MiniCPM-V-2.6-8B & 40.0 & 43.1 & 56.5 & 46.5 \\
InternVL2.5-8B & 51.2 & 45.7 & 54.2 & 50.4 \\
Qwen3-VL-8B & 57.8 & 55.2  & 67.3 & 60.1 \\
Qwen3-VL-Thinking-8B & 46.7 & 47.7  & 67.4 & 53.9 \\
Qwen2.5-Omni-7B & 49.7 & 46.6 & 64.9 & 53.7 \\
Emotion-LLaMA & 36.9 & 30.7 & 54.1 & 40.6 \\
AffectGPT & 47.3 & 50.6  & 33.9 & 43.9 \\
\midrule
Qwen2.5-VL-32B & 51.4 & 49.1 & 69.1 & 56.7 \\
InternVL2.5-38B & 57.6 & 48.9 & 56.6 & 54.4 \\
Video-LLaMA2-72B & 50.7 & 37.3 & 61.8 & 49.9 \\
InternVL2.5-78B & 53.0 & 44.5 & 59.8 & 52.4 \\
Qwen2.5-VL-72B & 54.8 & 47.9 & 72.5 & 57.8 \\
\midrule
\rowcolor{gray!25}\multicolumn{5}{c}{\textit{Closed-Source Model (API)}} \\\midrule
GLM-4V-PLUS & 56.1 & 47.3 & 69.6 & 57.7 \\
Gemini-1.5-Flash & 59.7 & 55.6 & 68.6 & 61.3 \\
Gemini-2.0-Flash & 61.4 & 53.4 & 72.0 & 62.3 \\
Gemini-2.0-Flash-Thinking & 57.7 & 54.2 & 70.0 & 60.6 \\
Gemini-3.0-Flash & 62.0 & 56.2 & \second{78.4} & 65.5 \\
Gemini-3.0-Pro & \best{69.3} & \best{62.3} & \best{80.0} & \best{70.5} \\
GPT-5.2 & \second{65.0} & \second{57.5} & 77.0 & \second{66.5} \\
\bottomrule
\end{tabular}
% }
\vspace{2mm}
\vspace{-7.7mm}
\label{tab:performance_avg}
\end{table}

\subsection{Results and Findings}

Tables~\ref{tab:performance_fer}, \ref{tab:performance_ceu}, and \ref{tab:performance_scea} summarize the performance of various models on EmoBench-M across three dimensions: Foundational Emotion Recognition (FER), Conversational Emotion Understanding (CEU), and Socially Complex Emotion Analysis (SCEA). Overall, closed-source models consistently outperform open-source models. Table~\ref{tab:performance_avg} shows that Gemini-3.0-Pro achieves the best performance across all dimensions, with FER 69.3, CEU 62.3, SCEA 80.0, and an average score of 70.5. Among open-source models, Qwen2-Audio-7B leads in FER (59.9), particularly excelling on SOER and SPER, due to the indoor video sources where facial expressions are subtle and audio cues dominate. Qwen2.5-VL-72B attains the top SCEA score (72.5) and demonstrates strong overall performance. In addition, AffectGPT performs relatively well on PEA, FCDEA, and OSA tasks, which can be explained by the similarity between its training data and these evaluation scenarios. Examining the progression across model versions highlights the impact of scaling and multimodal architectures. Successive Gemini models show consistent improvements in emotional intelligence, with Gemini-2.0-Flash achieving an average score of 62.3 and Gemini-3.0-Flash reaching 65.5, indicating a clear upward trend. In contrast, the Qwen series exhibits only modest variation across variants: Qwen2-Audio-7B scores 53.0, Qwen2.5-VL-7B achieves 52.9, and Qwen2.5-Omni-7B, as a full-modal model, reaches 53.7. This limited improvement highlights the challenges of effectively integrating multiple modalities for emotion understanding.

Analysis of these results reveals several factors driving the performance differences. 
\textbf{(1) Model Scale and Modality Focus:} At smaller scales (4-8B), audio-centric models like Qwen2-Audio excel in FER (Table \ref{tab:performance_fer}), highlighting the primacy of speech for basic emotion analysis. At larger scales (38-78B), models with stronger vision encoders, such as InternVL, gain an advantage, suggesting that effective visual-textual fusion becomes increasingly critical as model capacity increases.
\textbf{(2) Native Multimodal Architecture:} The superior performance of the Gemini series likely stems from its
native multimodal design, which synchronously processes
video and audio streams. This holistic approach is better
suited for capturing the dynamic nature of emotional expression compared to models that fuse modalities at later stages.
\textbf{(3) Perception over Complex Reasoning:} Gemini-2.0-Flash-Thinking and Qwen3-VL-Thinking do not consistently outperform their base versions and sometimes perform slightly worse (Table~\ref{tab:performance_avg}), suggesting that emotion-centric tasks rely more on effective cross-modal perception than on complex chain-of-thought reasoning.
\textbf{(4) Emotion-specialized MLLMs: } Despite targeted emotion training, Emotion-LLaMA and AffectGPT achieve improvements on specific emotion-related tasks. Emotion-specific priors alone are insufficient to achieve robust emotional intelligence in complex, multimodal, and context-dependent settings. 
\textbf{(5) Limits of Comprehensive Emotional Understanding: } MLLMs exhibit relatively stronger performance on SCEA compared to FER and CEU. However, despite this relative advantage, their performance across all three tasks remains markedly inferior to human benchmarks, underscoring persistent limitations in capturing nuanced social cues, long-range emotional dynamics, and context-dependent interactions.

% \subsection{Comparison with Human Performance}
% Table~\ref{tab:performance_avg} compares the performance of MLLMs and human performance on EmoBench-M across Foundational Emotion Recognition (FER), Conversational Emotion Understanding (CEU), and Socially Complex Emotion Analysis (SCEA), together with the average score (Avg.). The benchmark data are collected from established emotion datasets, and their annotations are treated as the gold-standard reference for evaluation.
% Among the evaluated models, Gemini-3.0-Pro achieves the highest average score of 70.5, followed by GPT-5.2 at 66.5 and Gemini-3.0-Flash at 65.5. Despite these results, differences between MLLMs and human performance remain across the three emotion dimensions.
% In FER, models such as Gemini-3.0-Pro (69.3) and GPT-5.2 (65.0) show competitive performance in recognizing basic emotions from multimodal inputs. In CEU, MLLMs exhibit a clear performance gap relative to human performance, reflecting the difficulty of modeling emotional dynamics and long-range conversational context.
% In SCEA, Gemini-3.0-Pro achieves a score of 80.0, approaching but still trailing human-level performance, indicating that large-scale multimodal pretraining enhances the modeling of socially complex emotional patterns such as humor and sarcasm, while humans retain an overall advantage. 

\begin{figure*}[!t]
    \centering

    % ===== Row 1 =====
    \begin{minipage}[b]{0.24\linewidth}
        \centering
        \includegraphics[width=\linewidth]{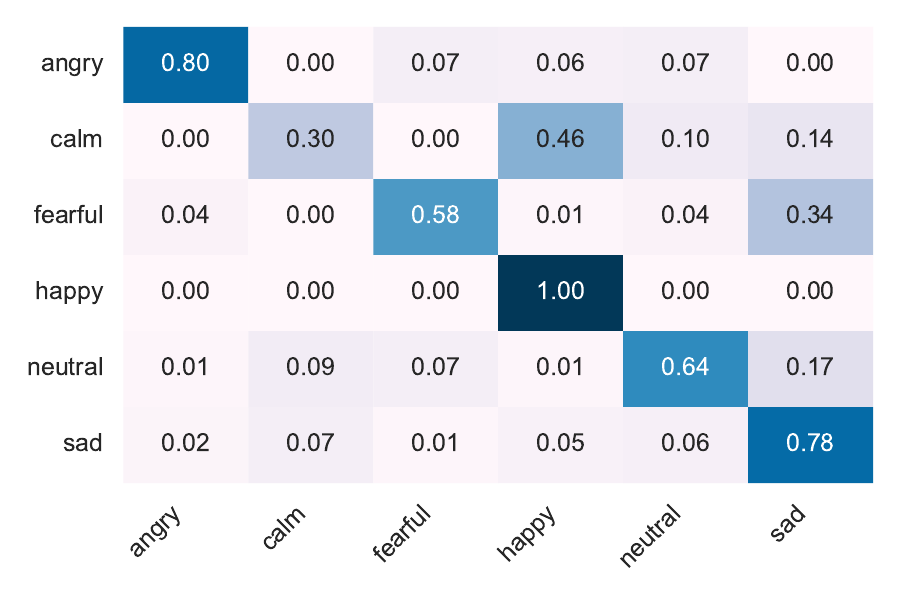}
        \subfloat{\small (a) SOER}
    \end{minipage} \hfill
    \begin{minipage}[b]{0.24\linewidth}
        \centering
        \includegraphics[width=\linewidth]{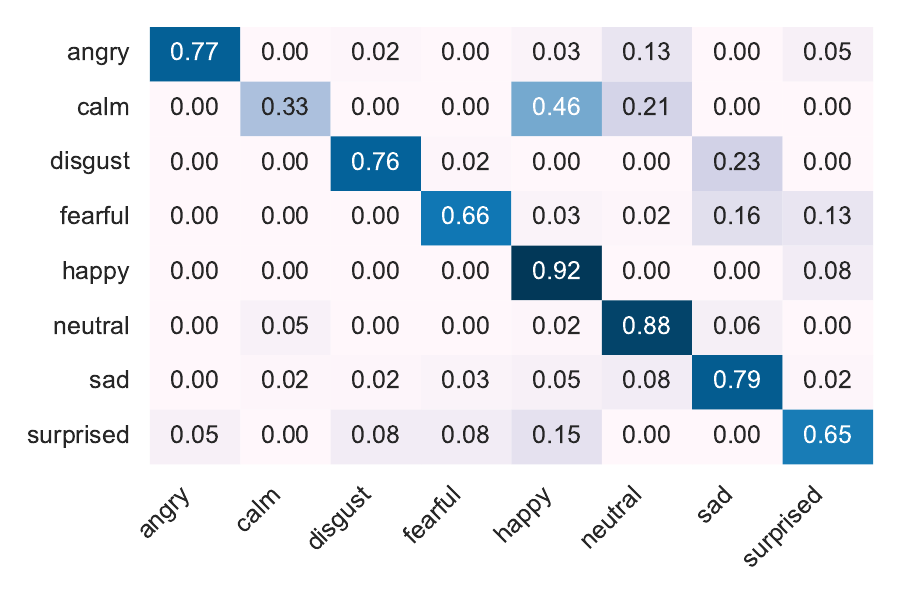}
        \subfloat{\small (b) SPER}
    \end{minipage} \hfill
    \begin{minipage}[b]{0.16\linewidth}
        \centering
        \includegraphics[width=\linewidth]{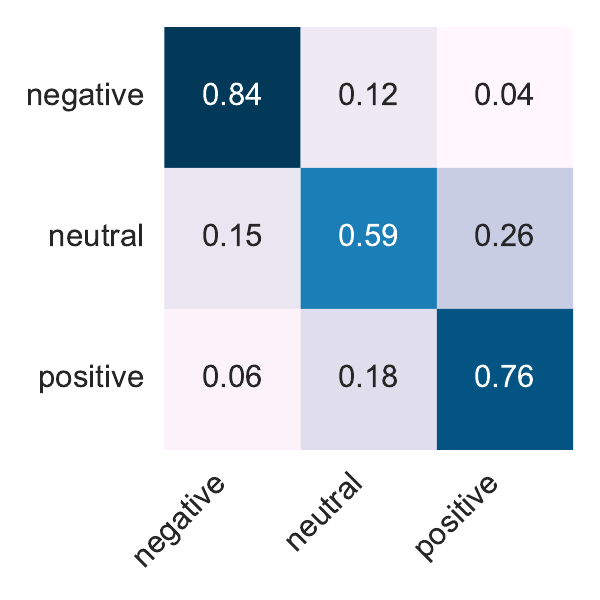}
        \subfloat{\small (d) OSA}
    \end{minipage} \hfill
    \begin{minipage}[b]{0.16\linewidth}
        \centering
        \includegraphics[width=\linewidth]{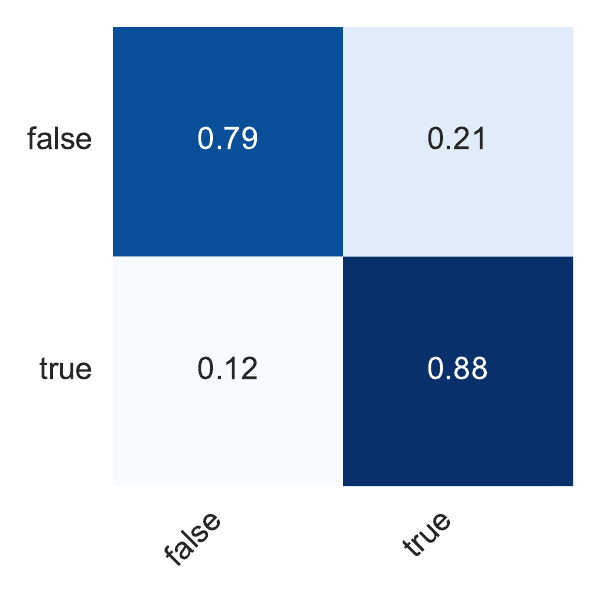}
        \subfloat{\small (l) HU}
    \end{minipage} \\

    % ===== Row 2 =====
    \begin{minipage}[b]{0.24\linewidth}
        \centering
        \includegraphics[width=\linewidth]{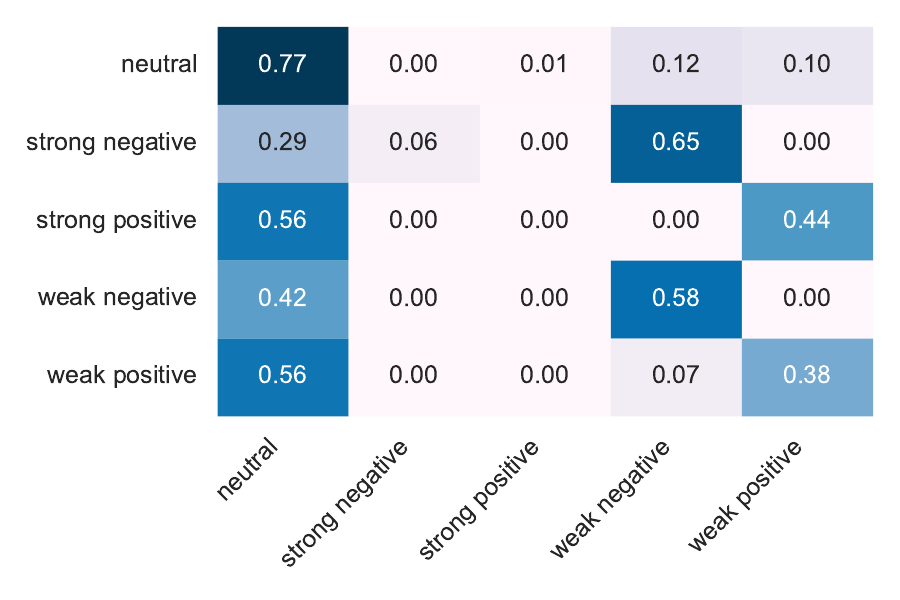}
        \subfloat{\small (c) SEA}
    \end{minipage} \hfill
    \begin{minipage}[b]{0.24\linewidth}
        \centering
        \includegraphics[width=\linewidth]{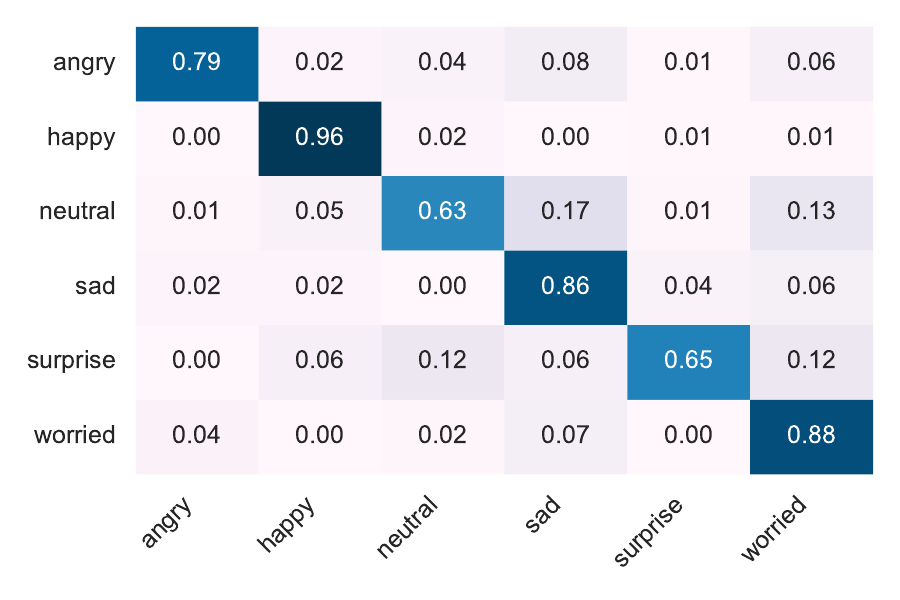}
        \subfloat{\small (f) FGDEA}
    \end{minipage} \hfill
    \begin{minipage}[b]{0.16\linewidth}
        \centering
        \includegraphics[width=\linewidth]{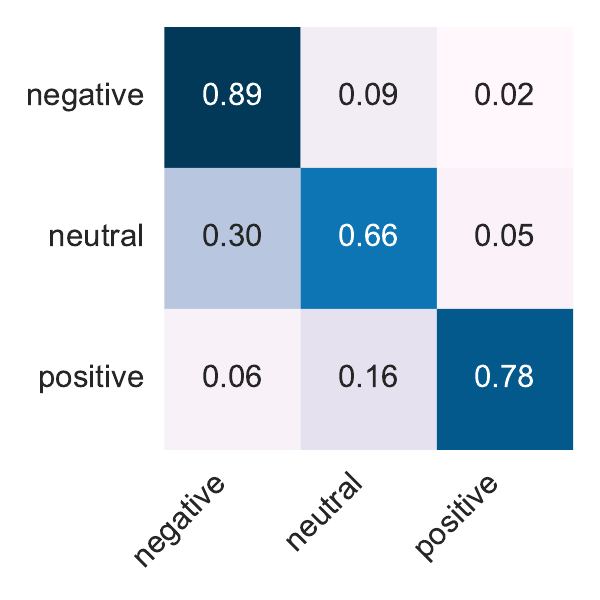}
        \subfloat{\small (g) FCDEA}
    \end{minipage} \hfill
    \begin{minipage}[b]{0.16\linewidth}
        \centering
        \includegraphics[width=\linewidth]{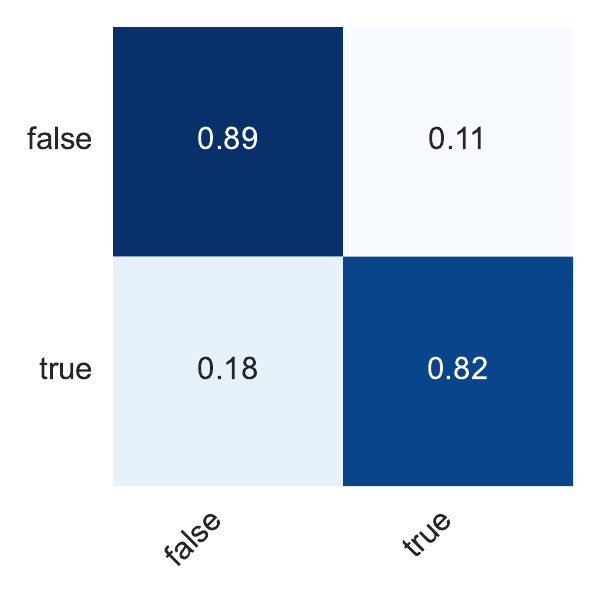}
        \subfloat{\small (m) SD}
    \end{minipage} \\

    % ===== Row 3 =====
    \begin{minipage}[b]{0.16\linewidth}
        \centering
        \includegraphics[width=\linewidth]{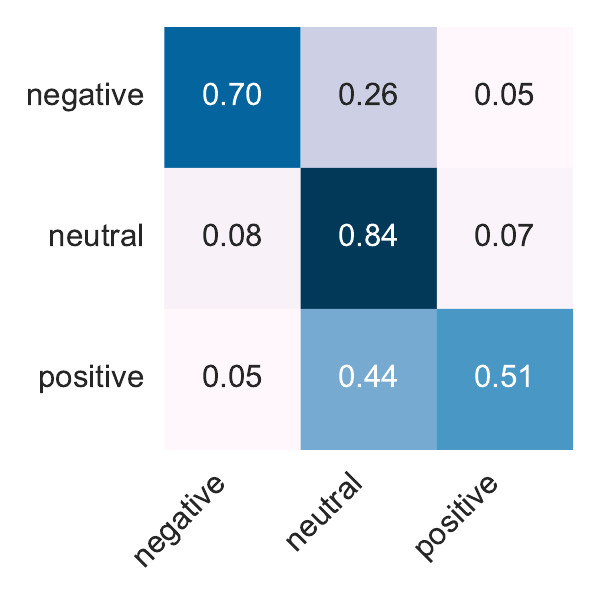}
        \subfloat{\small (e) EIA}
    \end{minipage} \hfill
    \begin{minipage}[b]{0.20\linewidth}
        \centering
        \includegraphics[width=\linewidth]{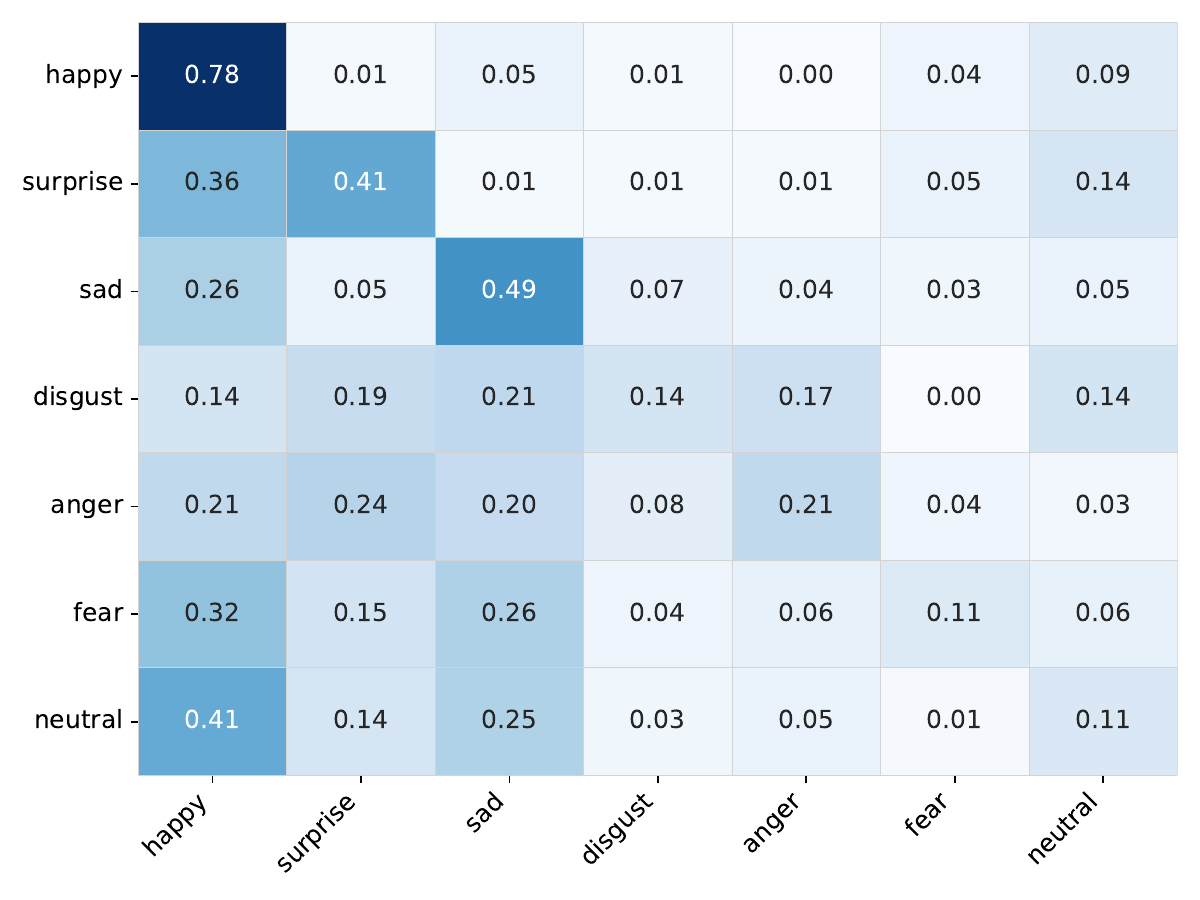}
        \subfloat{\small (i) CEIA (emo.)}
    \end{minipage} \hfill
    \begin{minipage}[b]{0.20\linewidth}
        \centering
        \includegraphics[width=\linewidth]{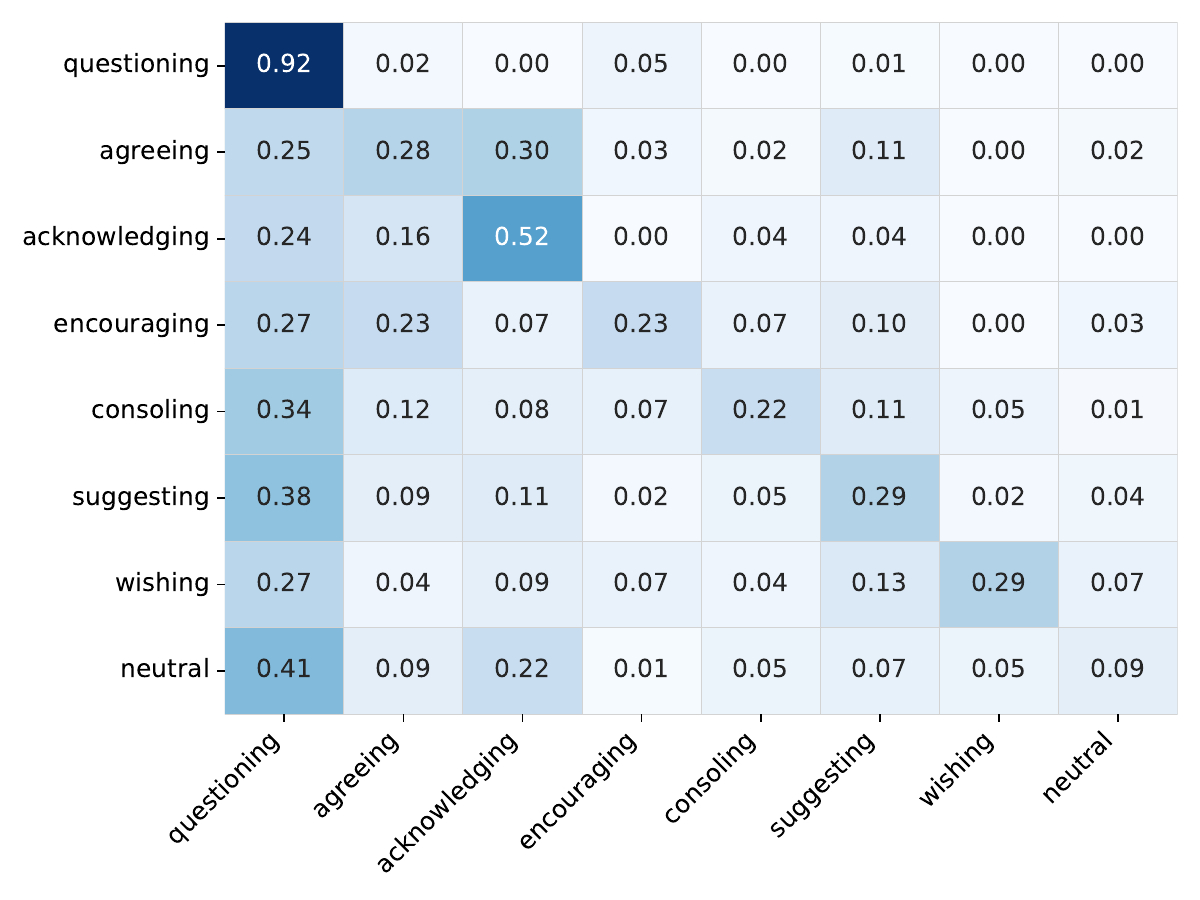}
        \subfloat{\small (j) CEIA (int.)}
    \end{minipage} \hfill
    \begin{minipage}[b]{0.24\linewidth}
        \centering
        \includegraphics[width=\linewidth]{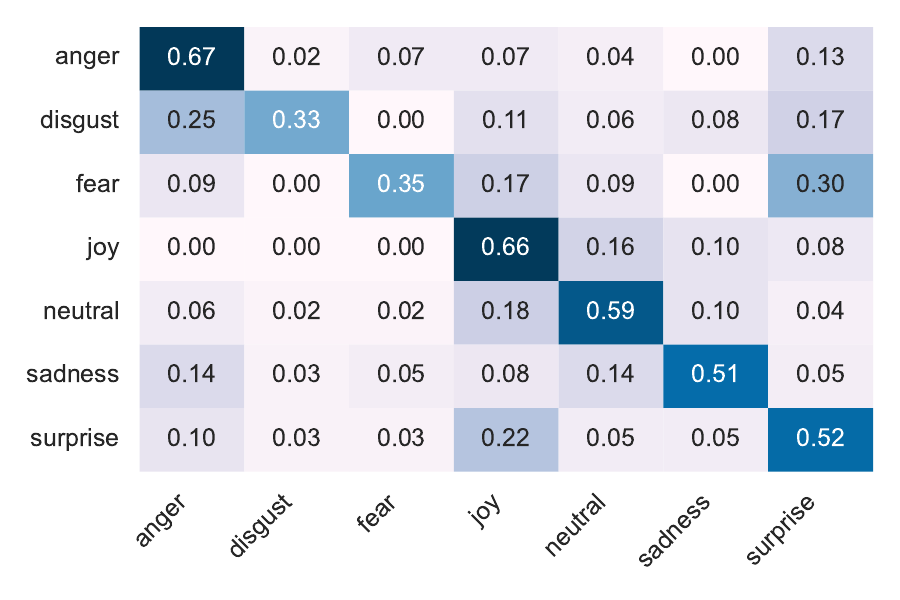}
        \subfloat{\small (k) MPDER}
    \end{minipage} \hfill
    \begin{minipage}[b]{0.16\linewidth}
        \centering
        \includegraphics[width=\linewidth]{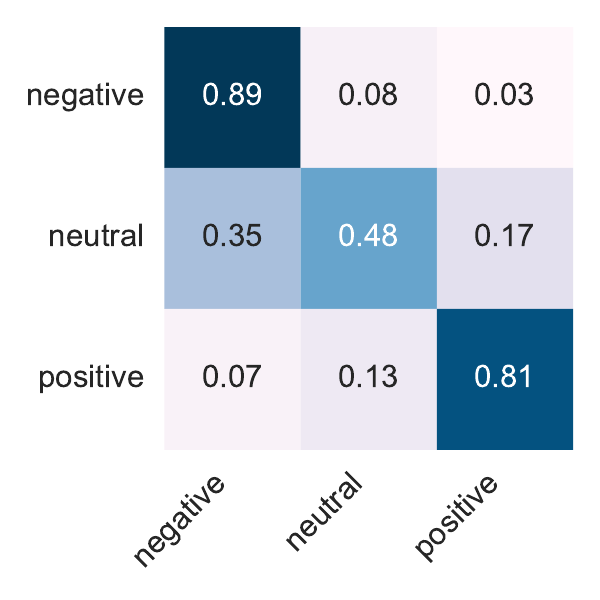}
        \subfloat{\small (h) PEA}
    \end{minipage}

    \caption{\small Confusion matrices for Gemini-3.0-Pro on each evaluation scenario of $\ours$.}
    \label{fig:confusion-Gemini-3.0-Pro}
\end{figure*}

\subsection{Analysis on Class-wise Performance}
As the best-performing model, Gemini-3.0-Pro serves as a representative case for analysis. Figure~\ref{fig:confusion-Gemini-3.0-Pro} shows the class-wise performance of Gemini-3.0-Pro across Foundational Emotion Recognition (FER), Conversational Emotion Understanding (CEU), and Socially Complex Emotion Analysis (SCEA). In FER (SOER, SPER, SEA, OSA, EIA), the model performs well on primary emotions such as angry, happy, sad, and neutral, as well as coarse sentiment categories like positive and negative, while confusions persist between semantically adjacent categories, including fearful versus sad, calm versus happy or neutral, and fine-grained sentiment intensities. In CEU (FGDEA, PEA, FCDEA, CEIA, MPDER), the model shows stable performance on common emotions, such as neutral, and handles structured intent categories reasonably well, whereas emotions with higher contextual dependency, such as surprise, and nuanced intent classes like encouraging and consoling, remain challenging. In SCEA (HU, SD), the model achieves strong performance in humor understanding, but sarcasm detection exhibits increased confusion, reflecting the difficulty of recognizing subtle social cues. Gemini-3.0-Pro demonstrates strong performance while continuing to struggle with nuanced and overlapping categories in socially and contextually complex scenarios.

\subsection{Analysis on Generation Metric}
Table \ref{tab:metric} shows cosine similarity and Pearson correlation results for traditional metrics (BLEU-4, ROUGE-L, BERTScore) and the open-source LLM Qwen2.5-72B with human judgments. Among traditional metrics, BERTScore showed the highest consistency with human evaluation, outperforming BLEU-4 and ROUGE-L. Notably, Qwen2.5-72B achieved superior results, highlighting its ability to better align with human assessments in laughter reasoning tasks. It reveals the limitations of traditional metrics and underscores the potential of LLM-based approaches for more reliable evaluation.

\begin{table}[!t]\small
\caption{\small Cosine similarity and Pearson correlation between metric scores and human evaluation scores in laughter reasoning. }
\centering
% \setlength{\tabcolsep}{0.95mm}
% \resizebox{\linewidth}{!}{
\begin{tabular}{lcc}
\toprule
\bf Metric & \textbf{Cosine Sim.} & \textbf{Pearson Corr.} \\
\midrule
BLEU-4  & 0.7381 & 0.2934 \\
ROUGE-L & 0.8563 & 0.2947 \\
BERTScore & 0.8762 & 0.3199 \\
\midrule
\rowcolor{blue!10}
Qwen2.5-72B & \bf 0.9353 & \bf 0.4042 \\
\bottomrule
\end{tabular}
% }
\vspace{4mm}
\label{tab:metric}
\vspace{-4mm}
\end{table}

\begin{table}[!t]\small
\caption{\small Stability experiment of Gemini-2.0-Flash on three dimensions, running predictions 1, 3, and 5 times. }
\centering
\setlength{\tabcolsep}{0.95mm}
% \resizebox{0.9\linewidth}{!}{
\begin{tabular}{lccc}
\toprule
\bf Dimensions                          & \bf 1\textsuperscript{st}     & \bf 3\textsuperscript{rd}     & \bf 5\textsuperscript{th}     \\ \midrule
Foundational Emotion Recognition        & \bf{61.4}  & 61.2  & 61.0  \\
Conversational Emotion Understanding    & 53.4  & 54.0  & \bf{54.1}  \\
Socially Complex Emotion Analysis       & 72.0  & 72.5  & \bf{72.8}  \\ \bottomrule
\end{tabular}
% }
\vspace{4mm}
\vspace{-8mm}
\label{fig:stable}
\end{table}

\subsection{Stability Analysis of MLLM}
We analyze the stability of Gemini-2.0-Flash across three dimensions: FER, CEU, and SCEA. Stability is evaluated by running predictions 1, 3, and 5 times, using majority voting for final results (shown in Table~\ref{fig:stable}). In FER, scores are 61.4, 61.2, and 61.0, with the highest at 61.4. For CEU, scores improve from 53.4 to 54.1 across iterations. In SCEA, results are 72.0, 72.5, and 72.8, considering only HU and SD scenarios due to the generative nature of this task. These results demonstrate the model's robust stability, with minor variations in complex emotional contexts.

% \begin{table}[!t]\small
% \centering
% \setlength{\tabcolsep}{3.5mm}
% \begin{tabular}{lcc}
% \toprule
% \bf Method & \bf Params & \bf Average \\
% \midrule
% Video-LLaMA2-7B-16F & 7B & 51.0 \\
% \textbf{Video-LLaMA2.1-7B-16F} & 7B & 51.5 {\color{red} ($\uparrow$ +0.5)} \\
% \midrule
% Qwen2.5-VL-7B & 7B & 52.9 \\
% \textbf{Qwen3-VL-8B} & 8B & 60.1 {\color{red} ($\uparrow$ +7.2)} \\
% \midrule
% Gemini-1.5-Flash & API & 61.3 \\
% \textbf{Gemini-2.0-Flash} & API & 62.3 {\color{red} ($\uparrow$ +1.0)} \\
% \bottomrule
% \end{tabular}
% \vspace{-1mm}
% \caption{\small Pairwise comparison with model parameter sizes. Red values indicate absolute improvements over the previous version.}
% \vspace{-2mm}
% \label{tab:avg_with_params}
% \end{table}

% \begin{table}[!t]\small
% \centering
% \setlength{\tabcolsep}{3mm}
% \renewcommand{\arraystretch}{1.05}
% \begin{tabular}{lcc}
% \toprule
% \bf Method & \bf Params & \bf Average
% \\\midrule
% Qwen3-VL & 8B & 60.1 \\
% \textbf{Qwen3-VL-Thinking} & 8B & 53.9 {\color{red} ($\downarrow$ 6.2)} \\
% \midrule
% Gemini-2.0-Flash & API & 62.3 \\
% \textbf{Gemini-2.0-Flash-Thinking} & API & 60.6 {\color{red} ($\downarrow$ 1.7)} \\
% \bottomrule
% \end{tabular}
% \vspace{-1mm}
% \caption{\small Pairwise comparison (average scores only) of reasoning variants.
% Red values indicate absolute differences (percentage points) compared to the corresponding base model.}
% \vspace{-2mm}
% \label{tab:pairwise_reasoning_avg}
% \end{table}

\section{Conclusion }
In this work, we present EmoBench-M, a multimodal benchmark for evaluating emotional intelligence in MLLMs. It combines classification metrics with generative and LLM-based evaluations on data curated and verified by human experts. Experiments on 27 models reveal a substantial gap between current MLLMs and human-level EI. Although emotion-specific models perform well on certain tasks, their generalization remains limited, highlighting the complexity of emotional understanding in multimodal settings. Future work should prioritize enhancing contextual understanding and social awareness in MLLMs. Key directions include improving dialogue modeling and social reasoning, strengthening multimodal fusion and cross-modal alignment, integrating psychologically informed priors and culturally diverse datasets, and exploring adaptive reasoning and reinforcement learning to support more dynamic and human-like emotional understanding.

% Future research can prioritize work that can focus on enhancing models' contextual understanding and social awareness in MLLMs. Key areas include: leveraging advanced dialogue modeling, incorporating social reasoning, leveraging mechanisms, and diverse datasets with encompassing varied cultural contexts, improving multimodal fusion, integrating techniques, and embedding psychological principles directly into model design architecture can further advance EI capabilities. 

% \section{Case Study}

% \section*{Limitations}
% The limitation of our study is the exclusion of MLLMs specifically fine-tuned on emotion-centric datasets. We focus on evaluating general-purpose MLLMs to understand their EI capabilities in realistic applications, where models are expected to generalize across diverse tasks and domains. This aligns with our objective of assessing the broader applicability of MLLMs in real-world scenarios rather than focusing on specialized models designed for emotion computation. 

% \section*{Ethical Considerations}
% In this study, we exclusively utilized publicly available datasets with open-source links. Furthermore, we ensured compliance with the licensing agreements associated with these datasets by formally obtaining usage permissions where required. As a result, our research does not raise any ethical concerns regarding data usage or handling.

% \clearpage
\bibliographystyle{ACM-Reference-Format}
\bibliography{acmart-primary/samples/sample-base.bib}
 
% \appendix

% \clearpage
% {\small
% \bibliographystyle{named}
% \bibliography{ijcai26}
% }
\clearpage
\appendix

\section{Details and Case of Datasets}\label{app:dataset}
\begin{figure*}[ht]
    \centering
    \includegraphics[width=\textwidth]{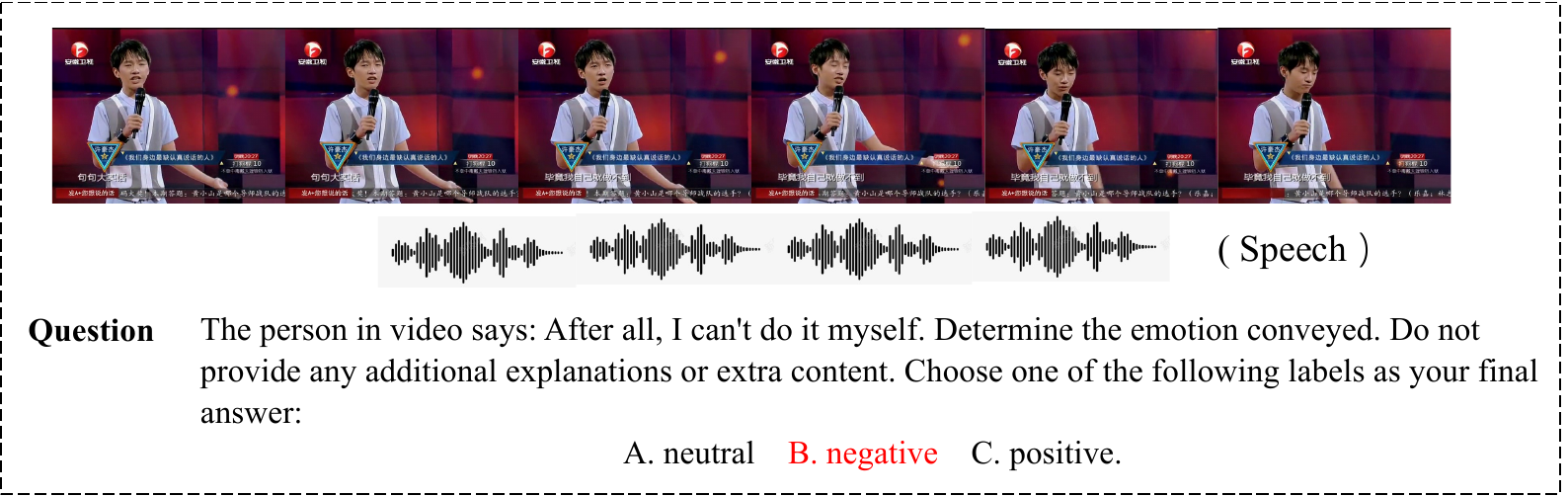} 
    \caption{\small Example of CH-SIMS dataset.}
    \label{fig:Ch_SIMS}
\end{figure*}

\begin{figure*}[ht]
    \centering
    \includegraphics[width=\textwidth]{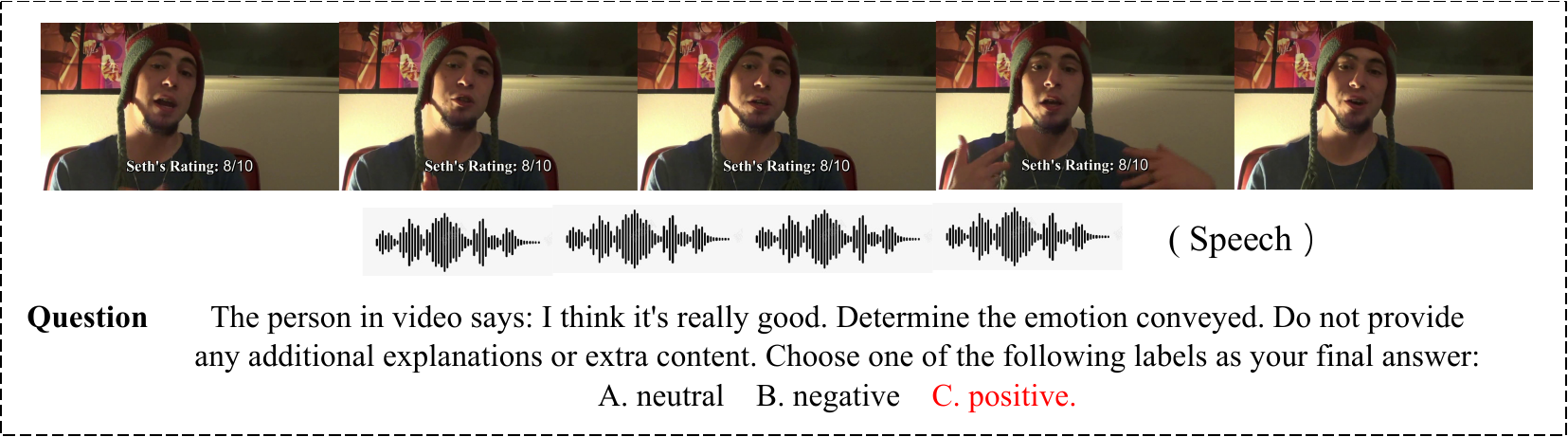} 
    \caption{\small Example of CMU-MOSEI dataset.}
    \label{fig:CMU_MOSEI}
\end{figure*}

\begin{figure*}[ht]
    \centering
    \includegraphics[width=\textwidth]{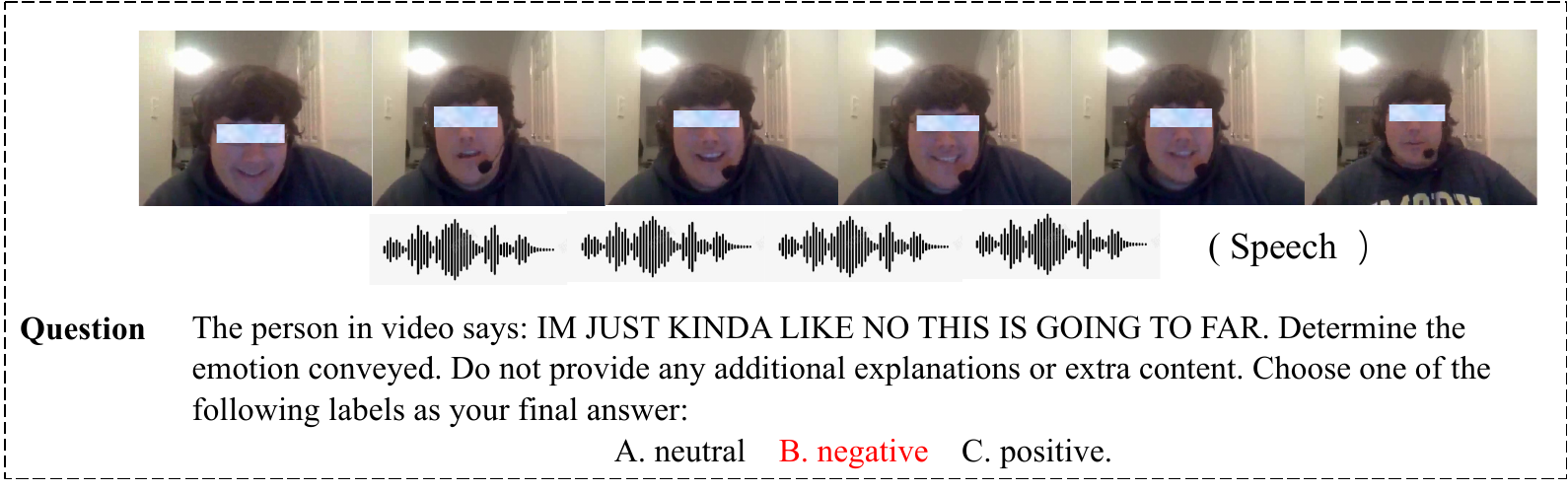} 
    \caption{\small Example of CMU-MOSI dataset.}
    \label{fig:CMU_MOSI}
\end{figure*}

\begin{figure*}[ht]
    \centering
    \includegraphics[width=\textwidth]{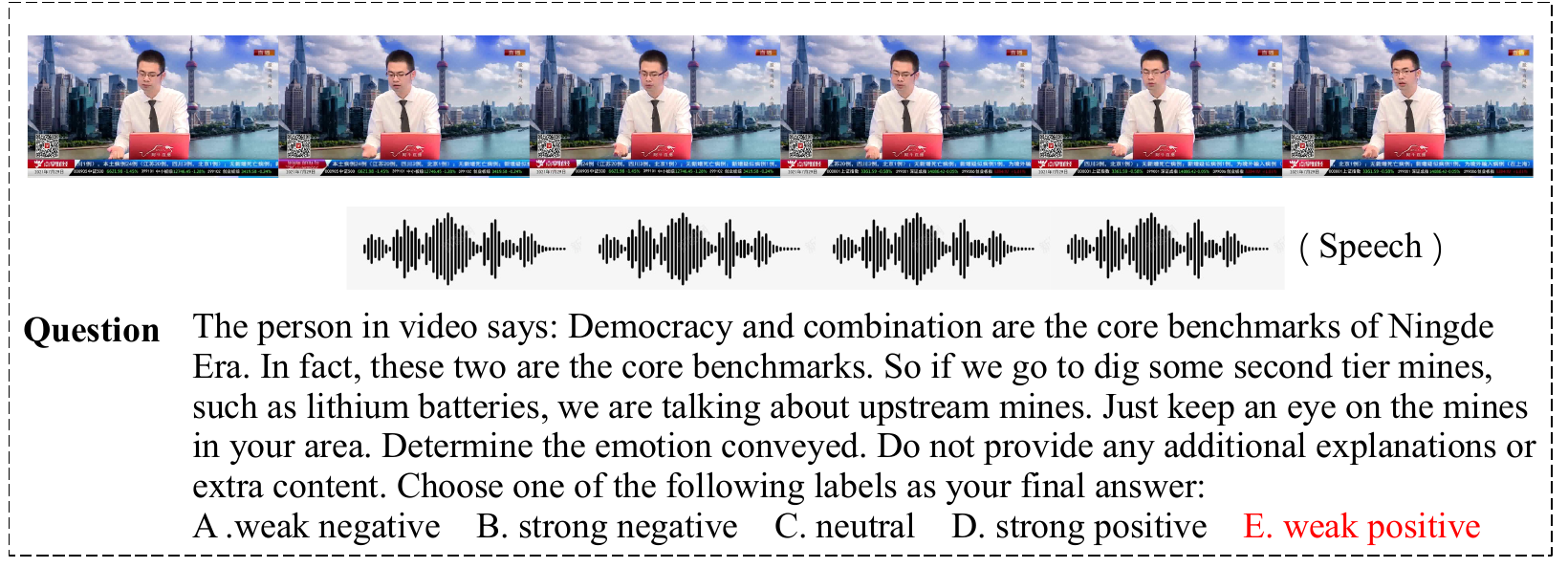} 
    \caption{\small Example of FMSA-SC dataset.}
    \label{fig:FMSA-SC}
\end{figure*}

\begin{figure*}[ht]
    \centering
    \includegraphics[width=\textwidth]{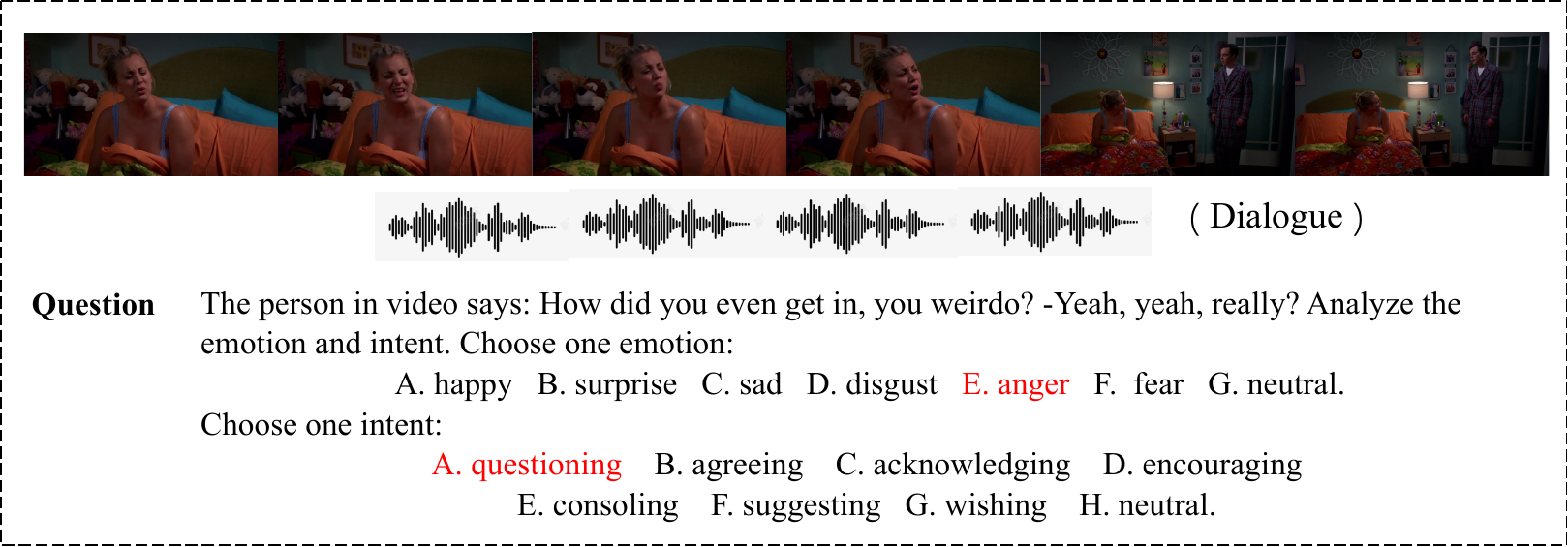} 
    \caption{\small Example of MC-EIU dataset.}
    \label{fig:MC-EIU}
\end{figure*}

\begin{figure*}[ht]
    \centering
    \includegraphics[width=\textwidth]{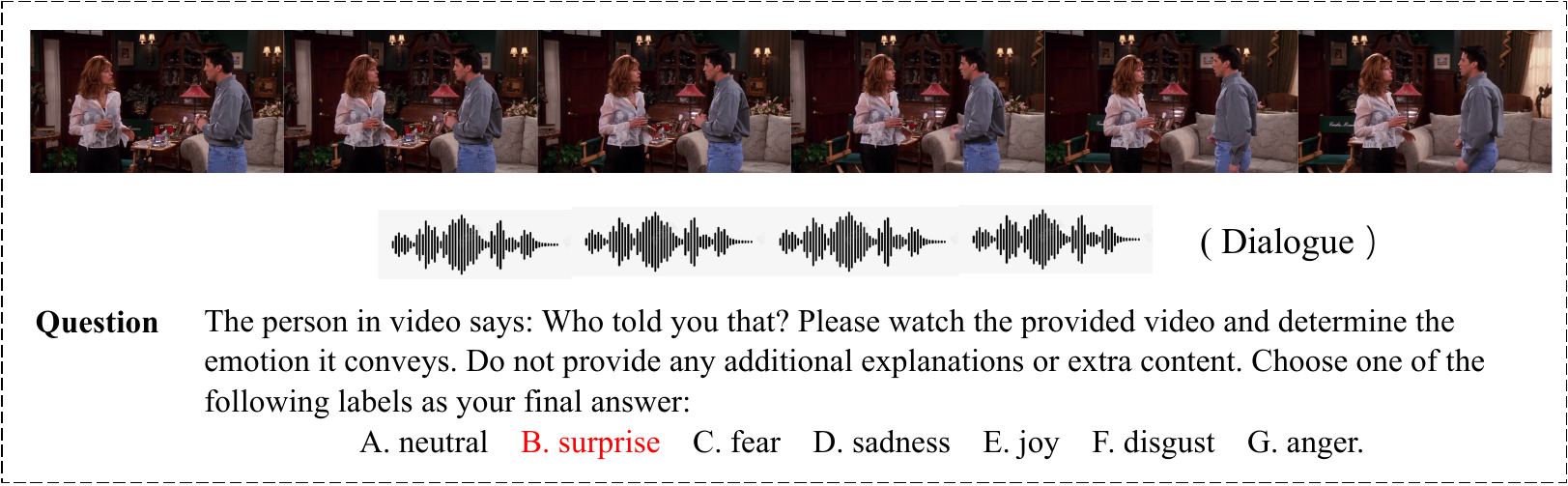} 
    \caption{\small Example of MELD dataset.}
    \label{fig:MELD}
\end{figure*}

\begin{figure*}[ht]
    \centering
    \includegraphics[width=\textwidth]{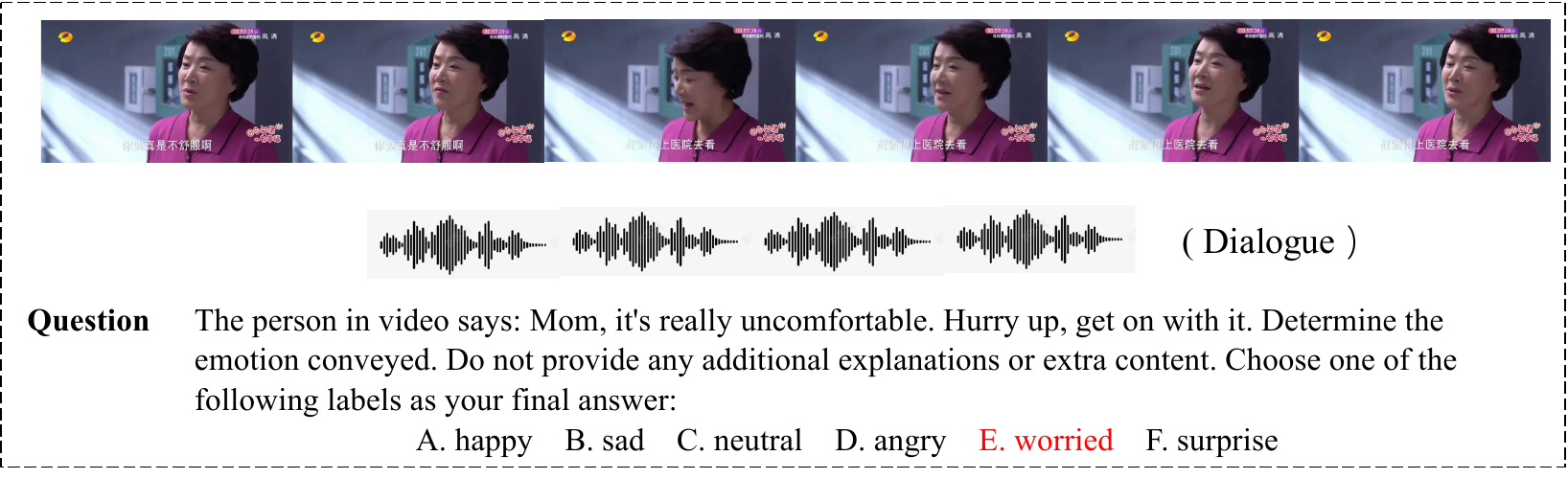} 
    \caption{\small Example of MER2023 dataset.}
    \label{fig:MER2023}
\end{figure*}

\begin{figure*}[ht]
    \centering
    \includegraphics[width=\textwidth]{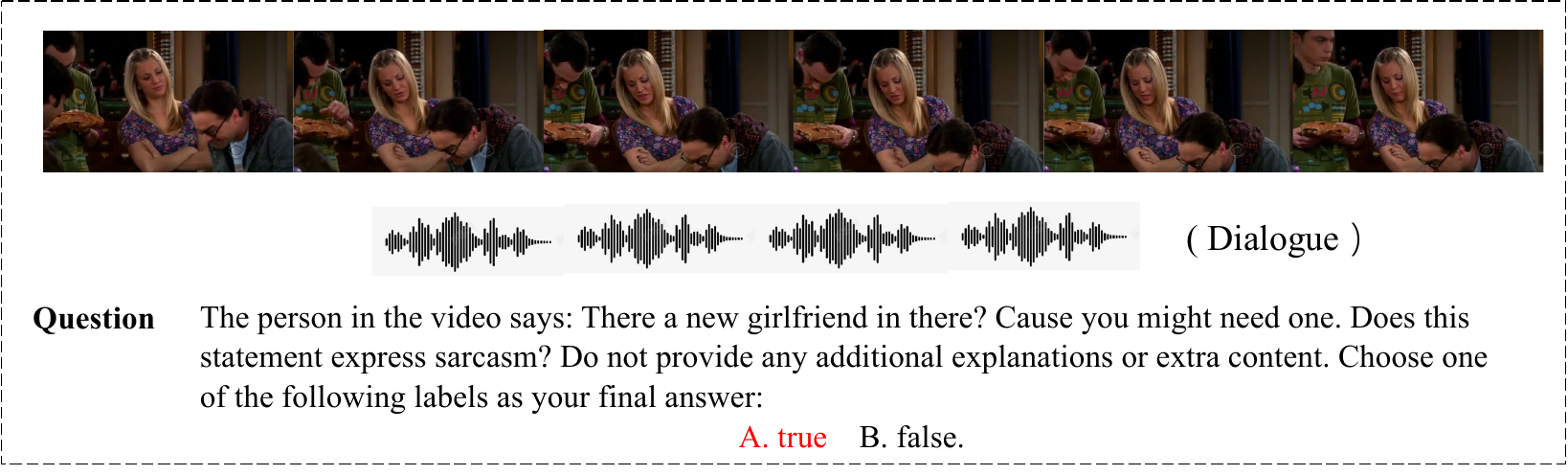} 
    \caption{\small Example of MUStARD dataset.}
    \label{fig:MUStARD}
\end{figure*}

\begin{figure*}[ht]
    \centering
    \includegraphics[width=\textwidth]{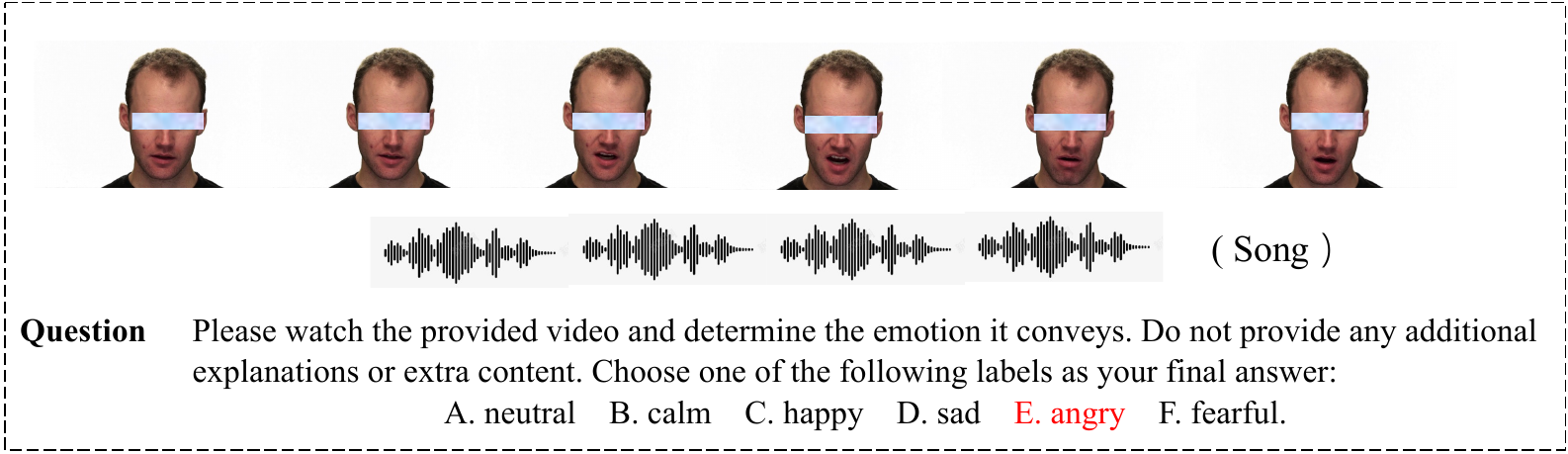} 
    \caption{\small Example of RAVDSS-song dataset.}
    \label{fig:RAVDSS-song}
\end{figure*}

\begin{figure*}[ht]
    \centering
    \includegraphics[width=\textwidth]{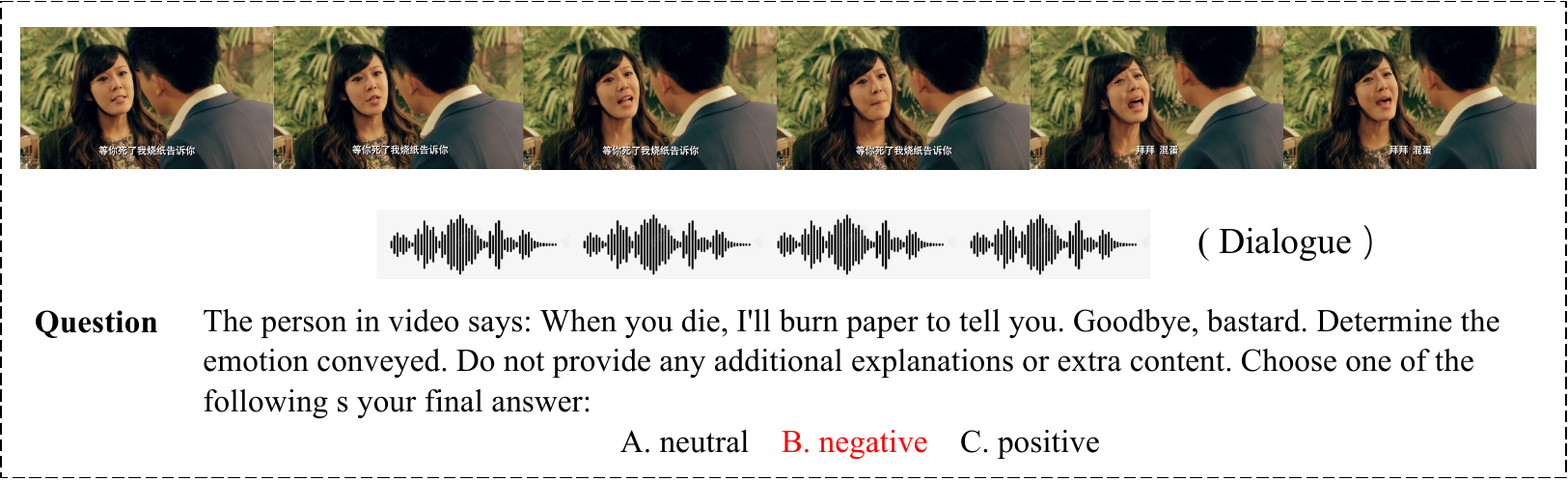} 
    \caption{\small Example of CH-SIMSv2 dataset.}
    \label{fig:Ch_SIMSv2}
\end{figure*}

\begin{figure*}[ht]
    \centering
    \includegraphics[width=\textwidth]{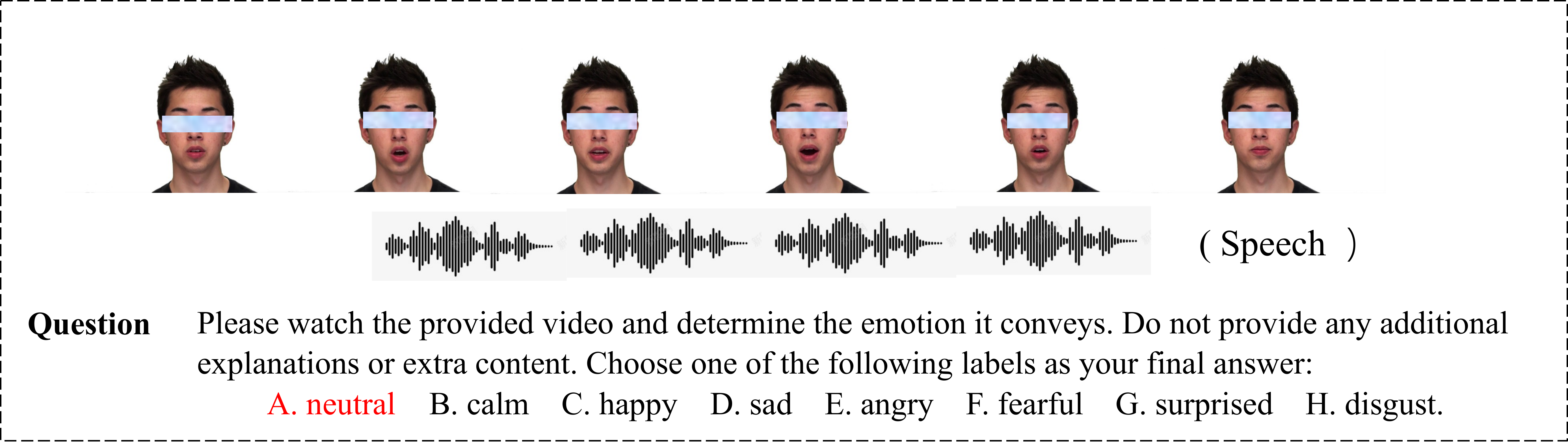} 
    \caption{\small Example of RAVDSS-speech dataset.}
    \label{fig:RAVDSS-speech}
\end{figure*}

\begin{figure*}[ht]
    \centering
    \includegraphics[width=\textwidth]{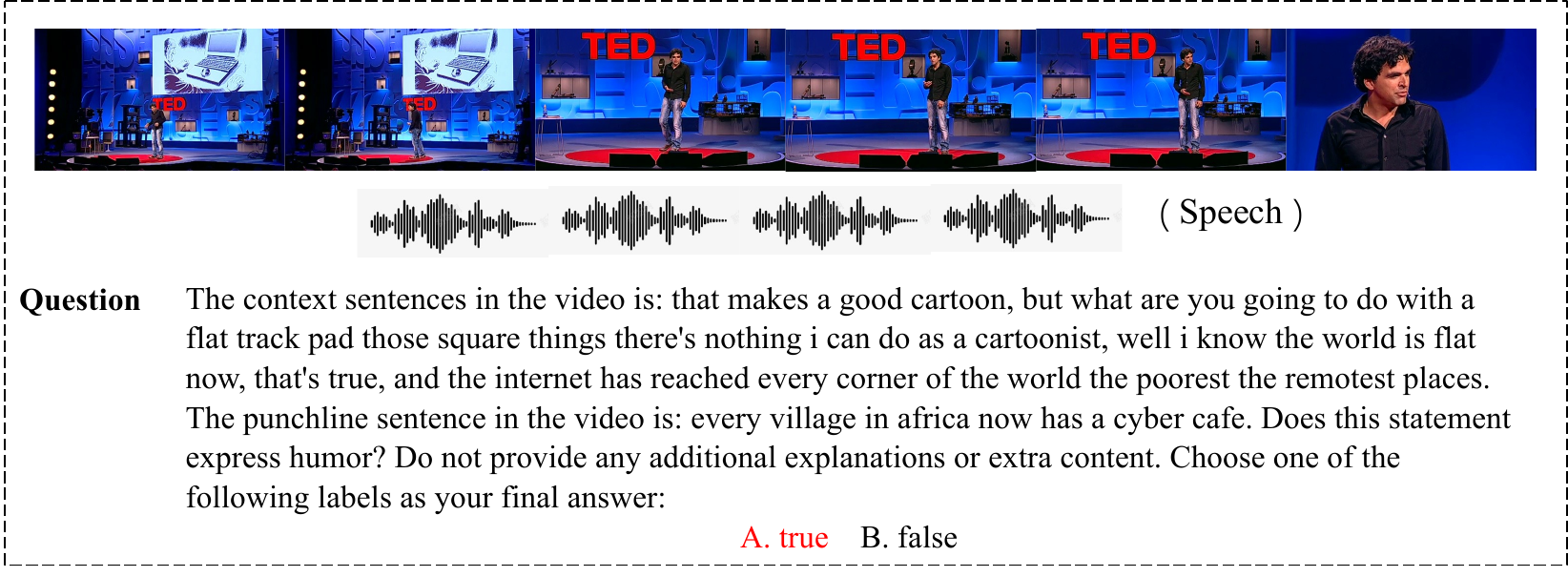} 
    \caption{\small Example of UR-FUNNY dataset.}
    \label{fig:UR-FUNNY}
\end{figure*}

\begin{figure*}[ht]
    \centering
    \includegraphics[width=\textwidth]{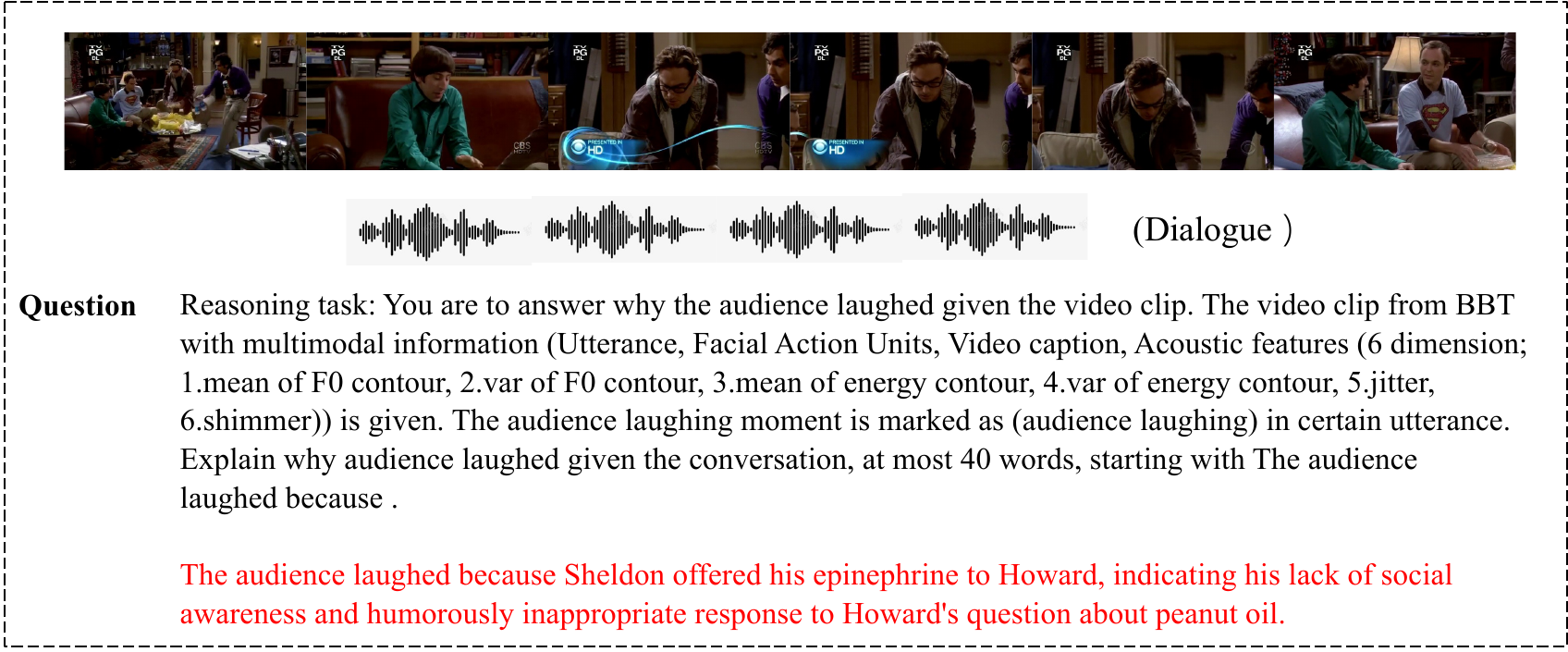} 
    \caption{\small Example of SMILE dataset.}
    \label{fig:SMILE}
\end{figure*}

This section provides an overview of the datasets used in our experiments, highlighting the number of test samples and the corresponding labels associated with each dataset. Table~\ref{tab:datasets} summarizes this information, covering a wide range of emotional and intent-based annotations across various modalities. Visual examples for each dataset are provided in Figure~\ref{fig:Ch_SIMS}-\ref{fig:SMILE}.
\begin{itemize}
    \item \textbf{RAVDESS (song \& speech) :} The RAVDESS dataset includes both speech and song audio files annotated with emotions such as \textit{neutral}, \textit{calm}, \textit{happy}, \textit{sad}, \textit{angry}, and \textit{fearful}. The speech subset contains additional labels, including \textit{surprised} and \textit{disgust}.
    
    \item \textbf{CMU-MOSI and CMU-MOSEI:} These datasets are designed for multimodal sentiment analysis and include three sentiment labels: \textit{neutral}, \textit{positive}, and \textit{negative}.
    
    \item \textbf{FMSA-SC:} This dataset captures fine-grained sentiment annotations with labels such as \textit{weak negative}, \textit{strong negative}, \textit{neutral}, \textit{weak positive}, and \textit{strong positive}.
    
    \item \textbf{MER2023:} Designed for emotion recognition, this dataset provides six emotion categories: \textit{happiness}, \textit{sadness}, \textit{anger}, \textit{surprise}, \textit{neutral}, and \textit{calm}.
    
    \item \textbf{CH-SIMSv2 and CH-SIMS:} These datasets, used for sentiment analysis in Chinese, are annotated with three labels: \textit{neutral}, \textit{negative}, and \textit{positive}.
    
    \item \textbf{MC-EIU:} This dataset offers both emotion annotations (\textit{happy}, \textit{surprise}, \textit{sad}, \textit{disgust}, \textit{anger}, \textit{fear}, \textit{neutral}) and intent annotations (\textit{questioning}, \textit{agreeing}, \textit{acknowledging}, \textit{encouraging}, \textit{consoling}, \textit{suggesting}, \textit{wishing}, \textit{neutral}).
    
    \item \textbf{MELD:} This multimodal dataset includes seven emotion labels: \textit{neutral}, \textit{surprise}, \textit{fear}, \textit{sadness}, \textit{joy}, \textit{disgust}, and \textit{anger}.
    
    \item \textbf{UR-FUNNY and MUStARD:} Both binary-labeled multimodal datasets. UR-FUNNY is used for humor detection, while MUStARD is designed for sarcasm detection, with annotations of \textit{true} or \textit{false}.
    
    \item \textbf{SMILE:} A small-scale dataset focusing on humor comprehension, annotated with explanations of why the audience laughed.
\end{itemize}
Table~\ref{tab:datasets} provides detailed statistics on the test samples and label distributions for each dataset.

\begin{table}[!t]\small
\caption{\small Model configuration evaluated on $\ours$.}
\centering
\resizebox{\linewidth}{!}{
\setlength{\tabcolsep}{2pt}
\begin{tabular}{lcccr}
\toprule
\textbf{Method} & \textbf{top-p} & \textbf{top-k} & \textbf{temp.} & \textbf{VRAM} \\
\midrule
InternVL2.5-4B \cite{InternVL}    & 1.0 & 50 & 1.0 & 14G \\
Video-LLaMA2-7B \cite{VideoLLaMA_2}      & 0.8 & 20 & 0.7 & 19G \\
Video-LLaMA2-7B-16F \cite{VideoLLaMA_2}    & 0.8 & 20 & 0.7 & 19G \\
Qwen2-Audio-7B-Instruct \cite{Qwen2-Audio}       & 0.5 & 20 & 0.7 & 33G \\
Video-LLaMA2.1-7B-16F \cite{VideoLLaMA_2}   & 0.8 & 20 & 0.7 & 22G \\
Video-LLaMA2.1-7B-AV \cite{VideoLLaMA_2} & 0.8 & 20 & 0.7 & 21G \\
LongVA-DPO-7B \cite{LongVA}      & 1.0 & 50 & 1.0 & 21G \\
Qwen2.5-VL-Instruct-7B \cite{Qwen-VL}      & 0.8 & 20 & 0.7 & 17G \\
InternVideo2-Chat-8B \cite{InternVideo2}  & 1.0 & 50 & 1.0 & 17G \\
MiniCPM-V-2.6-8B \cite{MiniCPM-V}    & 0.8 & 100 & 0.7 & 19G \\
InternVL2.5-8B \cite{InternVL}          & 1.0 & 50 & 1.0 & 24G \\
Qwen3-VL-Instruct-8B \cite{Qwen3}      & 0.95 & 20 & 0.6 & 18G \\
Qwen3-VL-Instruct-Thinking-8B \cite{Qwen3}      & 0.8 & 20 & 0.7 & 18G \\
Qwen2.5-Omni-7B \cite{Qwen2.5-Omni}      & 1.0 & 50 & 1.0 & 36G \\
Emotion-LLaMA \cite{Emotion-LLaMA}      & 0.6 & - & 0.9 & 16G \\
AffectGPT \cite{AffectGPT}      & 0.9 & - & 1.0 & 16G \\
Qwen2.5-VL-Instruct-32B \cite{Qwen-VL}      & 0.8 & 20 & 0.7 & 63G \\
InternVL2.5-38B \cite{InternVL}   & 1.0 & 50 & 1.0 & 73G \\
Video-LLaMA2-72B \cite{VideoLLaMA_2}      & 0.8 & 20 & 0.7 & 148G \\
InternVL2.5-78B \cite{InternVL}           & 1.0 & 50 & 1.0 & 168G \\
Qwen2.5-VL-Instruct-72B \cite{Qwen-VL}      & 0.8 & 20 & 0.7 & 147G \\
\midrule
GLM-4V-PLUS \cite{GLM-4}            & 0.6 & - & 0.8 & API \\
Gemini-1.5-Flash \cite{Gemini}            & 0.95 & 40 & 1.0  & API \\
Gemini-2.0-Flash \cite{Gemini2}        & 0.95 & 40 & 1.0 & API \\
Gemini-2.0-Flash-Thinking \cite{Gemini2}        & 0.95 & 64 & 1.0 & API \\
Gemini-3.0-Flash \cite{gemini3}        & 0.95 & 64 & 1.0 & API \\
Gemini-3.0-Pro \cite{gemini3}        & 0.95 & 64 & 1.0 & API \\
GPT-5.2  \cite{gpt4o}        & 0.9 & - & 1.0 & API \\
\bottomrule
\end{tabular}}
\vspace{2mm}
\label{tab:config}
\vspace{-5mm}
\end{table}

\section{Model Configuration}\label{app:modelconfig}
The configuration details of the models evaluated on $\ours$ are summarized in Table~\ref{tab:config}. The table provides a detailed comparison of key hyperparameters, including top-p and top-k sampling values, temperature settings, and VRAM requirements. For models accessed via APIs, the VRAM is denoted as ``API'', reflecting their closed-source nature and cloud-based deployment.
The evaluated models encompass a range of architectures, from smaller-scale models, such as InternVL2.5-4B~\cite{InternVL}, to large-scale variants like InternVL2.5-78B~\cite{InternVL}. Noteworthy configurations include Video-LLaMA2.1-AV-7B~\cite{VideoLLaMA_2}, which incorporates audiovisual processing, and Gemini-2.0-Flash-Thinking~\cite{Gemini}, which features enhanced reasoning capabilities.
Most models exhibit consistent sampling parameters (e.g., top-p and temperature), ensuring a standardized basis for comparison. The VRAM requirements reflect the computational demands of these models, ranging from 14 GB for smaller models to 168 GB for the largest configurations. In contrast, API-based models provide an accessible option for users, particularly when local resources are constrained, albeit at the cost of reduced transparency due to their closed-source nature.

\begin{table*}[!t]\small
\caption{\small Overview of datasets, test samples, and emotion/intent labels.}
\centering
\begin{tabular}{@{}p{3cm}p{2cm}p{9cm}@{}}
\toprule
\textbf{Dataset}          & \textbf{\# Test samples} & \textbf{Labels}                                                                 \\ \midrule
RAVDESS (song)            & 500                      & neutral, calm, happy, sad, angry, fearful                                       \\
RAVDESS (speech)          & 500                      & neutral, calm, happy, sad, angry, fearful, surprised, disgust                  \\
CMU-MOSI                  & 500                      & neutral, negative, positive                                                    \\
CMU-MOSEI                 & 500                      & neutral, negative, positive                                                    \\
FMSA-SC                   & 250                      & weak negative, strong negative, neutral, strong positive, weak positive        \\
MER2023                   & 411                      & happiness, sadness, anger, surprise, neutral, calm                             \\
CH-SIMSv2                 & 500                      & neutral, negative, positive                                                    \\
CH-SIMS                   & 457                      & neutral, negative, positive                                                    \\
MC-EIU                    & 500                      & emotion: happy, surprise, sad, disgust, anger, fear, neutral                  \\ 
                          &                          & intent: questioning, agreeing, acknowledging, encouraging, consoling,      suggesting, wishing, neutral                                                   \\
MELD                      & 500                      & neutral, surprise, fear, sadness, joy, disgust, anger                          \\
UR-FUNNY                  & 448                      & true, false                                                                    \\
MUStARD                   & 500                      & true, false                                                                    \\
SMILE                     & 80                       & The audience laughed because...                                                \\ \bottomrule
\end{tabular}
\vspace{2mm}
\label{tab:datasets}
\end{table*}

% \begin{table*}[!t]\small
% \centering
% \setlength{\tabcolsep}{0.9mm}
% \begin{tabular}{lccccccccccccc}
% \toprule
% \bf Method & \bf SOER & \bf SPER & \bf OSA & \bf EIA & \bf SCEA & \bf FGDEA & \bf PEA & \bf FCDEA & \bf CEIA & \bf MPDER & \bf HU & \bf SD & \bf LR \\
% \midrule
% Random & 16.6 & 12.5 & 33.3 & 33.3 & 20.0 & 16.6 & 33.3 & 33.3 & 1.8 & 14.2 & 50.0 & 50.0 & 0.0 \\
% InternVL2.5-78B & 48.8 & 41.2 & 63.2 & 59.4 & 52.4 & 52.7 & 56.8 & 56.7 & 12.6 & 43.5 & 76.8 & 64.4 & 38.2 \\
% GLM-4V-PLUS & 54.9 & 43.7 & 70.0 & 61.2 & 50.8 & 51.8 & 62.8 & 65.4 & 14.7 & 41.6 & 74.7 & 59.8 & 74.2 \\
% Gemini-3.0-Flash & 58.6 & 59.6 & 69.6 & 63.5 & 58.8 & 72.5 & 73.0 & 73.7 & 14.6 & 47.0 & 79.0 & 80.0 & 76.1 \\
% GPT-5.2 & 65.8 & 59.4 & 80.8 & 66.0 & 53.2 & 68.4 & 75.4 & 75.9 & 13.1 & 54.6 & 81.7 & 73.0 & 76.2 \\
% \bottomrule
% \end{tabular}
% \caption{\small Performance comparison of MLLMs and humans across different scenarios.}
% \label{fig:scenario_comp}
% \end{table*}

\begin{table*}[!t]\small
\caption{\small Stability experiment of Gemini-2.0-Flash on different scenarios, running predictions 1, 3, and 5 times, with the majority voting to obtain the final result.}
\setlength{\tabcolsep}{0.9mm}
\centering
\begin{tabular}{ccccccccccccc}
\toprule
\bf Scenarios & \bf SOER & \bf SPER & \bf OSA & \bf EIA & \bf SCEA & \bf FGDEA & \bf PEA & \bf FCDEA & \bf CEIA & \bf MPDER & \bf HU & \bf SD \\ \midrule
\bf 1 & 63.3 & 55.8 & 68.8 & 63.5 & 55.6 & 64.2 & 70.9 & 71.9 & 11.1 & 48.7 & 79.2 & 64.8 \\
\bf 3 & 64.6 & 55.4 & 67.8 & 63.4 & 54.8 & 66.4 & 71.3 & 71.9 & 11.1 & 49.5 & 79.2 & 65.8 \\
\bf 5 & 63.8 & 54.8 & 69.0 & 63.4 & 54.0 & 66.2 & 72.1 & 72.1 & 11.1 & 48.9 & 79.7 & 66.0 \\ \bottomrule
\end{tabular}
\vspace{2mm}
\label{fig:more_stable_transposed}
\end{table*}

\section{Details on Stability Analysis of MLLM}\label{app:stability}
Table~\ref{fig:more_stable_transposed} shows the results of stability experiments conducted on Gemini-2.0-Flash. The analysis evaluates the model's performance across different numbers of prediction iterations (1, 3, and 5), with the final output determined through a majority voting mechanism. Scenarios assessed include SOER, SPER, OSA, EIA, SCEA, FGDEA, PEA, FCDEA, CEIA, MPDER, HU, and SD.
The results demonstrate consistent performance across different iteration counts, with minimal variation observed in most metrics. For instance, the performance on SOER shows only a slight fluctuation, ranging from 63.3 in the single iteration setup to 63.8 in the five-iteration case. Similarly, performance on SPER and EIA exhibit stable trends, reinforcing the model's robustness against runtime variability.

\section{More Analysis on Class-wise Performance}\label{app:confusion}
\begin{figure*}[!t]
    \centering
    \begin{minipage}[b]{0.24\linewidth}
        \centering
        \includegraphics[width=\linewidth]{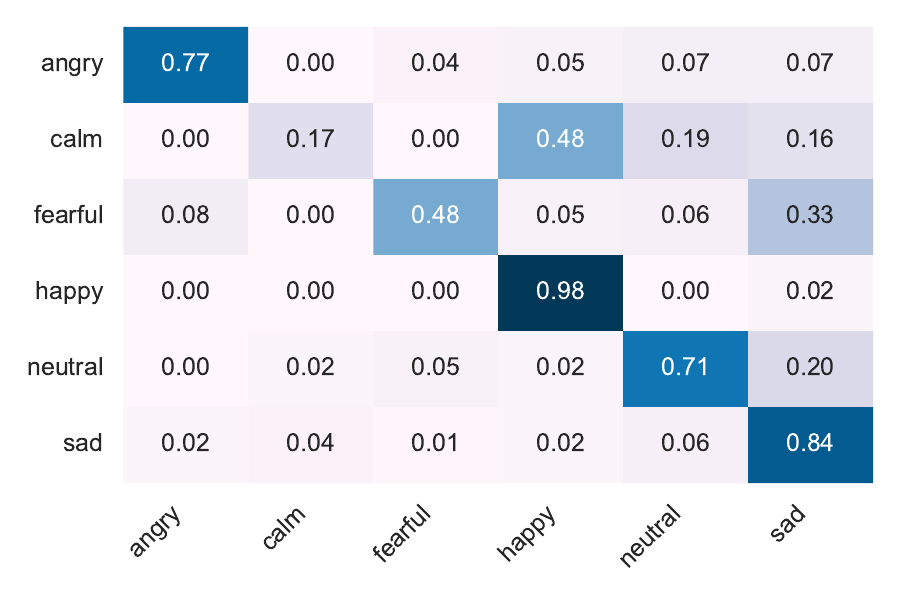}
        \subfloat{\small (a) SOER}
    \end{minipage} \hfill
    \begin{minipage}[b]{0.24\linewidth}
        \centering
        \includegraphics[width=\linewidth]{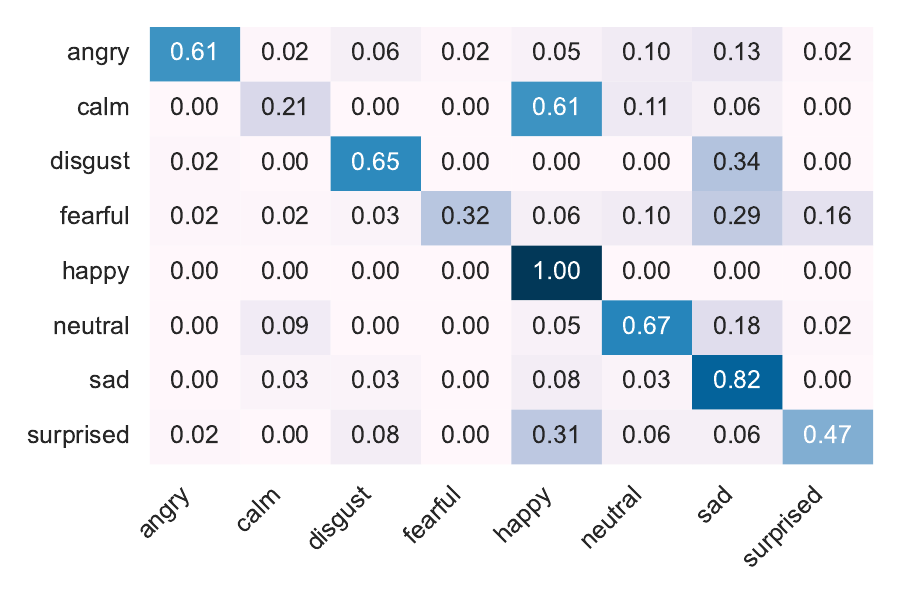}
        \subfloat{\small (b) SPER}
    \end{minipage} \hfill
    \begin{minipage}[b]{0.16\linewidth}
        \centering
        \includegraphics[width=\linewidth]{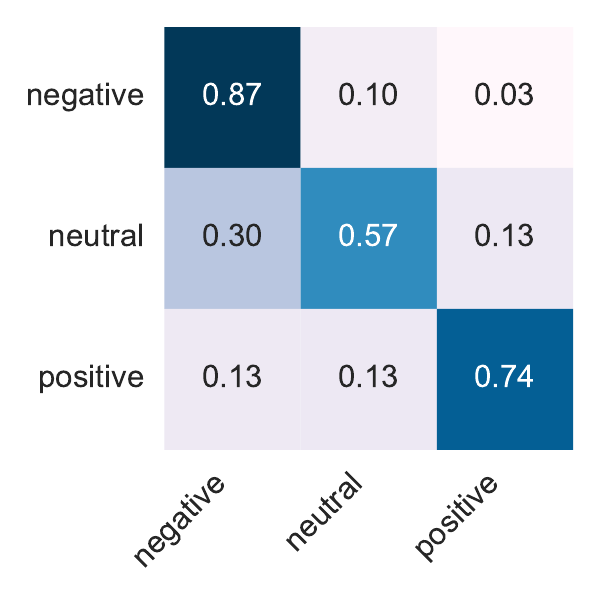}
        \subfloat{\small (d) OSA}
    \end{minipage} \hfill
    \begin{minipage}[b]{0.16\linewidth}
        \centering
        \includegraphics[width=\linewidth]{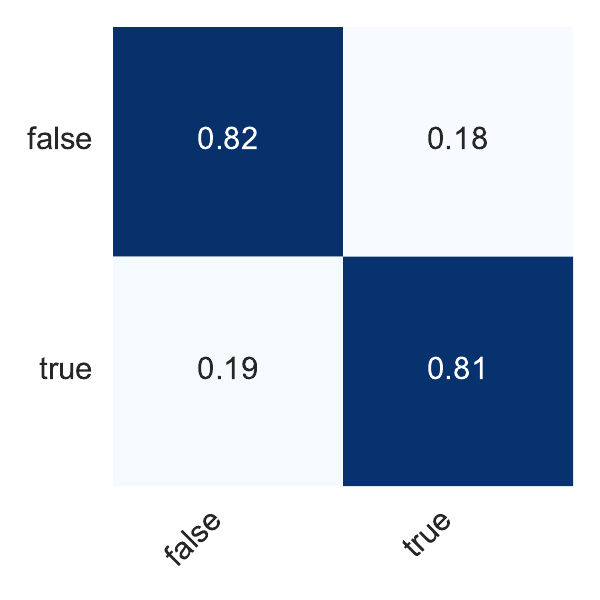}
        \subfloat{\small (l) HU}
    \end{minipage} \\
    
    \begin{minipage}[b]{0.24\linewidth}
        \centering
        \includegraphics[width=\linewidth]{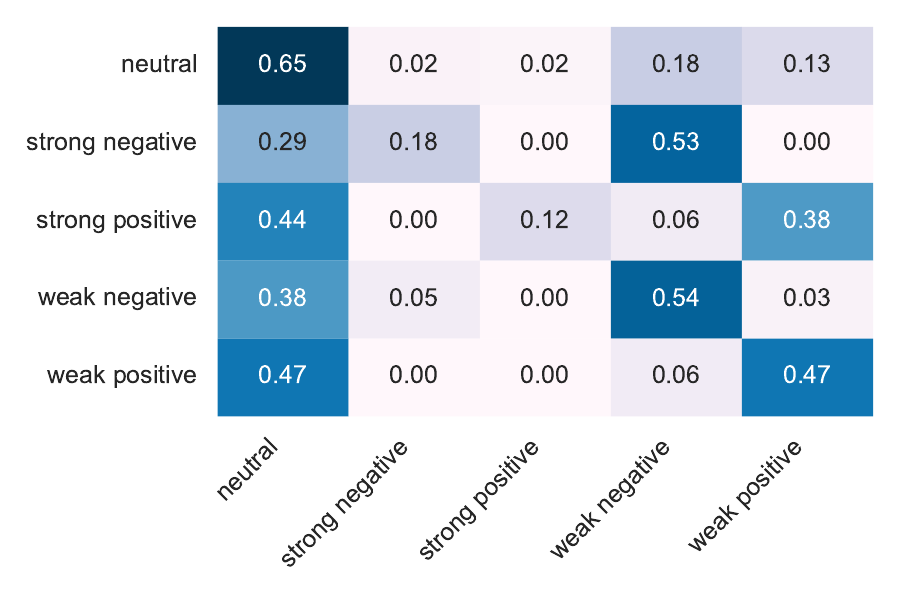}
        \subfloat{\small (c) SEA}
    \end{minipage} \hfill
    \begin{minipage}[b]{0.24\linewidth}
        \centering
        \includegraphics[width=\linewidth]{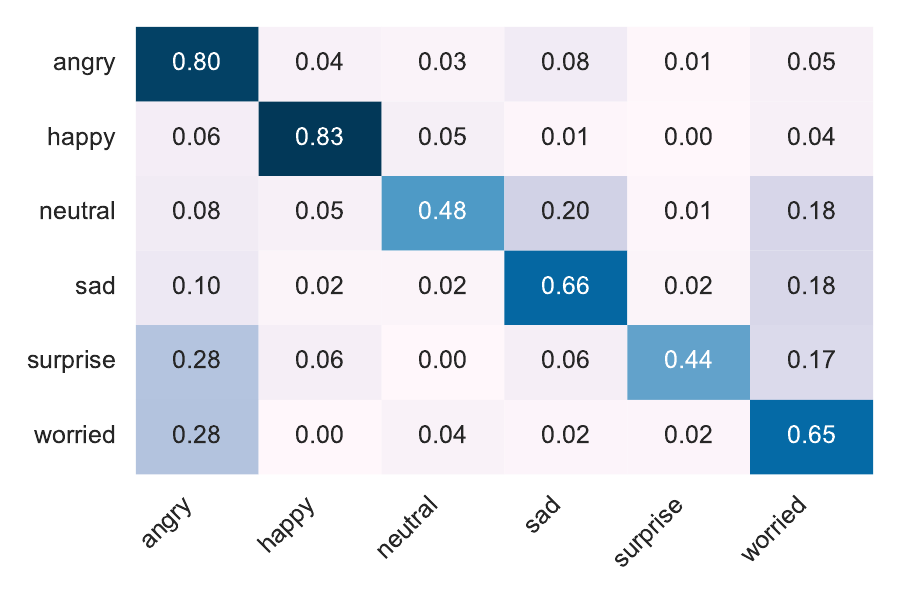}
        \subfloat{\small (f) FGDEA}
    \end{minipage} \hfill
    \begin{minipage}[b]{0.16\linewidth}
        \centering
        \includegraphics[width=\linewidth]{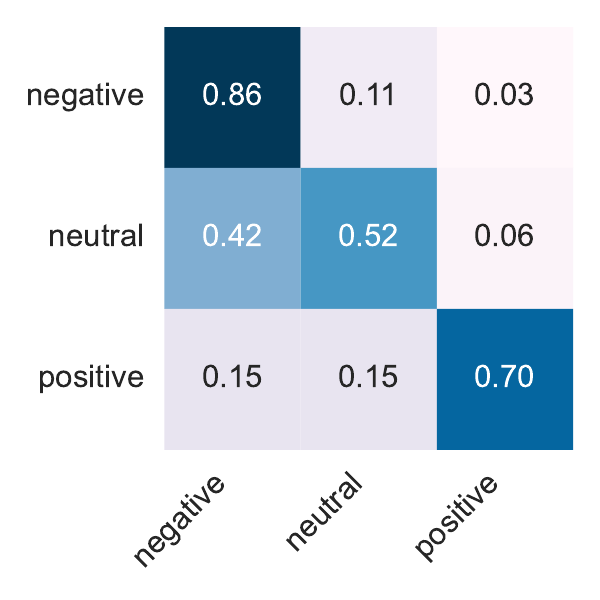}
        \subfloat{\small (g) FCDEA}
    \end{minipage} \hfill
    \begin{minipage}[b]{0.16\linewidth}
        \centering
        \includegraphics[width=\linewidth]{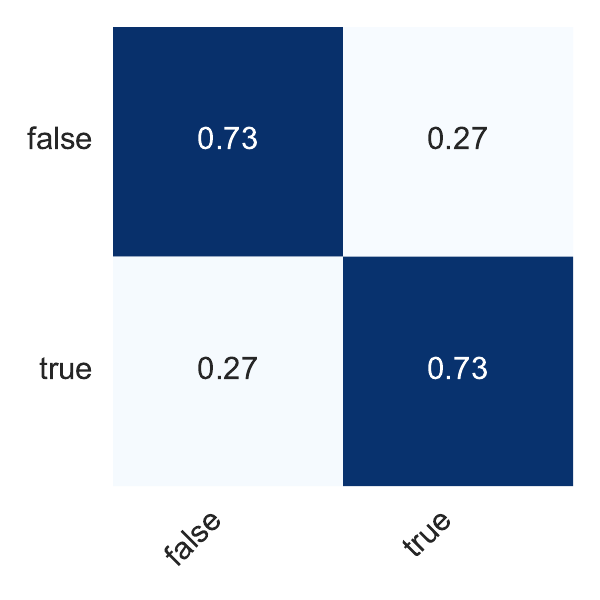}
        \subfloat{\small (m) SD}
    \end{minipage} \\
    
    \begin{minipage}[b]{0.16\linewidth}
        \centering
        \includegraphics[width=\linewidth]{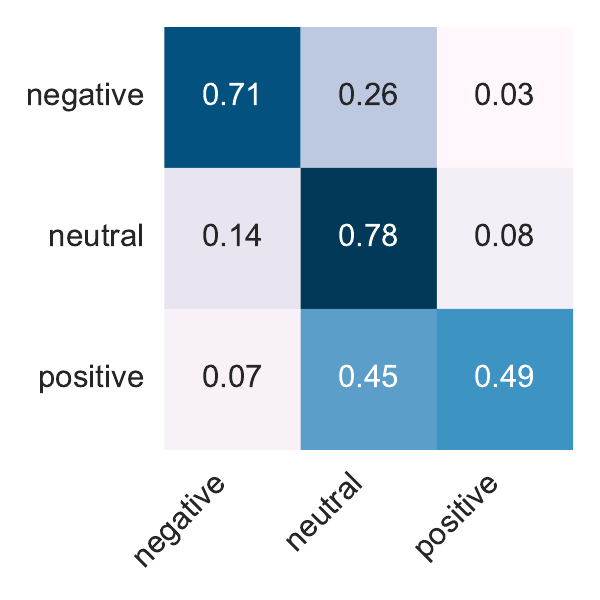}
        \subfloat{\small (e) EIA}
    \end{minipage} \hfill
    \begin{minipage}[b]{0.20\linewidth}
        \centering
        \includegraphics[width=\linewidth]{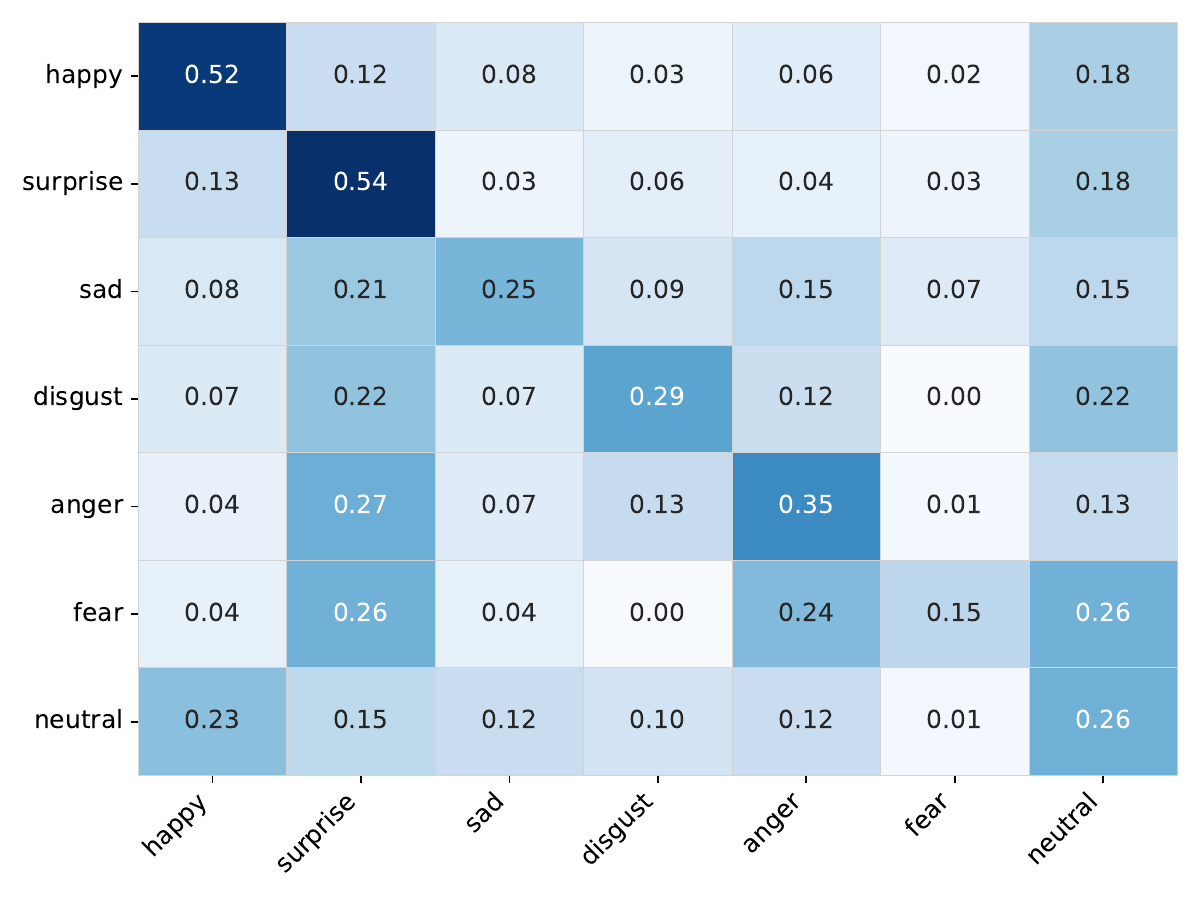}
        \subfloat{\small (i) CEIA (emo.)}
    \end{minipage} \hfill
    \begin{minipage}[b]{0.20\linewidth}
        \centering
        \includegraphics[width=\linewidth]{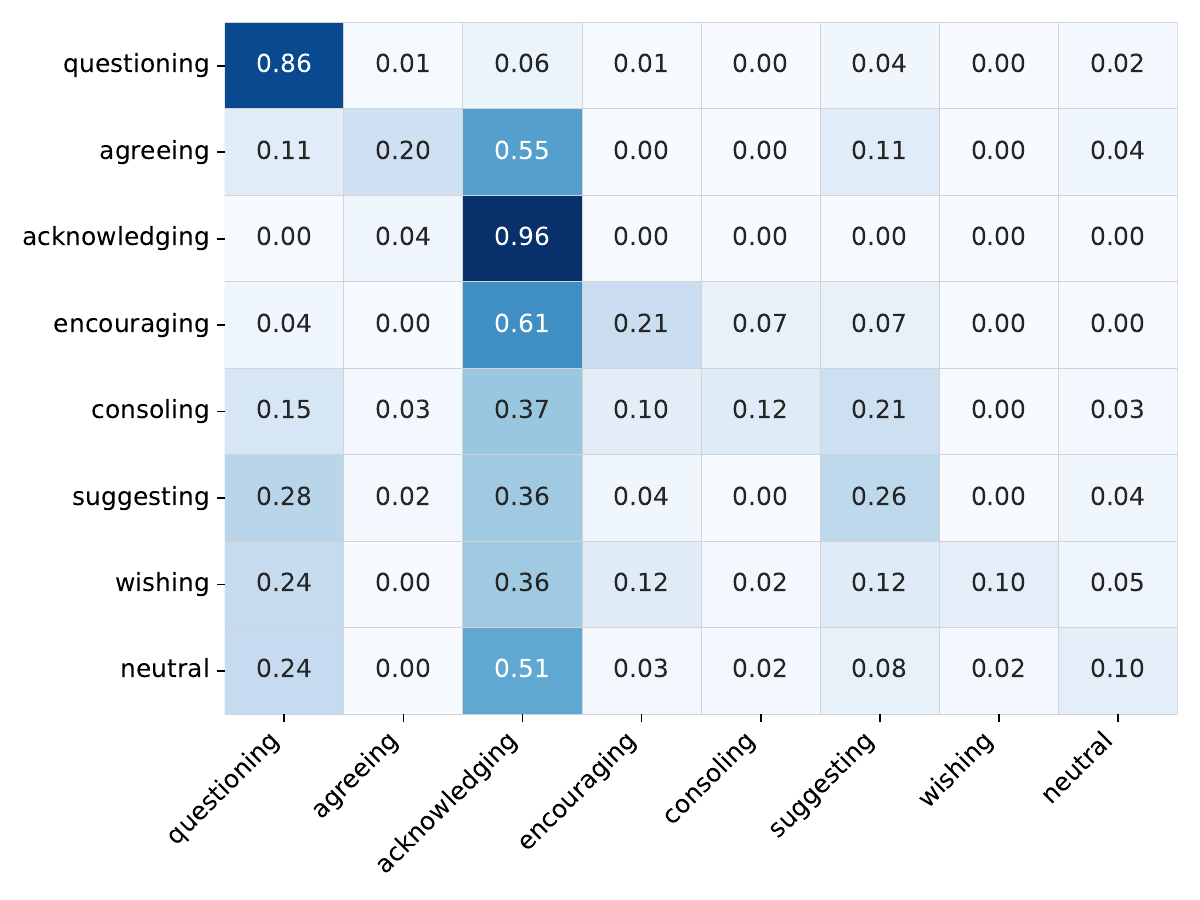}
        \subfloat{\small (j) CEIA (int.)}
    \end{minipage} \hfill
    \begin{minipage}[b]{0.24\linewidth}
        \centering
        \includegraphics[width=\linewidth]{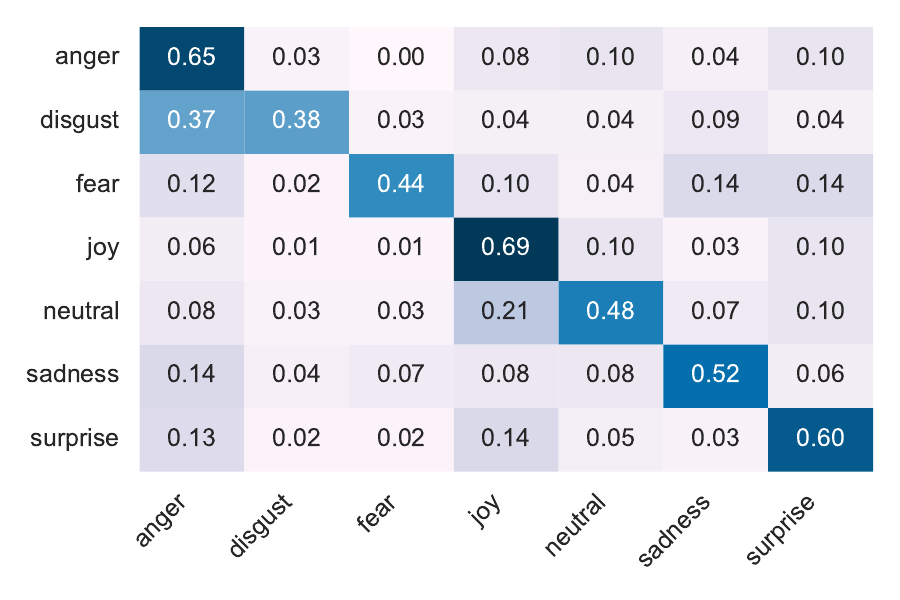}
        \subfloat{\small (k) MPDER}
    \end{minipage} \hfill
    \begin{minipage}[b]{0.16\linewidth}
        \centering
        \includegraphics[width=\linewidth]{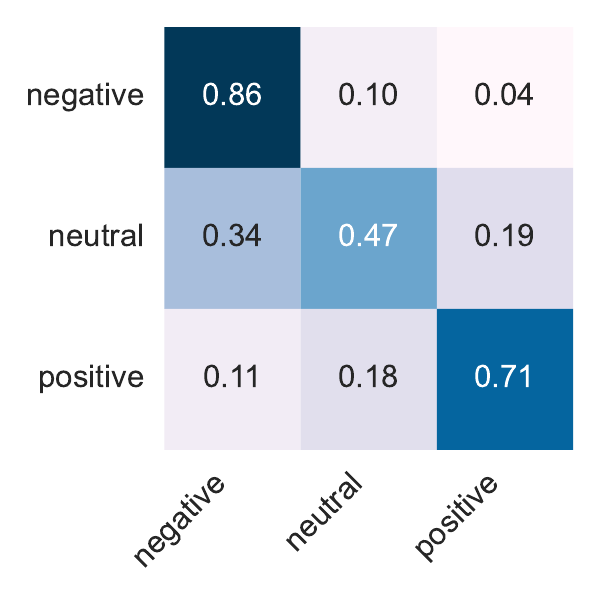}
        \subfloat{\small (h) PEA}
    \end{minipage}
    % \vspace{-2mm}
    \caption{\small Confusion matrices for GPT-5.2 on each evaluation scenario of $\ours$.}
    \label{fig:confusion-GPT5.2}
    \vspace{-2mm}
\end{figure*}

\begin{figure*}[!t]
    \centering

    % ===== Row 1 =====
    \begin{minipage}[b]{0.24\linewidth}
        \centering
        \includegraphics[width=\linewidth]{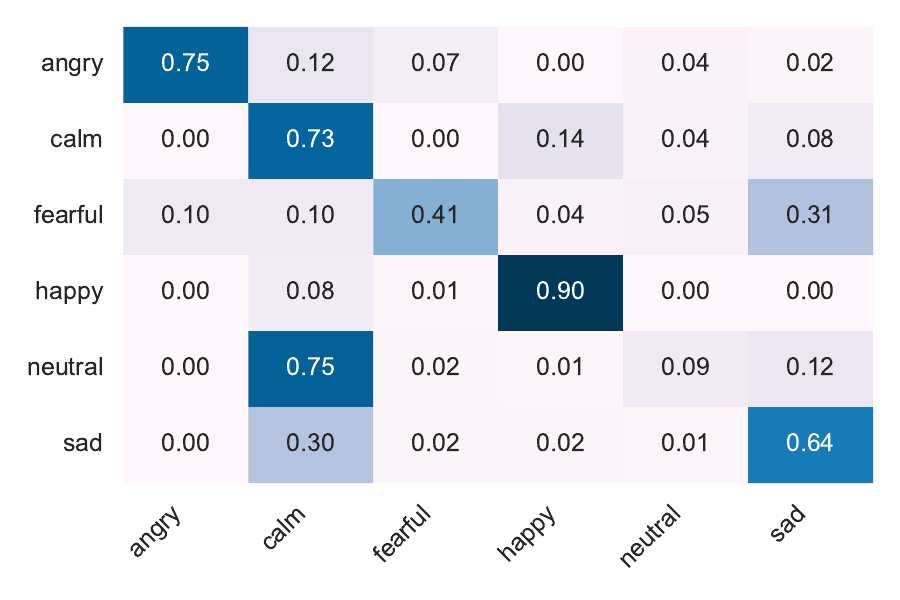}
        \subfloat{\small (a) SOER}
    \end{minipage} \hfill
    \begin{minipage}[b]{0.24\linewidth}
        \centering
        \includegraphics[width=\linewidth]{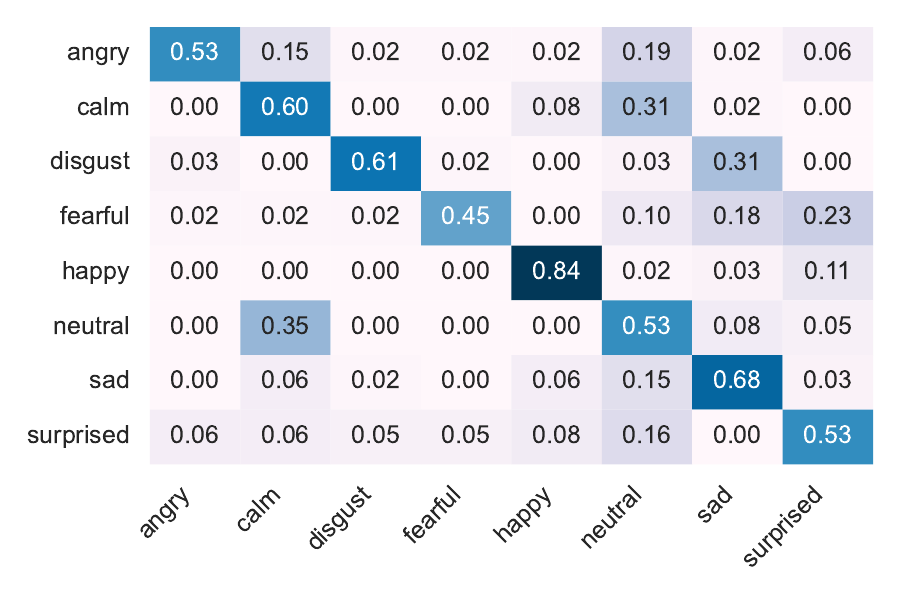}
        \subfloat{\small (b) SPER}
    \end{minipage} \hfill
    \begin{minipage}[b]{0.16\linewidth}
        \centering
        \includegraphics[width=\linewidth]{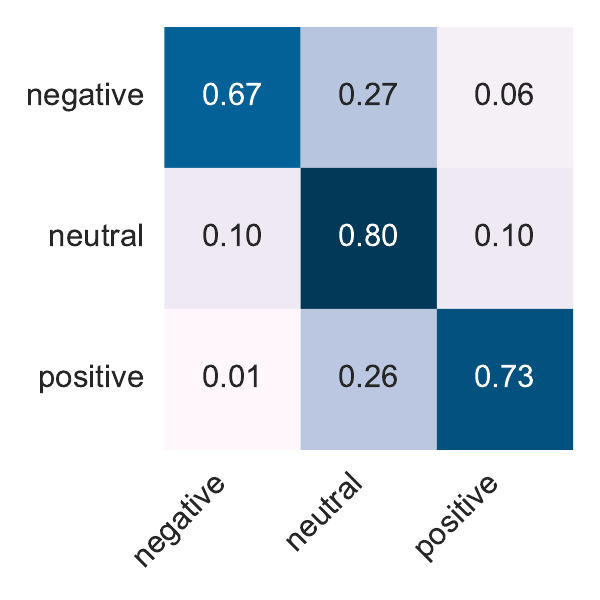}
        \subfloat{\small (d) OSA}
    \end{minipage} \hfill
    \begin{minipage}[b]{0.16\linewidth}
        \centering
        \includegraphics[width=\linewidth]{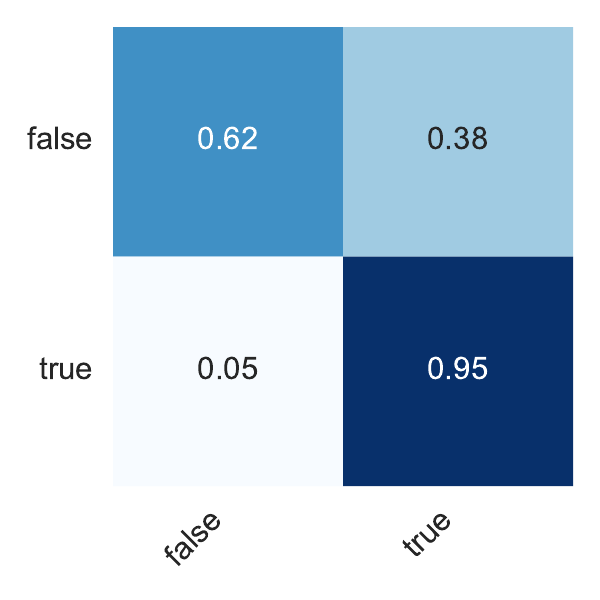}
        \subfloat{\small (l) HU}
    \end{minipage} \\

    % ===== Row 2 =====
    \begin{minipage}[b]{0.24\linewidth}
        \centering
        \includegraphics[width=\linewidth]{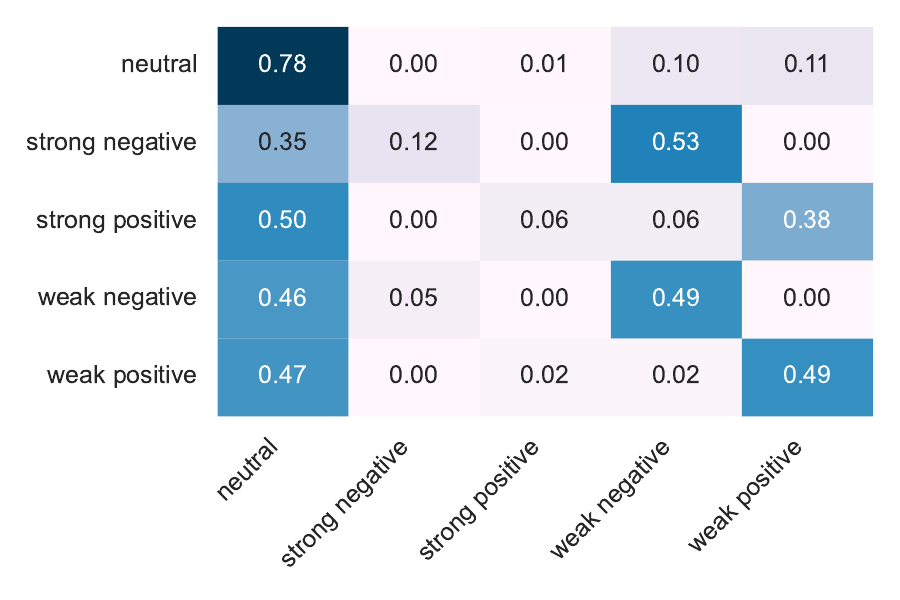}
        \subfloat{\small (c) SCEA}
    \end{minipage} \hfill
    \begin{minipage}[b]{0.24\linewidth}
        \centering
        \includegraphics[width=\linewidth]{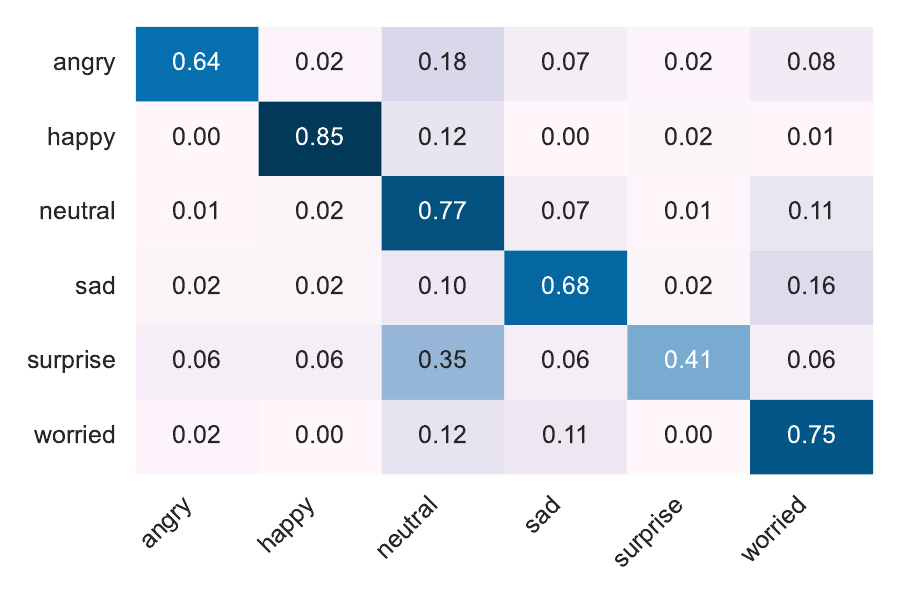}
        \subfloat{\small (f) FGDEA}
    \end{minipage} \hfill
    \begin{minipage}[b]{0.16\linewidth}
        \centering
        \includegraphics[width=\linewidth]{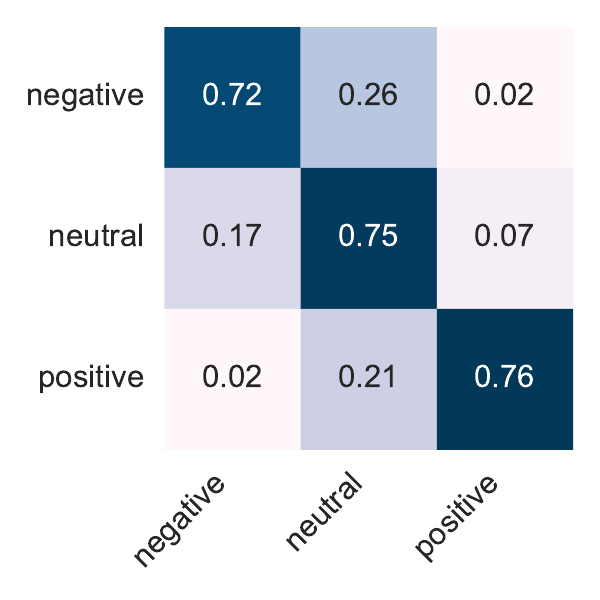}
        \subfloat{\small (g) FCDEA}
    \end{minipage} \hfill
    \begin{minipage}[b]{0.16\linewidth}
        \centering
        \includegraphics[width=\linewidth]{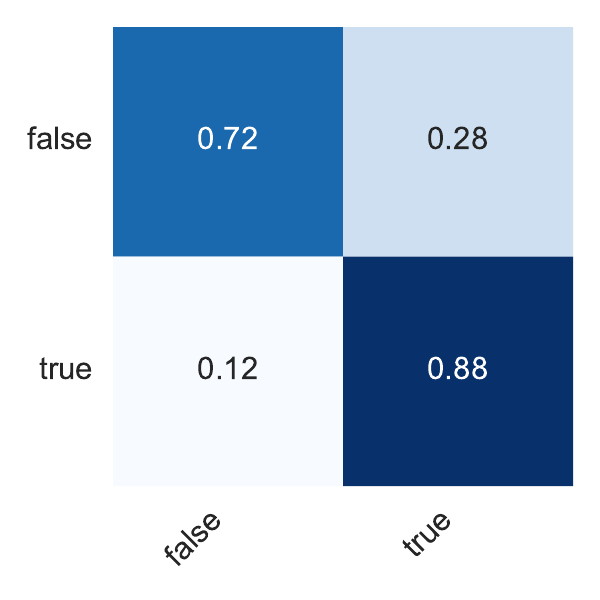}
        \subfloat{\small (m) SD}
    \end{minipage} \\

    % ===== Row 3 =====
    \begin{minipage}[b]{0.16\linewidth}
        \centering
        \includegraphics[width=\linewidth]{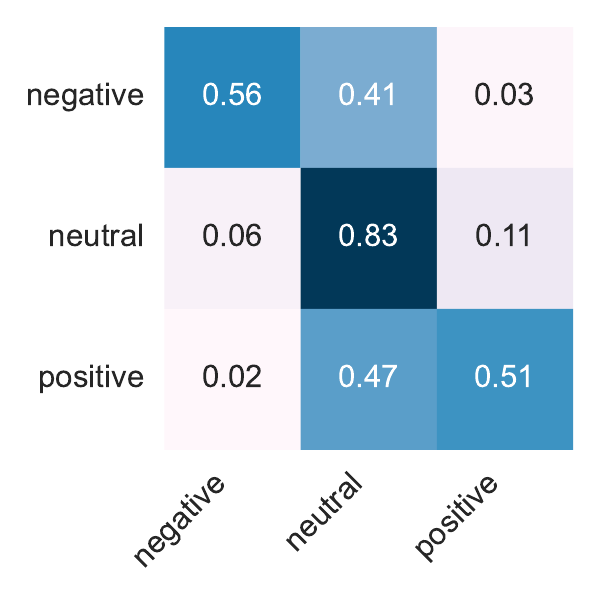}
        \subfloat{\small (e) EIA}
    \end{minipage} \hfill
    \begin{minipage}[b]{0.20\linewidth}
        \centering
        \includegraphics[width=\linewidth]{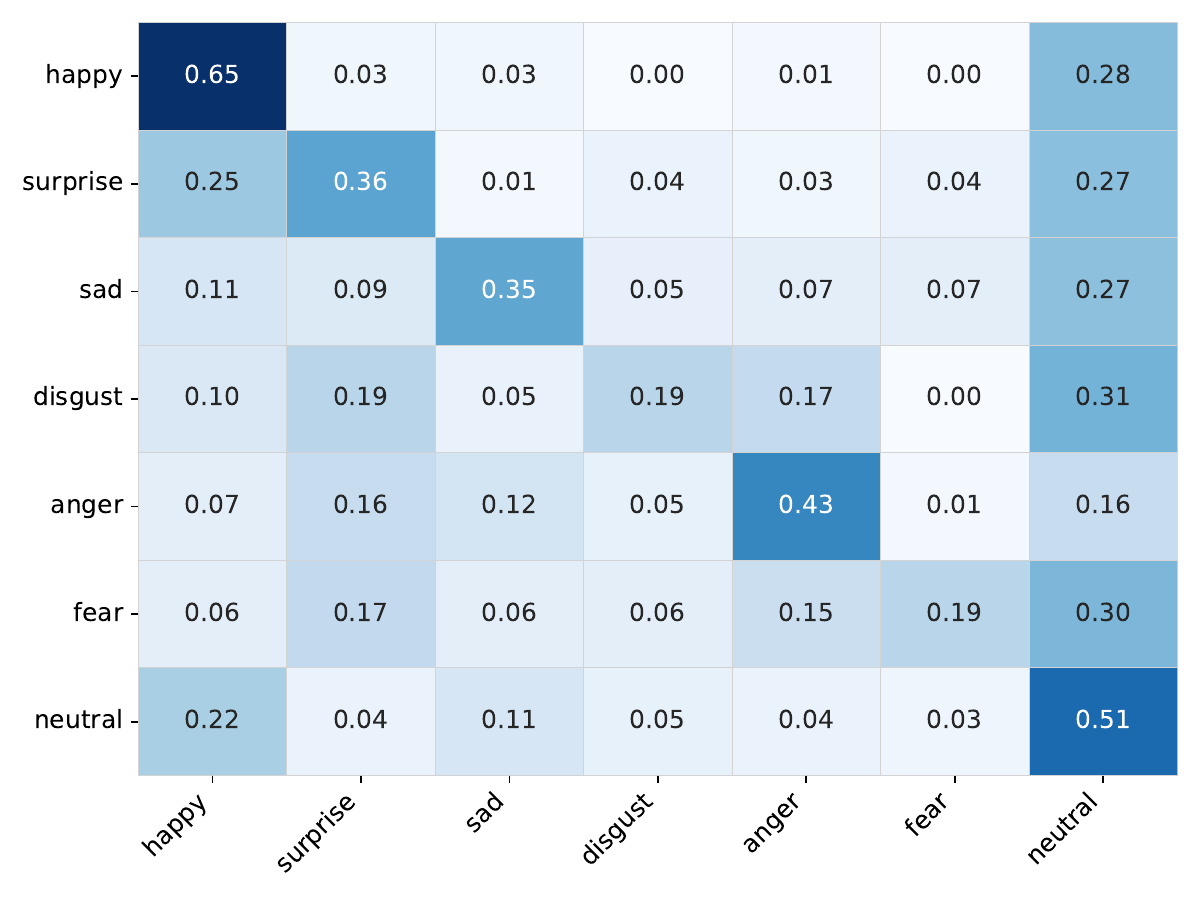}
        \subfloat{\small (i) CEIA (emo.)}
    \end{minipage} \hfill
    \begin{minipage}[b]{0.20\linewidth}
        \centering
        \includegraphics[width=\linewidth]{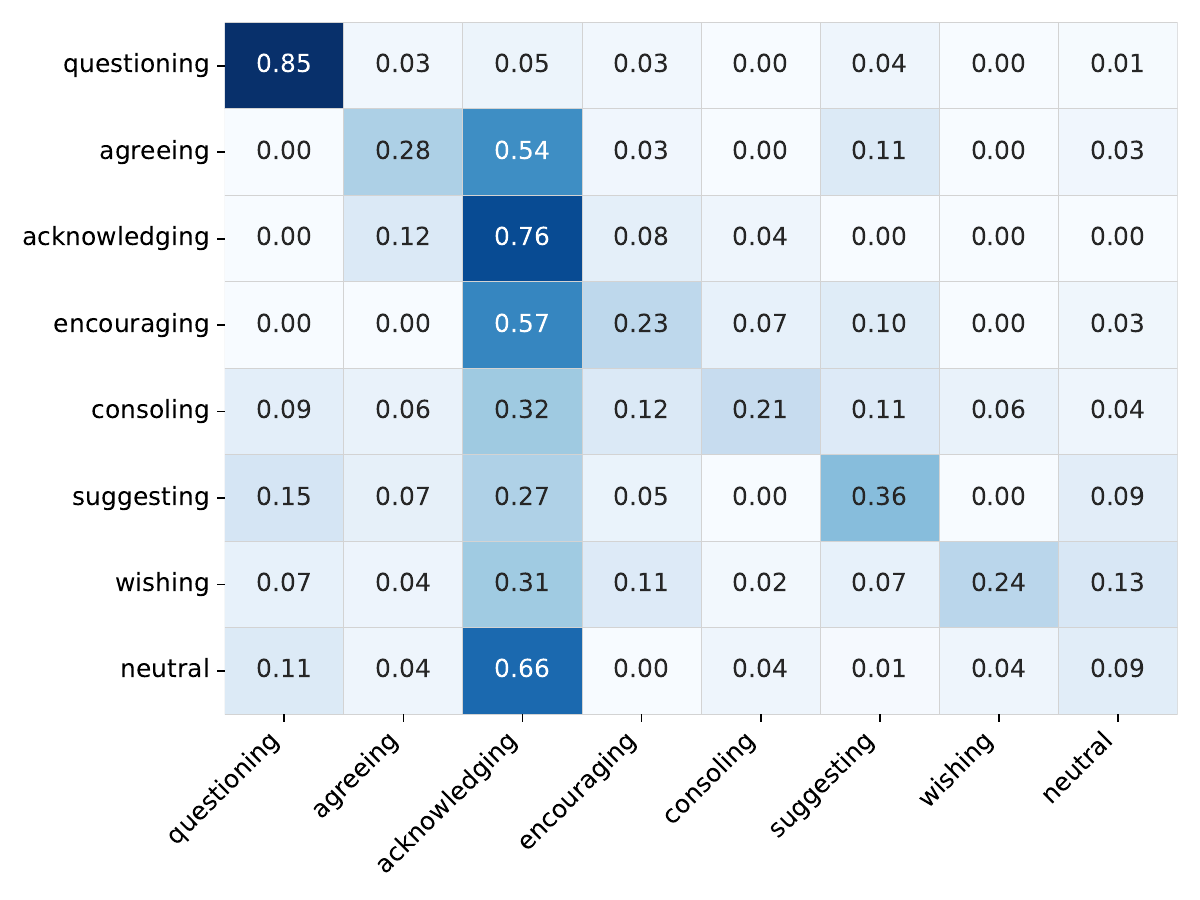}
        \subfloat{\small (j) CEIA (int.)}
    \end{minipage} \hfill
    \begin{minipage}[b]{0.24\linewidth}
        \centering
        \includegraphics[width=\linewidth]{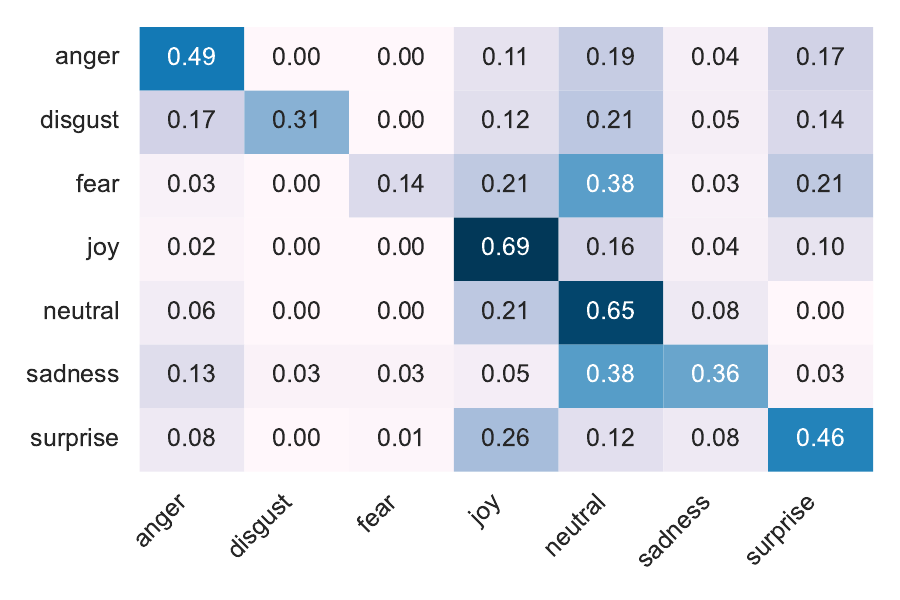}
        \subfloat{\small (k) MPDER}
    \end{minipage} \hfill
    \begin{minipage}[b]{0.16\linewidth}
        \centering
        \includegraphics[width=\linewidth]{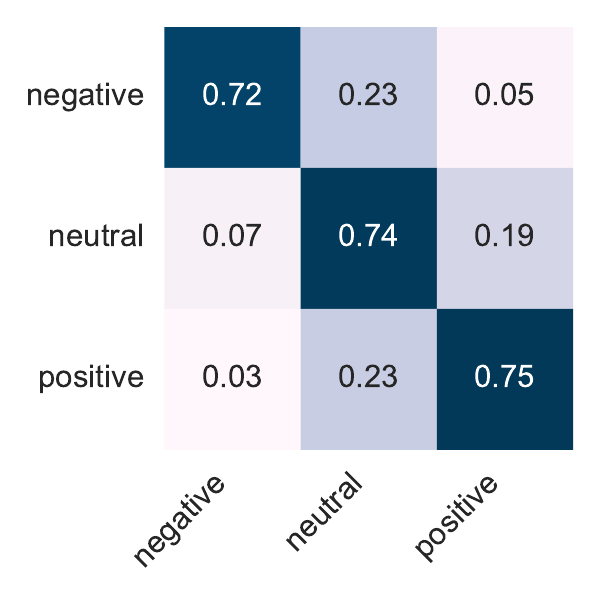}
        \subfloat{\small (h) PEA}
    \end{minipage}

    \caption{\small Confusion matrices for Gemini-3.0-Flash on each evaluation scenario of $\ours$.}
    \label{fig:confusion-Gemini-3.0-Flash}
\end{figure*}

\begin{figure*}[!t]
    \centering

    % ===== Row 1 =====
    \begin{minipage}[b]{0.24\linewidth}
        \centering
        \includegraphics[width=\linewidth]{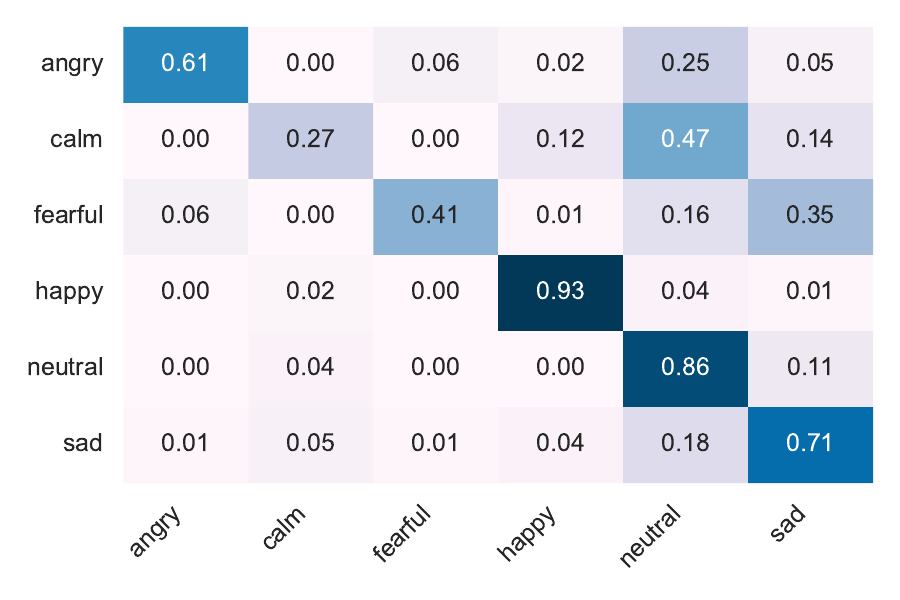}
        \subfloat{\small (a) SOER}
    \end{minipage} \hfill
    \begin{minipage}[b]{0.24\linewidth}
        \centering
        \includegraphics[width=\linewidth]{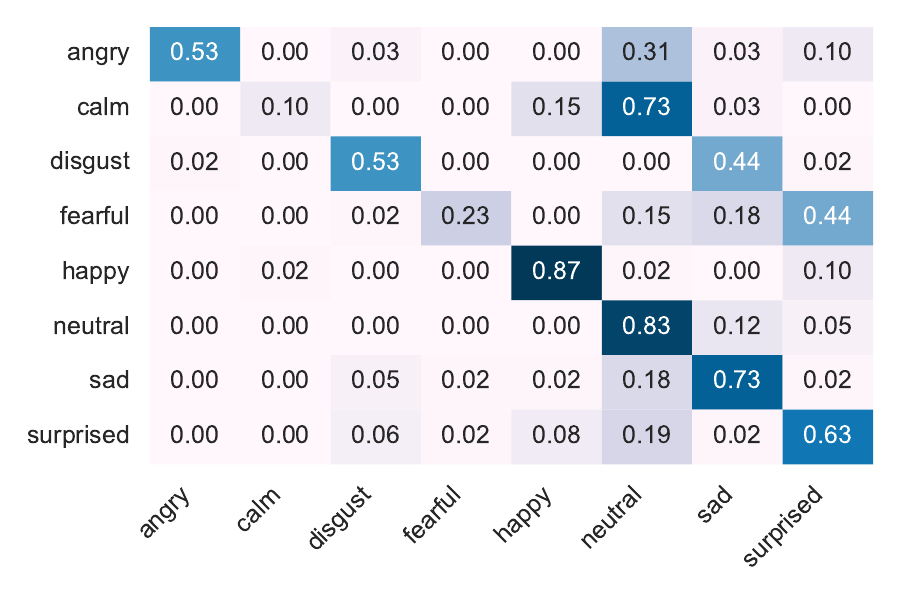}
        \subfloat{\small (b) SPER}
    \end{minipage} \hfill
    \begin{minipage}[b]{0.16\linewidth}
        \centering
        \includegraphics[width=\linewidth]{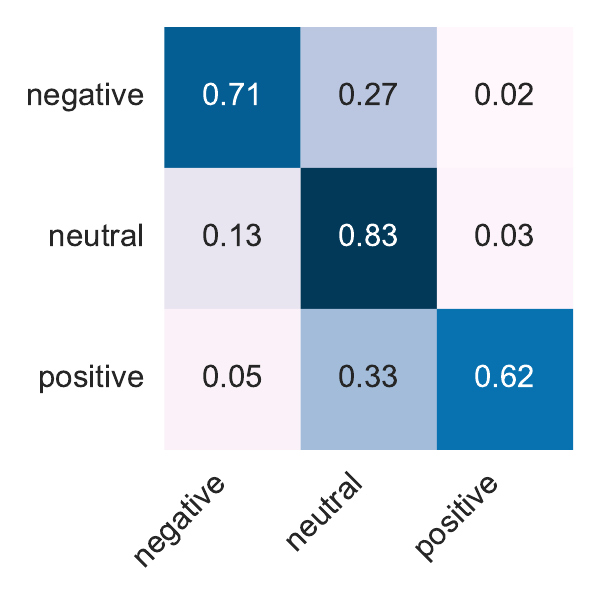}
        \subfloat{\small (d) OSA}
    \end{minipage} \hfill
    \begin{minipage}[b]{0.16\linewidth}
        \centering
        \includegraphics[width=\linewidth]{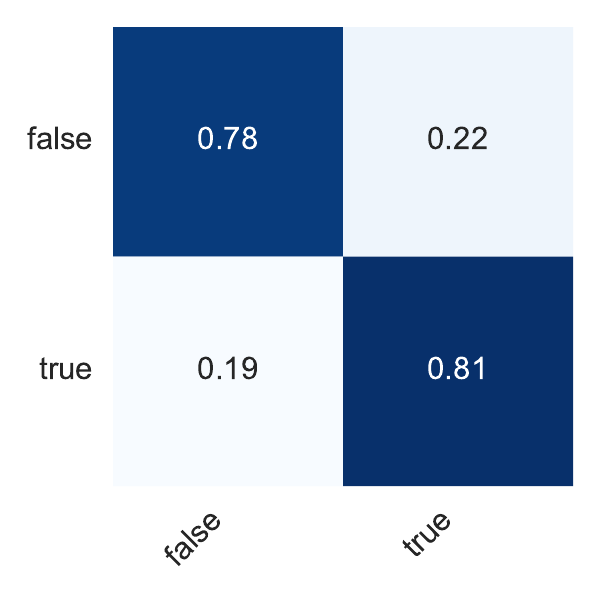}
        \subfloat{\small (l) HU}
    \end{minipage} \\

    % ===== Row 2 =====
    \begin{minipage}[b]{0.24\linewidth}
        \centering
        \includegraphics[width=\linewidth]{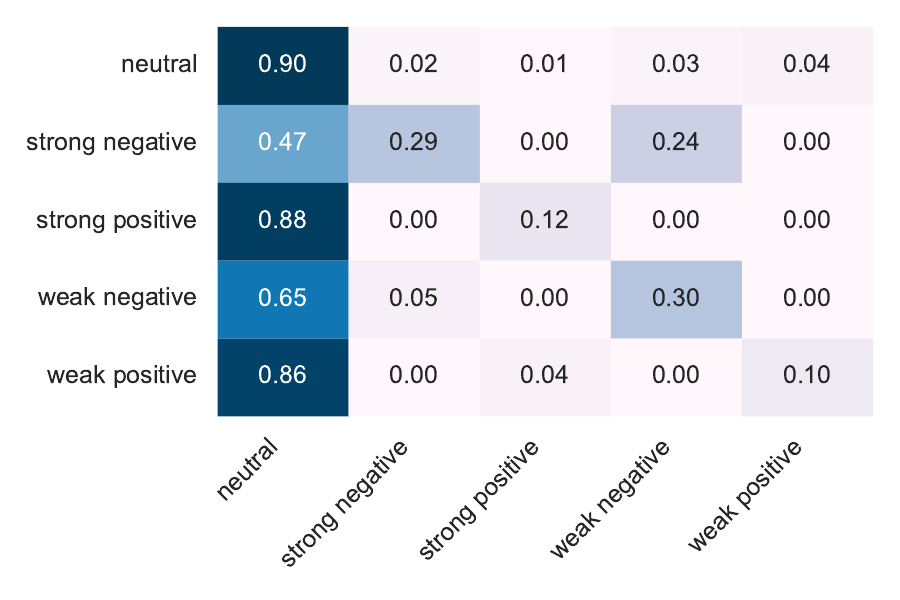}
        \subfloat{\small (c) SCEA}
    \end{minipage} \hfill
    \begin{minipage}[b]{0.24\linewidth}
        \centering
        \includegraphics[width=\linewidth]{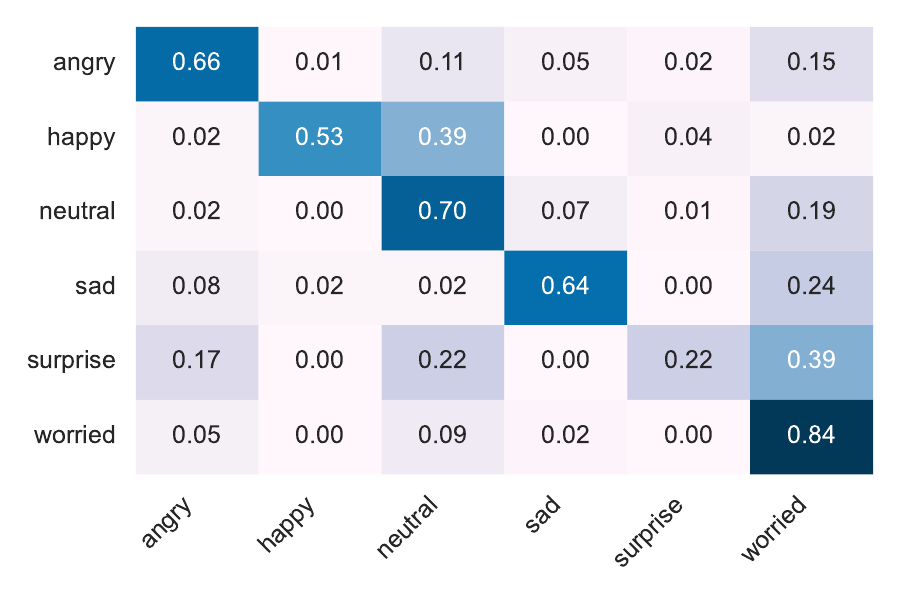}
        \subfloat{\small (f) FGDEA}
    \end{minipage} \hfill
    \begin{minipage}[b]{0.16\linewidth}
        \centering
        \includegraphics[width=\linewidth]{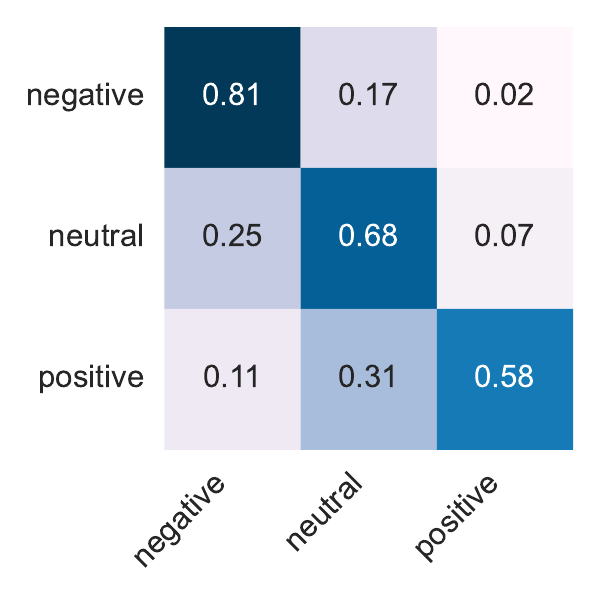}
        \subfloat{\small (g) FCDEA}
    \end{minipage} \hfill
    \begin{minipage}[b]{0.16\linewidth}
        \centering
        \includegraphics[width=\linewidth]{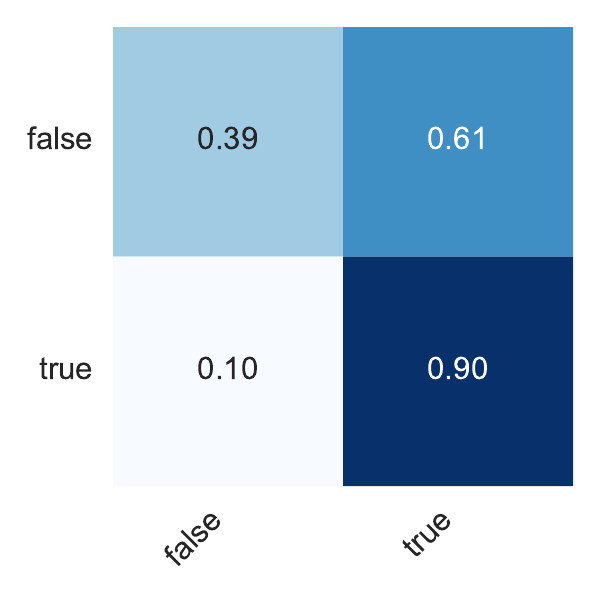}
        \subfloat{\small (m) SD}
    \end{minipage} \\

    % ===== Row 3 =====
    \begin{minipage}[b]{0.16\linewidth}
        \centering
        \includegraphics[width=\linewidth]{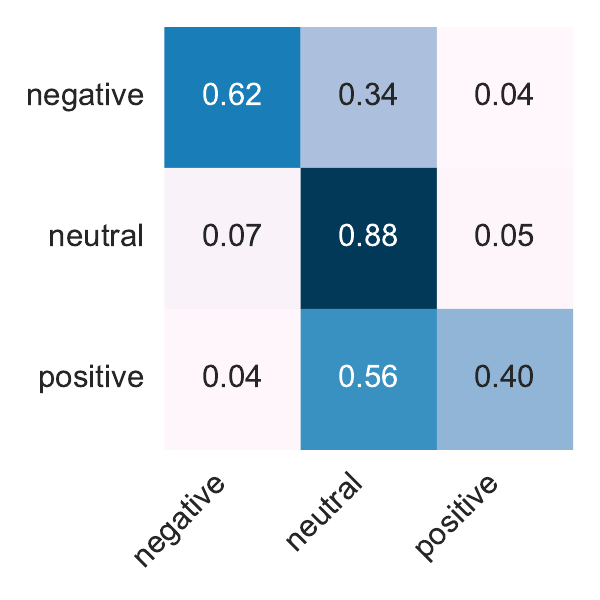}
        \subfloat{\small (e) EIA}
    \end{minipage} \hfill
    \begin{minipage}[b]{0.20\linewidth}
        \centering
        \includegraphics[width=\linewidth]{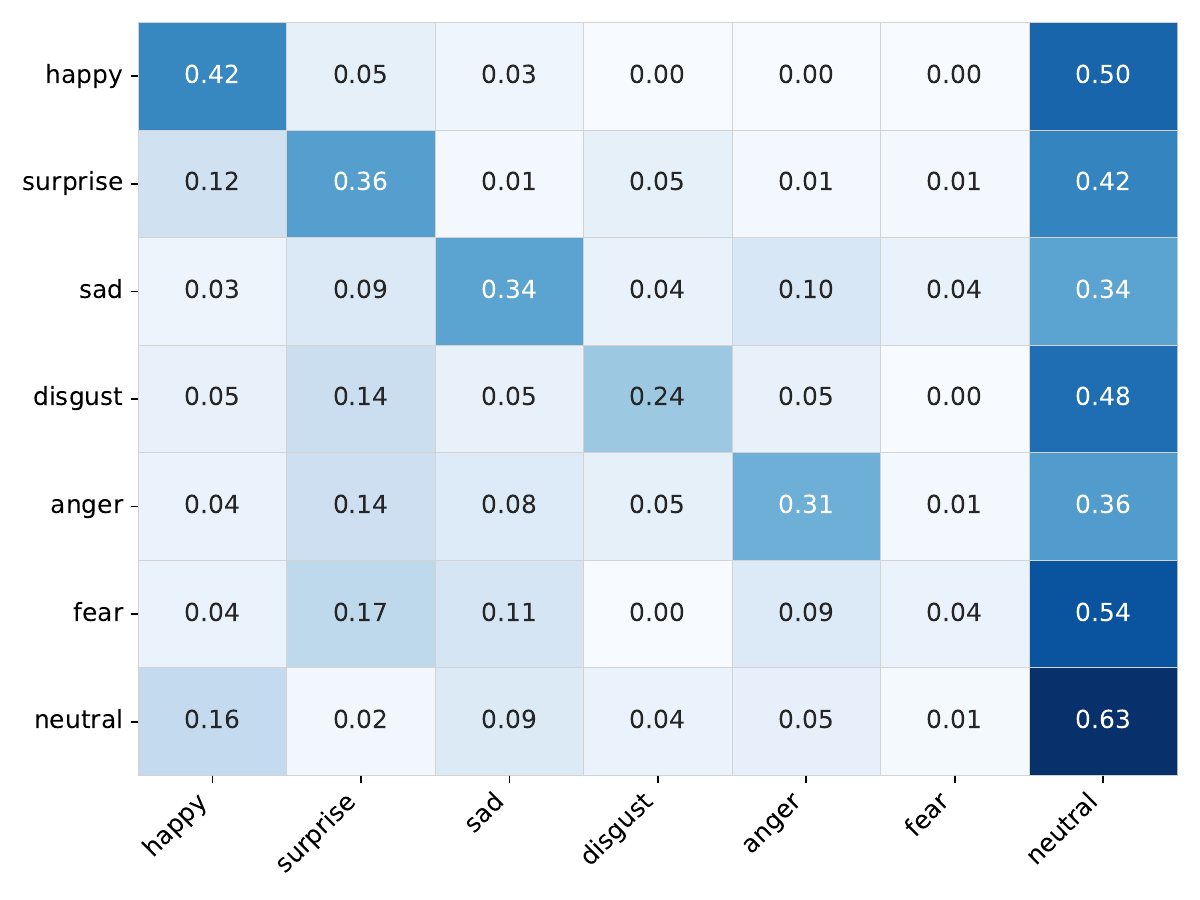}
        \subfloat{\small (i) CEIA (emo.)}
    \end{minipage} \hfill
    \begin{minipage}[b]{0.20\linewidth}
        \centering
        \includegraphics[width=\linewidth]{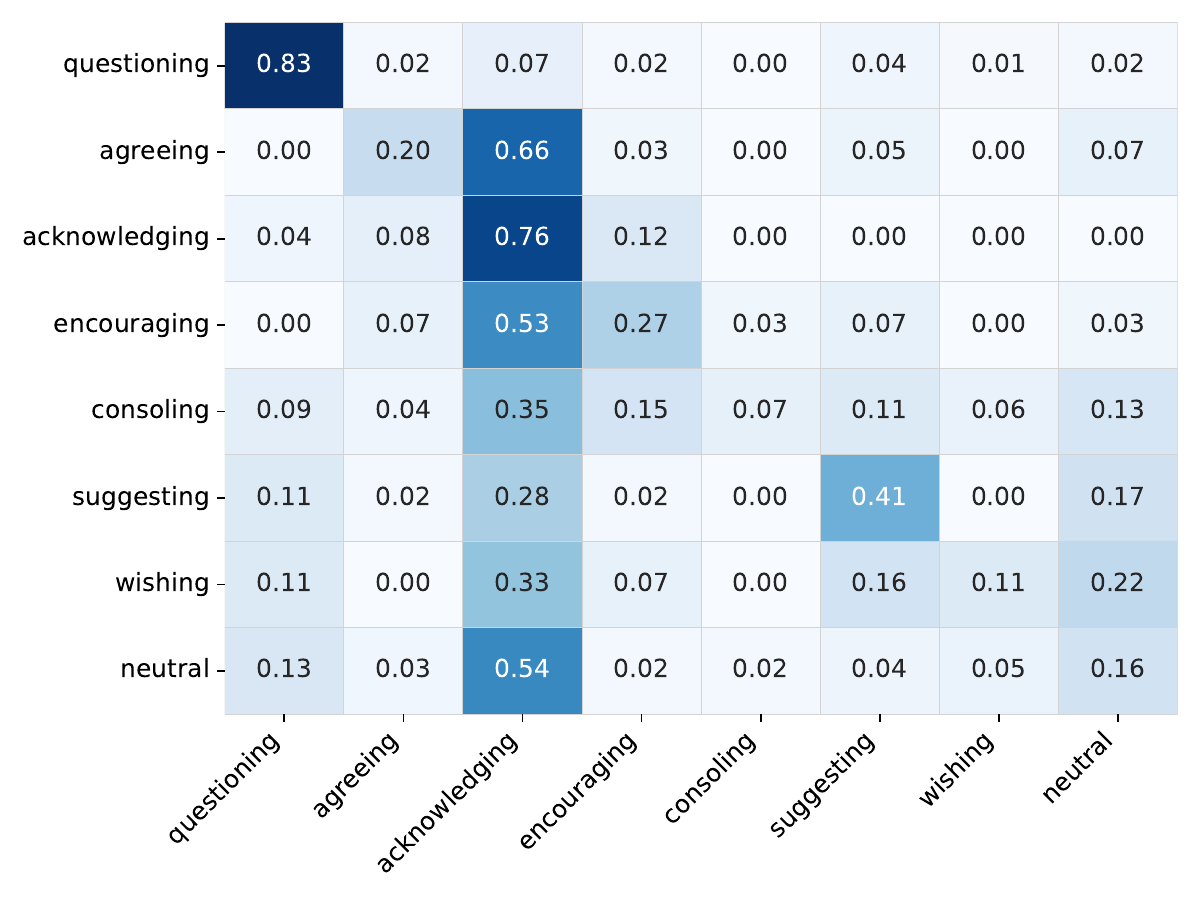}
        \subfloat{\small (j) CEIA (int.)}
    \end{minipage} \hfill
    \begin{minipage}[b]{0.24\linewidth}
        \centering
        \includegraphics[width=\linewidth]{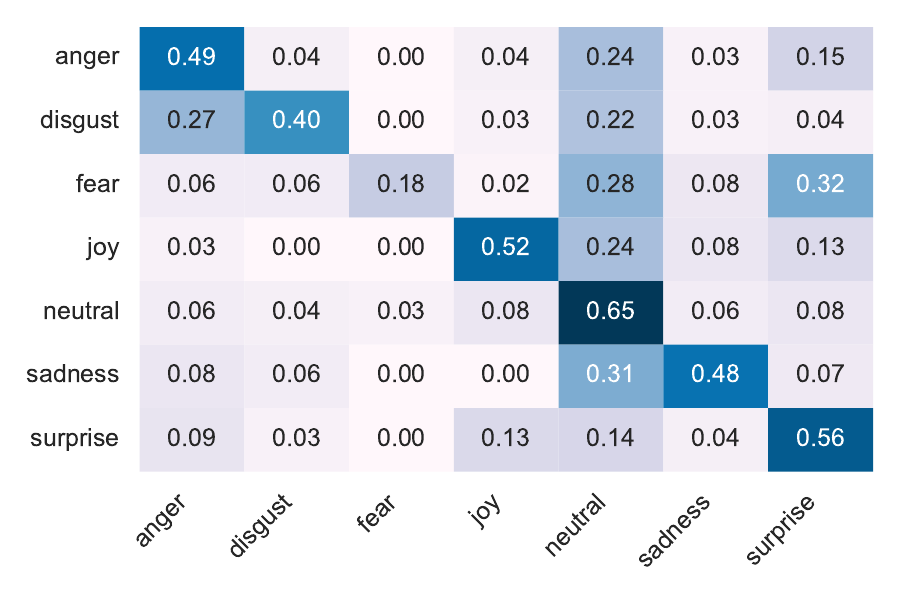}
        \subfloat{\small (k) MPDER}
    \end{minipage} \hfill
    \begin{minipage}[b]{0.16\linewidth}
        \centering
        \includegraphics[width=\linewidth]{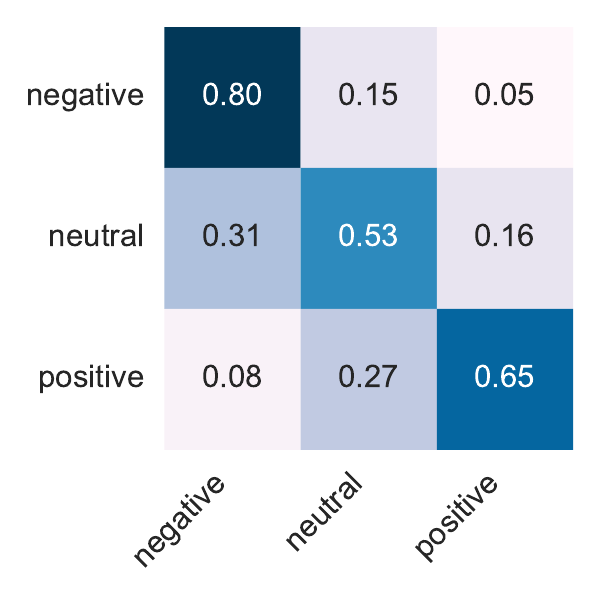}
        \subfloat{\small (h) PEA}
    \end{minipage}

    \caption{\small Confusion matrices for Gemini-2.0-Flash on each evaluation scenario of $\ours$.}
    \label{fig:confusion-Gemini-2.0-Flash}
\end{figure*}

\begin{figure*}[!t]
    \centering
    
    % ===== Row 1 =====
    \begin{minipage}[b]{0.24\linewidth}
        \centering
        \includegraphics[width=\linewidth]{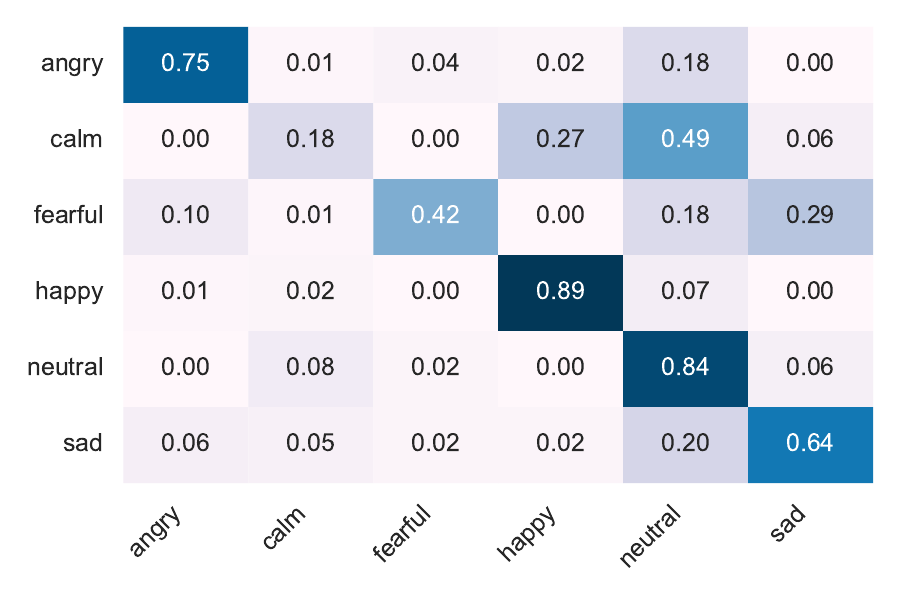}
        \subcaption{\small SOER}
    \end{minipage} \hfill
    \begin{minipage}[b]{0.24\linewidth}
        \centering
        \includegraphics[width=\linewidth]{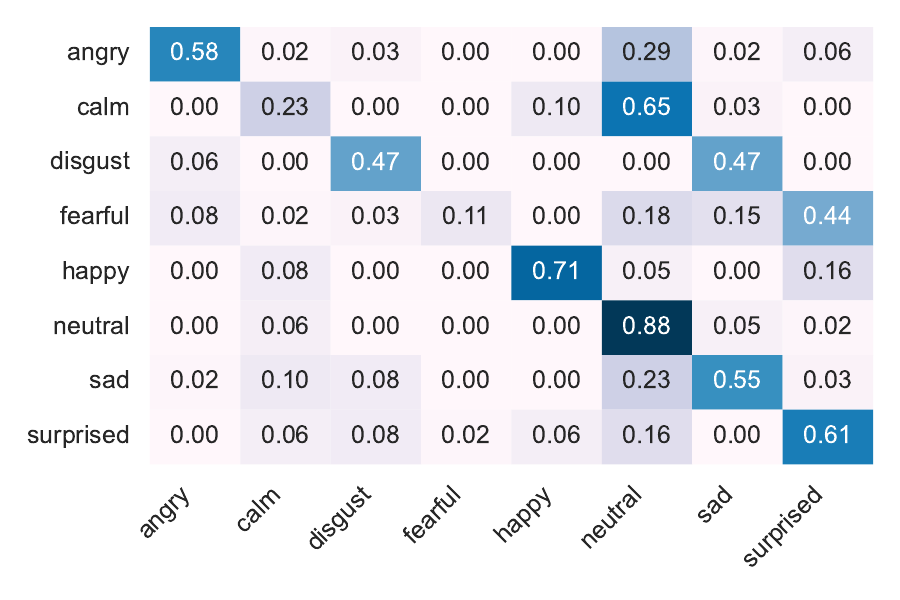}
        \subcaption{\small SPER}
    \end{minipage} \hfill
    \begin{minipage}[b]{0.16\linewidth}
        \centering
        \includegraphics[width=\linewidth]{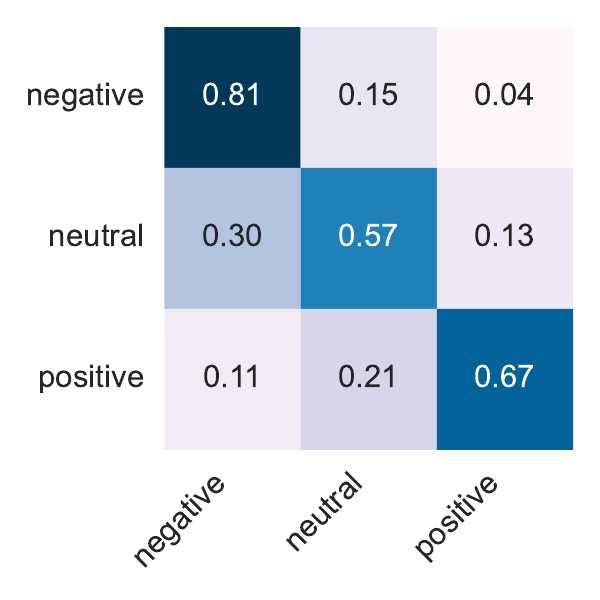}
        \subcaption{\small OSA}
    \end{minipage} \hfill
    \begin{minipage}[b]{0.16\linewidth}
        \centering
        \includegraphics[width=\linewidth]{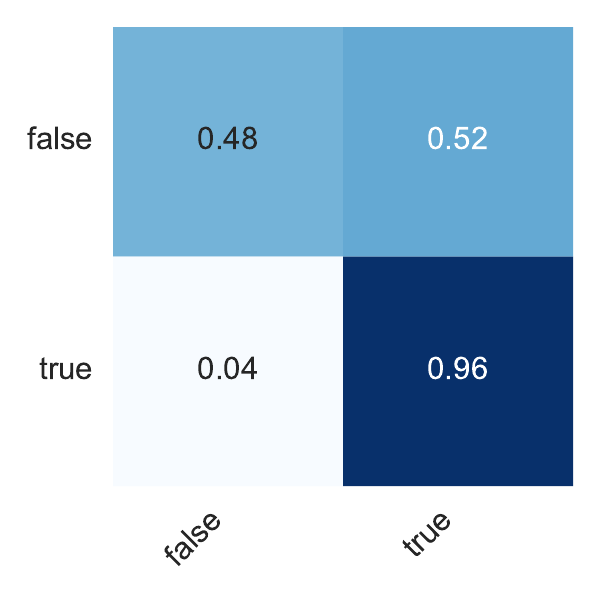}
        \subcaption{\small HU}
    \end{minipage} \\

    % ===== Row 2 =====
    \begin{minipage}[b]{0.24\linewidth}
        \centering
        \includegraphics[width=\linewidth]{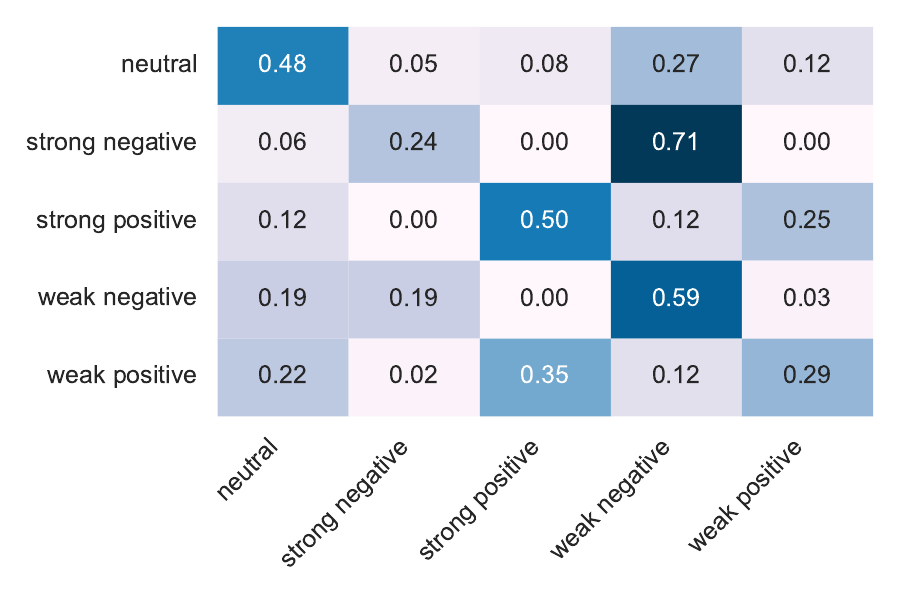}
        \subcaption{\small SCEA}
    \end{minipage} \hfill
    \begin{minipage}[b]{0.24\linewidth}
        \centering
        \includegraphics[width=\linewidth]{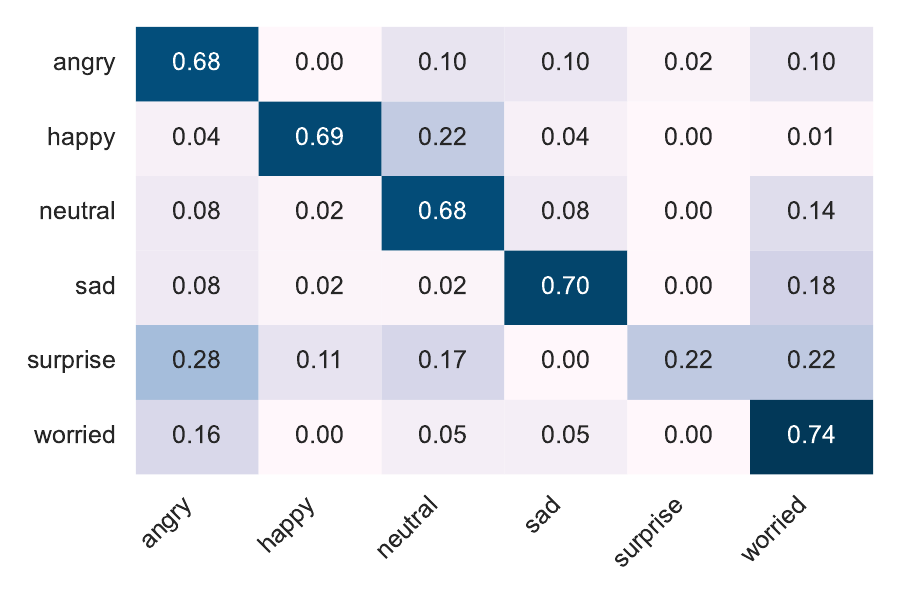}
        \subcaption{\small FGDEA}
    \end{minipage} \hfill
    \begin{minipage}[b]{0.16\linewidth}
        \centering
        \includegraphics[width=\linewidth]{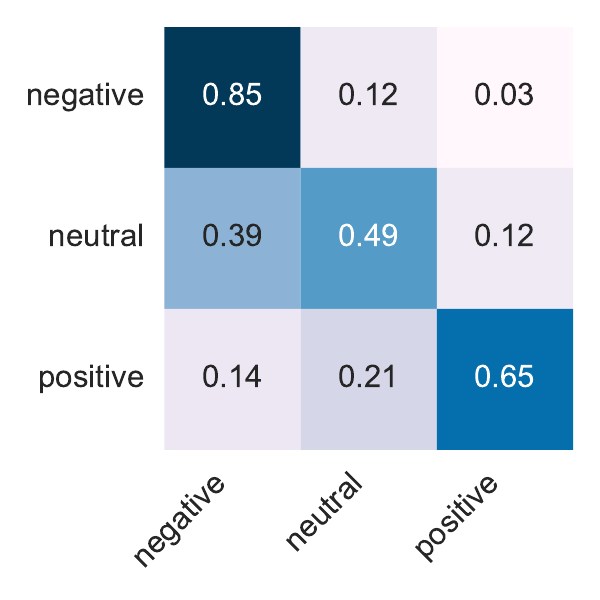}
        \subcaption{\small FCDEA}
    \end{minipage} \hfill
    \begin{minipage}[b]{0.16\linewidth}
        \centering
        \includegraphics[width=\linewidth]{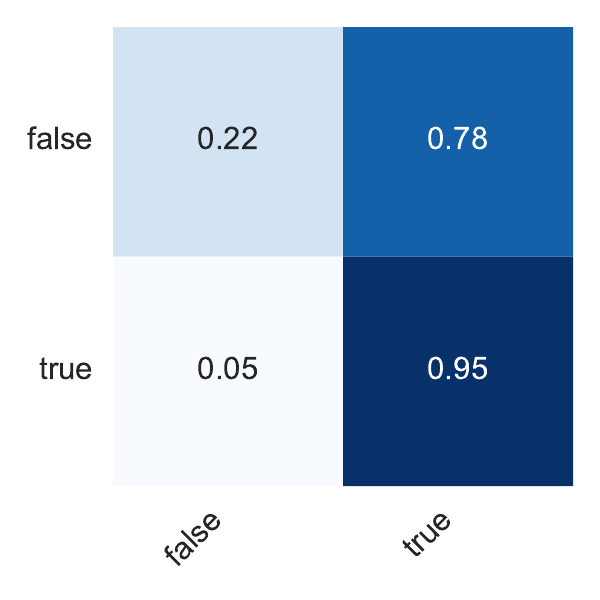}
        \subcaption{\small SD}
    \end{minipage} \\

    % ===== Row 3 =====
    \begin{minipage}[b]{0.16\linewidth}
        \centering
        \includegraphics[width=\linewidth]{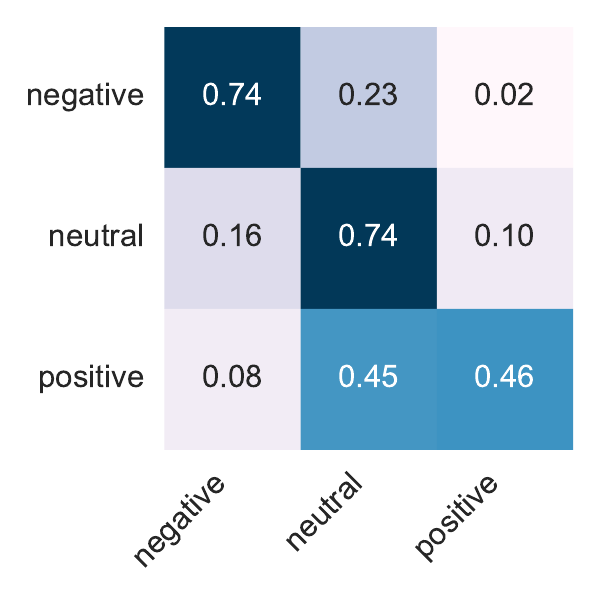}
        \subcaption{\small EIA}
    \end{minipage} \hfill
    \begin{minipage}[b]{0.20\linewidth}
        \centering
        \includegraphics[width=\linewidth]{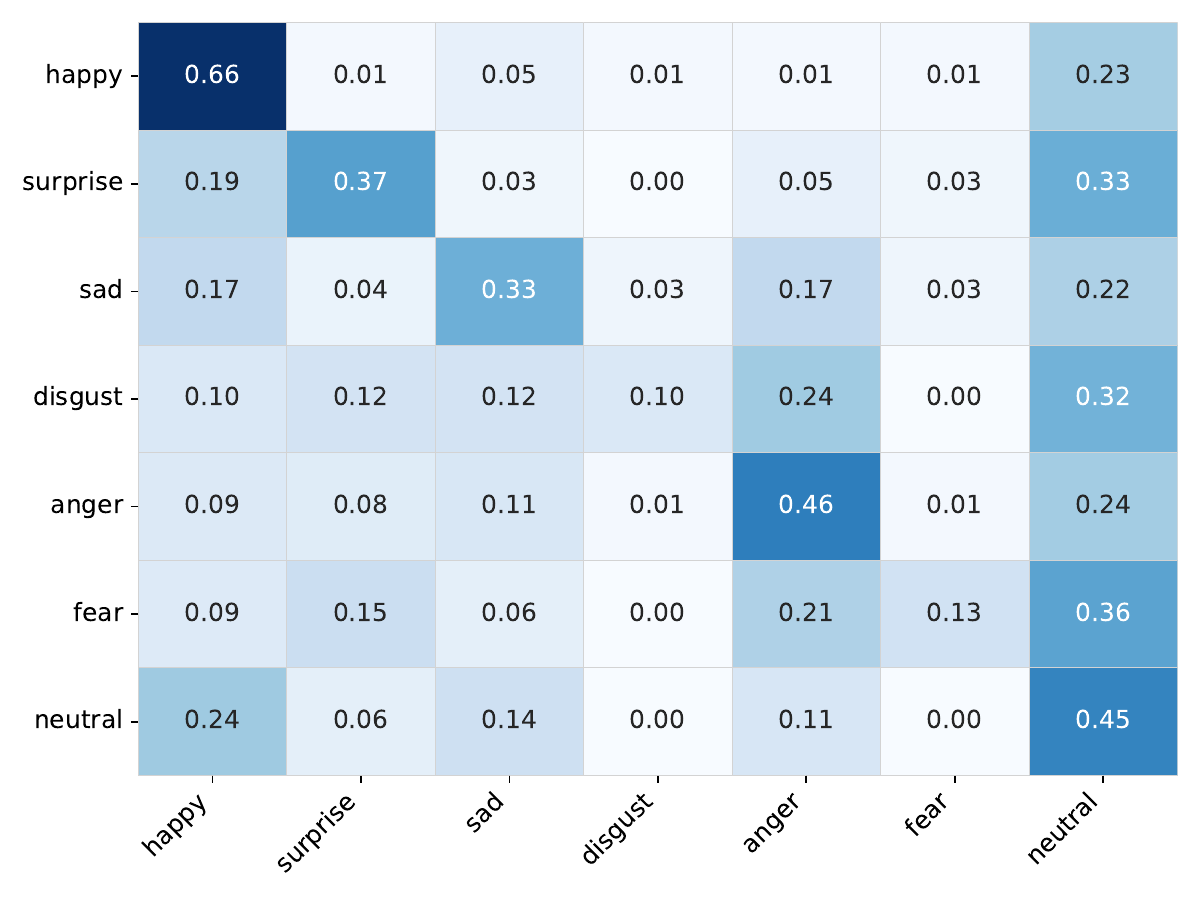}
        \subcaption{\small CEIA (emo.)}
    \end{minipage} \hfill
    \begin{minipage}[b]{0.20\linewidth}
        \centering
        \includegraphics[width=\linewidth]{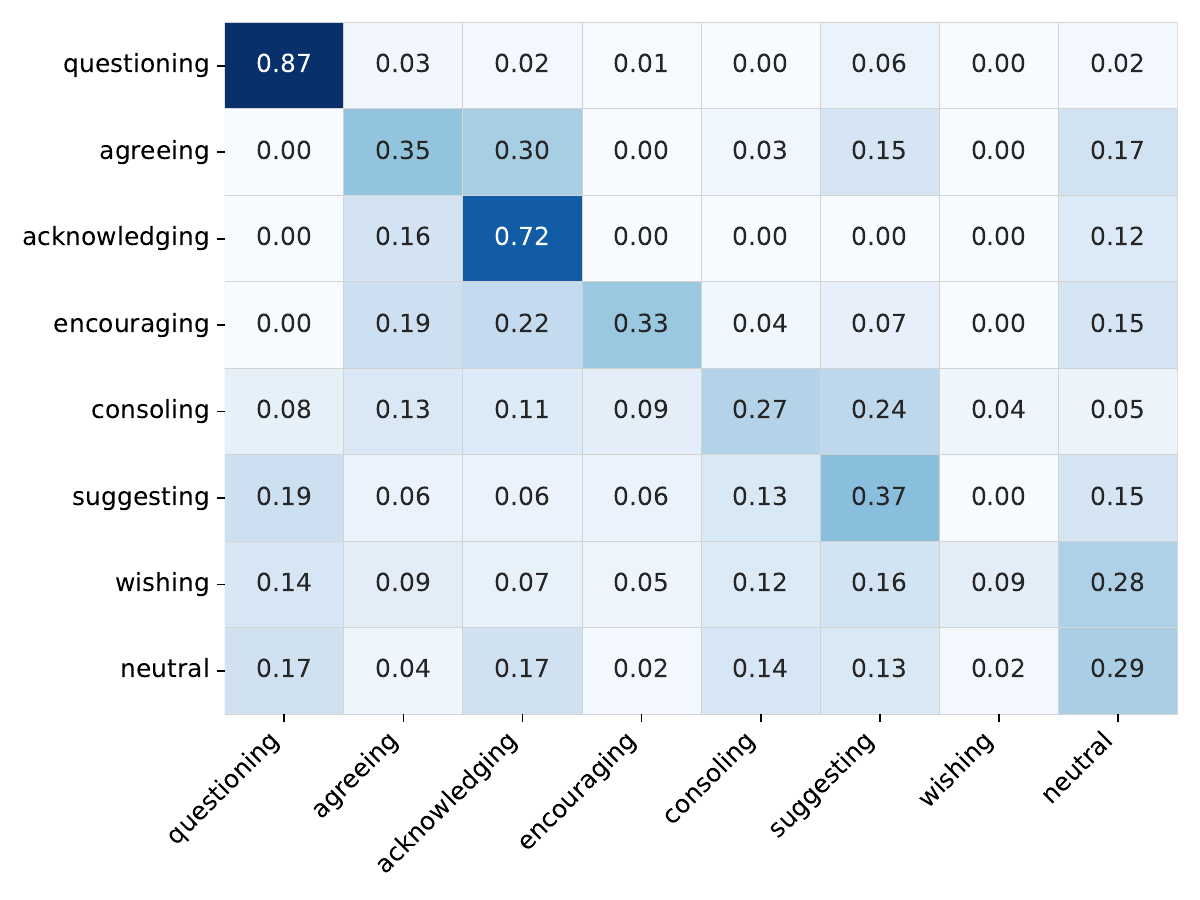}
        \subcaption{\small CEIA (int.)}
    \end{minipage} \hfill
    \begin{minipage}[b]{0.24\linewidth}
        \centering
        \includegraphics[width=\linewidth]{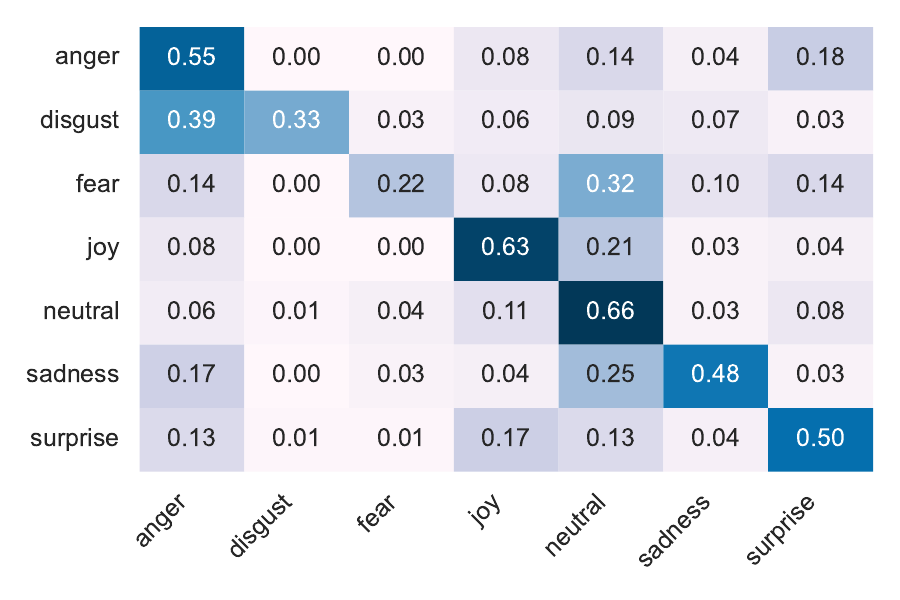}
        \subcaption{\small MPDER}
    \end{minipage} \hfill
    \begin{minipage}[b]{0.16\linewidth}
        \centering
        \includegraphics[width=\linewidth]{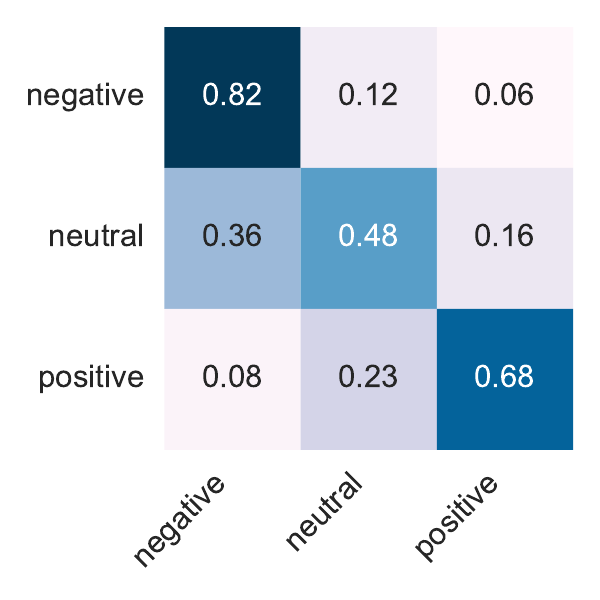}
        \subcaption{\small PEA}
    \end{minipage}

    \caption{\small Confusion matrices for Gemini-1.5-Flash on each evaluation scenario of $\ours$.}
    \label{fig:confusion-Gemini-1.5-Flash}
\end{figure*}

\begin{figure*}[!t]
    \centering
    
    % ===== Row 1 =====
    \begin{minipage}[b]{0.24\linewidth}
        \centering
        \includegraphics[width=\linewidth]{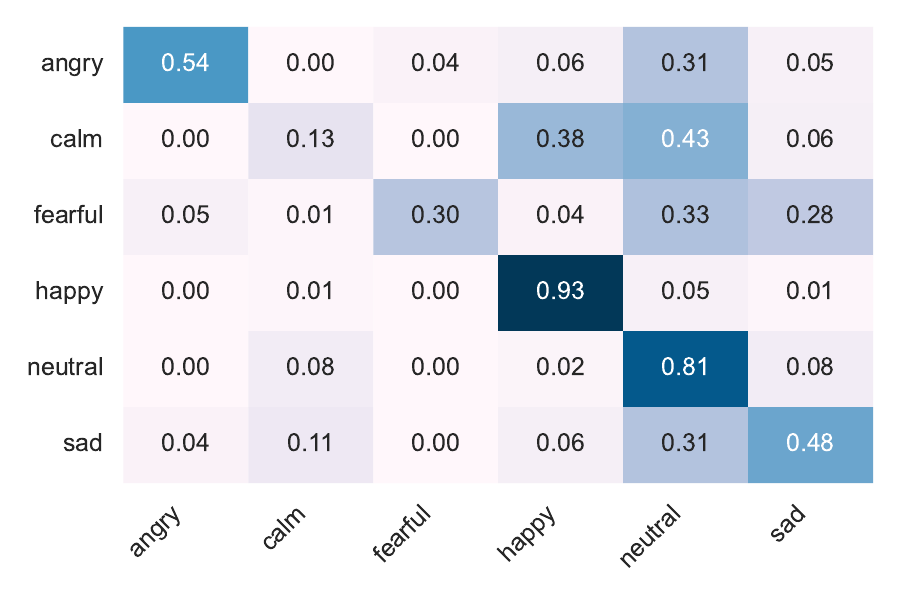}
        \subcaption{\small SOER}
    \end{minipage} \hfill
    \begin{minipage}[b]{0.24\linewidth}
        \centering
        \includegraphics[width=\linewidth]{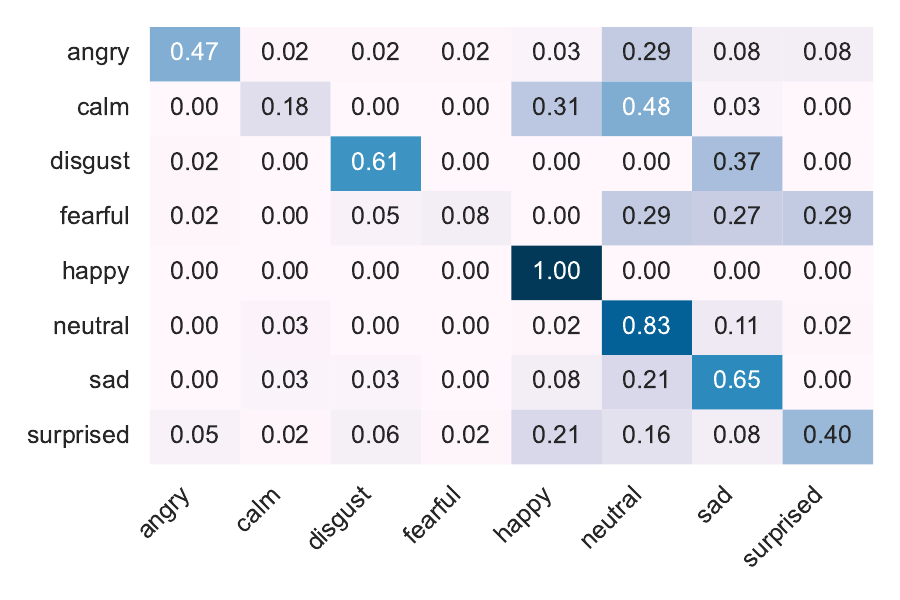}
        \subcaption{\small SPER}
    \end{minipage} \hfill
    \begin{minipage}[b]{0.16\linewidth}
        \centering
        \includegraphics[width=\linewidth]{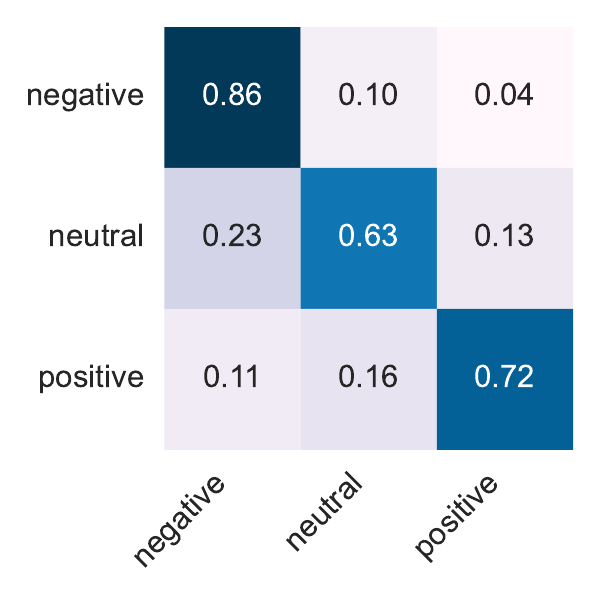}
        \subcaption{\small OSA}
    \end{minipage} \hfill
    \begin{minipage}[b]{0.16\linewidth}
        \centering
        \includegraphics[width=\linewidth]{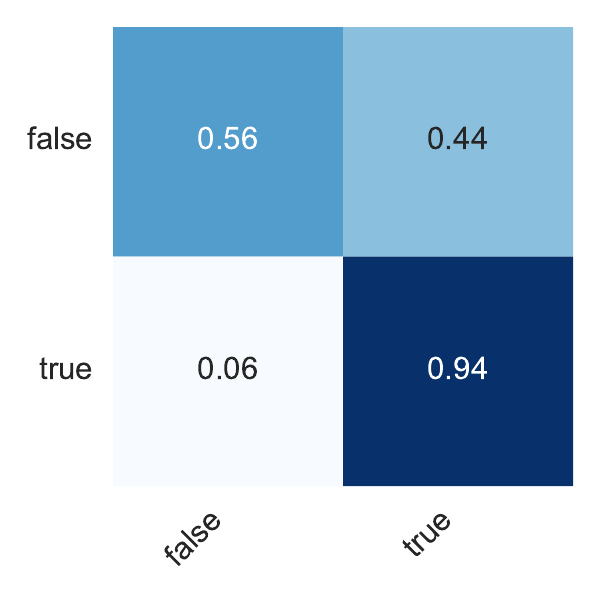}
        \subcaption{\small HU}
    \end{minipage} \\

    % ===== Row 2 =====
    \begin{minipage}[b]{0.24\linewidth}
        \centering
        \includegraphics[width=\linewidth]{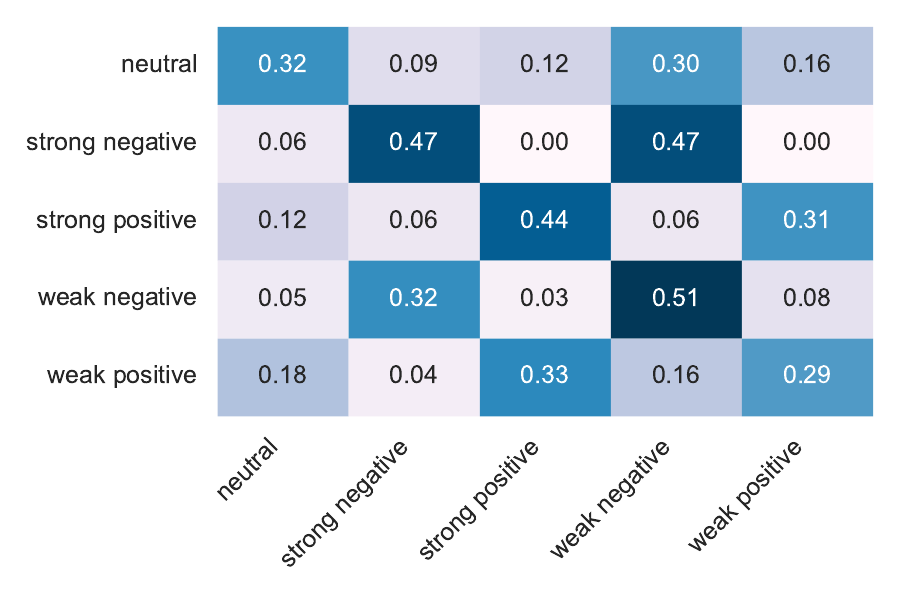}
        \subcaption{\small SCEA}
    \end{minipage} \hfill
    \begin{minipage}[b]{0.24\linewidth}
        \centering
        \includegraphics[width=\linewidth]{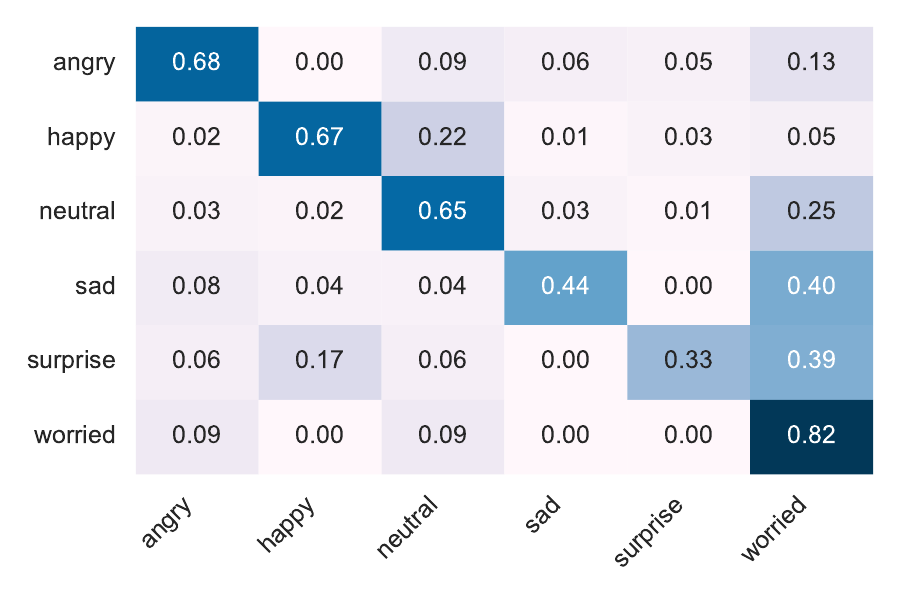}
        \subcaption{\small FGDEA}
    \end{minipage} \hfill
    \begin{minipage}[b]{0.16\linewidth}
        \centering
        \includegraphics[width=\linewidth]{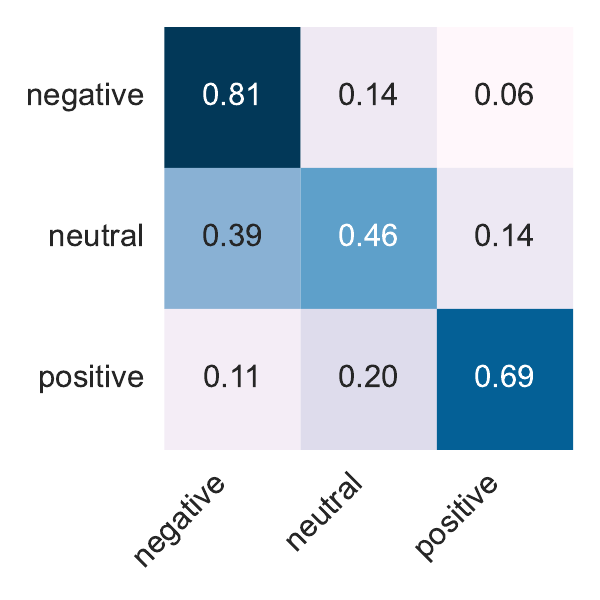}
        \subcaption{\small FCDEA}
    \end{minipage} \hfill
    \begin{minipage}[b]{0.16\linewidth}
        \centering
        \includegraphics[width=\linewidth]{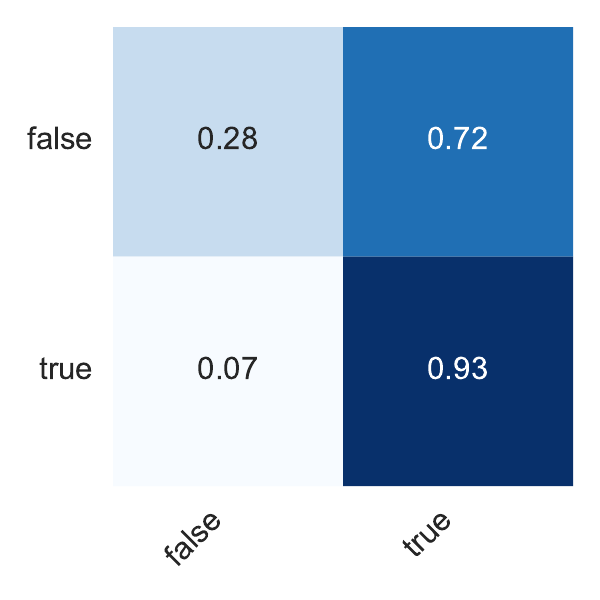}
        \subcaption{\small SD}
    \end{minipage} \\

    % ===== Row 3 =====
    \begin{minipage}[b]{0.16\linewidth}
        \centering
        \includegraphics[width=\linewidth]{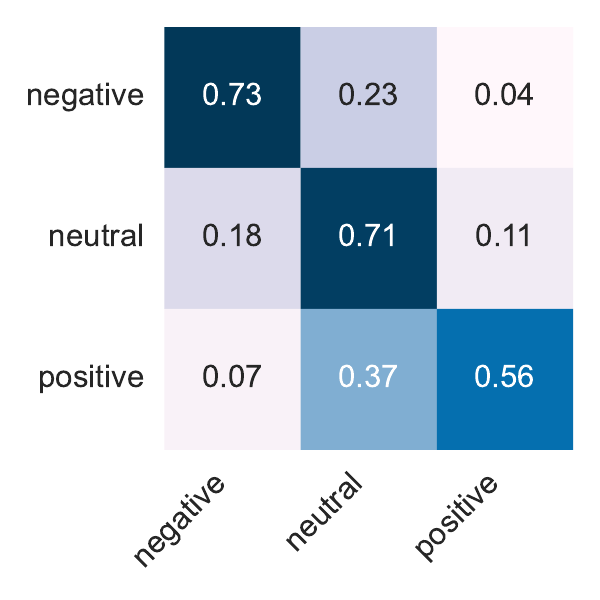}
        \subcaption{\small EIA}
    \end{minipage} \hfill
    \begin{minipage}[b]{0.20\linewidth}
        \centering
        \includegraphics[width=\linewidth]{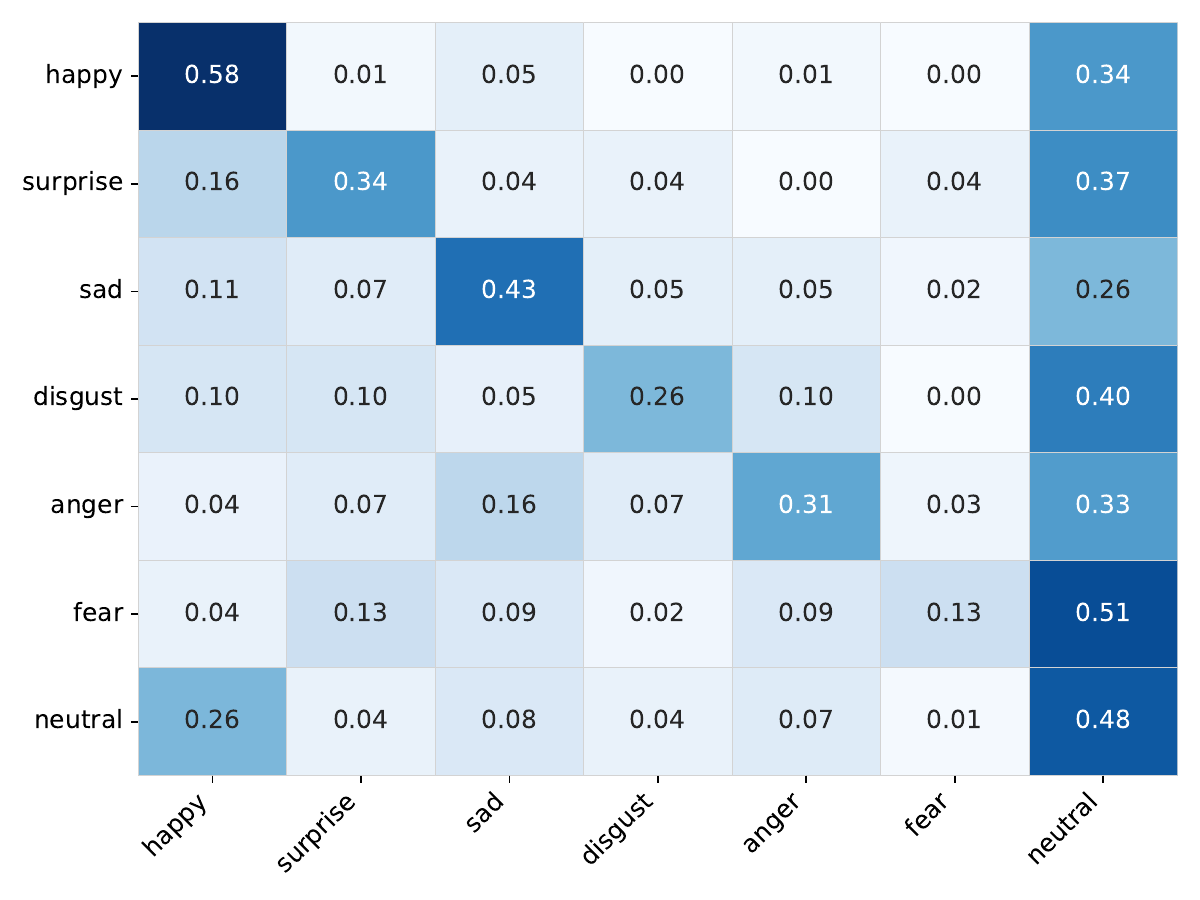}
        \subcaption{\small CEIA (emo.)}
    \end{minipage} \hfill
    \begin{minipage}[b]{0.20\linewidth}
        \centering
        \includegraphics[width=\linewidth]{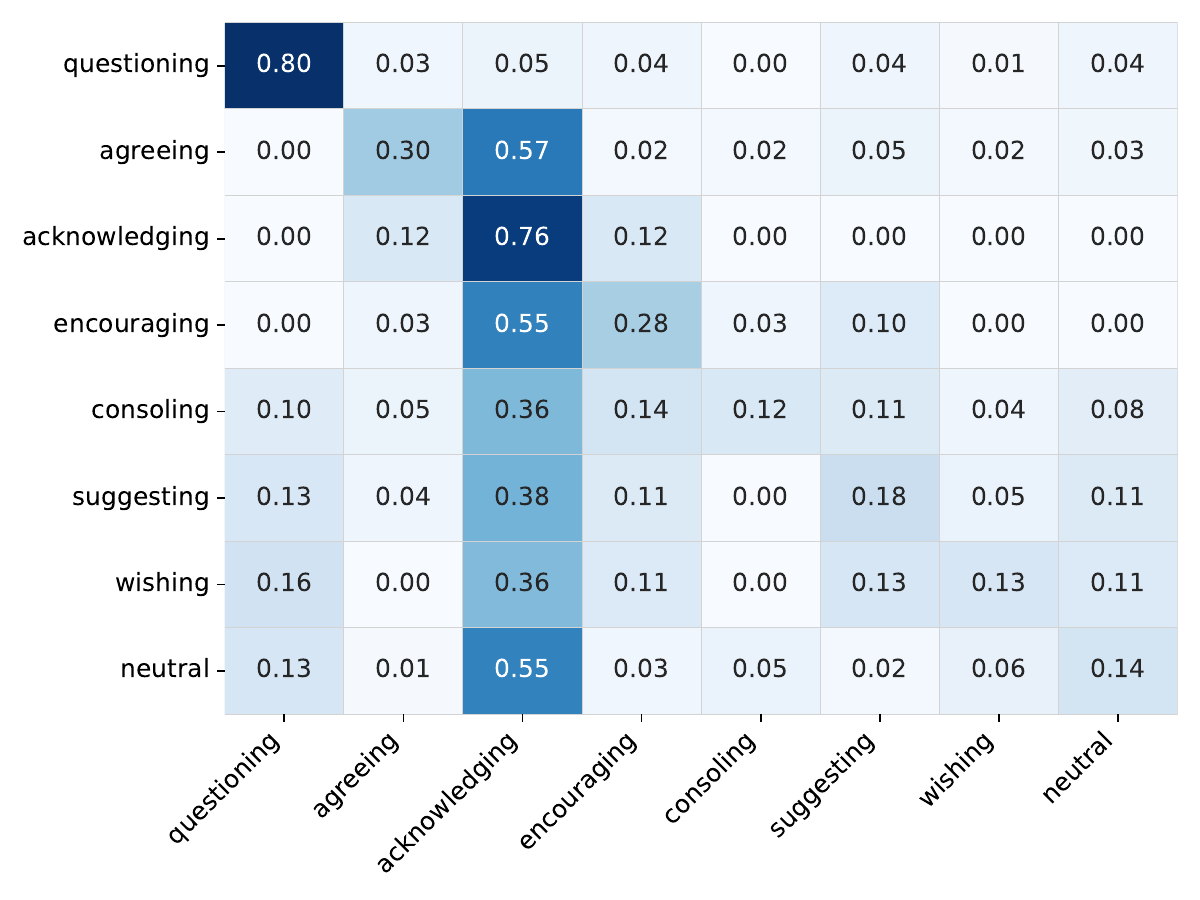}
        \subcaption{\small CEIA (int.)}
    \end{minipage} \hfill
    \begin{minipage}[b]{0.24\linewidth}
        \centering
        \includegraphics[width=\linewidth]{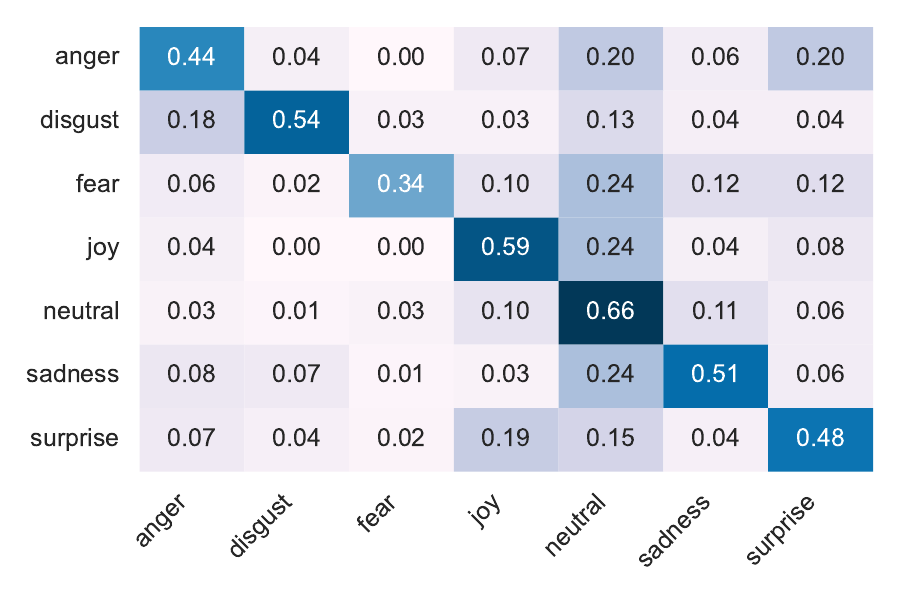}
        \subcaption{\small MPDER}
    \end{minipage} \hfill
    \begin{minipage}[b]{0.16\linewidth}
        \centering
        \includegraphics[width=\linewidth]{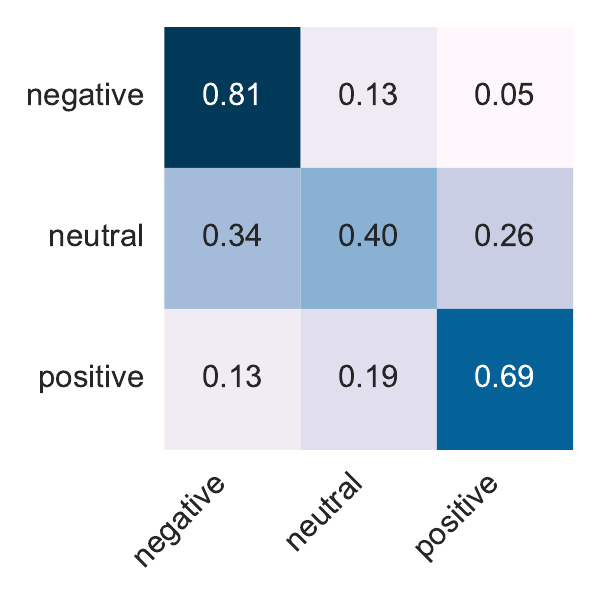}
        \subcaption{\small PEA}
    \end{minipage}

    \vspace{-3mm}
    \caption{\small Confusion matrices for Gemini-2.0-Flash-Thinking on each evaluation scenario of $\ours$.}
    \label{fig:confusion-Gemini-2.0-Flash-Thinking}
\end{figure*}

\begin{figure*}[!t]
    \centering
    
    % ===== Row 1 =====
    \begin{minipage}[b]{0.24\linewidth}
        \centering
        \includegraphics[width=\linewidth]{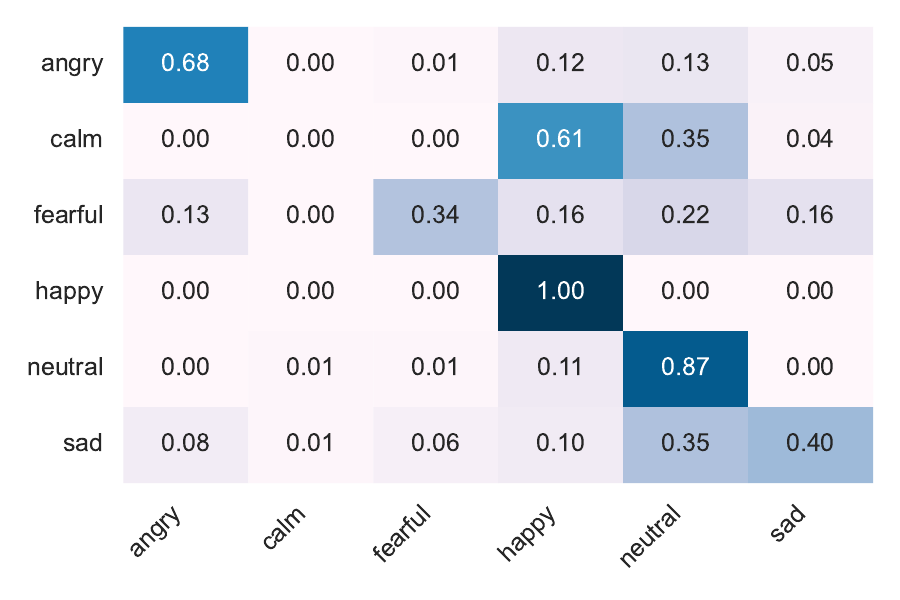}
        \subcaption{\small SOER}
    \end{minipage} \hfill
    \begin{minipage}[b]{0.24\linewidth}
        \centering
        \includegraphics[width=\linewidth]{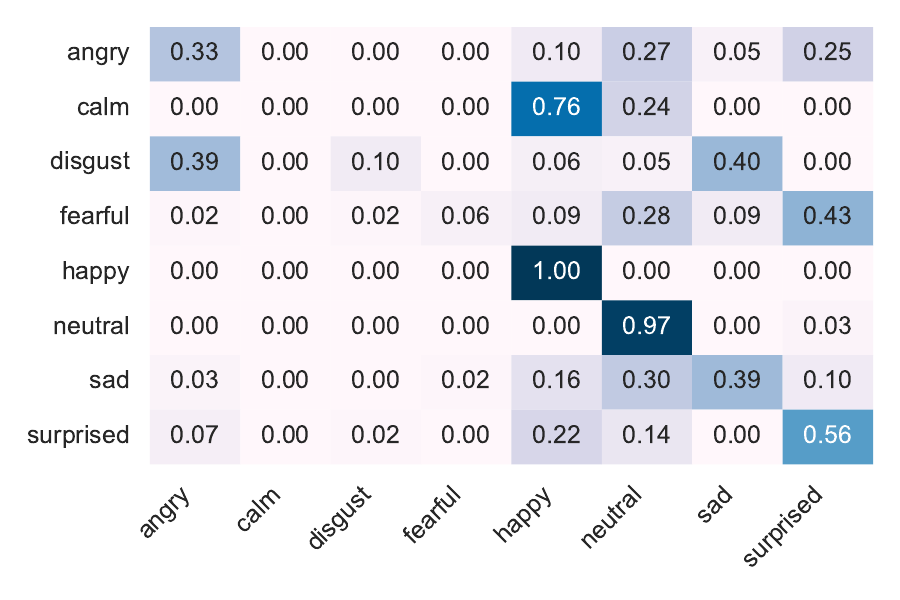}
        \subcaption{\small SPER}
    \end{minipage} \hfill
    \begin{minipage}[b]{0.16\linewidth}
        \centering
        \includegraphics[width=\linewidth]{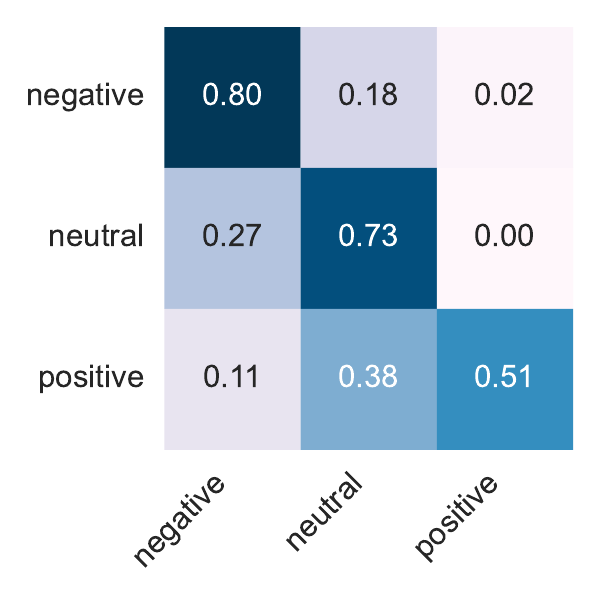}
        \subcaption{\small OSA}
    \end{minipage} \hfill
    \begin{minipage}[b]{0.16\linewidth}
        \centering
        \includegraphics[width=\linewidth]{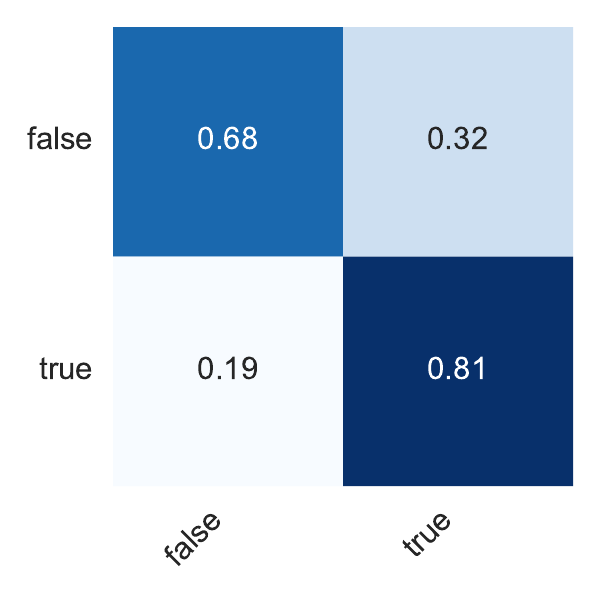}
        \subcaption{\small HU}
    \end{minipage} \\

    % ===== Row 2 =====
    \begin{minipage}[b]{0.24\linewidth}
        \centering
        \includegraphics[width=\linewidth]{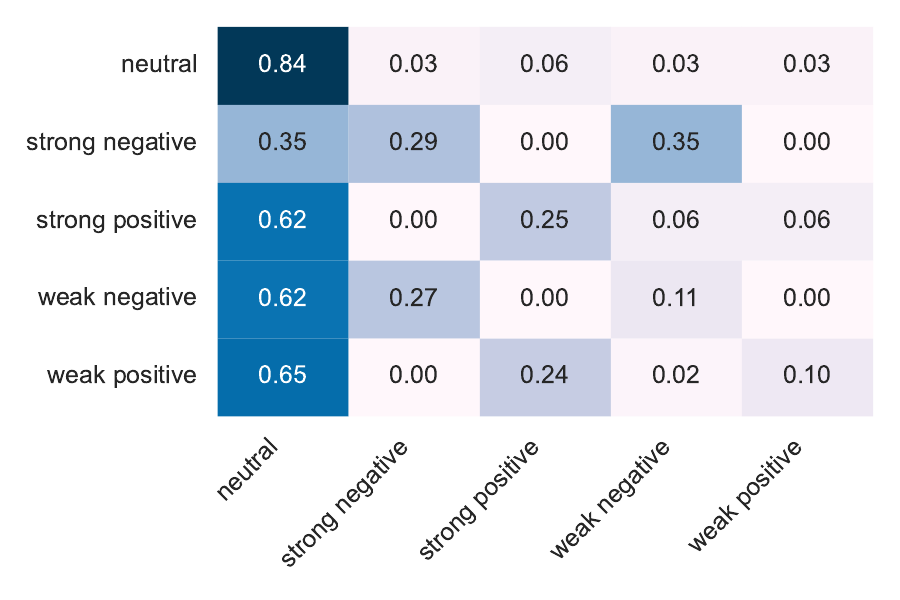}
        \subcaption{\small SCEA}
    \end{minipage} \hfill
    \begin{minipage}[b]{0.24\linewidth}
        \centering
        \includegraphics[width=\linewidth]{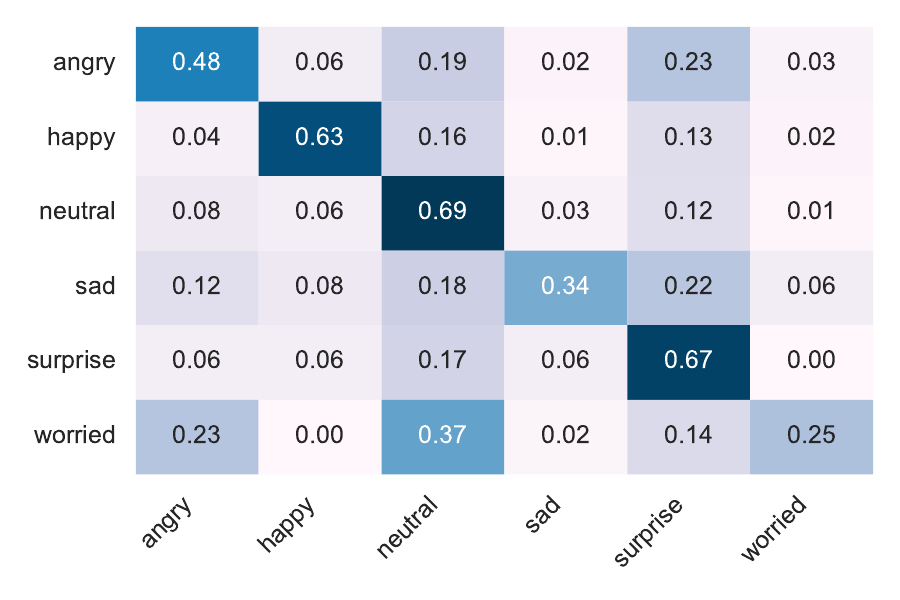}
        \subcaption{\small FGDEA}
    \end{minipage} \hfill
    \begin{minipage}[b]{0.16\linewidth}
        \centering
        \includegraphics[width=\linewidth]{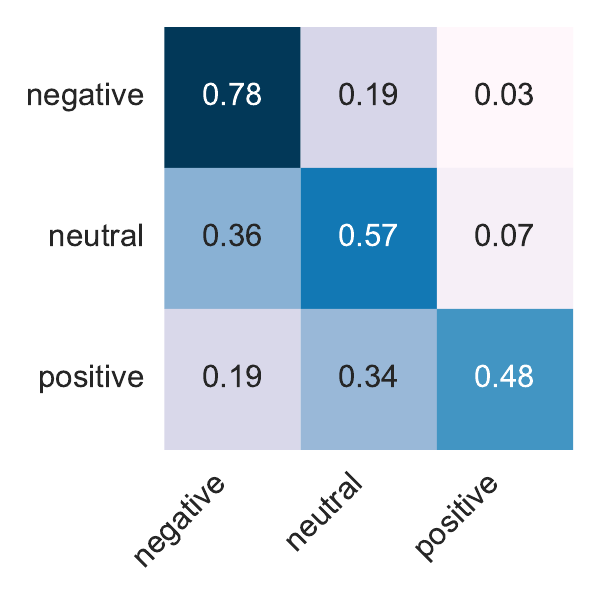}
        \subcaption{\small FCDEA}
    \end{minipage} \hfill
    \begin{minipage}[b]{0.16\linewidth}
        \centering
        \includegraphics[width=\linewidth]{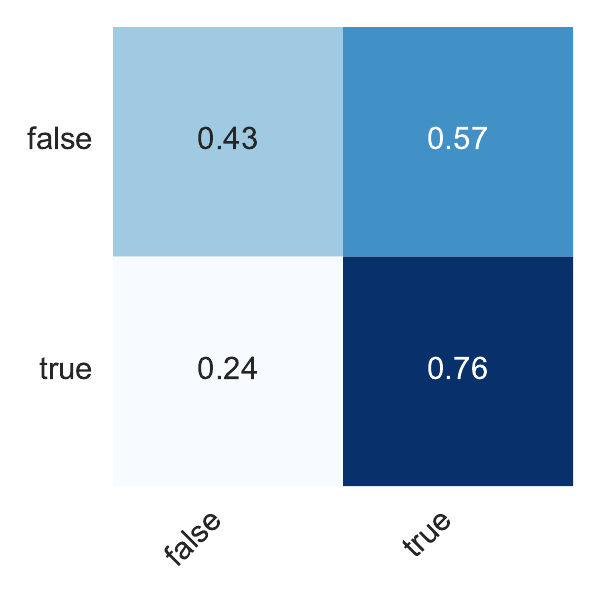}
        \subcaption{\small SD}
    \end{minipage} \\

    % ===== Row 3 =====
    \begin{minipage}[b]{0.16\linewidth}
        \centering
        \includegraphics[width=\linewidth]{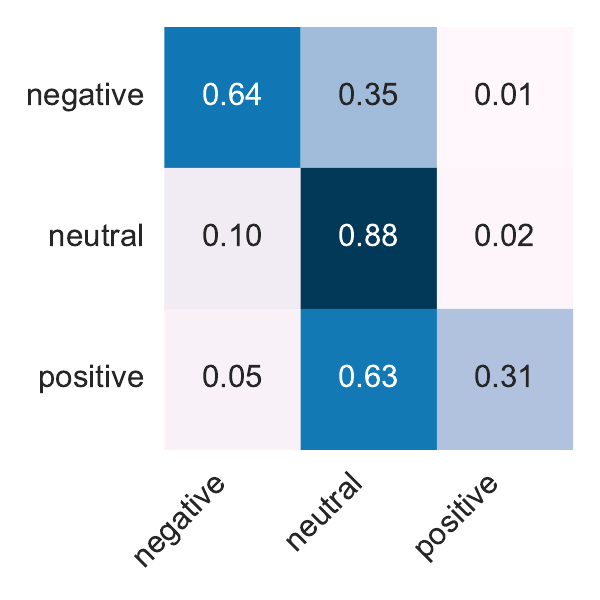}
        \subcaption{\small EIA}
    \end{minipage} \hfill
    \begin{minipage}[b]{0.20\linewidth}
        \centering
        \includegraphics[width=\linewidth]{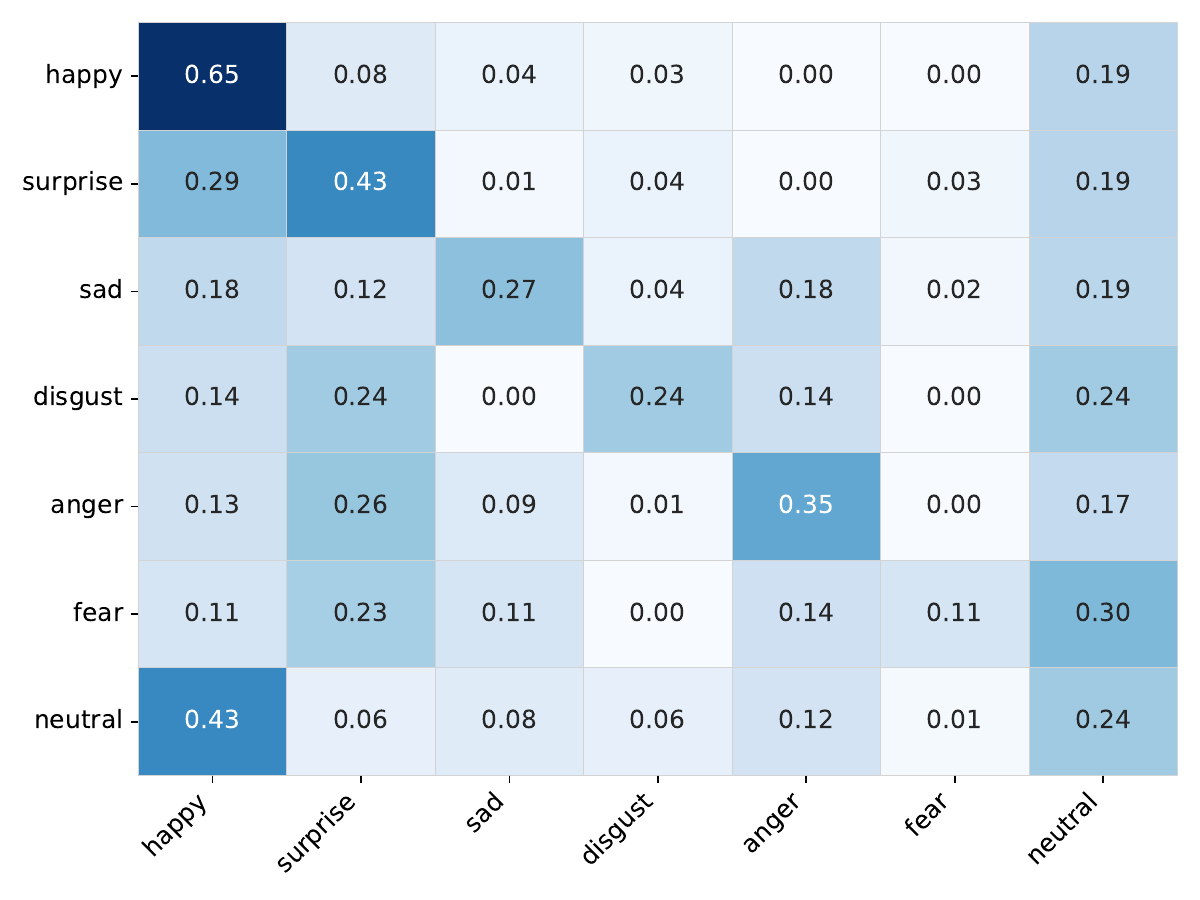}
        \subcaption{\small CEIA (emo.)}
    \end{minipage} \hfill
    \begin{minipage}[b]{0.20\linewidth}
        \centering
        \includegraphics[width=\linewidth]{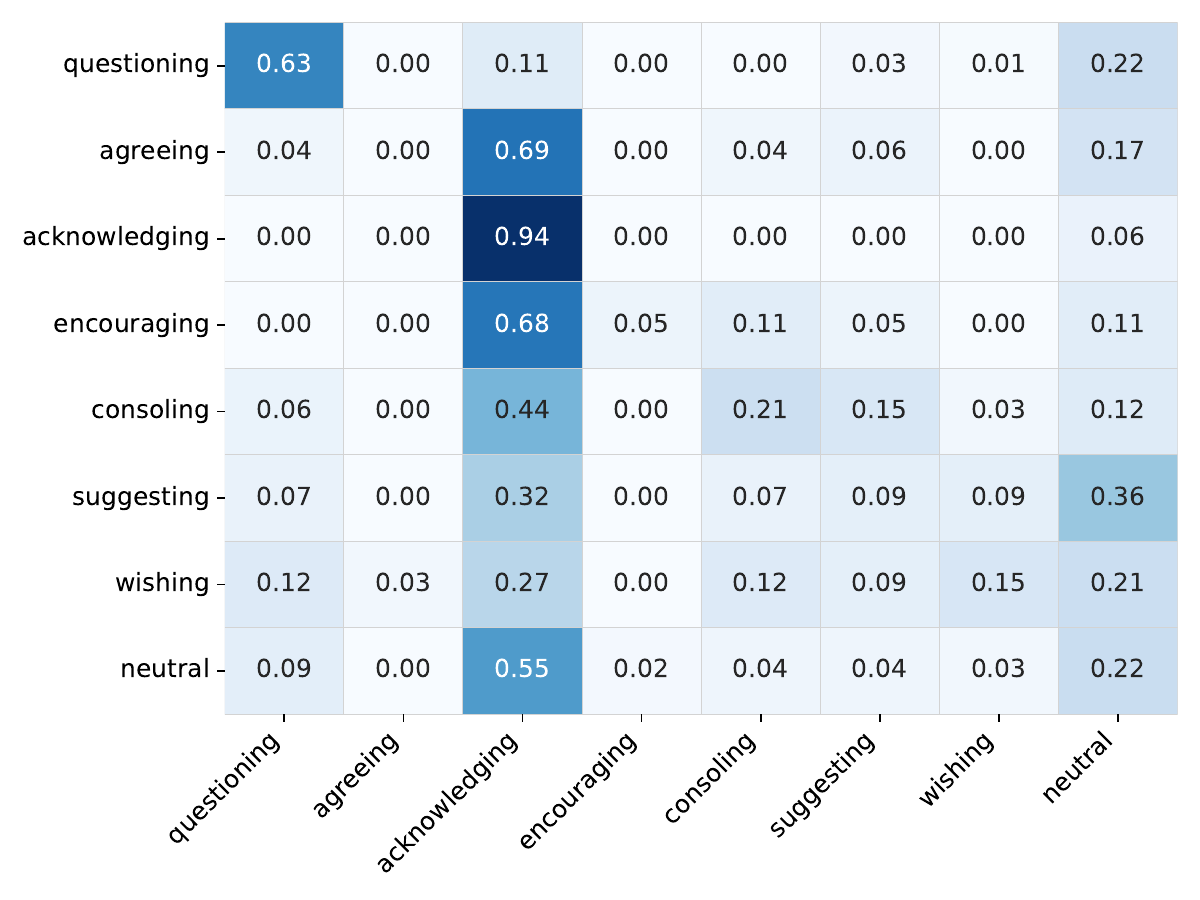}
        \subcaption{\small CEIA (int.)}
    \end{minipage} \hfill
    \begin{minipage}[b]{0.24\linewidth}
        \centering
        \includegraphics[width=\linewidth]{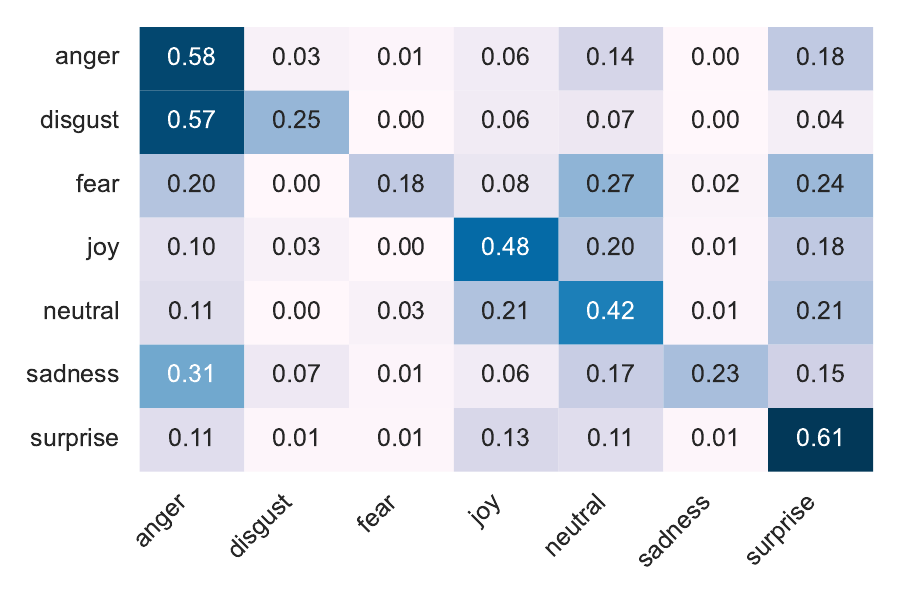}
        \subcaption{\small MPDER}
    \end{minipage} \hfill
    \begin{minipage}[b]{0.16\linewidth}
        \centering
        \includegraphics[width=\linewidth]{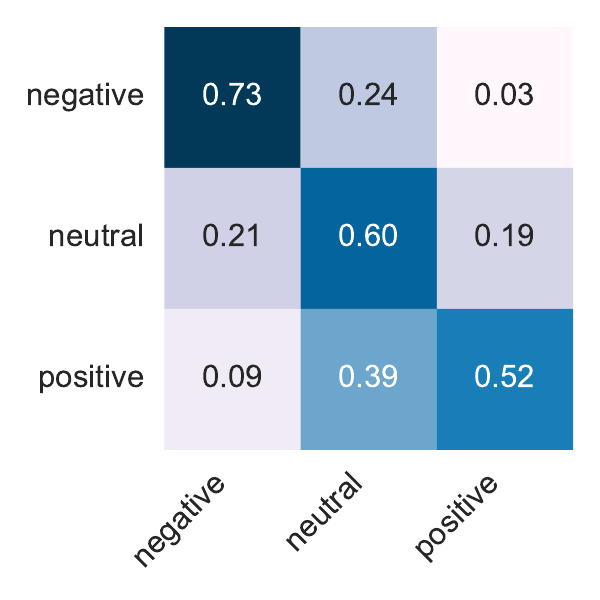}
        \subcaption{\small PEA}
    \end{minipage}

    \vspace{-2mm}
    \caption{\small Confusion matrices for GLM-4V-PLUS on each evaluation scenario of $\ours$.}
    \label{fig:confusion-GLM-4V-PLUS}
\end{figure*}

\begin{figure*}[!t]
    \centering
    
    % ===== Row 1 =====
    \begin{minipage}[b]{0.24\linewidth}
        \centering
        \includegraphics[width=\linewidth]{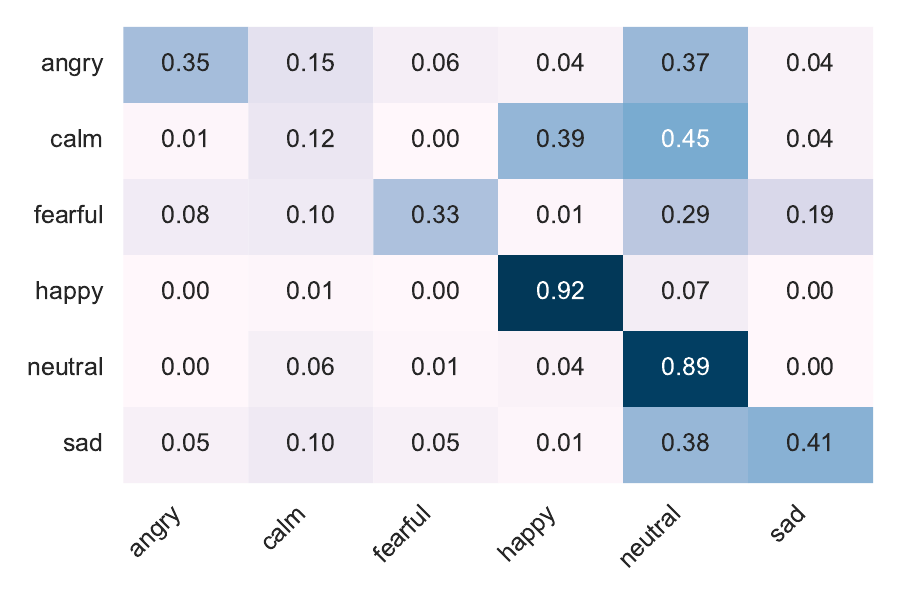}
        \subcaption{\small SOER}
    \end{minipage} \hfill
    \begin{minipage}[b]{0.24\linewidth}
        \centering
        \includegraphics[width=\linewidth]{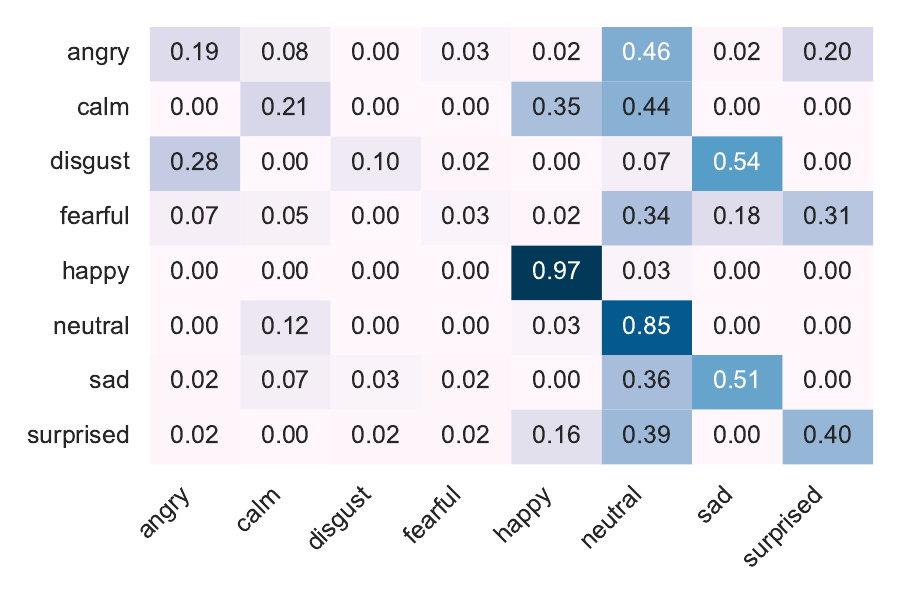}
        \subcaption{\small SPER}
    \end{minipage} \hfill
    \begin{minipage}[b]{0.16\linewidth}
        \centering
        \includegraphics[width=\linewidth]{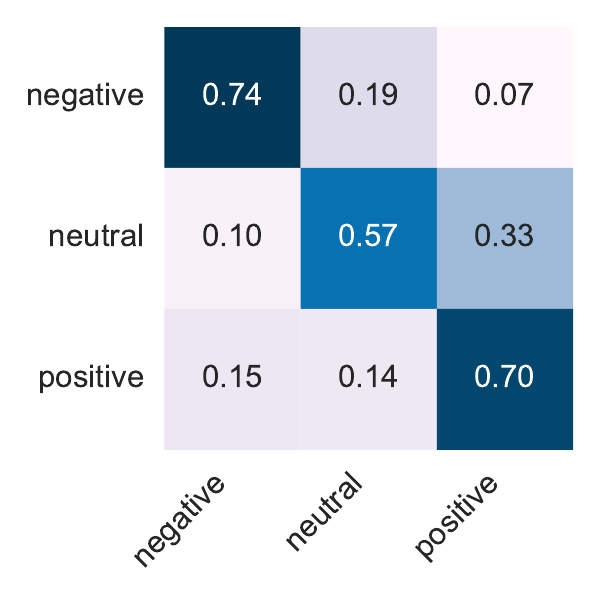}
        \subcaption{\small OSA}
    \end{minipage} \hfill
    \begin{minipage}[b]{0.16\linewidth}
        \centering
        \includegraphics[width=\linewidth]{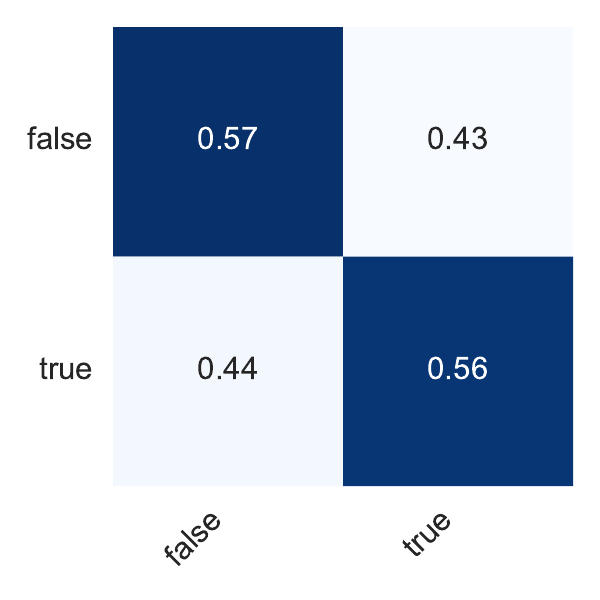}
        \subcaption{\small HU}
    \end{minipage} \\

    % ===== Row 2 =====
    \begin{minipage}[b]{0.24\linewidth}
        \centering
        \includegraphics[width=\linewidth]{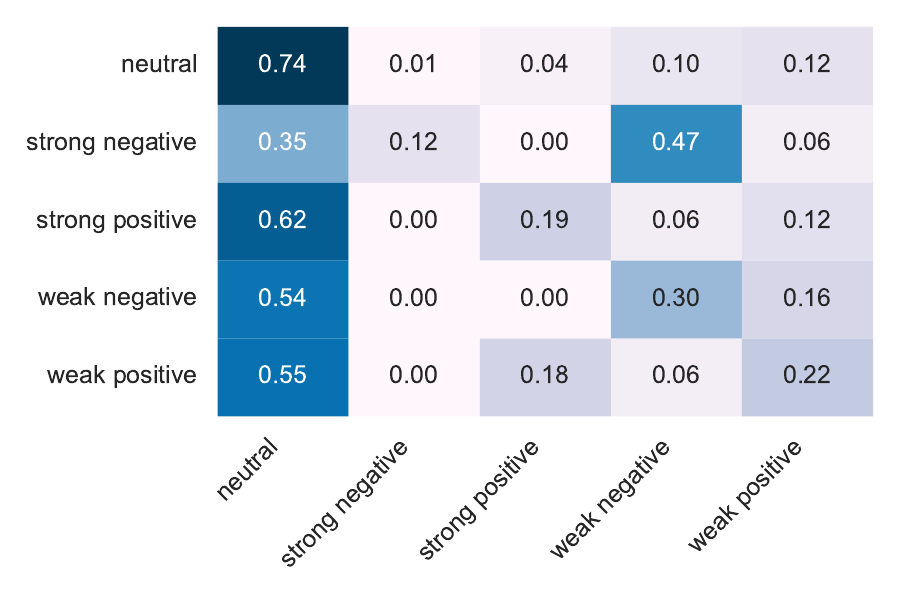}
        \subcaption{\small SCEA}
    \end{minipage} \hfill
    \begin{minipage}[b]{0.24\linewidth}
        \centering
        \includegraphics[width=\linewidth]{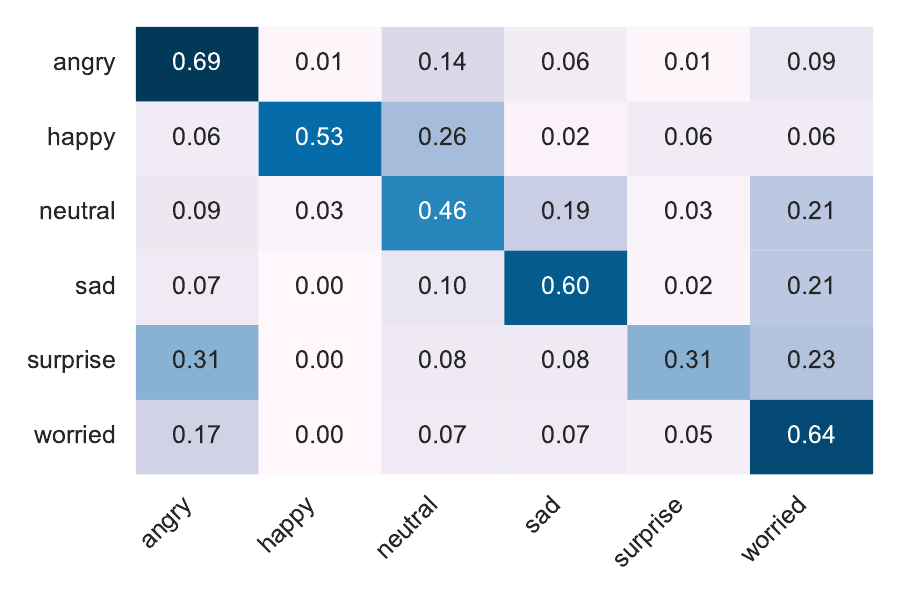}
        \subcaption{\small FGDEA}
    \end{minipage} \hfill
    \begin{minipage}[b]{0.16\linewidth}
        \centering
        \includegraphics[width=\linewidth]{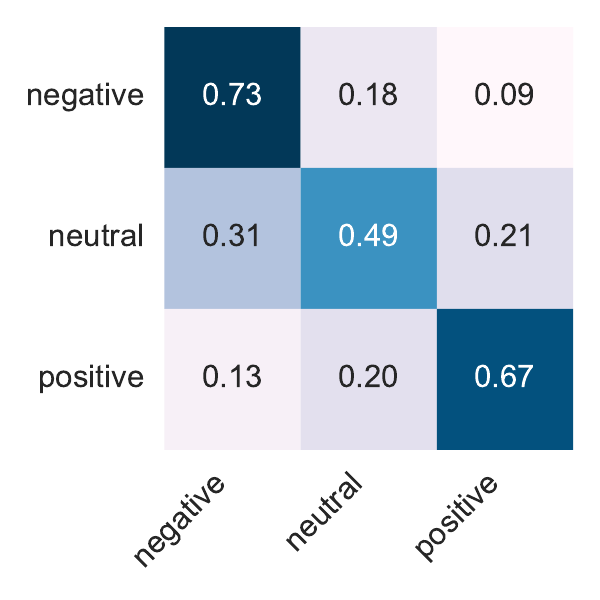}
        \subcaption{\small FCDEA}
    \end{minipage} \hfill
    \begin{minipage}[b]{0.16\linewidth}
        \centering
        \includegraphics[width=\linewidth]{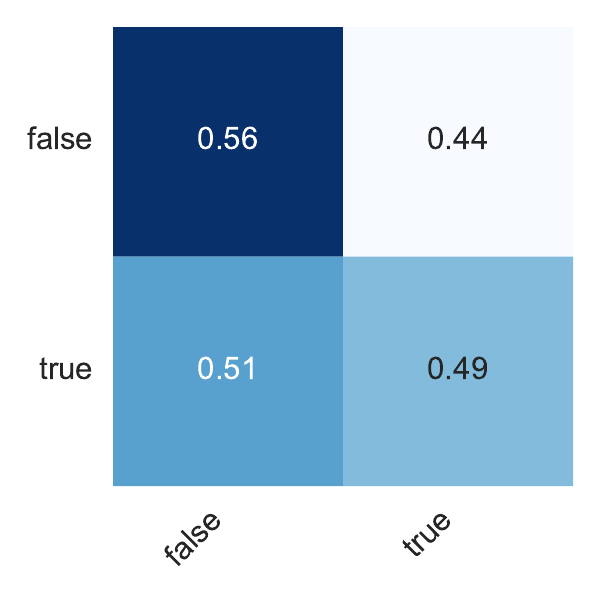}
        \subcaption{\small SD}
    \end{minipage} \\

    % ===== Row 3 =====
    \begin{minipage}[b]{0.16\linewidth}
        \centering
        \includegraphics[width=\linewidth]{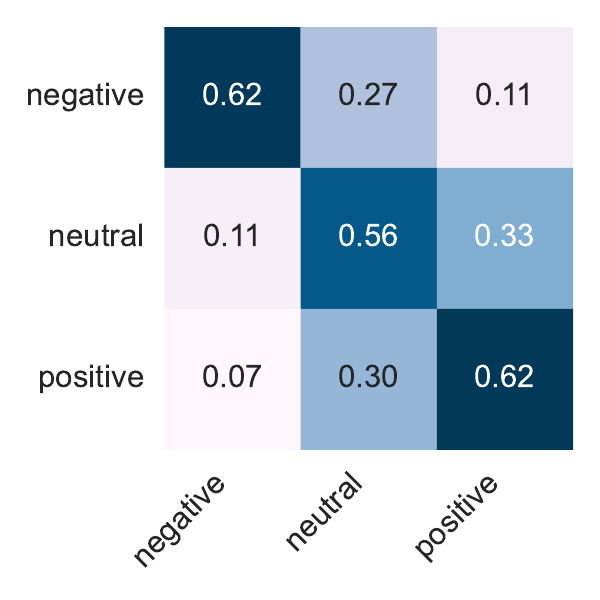}
        \subcaption{\small EIA}
    \end{minipage} \hfill
    \begin{minipage}[b]{0.20\linewidth}
        \centering
        \includegraphics[width=\linewidth]{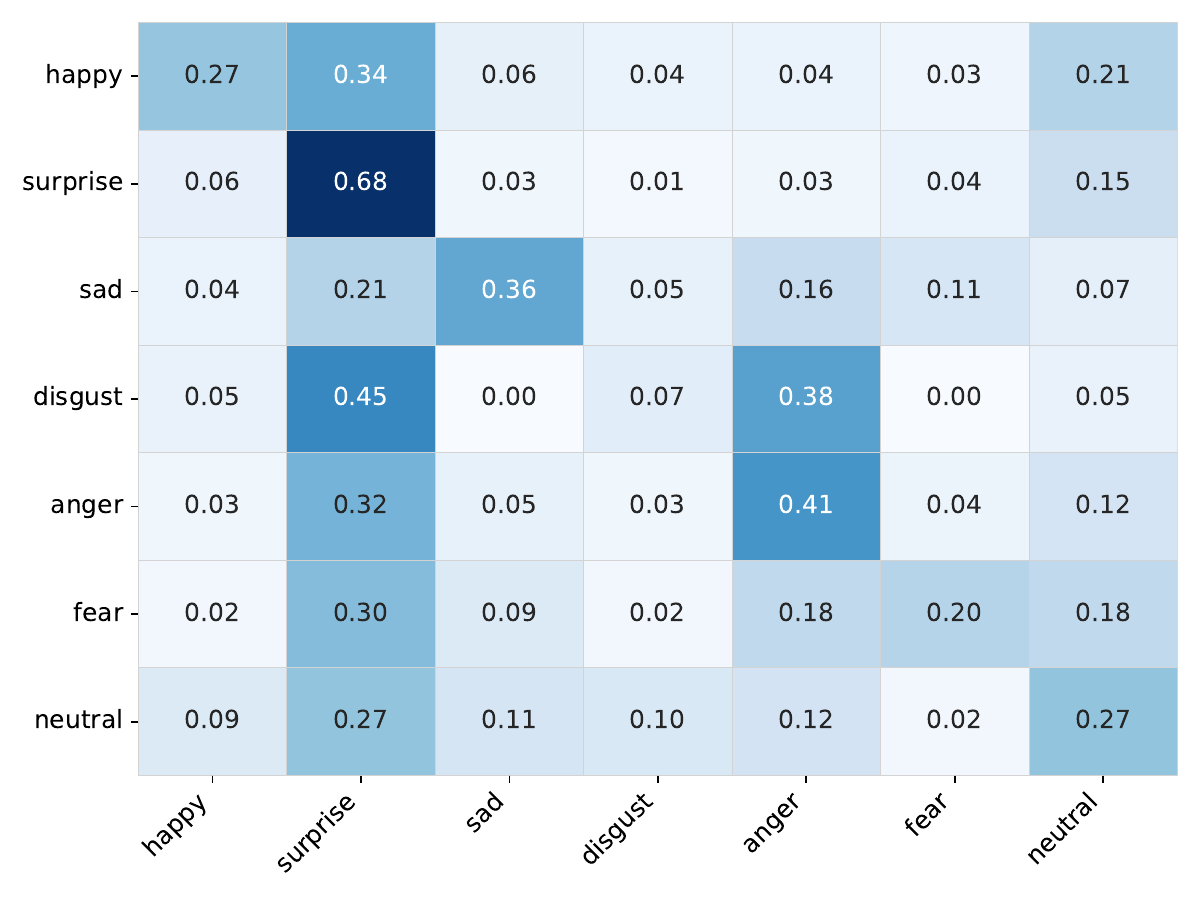}
        \subcaption{\small CEIA (emo.)}
    \end{minipage} \hfill
    \begin{minipage}[b]{0.20\linewidth}
        \centering
        \includegraphics[width=\linewidth]{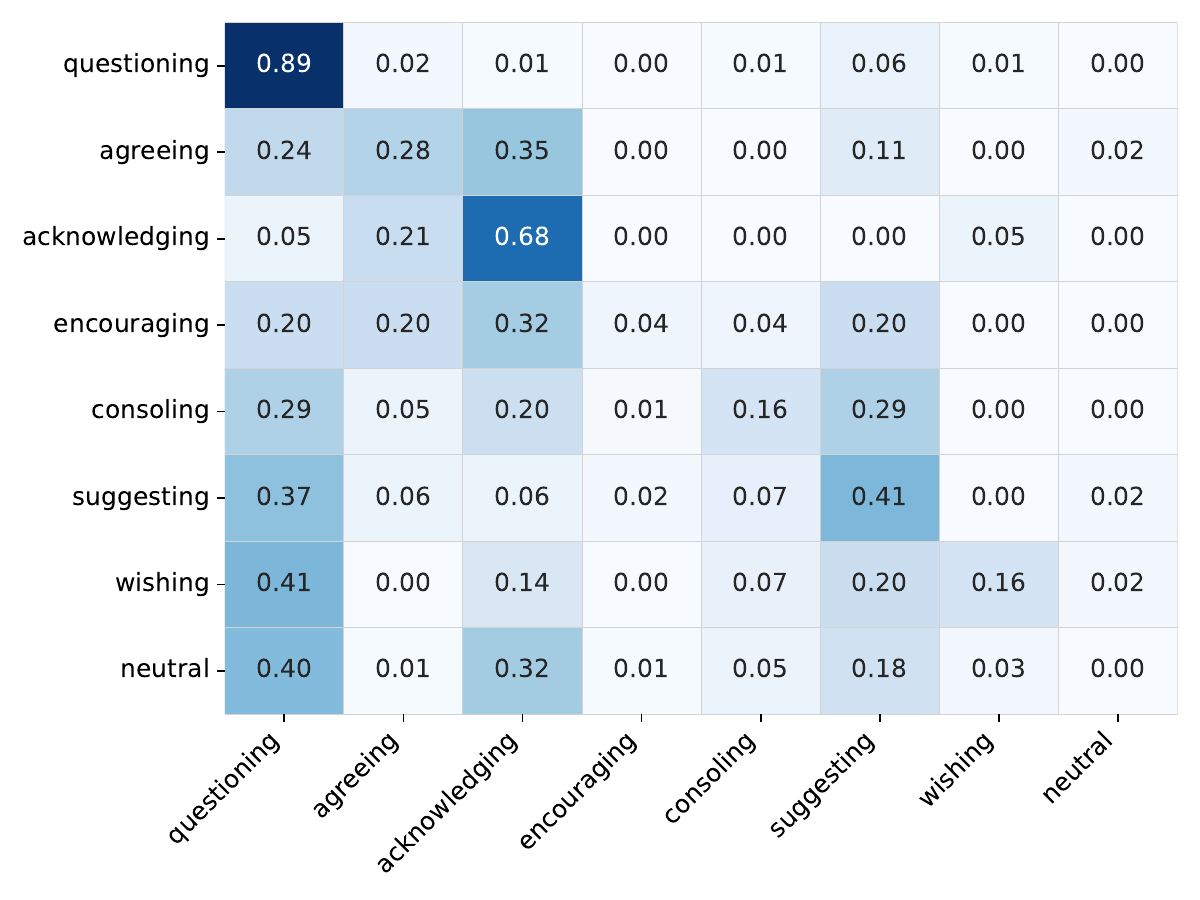}
        \subcaption{\small CEIA (int.)}
    \end{minipage} \hfill
    \begin{minipage}[b]{0.24\linewidth}
        \centering
        \includegraphics[width=\linewidth]{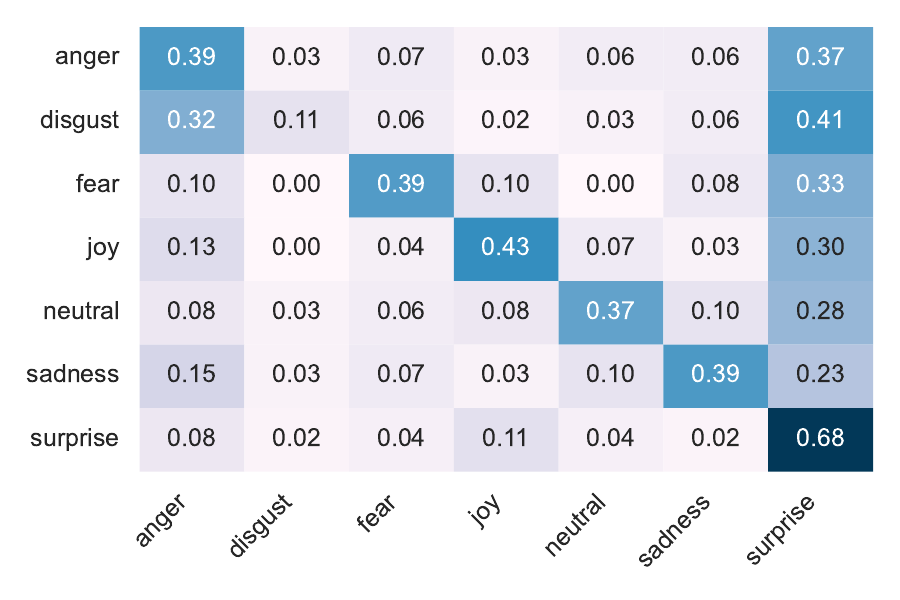}
        \subcaption{\small MPDER}
    \end{minipage} \hfill
    \begin{minipage}[b]{0.16\linewidth}
        \centering
        \includegraphics[width=\linewidth]{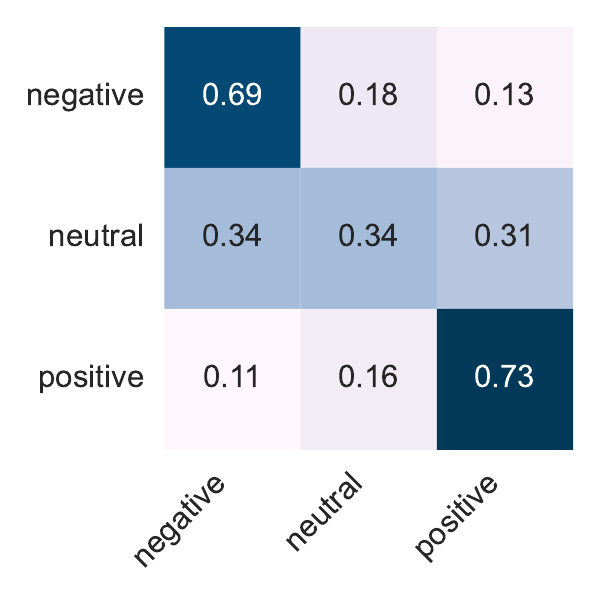}
        \subcaption{\small PEA}
    \end{minipage}

    \vspace{-3mm}
    \caption{\small Confusion matrices for InternVL2.5-4B on each evaluation scenario of $\ours$.}
    \label{fig:confusion-InternVL2.5-4B}
\end{figure*}

\begin{figure*}[!t]
    \centering
    
    % ===== Row 1 =====
    \begin{minipage}[t]{0.24\linewidth}
        \centering
        \includegraphics[width=\linewidth]{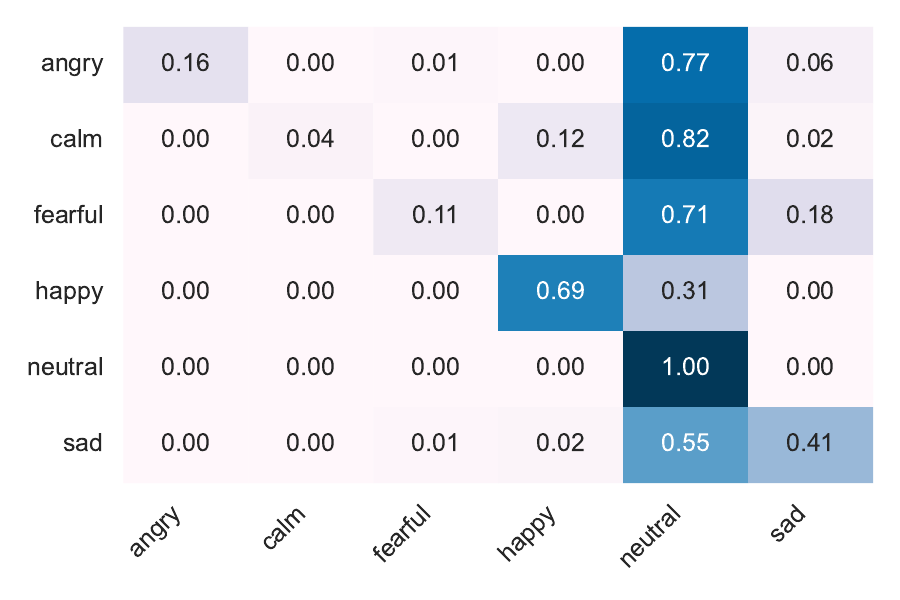}
        \subcaption{\small SOER}
    \end{minipage} \hfill
    \begin{minipage}[t]{0.24\linewidth}
        \centering
        \includegraphics[width=\linewidth]{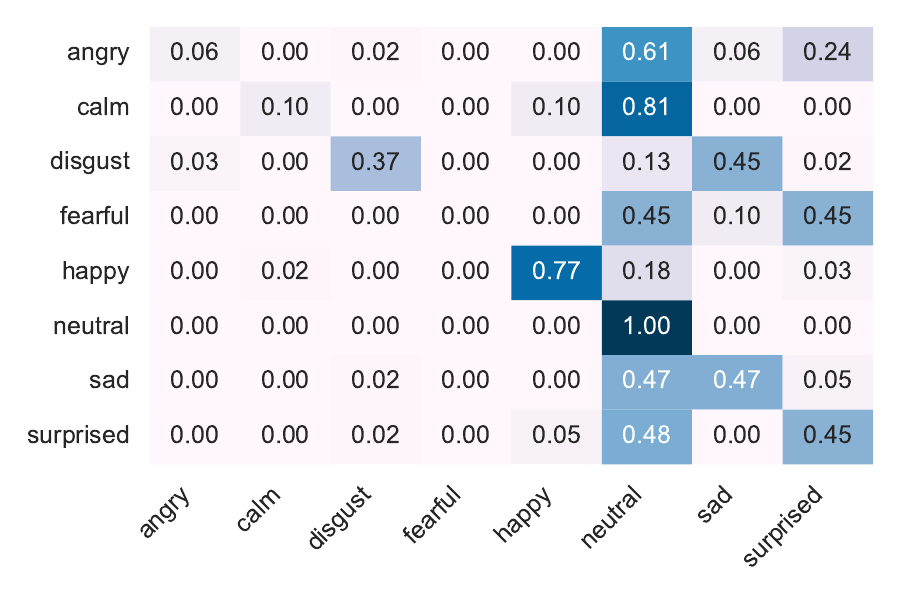}
        \subcaption{\small SPER}
    \end{minipage} \hfill
    \begin{minipage}[t]{0.16\linewidth}
        \centering
        \includegraphics[width=\linewidth]{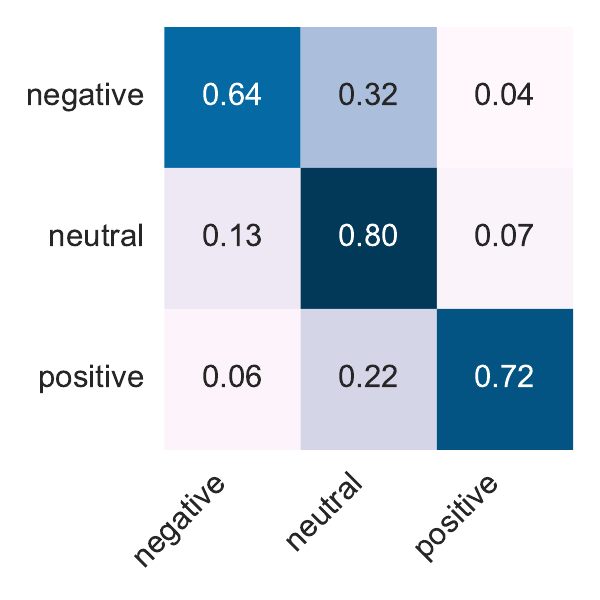}
        \subcaption{\small OSA}
    \end{minipage} \hfill
    \begin{minipage}[t]{0.16\linewidth}
        \centering
        \includegraphics[width=\linewidth]{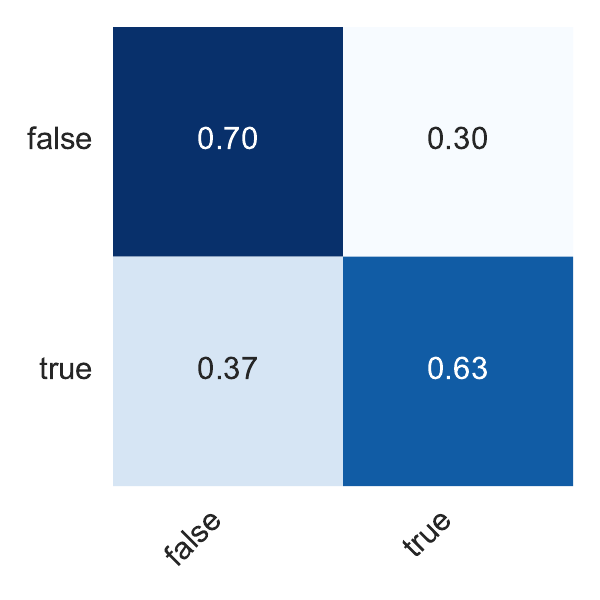}
        \subcaption{\small HU}
    \end{minipage} \\

    % ===== Row 2 =====
    \begin{minipage}[t]{0.24\linewidth}
        \centering
        \includegraphics[width=\linewidth]{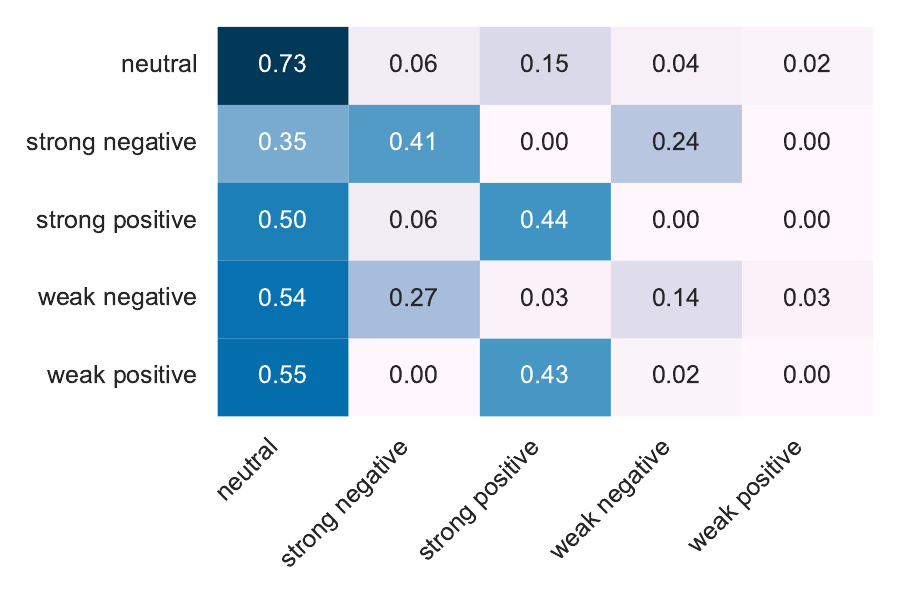}
        \subcaption{\small SCEA}
    \end{minipage} \hfill
    \begin{minipage}[t]{0.24\linewidth}
        \centering
        \includegraphics[width=\linewidth]{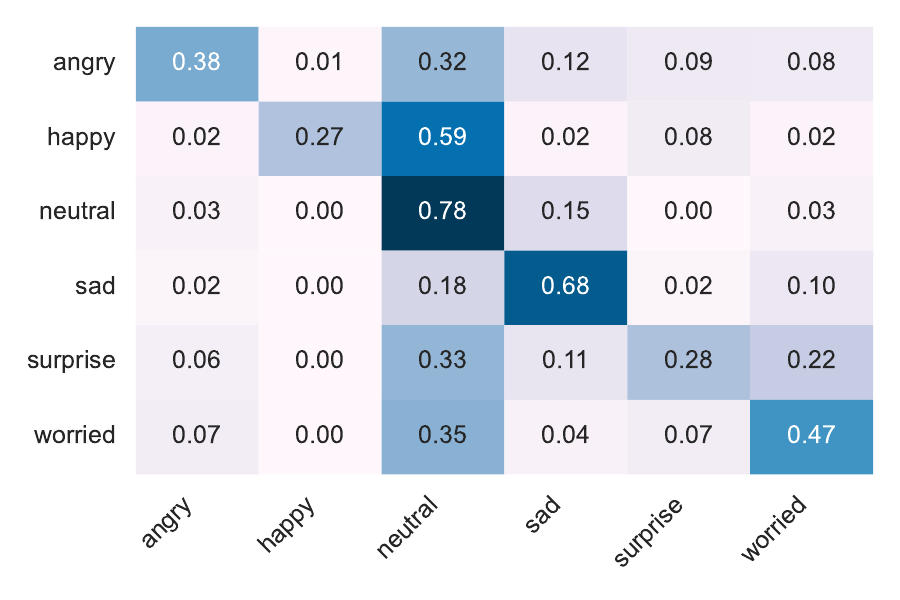}
        \subcaption{\small FGDEA}
    \end{minipage} \hfill
    \begin{minipage}[t]{0.16\linewidth}
        \centering
        \includegraphics[width=\linewidth]{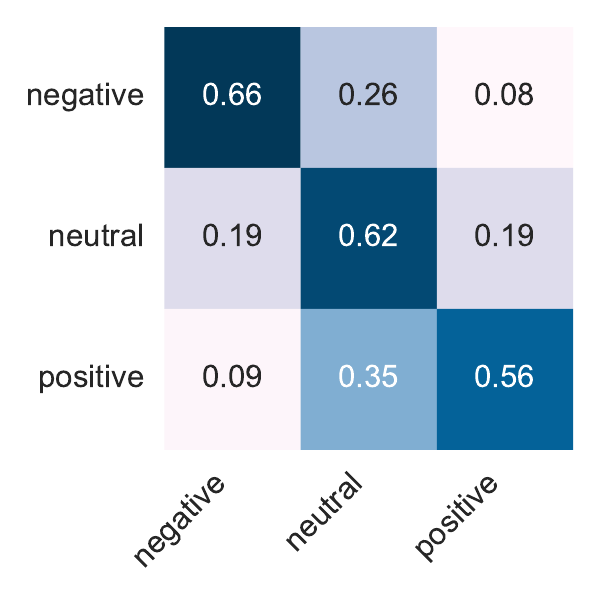}
        \subcaption{\small FCDEA}
    \end{minipage} \hfill
    \begin{minipage}[t]{0.16\linewidth}
        \centering
        \includegraphics[width=\linewidth]{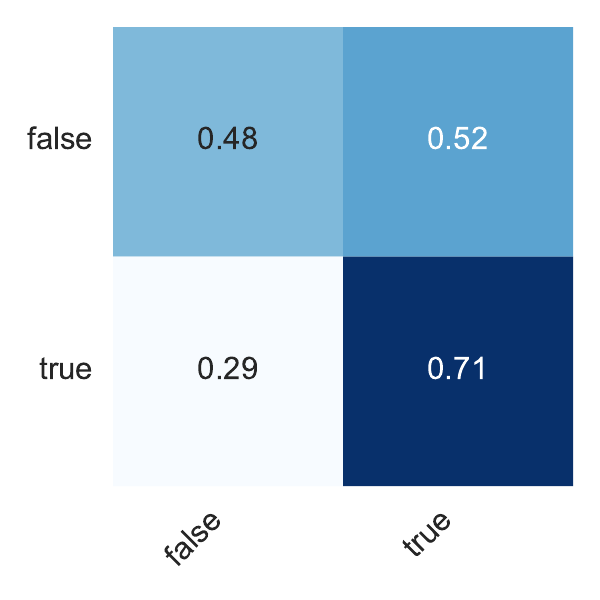}
        \subcaption{\small SD}
    \end{minipage} \\

    % ===== Row 3 =====
    \begin{minipage}[t]{0.16\linewidth}
        \centering
        \includegraphics[width=\linewidth]{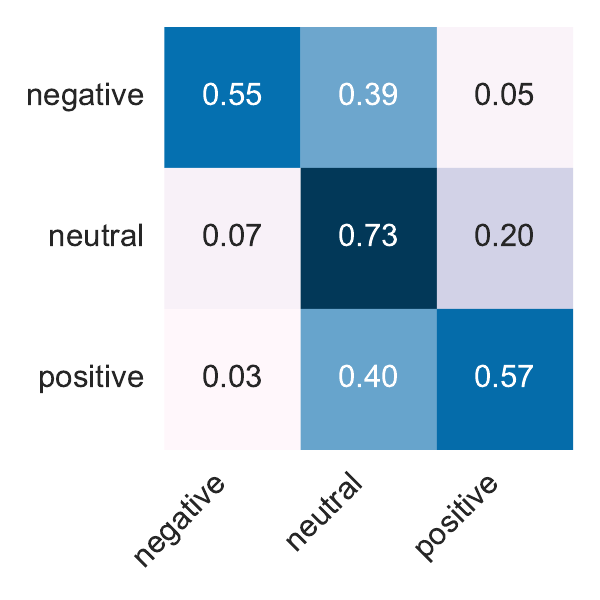}
        \subcaption{\small EIA}
    \end{minipage} \hfill
    \begin{minipage}[t]{0.20\linewidth}
        \centering
        \includegraphics[width=\linewidth]{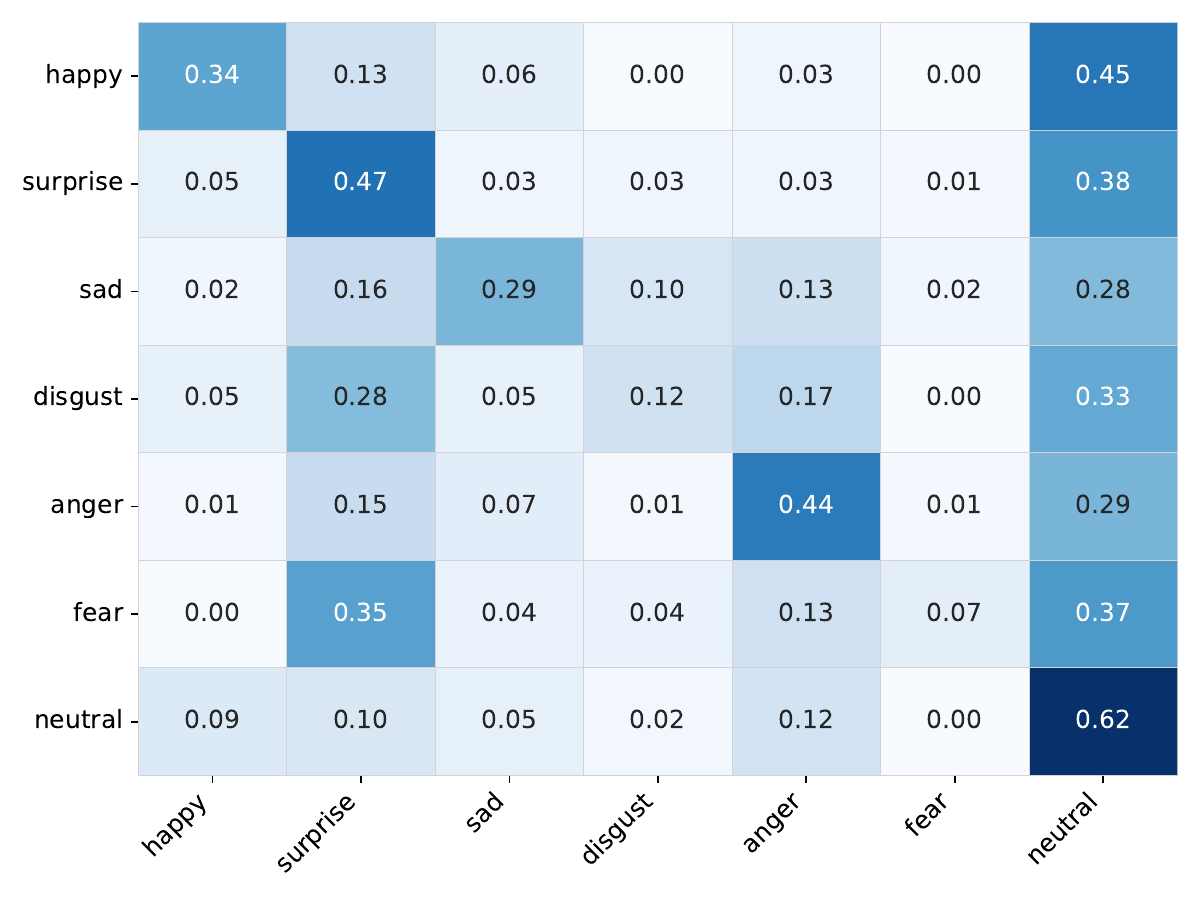}
        \subcaption{\small CEIA (emo.)}
    \end{minipage} \hfill
    \begin{minipage}[t]{0.20\linewidth}
        \centering
        \includegraphics[width=\linewidth]{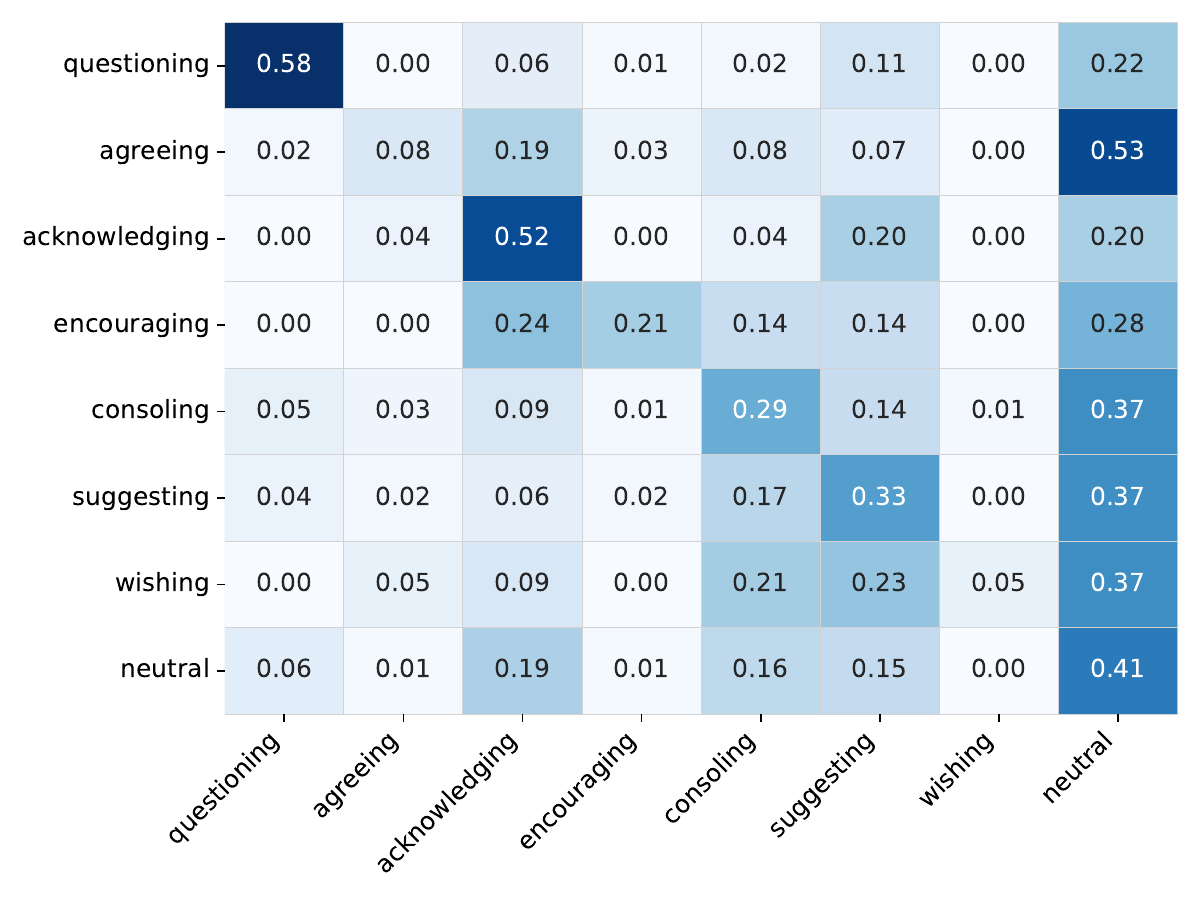}
        \subcaption{\small CEIA (int.)}
    \end{minipage} \hfill
    \begin{minipage}[t]{0.24\linewidth}
        \centering
        \includegraphics[width=\linewidth]{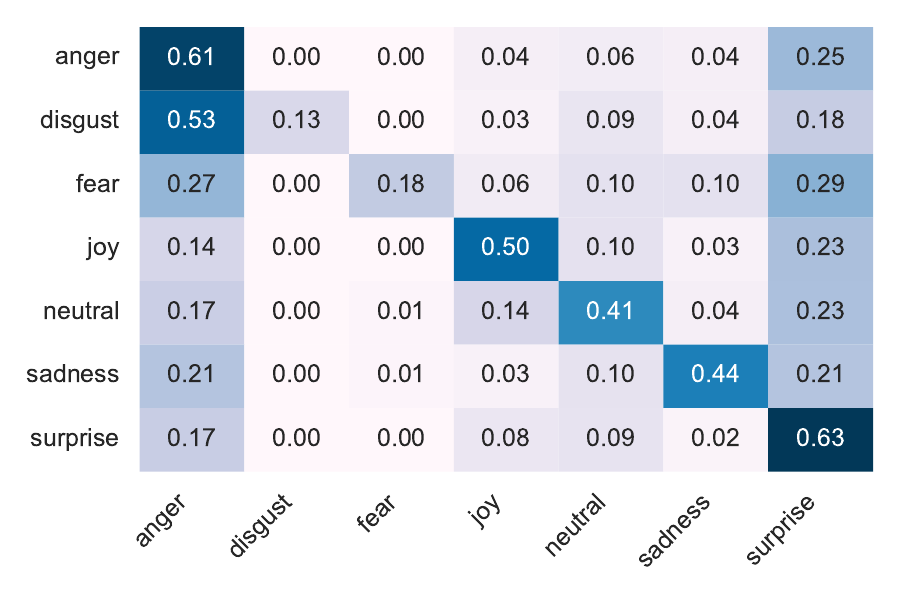}
        \subcaption{\small MPDER}
    \end{minipage} \hfill
    \begin{minipage}[t]{0.16\linewidth}
        \centering
        \includegraphics[width=\linewidth]{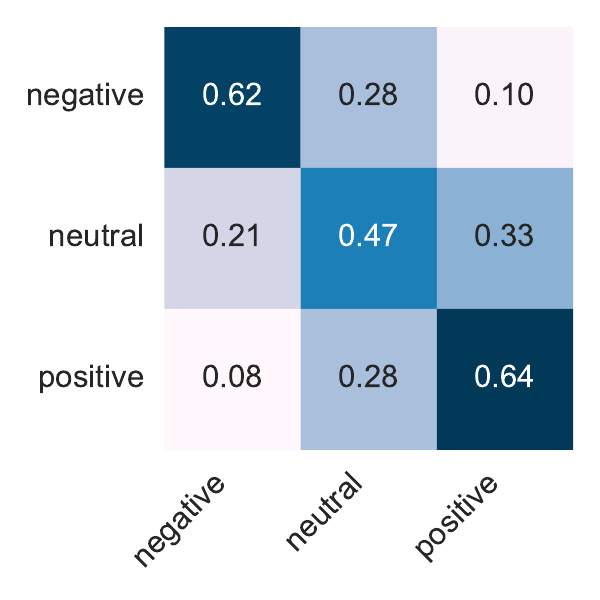}
        \subcaption{\small PEA}
    \end{minipage}

    \vspace{-2mm}
    \caption{\small Confusion matrices for InternVL2.5-8B on each evaluation scenario of $\ours$.}
    \label{fig:confusion-InternVL2.5-8B}
\end{figure*}

\begin{figure*}[!t]
    \centering
    
    % ===== Row 1 =====
    \begin{minipage}[t]{0.24\linewidth}
        \centering
        \includegraphics[width=\linewidth]{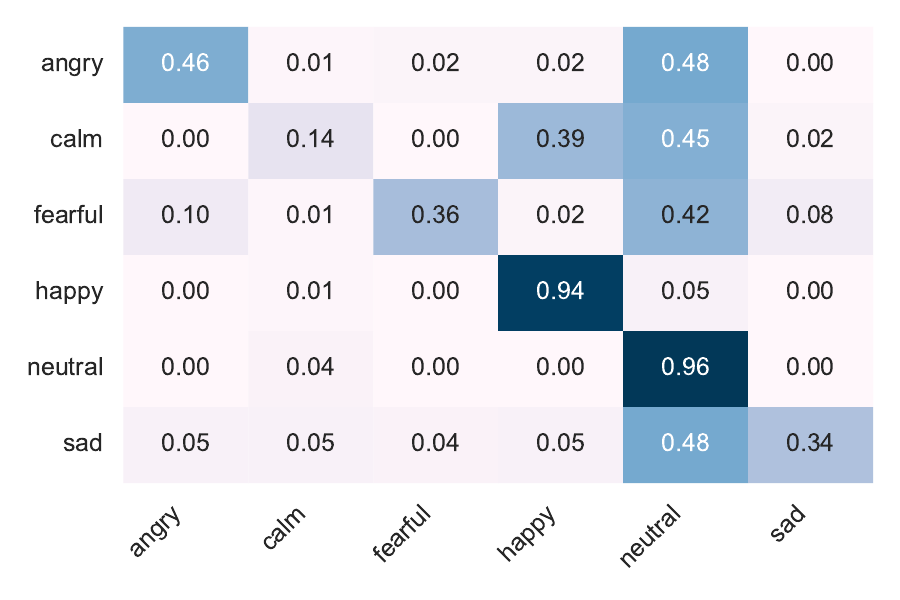}
        \subcaption{\small SOER}
    \end{minipage} \hfill
    \begin{minipage}[t]{0.24\linewidth}
        \centering
        \includegraphics[width=\linewidth]{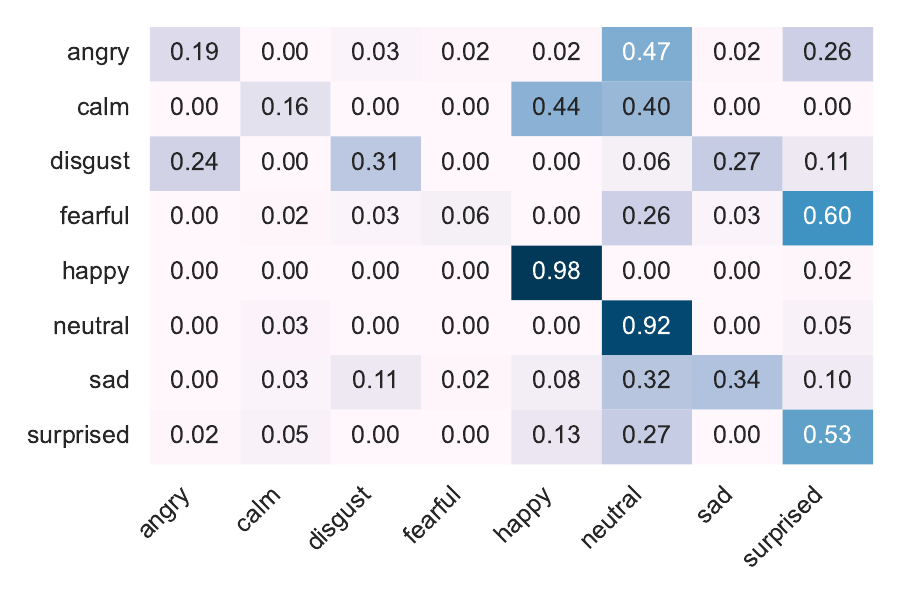}
        \subcaption{\small SPER}
    \end{minipage} \hfill
    \begin{minipage}[t]{0.16\linewidth}
        \centering
        \includegraphics[width=\linewidth]{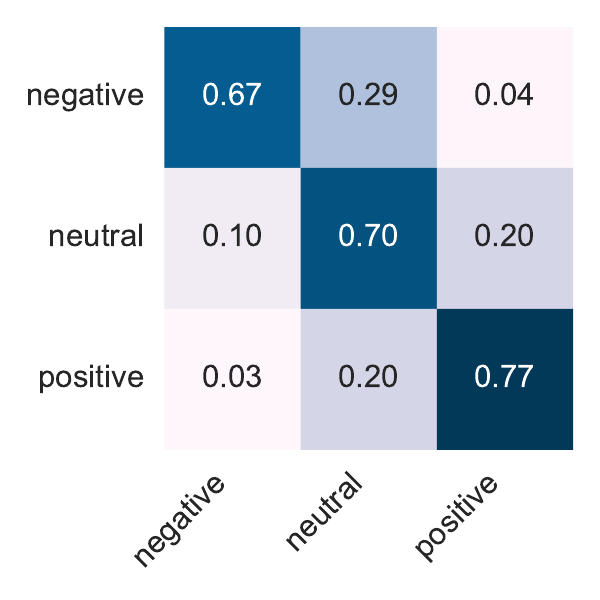}
        \subcaption{\small OSA}
    \end{minipage} \hfill
    \begin{minipage}[t]{0.16\linewidth}
        \centering
        \includegraphics[width=\linewidth]{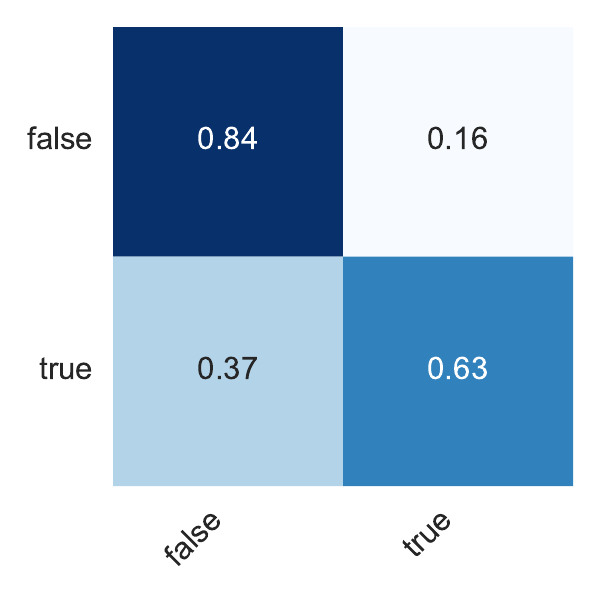}
        \subcaption{\small HU}
    \end{minipage} \\

    % ===== Row 2 =====
    \begin{minipage}[t]{0.24\linewidth}
        \centering
        \includegraphics[width=\linewidth]{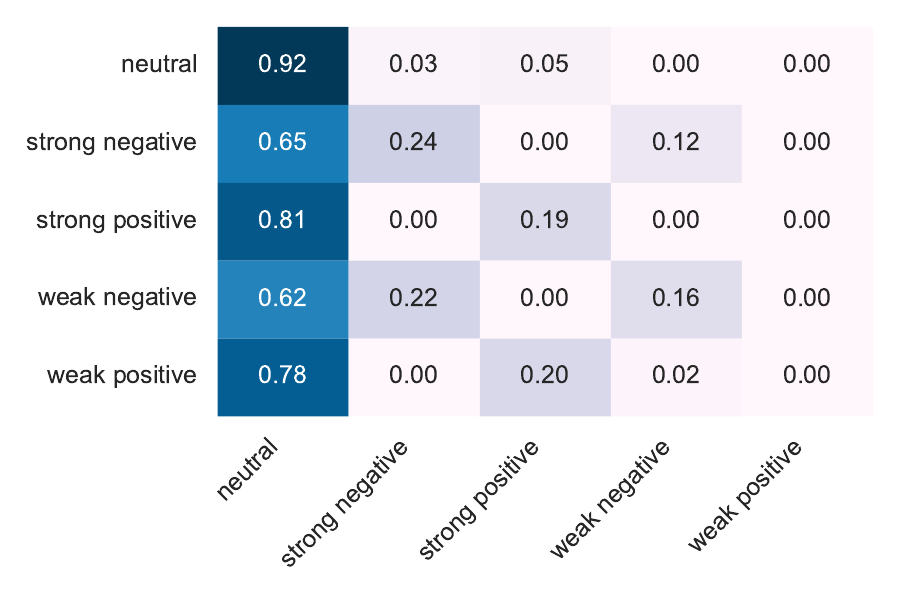}
        \subcaption{\small SCEA}
    \end{minipage} \hfill
    \begin{minipage}[t]{0.24\linewidth}
        \centering
        \includegraphics[width=\linewidth]{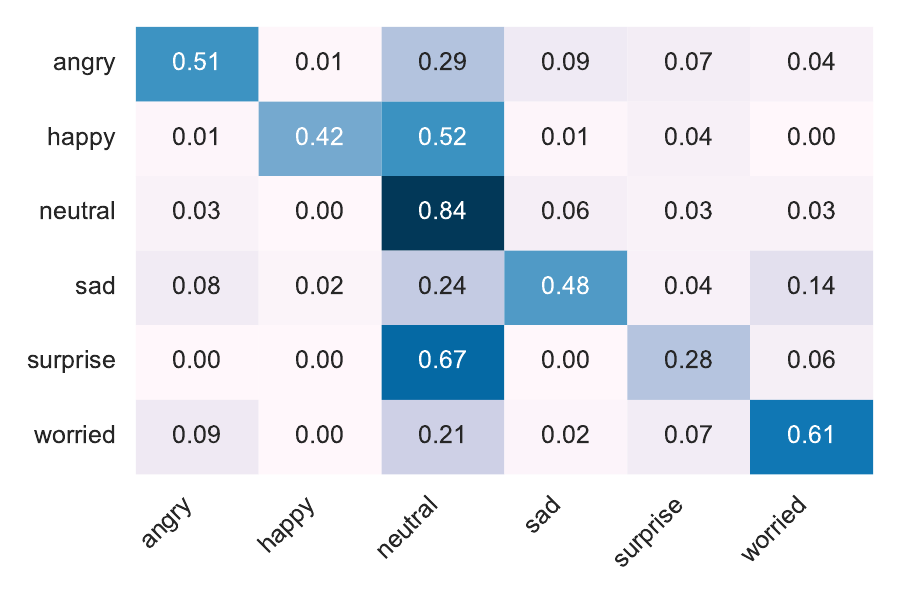}
        \subcaption{\small FGDEA}
    \end{minipage} \hfill
    \begin{minipage}[t]{0.16\linewidth}
        \centering
        \includegraphics[width=\linewidth]{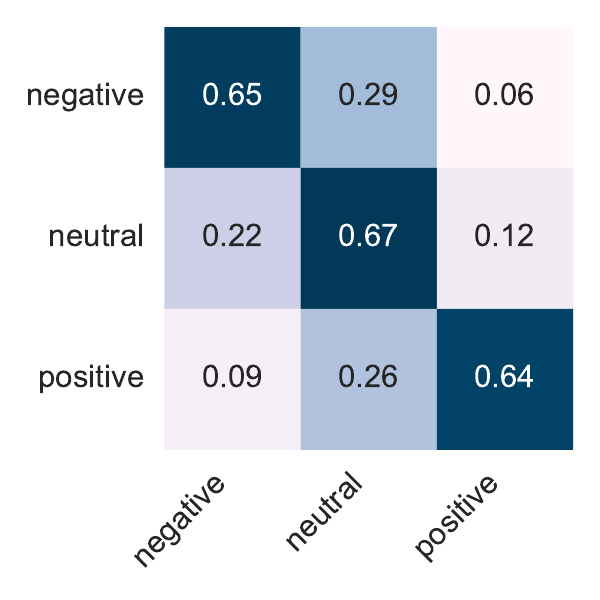}
        \subcaption{\small FCDEA}
    \end{minipage} \hfill
    \begin{minipage}[t]{0.16\linewidth}
        \centering
        \includegraphics[width=\linewidth]{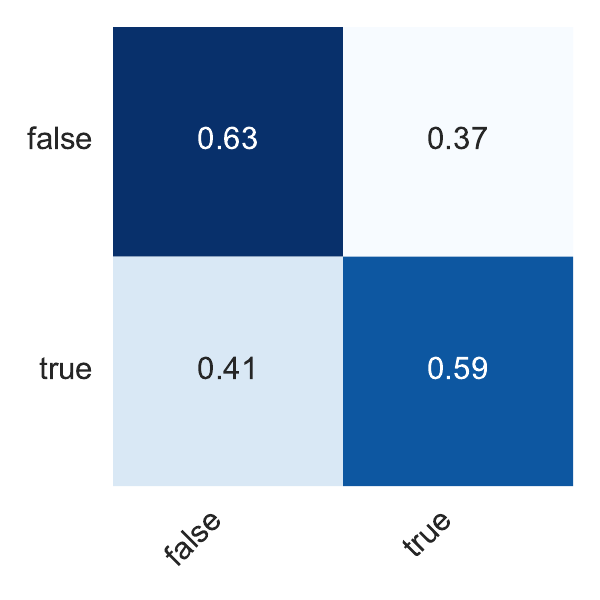}
        \subcaption{\small SD}
    \end{minipage} \\

    % ===== Row 3 =====
    \begin{minipage}[t]{0.16\linewidth}
        \centering
        \includegraphics[width=\linewidth]{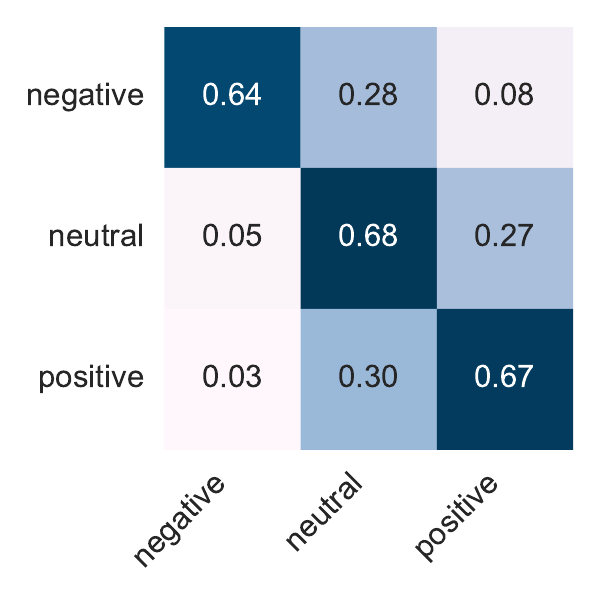}
        \subcaption{\small EIA}
    \end{minipage} \hfill
    \begin{minipage}[t]{0.20\linewidth}
        \centering
        \includegraphics[width=\linewidth]{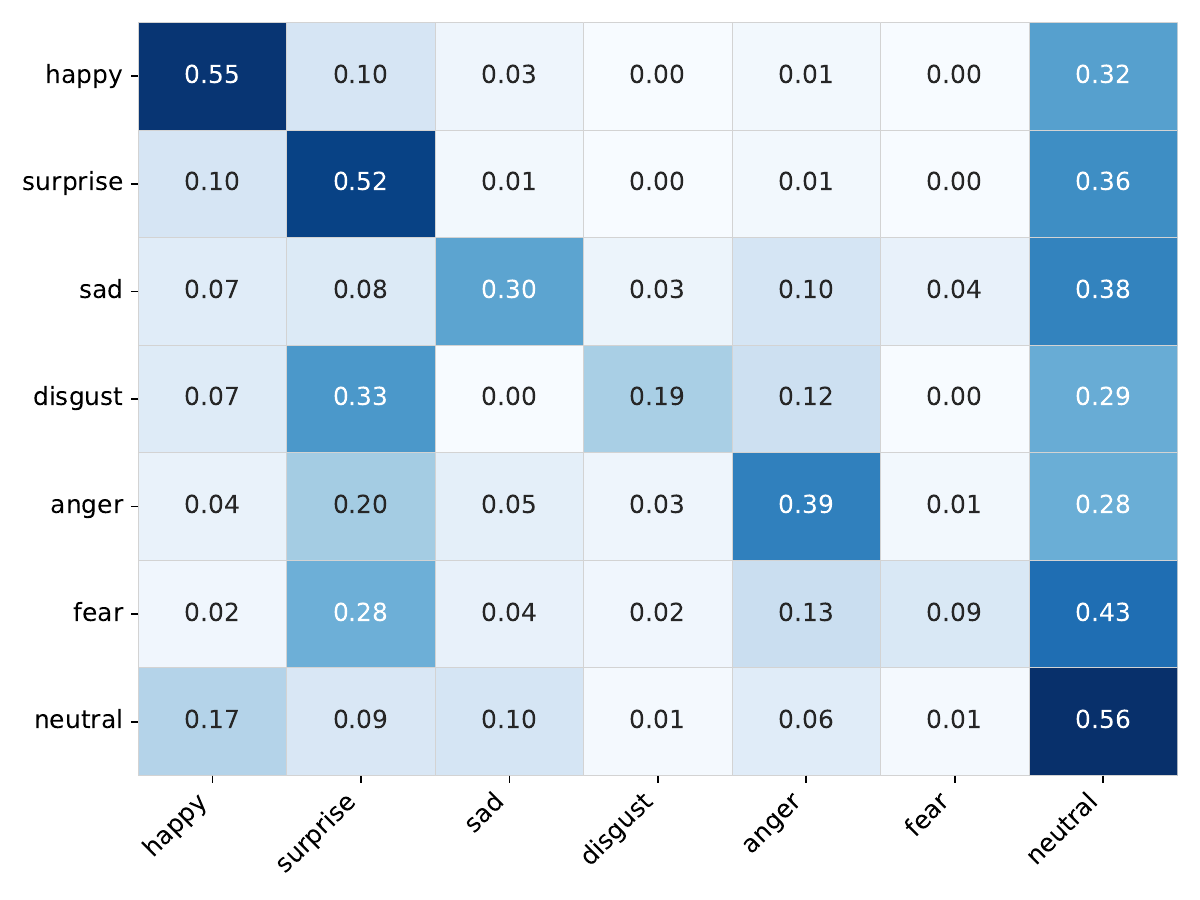}
        \subcaption{\small CEIA (emo.)}
    \end{minipage} \hfill
    \begin{minipage}[t]{0.20\linewidth}
        \centering
        \includegraphics[width=\linewidth]{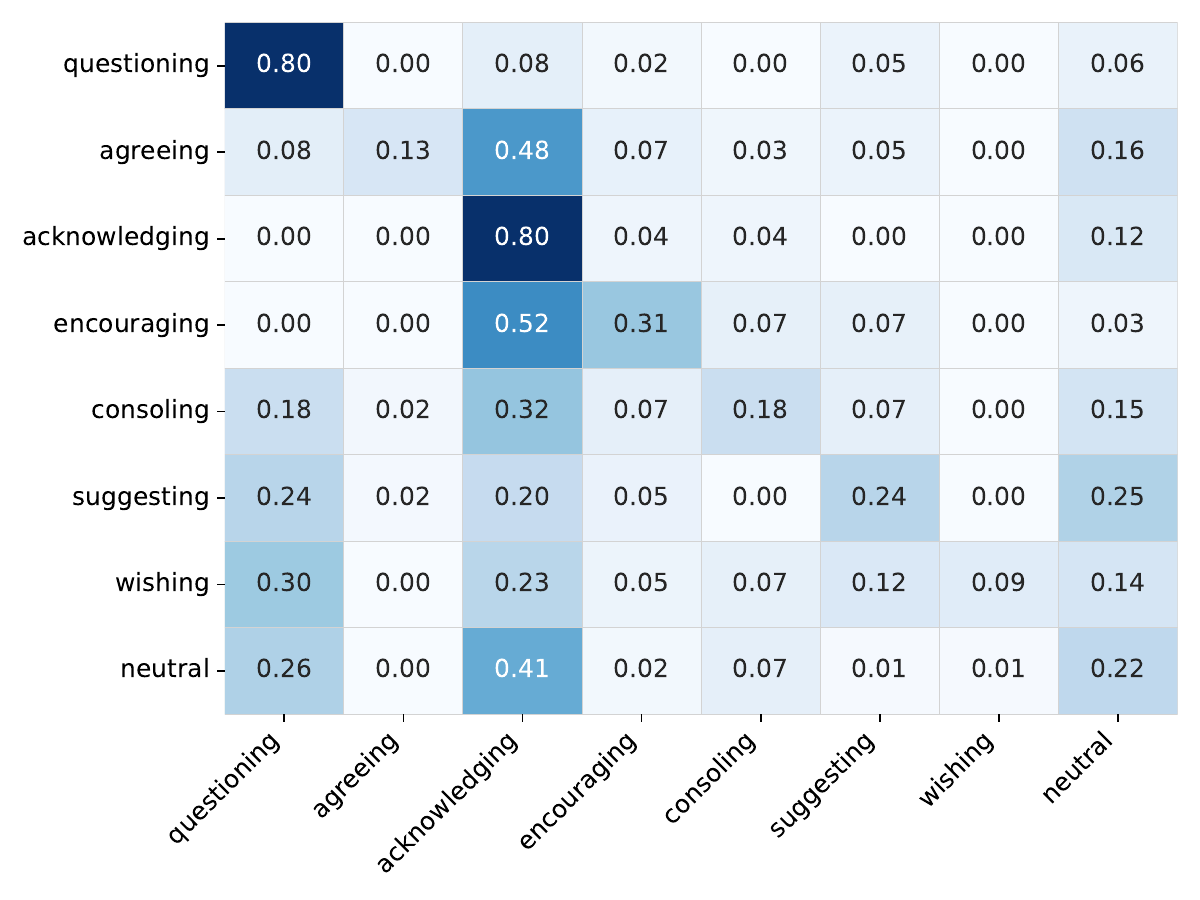}
        \subcaption{\small CEIA (int.)}
    \end{minipage} \hfill
    \begin{minipage}[t]{0.24\linewidth}
        \centering
        \includegraphics[width=\linewidth]{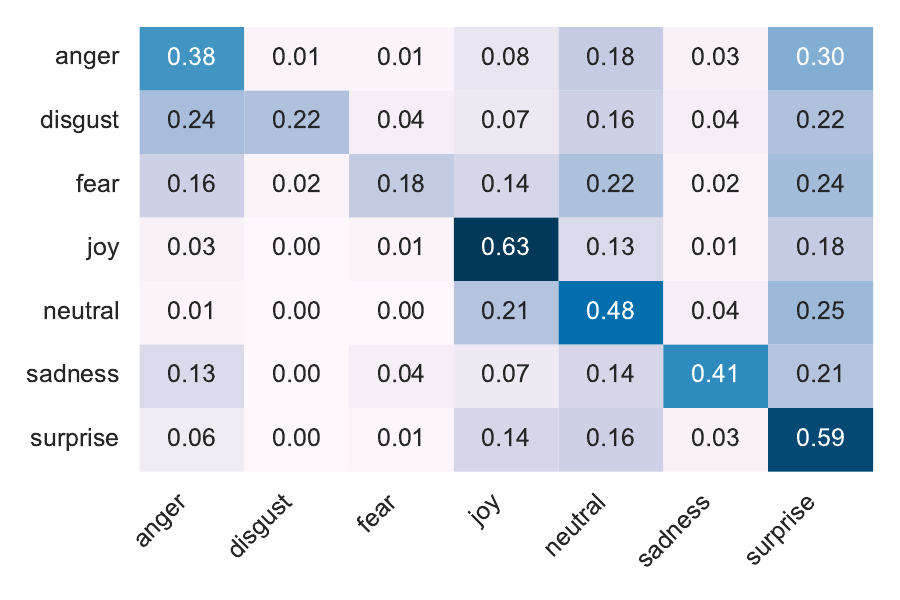}
        \subcaption{\small MPDER}
    \end{minipage} \hfill
    \begin{minipage}[t]{0.16\linewidth}
        \centering
        \includegraphics[width=\linewidth]{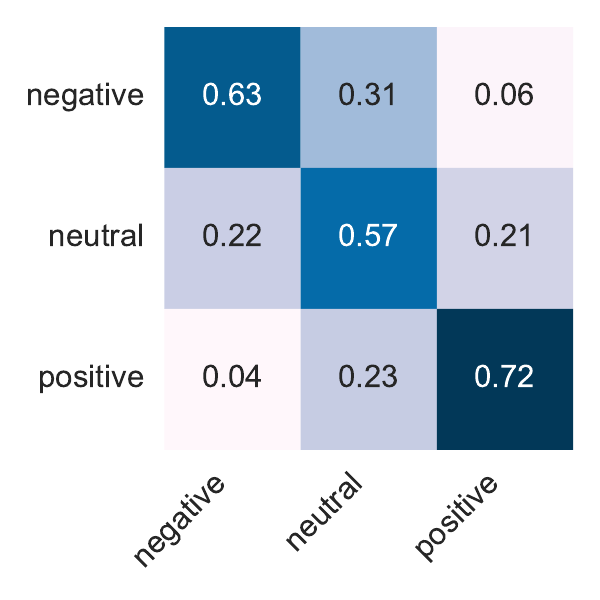}
        \subcaption{\small PEA}
    \end{minipage}
    
    \vspace{-2mm}
    \caption{\small Confusion matrices for InternVL2.5-38B on each evaluation scenario of $\ours$.}
    \label{fig:confusion-InternVL2.5-38B}
\end{figure*}

\begin{figure*}[!t]
    \centering
    
    % ===== Row 1 =====
    \begin{minipage}[t]{0.24\linewidth}
        \centering
        \includegraphics[width=\linewidth]{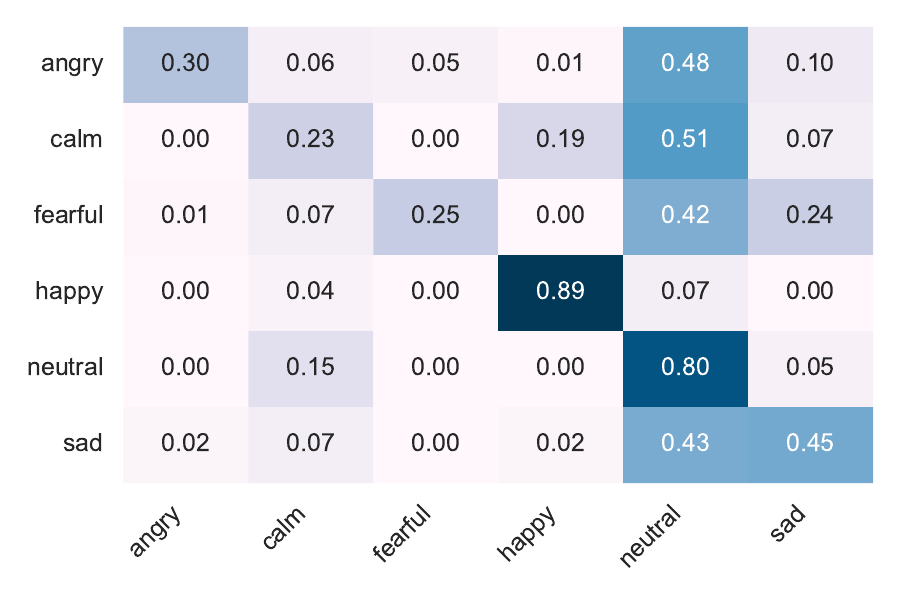}
        \subcaption{\small SOER}
    \end{minipage} \hfill
    \begin{minipage}[t]{0.24\linewidth}
        \centering
        \includegraphics[width=\linewidth]{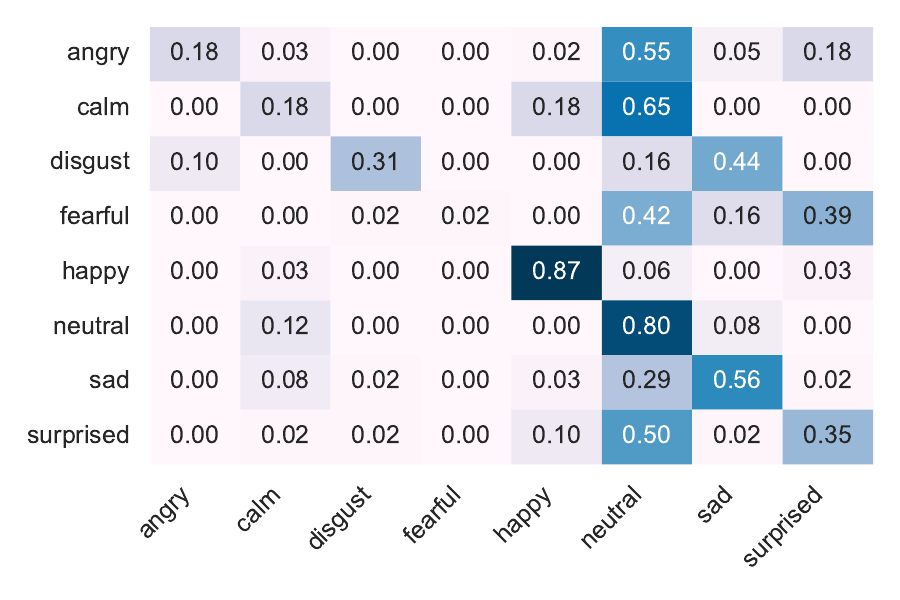}
        \subcaption{\small SPER}
    \end{minipage} \hfill
    \begin{minipage}[t]{0.16\linewidth}
        \centering
        \includegraphics[width=\linewidth]{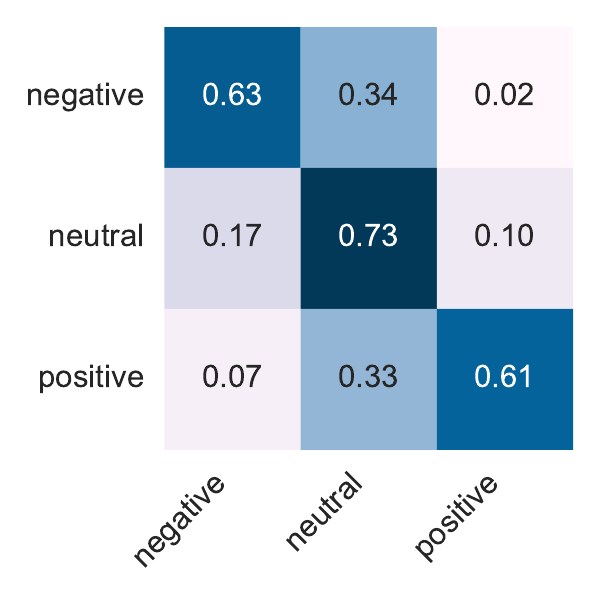}
        \subcaption{\small OSA}
    \end{minipage} \hfill
    \begin{minipage}[t]{0.16\linewidth}
        \centering
        \includegraphics[width=\linewidth]{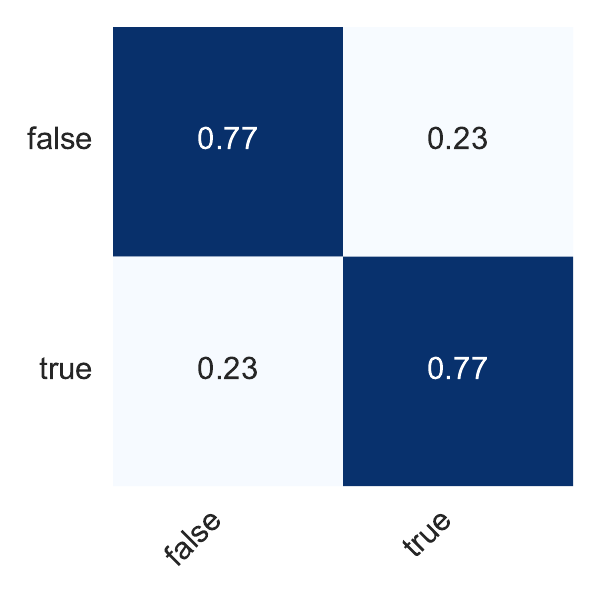}
        \subcaption{\small HU}
    \end{minipage} \\

    % ===== Row 2 =====
    \begin{minipage}[t]{0.24\linewidth}
        \centering
        \includegraphics[width=\linewidth]{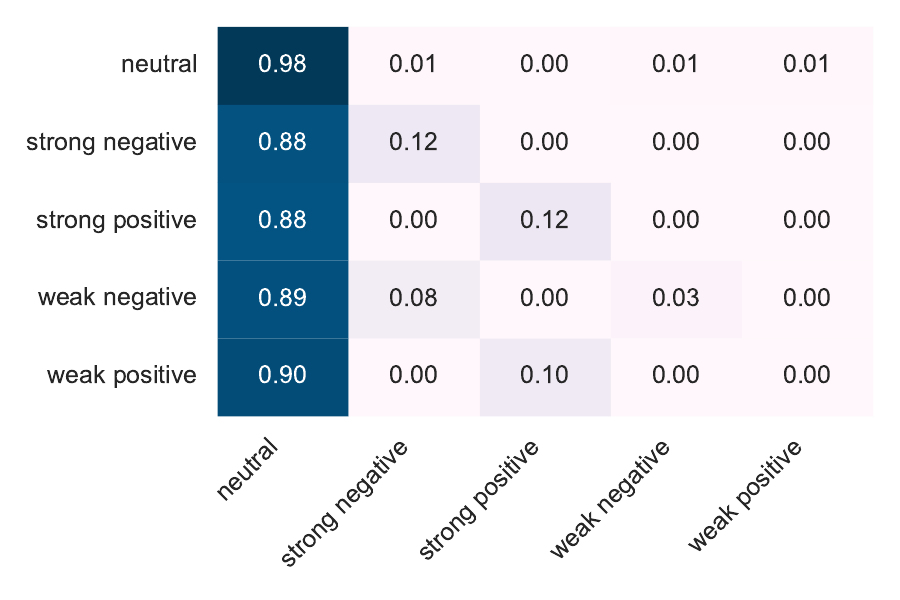}
        \subcaption{\small SCEA}
    \end{minipage} \hfill
    \begin{minipage}[t]{0.24\linewidth}
        \centering
        \includegraphics[width=\linewidth]{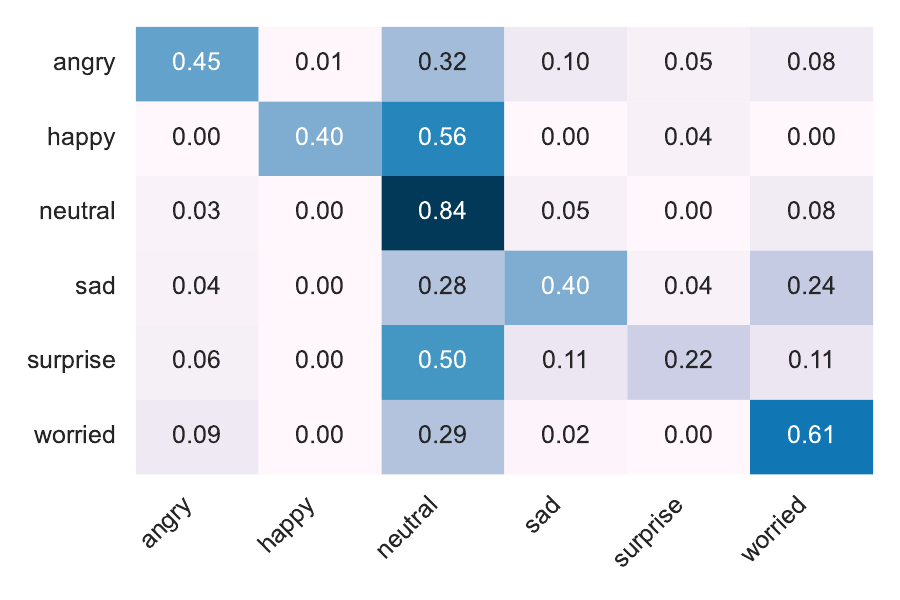}
        \subcaption{\small FGDEA}
    \end{minipage} \hfill
    \begin{minipage}[t]{0.16\linewidth}
        \centering
        \includegraphics[width=\linewidth]{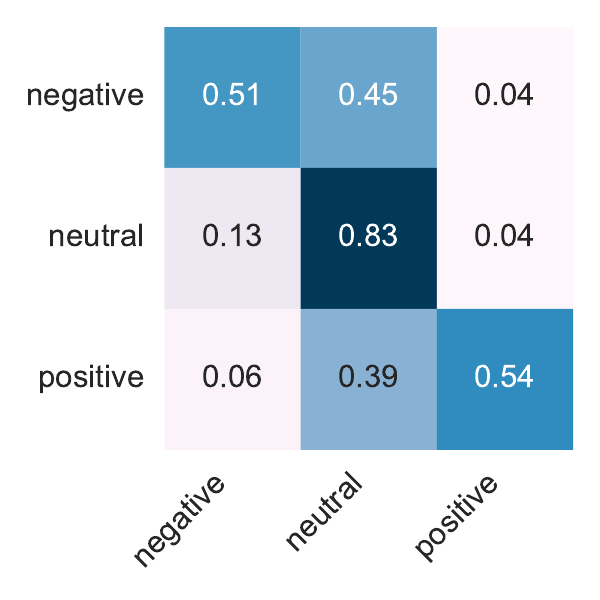}
        \subcaption{\small FCDEA}
    \end{minipage} \hfill
    \begin{minipage}[t]{0.16\linewidth}
        \centering
        \includegraphics[width=\linewidth]{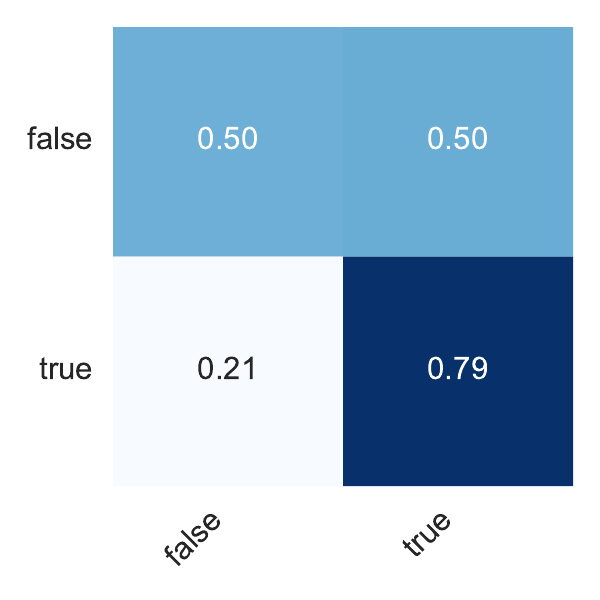}
        \subcaption{\small SD}
    \end{minipage} \\

    % ===== Row 3 =====
    \begin{minipage}[t]{0.16\linewidth}
        \centering
        \includegraphics[width=\linewidth]{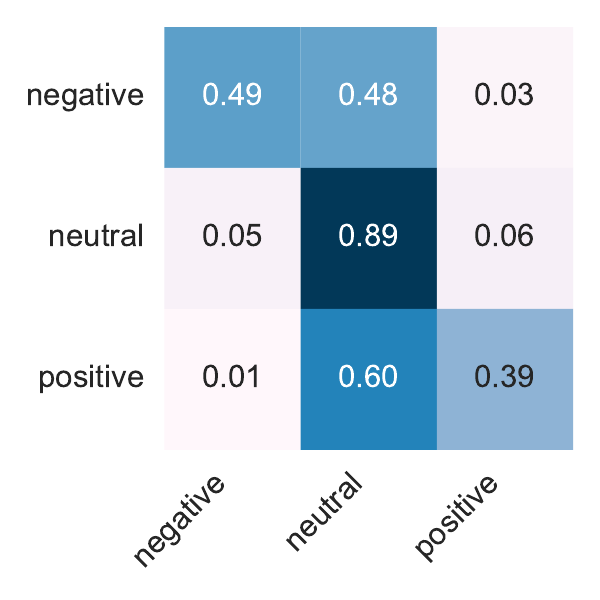}
        \subcaption{\small EIA}
    \end{minipage} \hfill
    \begin{minipage}[t]{0.20\linewidth}
        \centering
        \includegraphics[width=\linewidth]{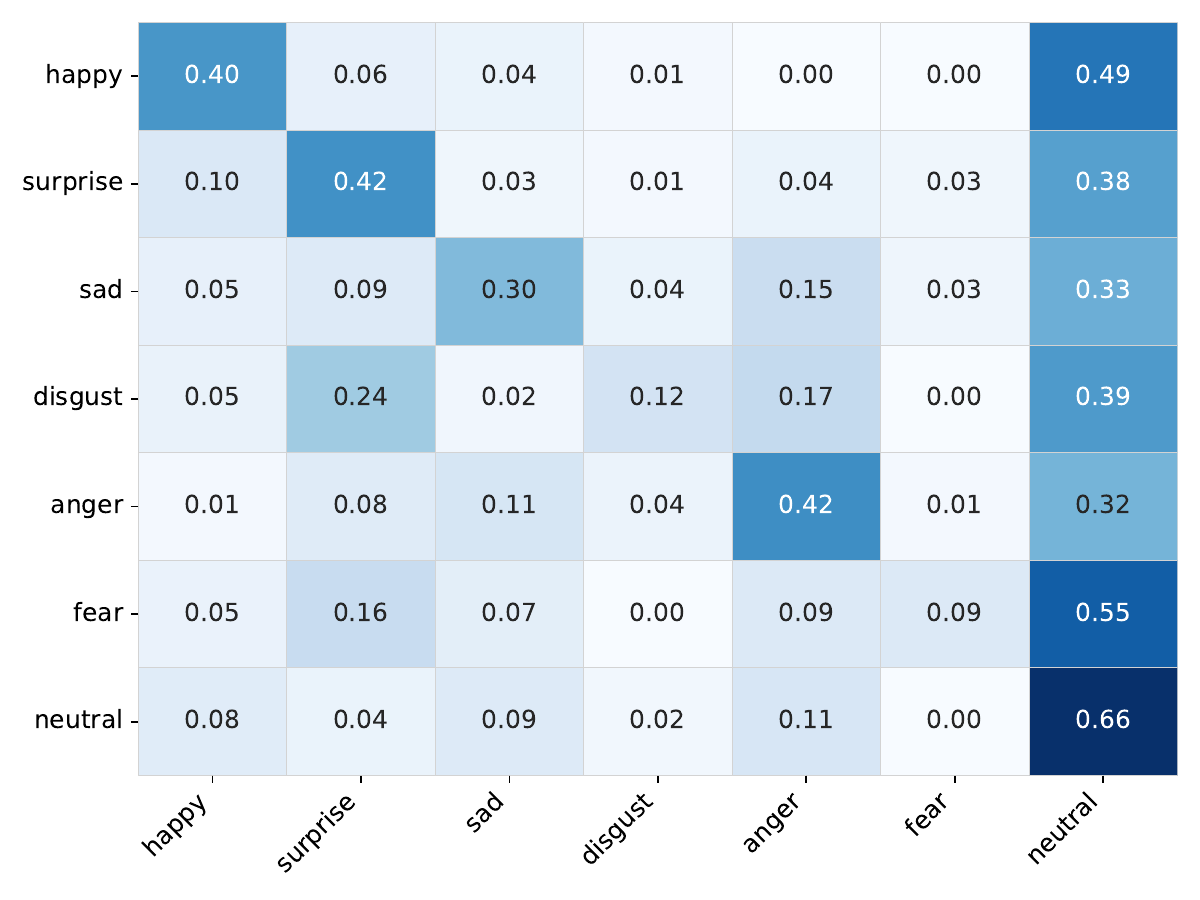}
        \subcaption{\small CEIA (emo.)}
    \end{minipage} \hfill
    \begin{minipage}[t]{0.20\linewidth}
        \centering
        \includegraphics[width=\linewidth]{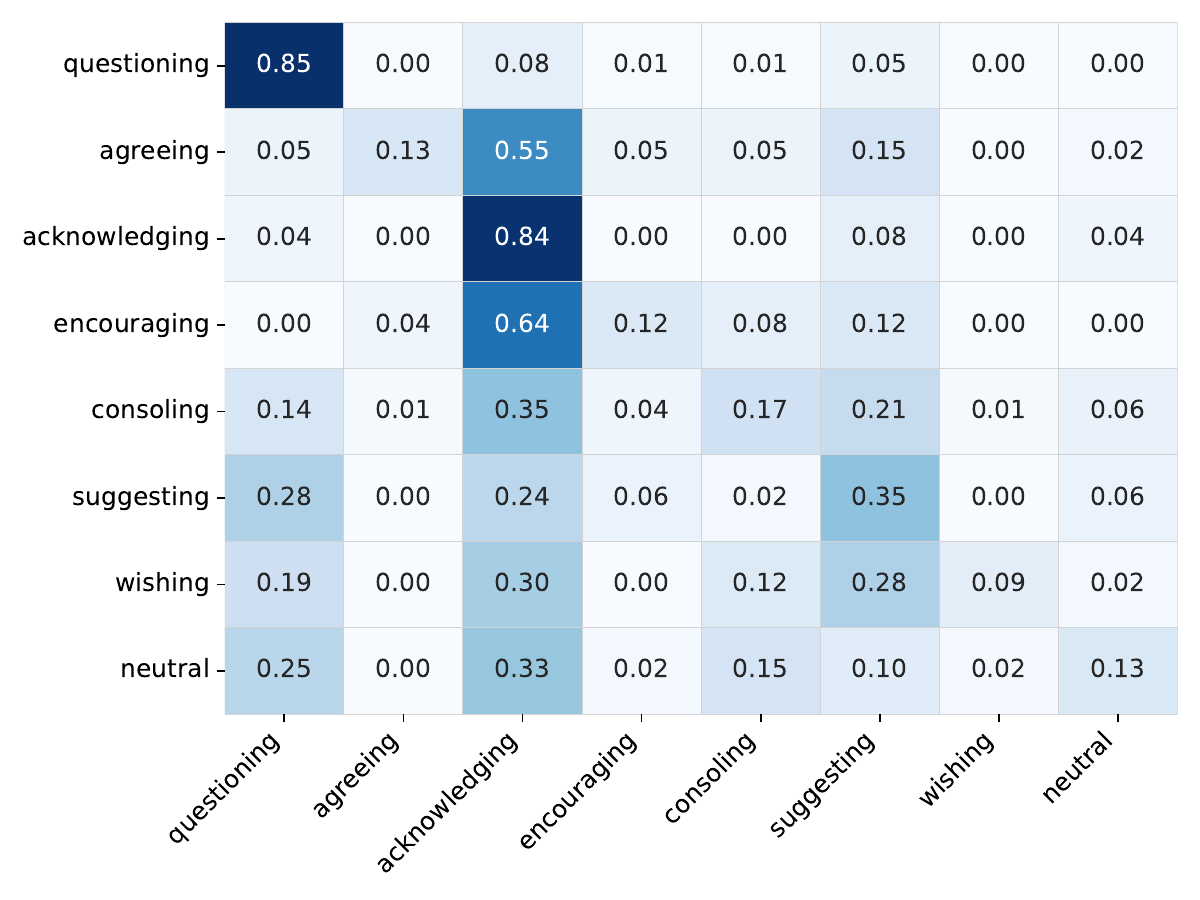}
        \subcaption{\small CEIA (int.)}
    \end{minipage} \hfill
    \begin{minipage}[t]{0.24\linewidth}
        \centering
        \includegraphics[width=\linewidth]{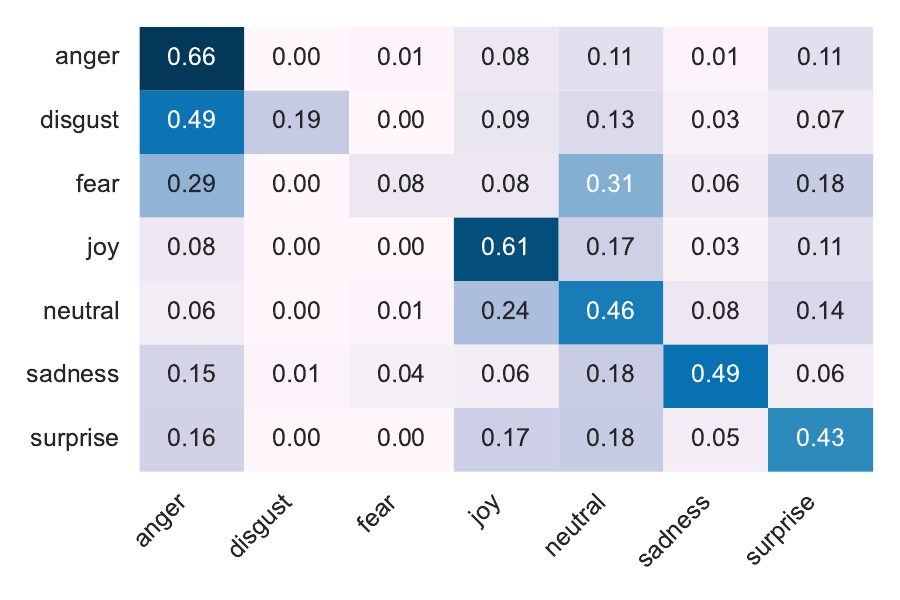}
        \subcaption{\small MPDER}
    \end{minipage} \hfill
    \begin{minipage}[t]{0.16\linewidth}
        \centering
        \includegraphics[width=\linewidth]{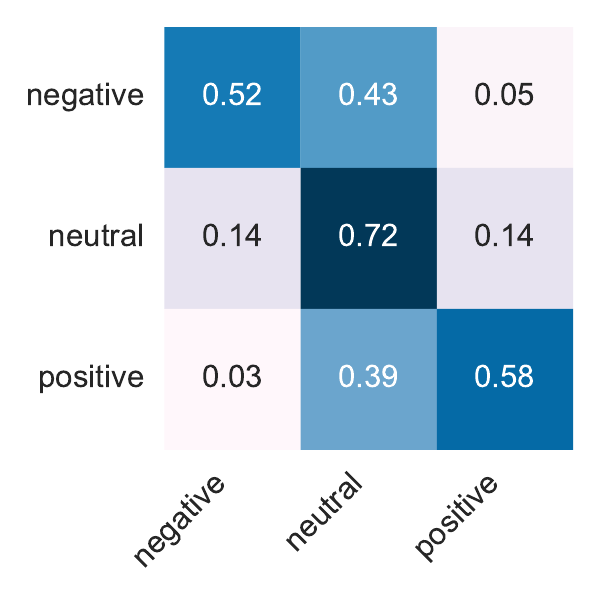}
        \subcaption{\small PEA}
    \end{minipage}

    \vspace{-3mm}
    \caption{\small Confusion matrices for InternVL2.5-78B on each evaluation scenario of $\ours$.}
    \label{fig:confusion-InternVL2.5-78B}
\end{figure*}

\begin{figure*}[!t]
    \centering
    
    % ===== Row 1 =====
    \begin{minipage}[t]{0.24\linewidth}
        \centering
        \includegraphics[width=\linewidth]{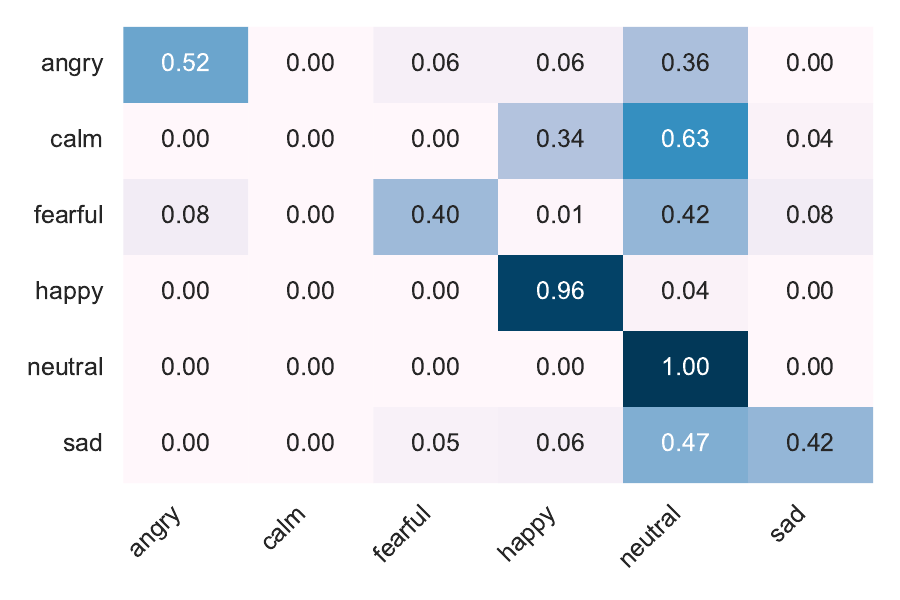}
        \subcaption{\small SOER}
    \end{minipage} \hfill
    \begin{minipage}[t]{0.24\linewidth}
        \centering
        \includegraphics[width=\linewidth]{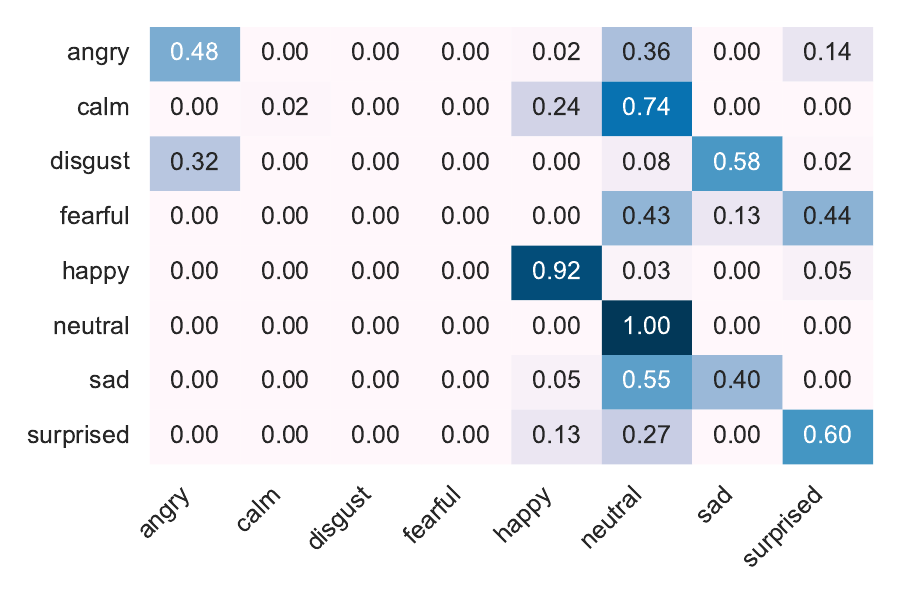}
        \subcaption{\small SPER}
    \end{minipage} \hfill
    \begin{minipage}[t]{0.16\linewidth}
        \centering
        \includegraphics[width=\linewidth]{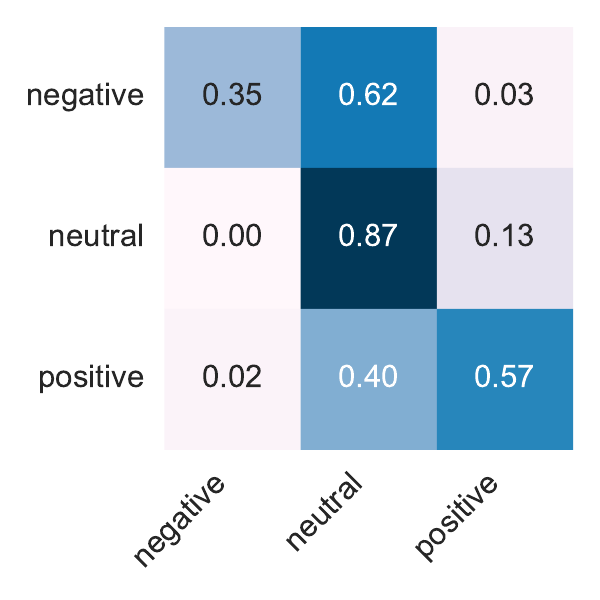}
        \subcaption{\small OSA}
    \end{minipage} \hfill
    \begin{minipage}[t]{0.16\linewidth}
        \centering
        \includegraphics[width=\linewidth]{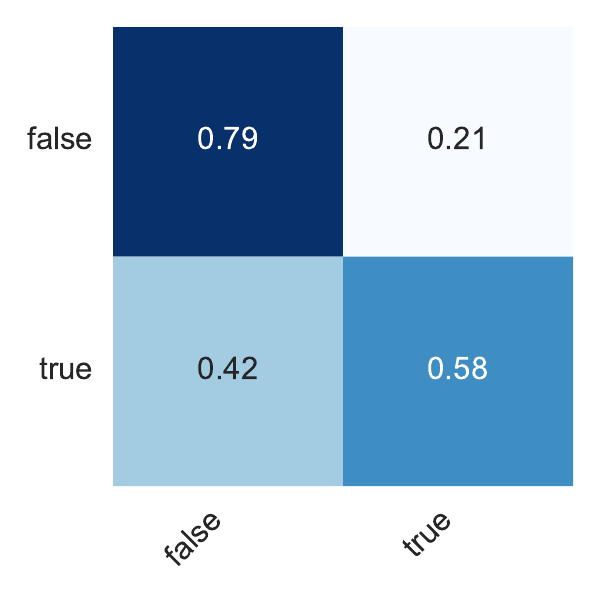}
        \subcaption{\small HU}
    \end{minipage} \\

    % ===== Row 2 =====
    \begin{minipage}[t]{0.24\linewidth}
        \centering
        \includegraphics[width=\linewidth]{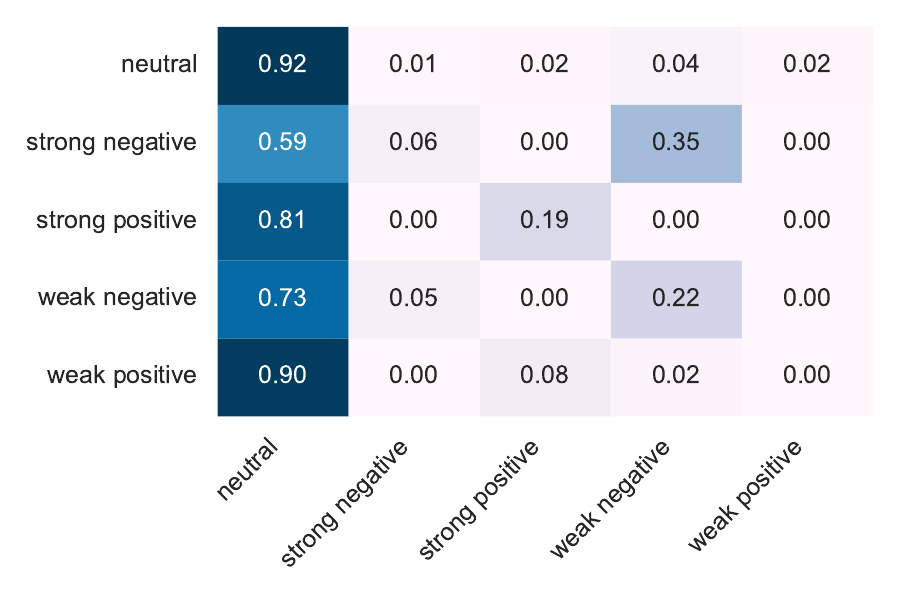}
        \subcaption{\small SCEA}
    \end{minipage} \hfill
    \begin{minipage}[t]{0.24\linewidth}
        \centering
        \includegraphics[width=\linewidth]{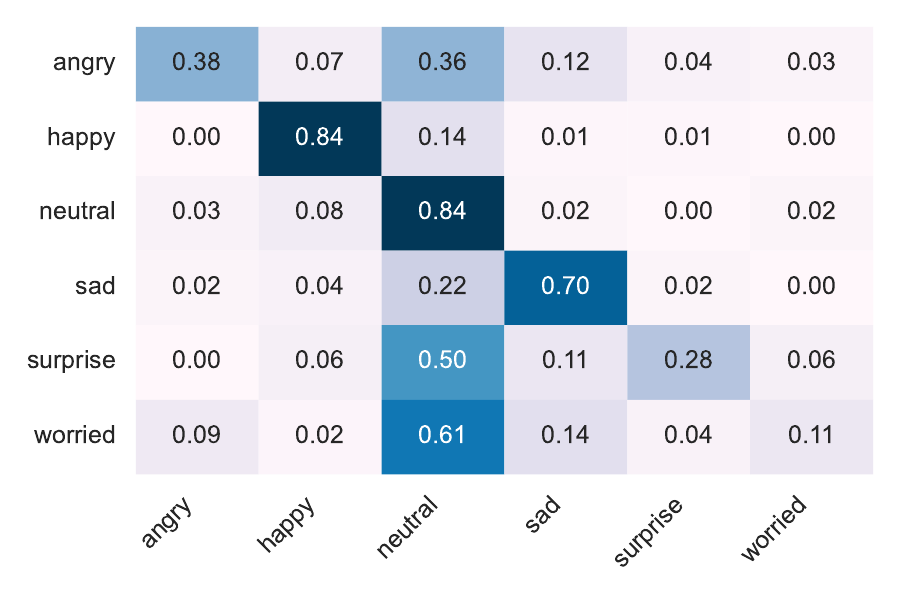}
        \subcaption{\small FGDEA}
    \end{minipage} \hfill
    \begin{minipage}[t]{0.16\linewidth}
        \centering
        \includegraphics[width=\linewidth]{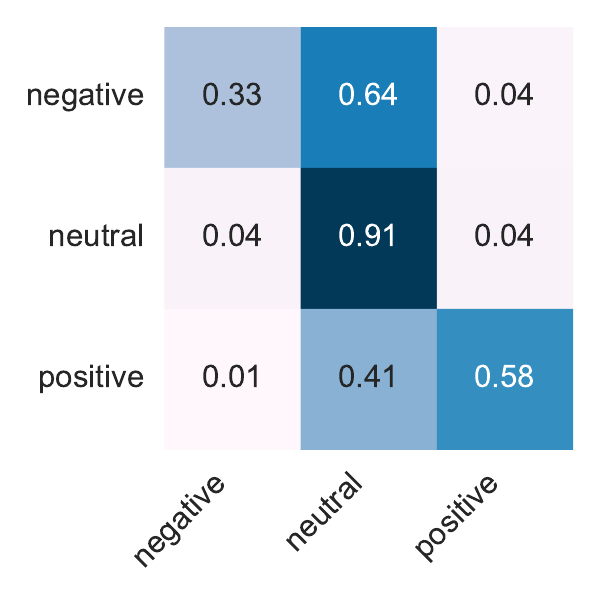}
        \subcaption{\small FCDEA}
    \end{minipage} \hfill
    \begin{minipage}[t]{0.16\linewidth}
        \centering
        \includegraphics[width=\linewidth]{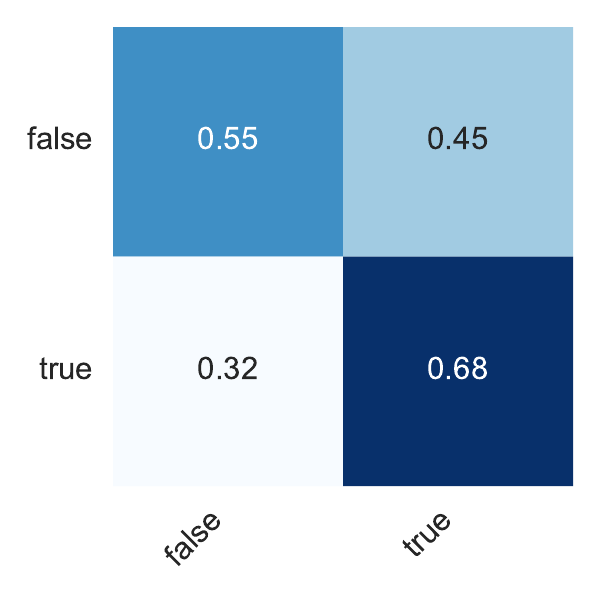}
        \subcaption{\small SD}
    \end{minipage} \\

    % ===== Row 3 =====
    \begin{minipage}[t]{0.16\linewidth}
        \centering
        \includegraphics[width=\linewidth]{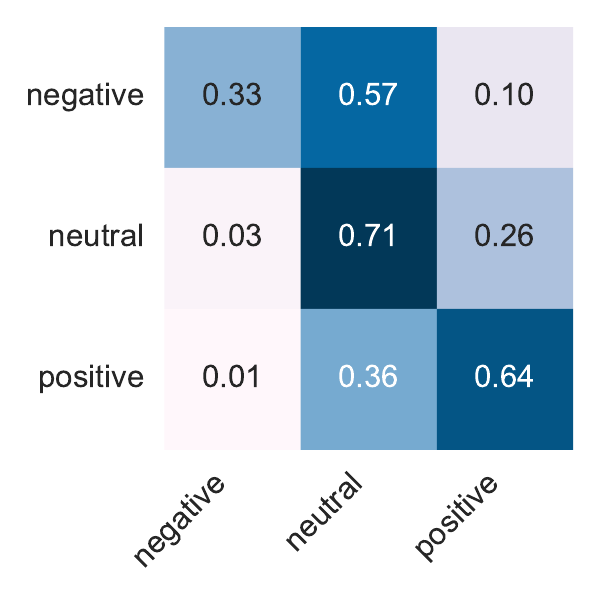}
        \subcaption{\small EIA}
    \end{minipage} \hfill
    \begin{minipage}[t]{0.20\linewidth}
        \centering
        \includegraphics[width=\linewidth]{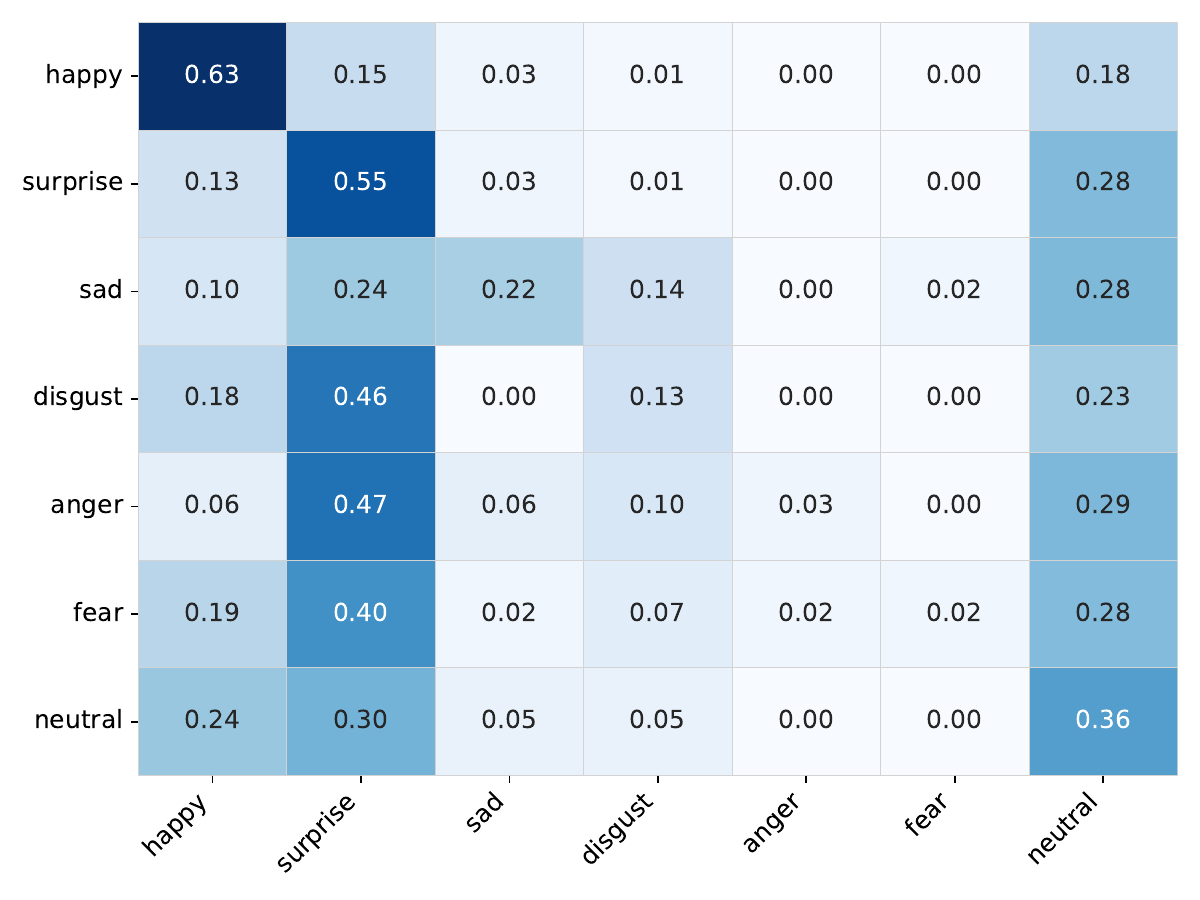}
        \subcaption{\small CEIA (emo.)}
    \end{minipage} \hfill
    \begin{minipage}[t]{0.20\linewidth}
        \centering
        \includegraphics[width=\linewidth]{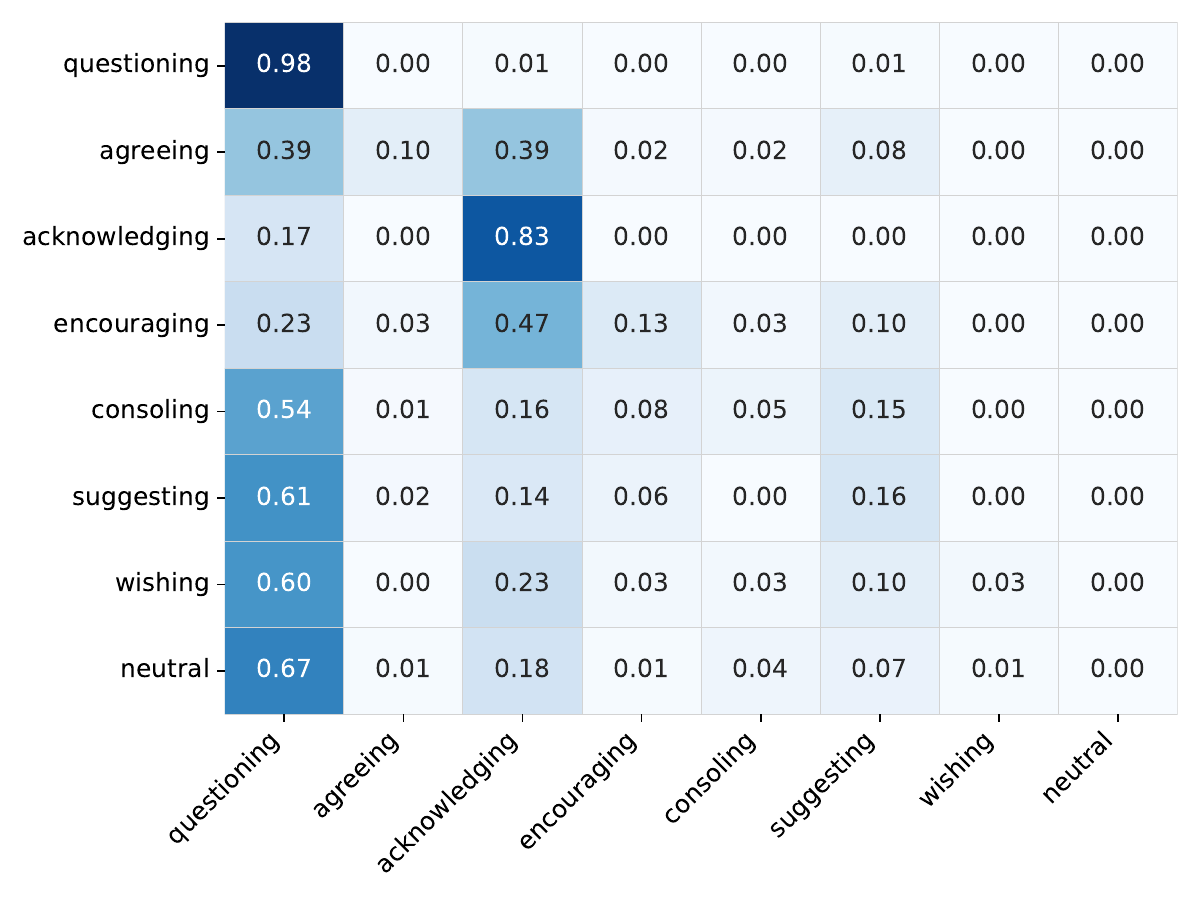}
        \subcaption{\small CEIA (int.)}
    \end{minipage} \hfill
    \begin{minipage}[t]{0.24\linewidth}
        \centering
        \includegraphics[width=\linewidth]{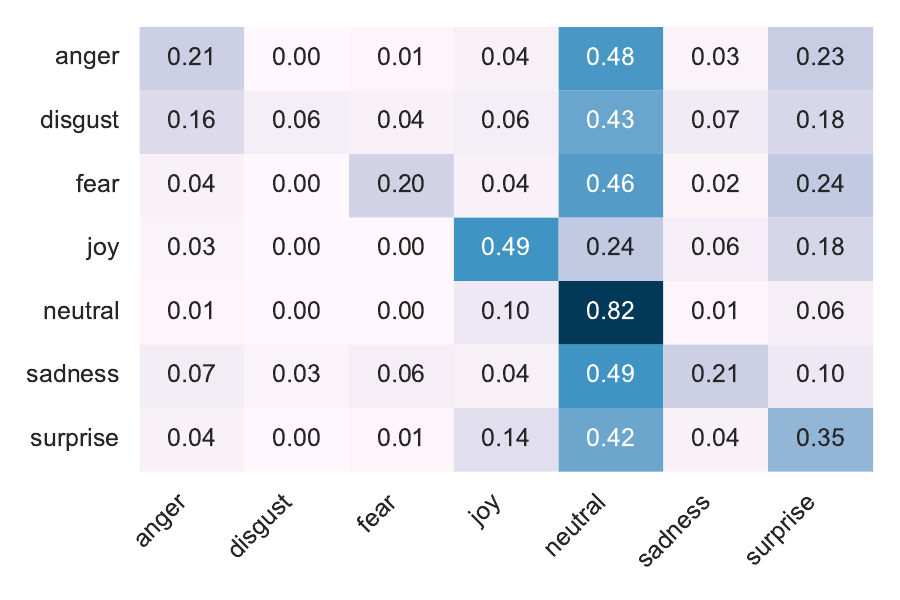}
        \subcaption{\small MPDER}
    \end{minipage} \hfill
    \begin{minipage}[t]{0.16\linewidth}
        \centering
        \includegraphics[width=\linewidth]{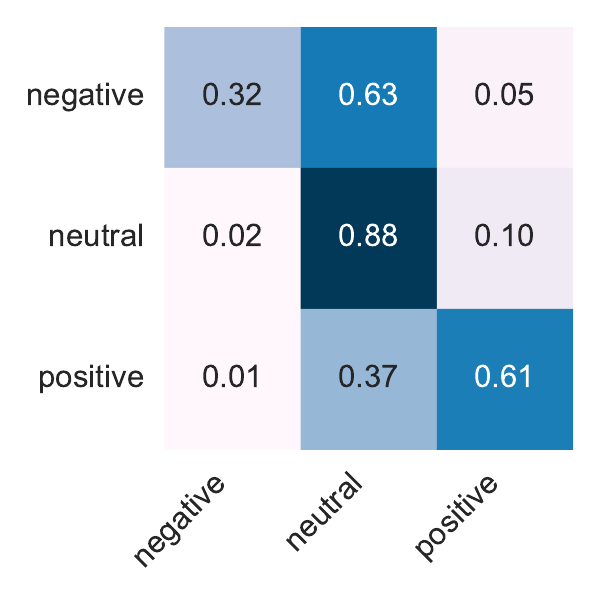}
        \subcaption{\small PEA}
    \end{minipage}

    \vspace{-3mm}
    \caption{\small Confusion matrices for InternVideo2-Chat-8B on each evaluation scenario of $\ours$.}
    \label{fig:confusion-InternVideo2-Chat-8B}
\end{figure*}

\begin{figure*}[!t]
    \centering
    
    % ===== Row 1 =====
    \begin{minipage}[t]{0.24\linewidth}
        \centering
        \includegraphics[width=\linewidth]{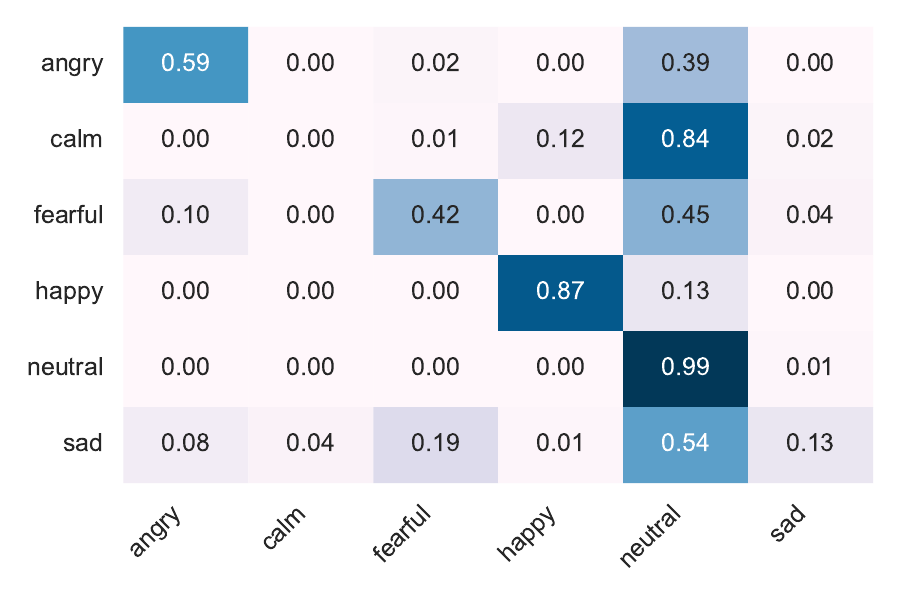}
        \subcaption{\small SOER}
    \end{minipage} \hfill
    \begin{minipage}[t]{0.24\linewidth}
        \centering
        \includegraphics[width=\linewidth]{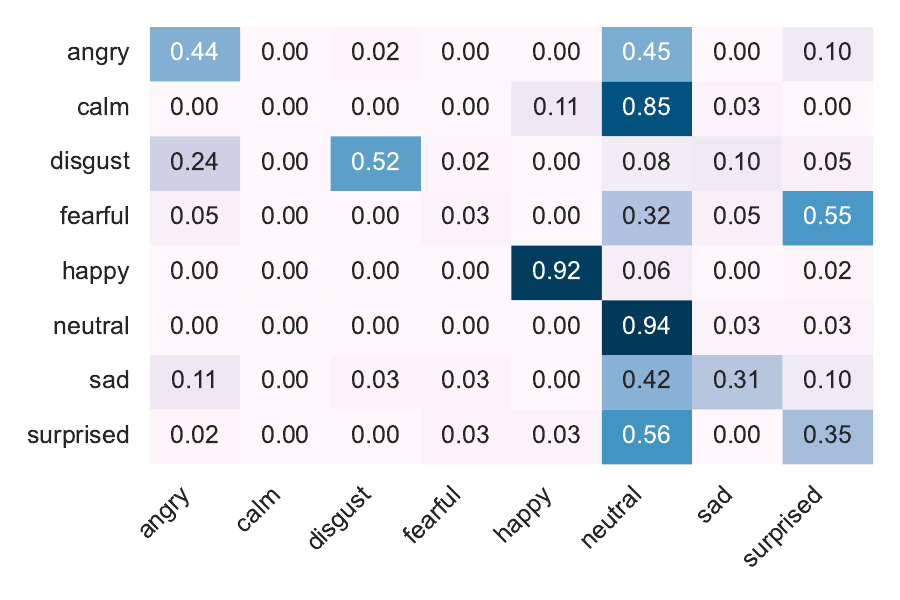}
        \subcaption{\small SPER}
    \end{minipage} \hfill
    \begin{minipage}[t]{0.16\linewidth}
        \centering
        \includegraphics[width=\linewidth]{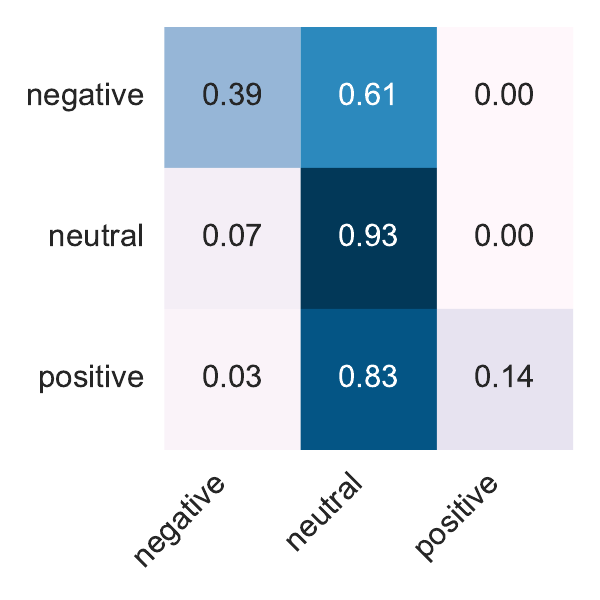}
        \subcaption{\small OSA}
    \end{minipage} \hfill
    \begin{minipage}[t]{0.16\linewidth}
        \centering
        \includegraphics[width=\linewidth]{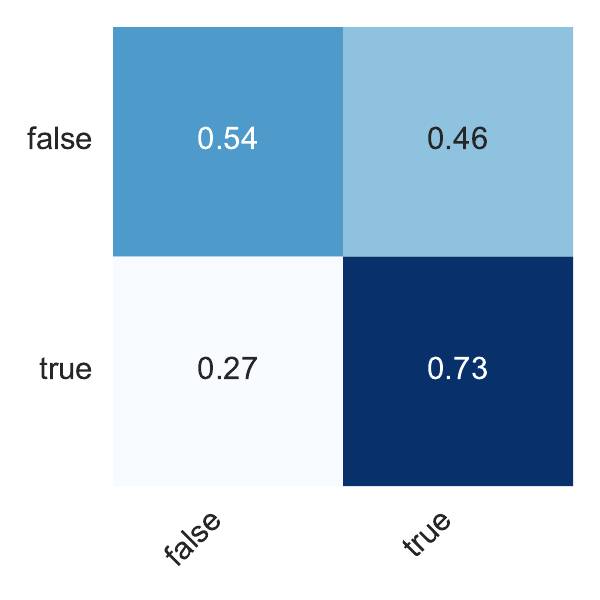}
        \subcaption{\small HU}
    \end{minipage} \\

    % ===== Row 2 =====
    \begin{minipage}[t]{0.24\linewidth}
        \centering
        \includegraphics[width=\linewidth]{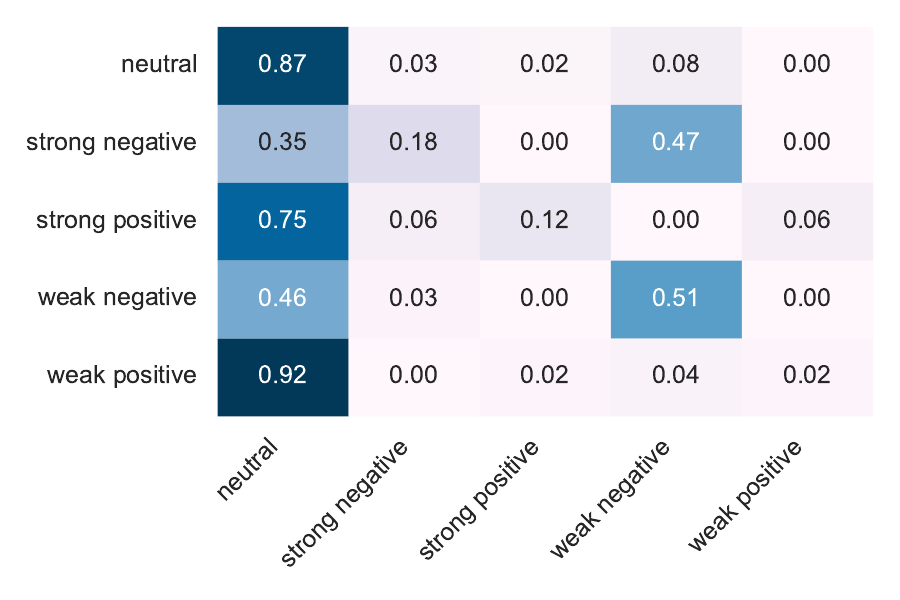}
        \subcaption{\small SCEA}
    \end{minipage} \hfill
    \begin{minipage}[t]{0.24\linewidth}
        \centering
        \includegraphics[width=\linewidth]{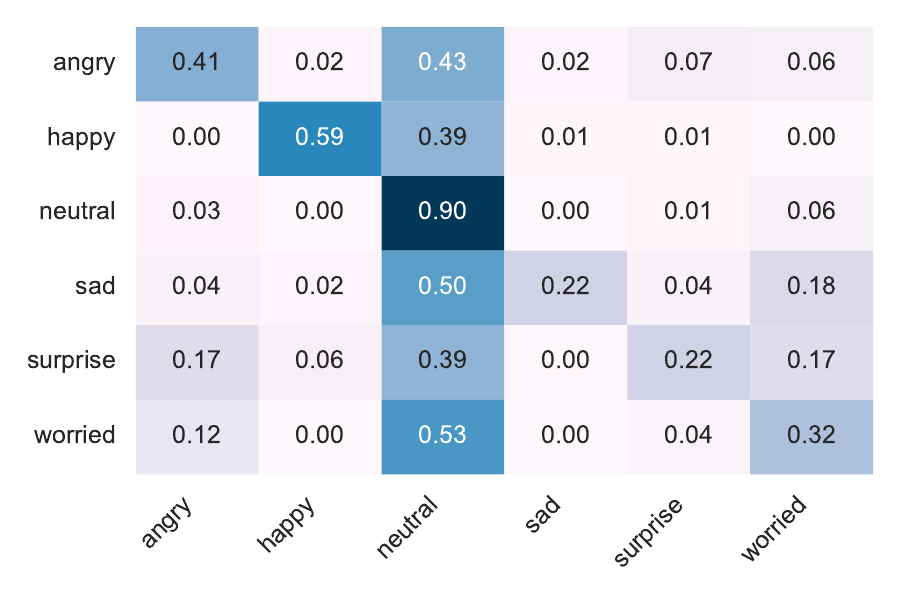}
        \subcaption{\small FGDEA}
    \end{minipage} \hfill
    \begin{minipage}[t]{0.16\linewidth}
        \centering
        \includegraphics[width=\linewidth]{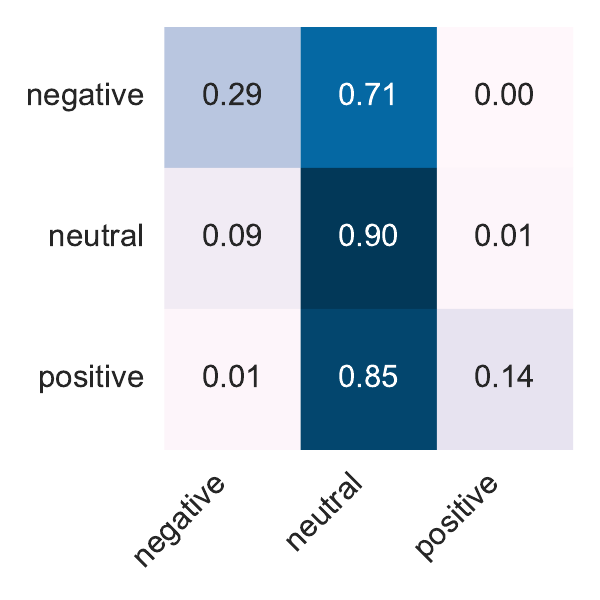}
        \subcaption{\small FCDEA}
    \end{minipage} \hfill
    \begin{minipage}[t]{0.16\linewidth}
        \centering
        \includegraphics[width=\linewidth]{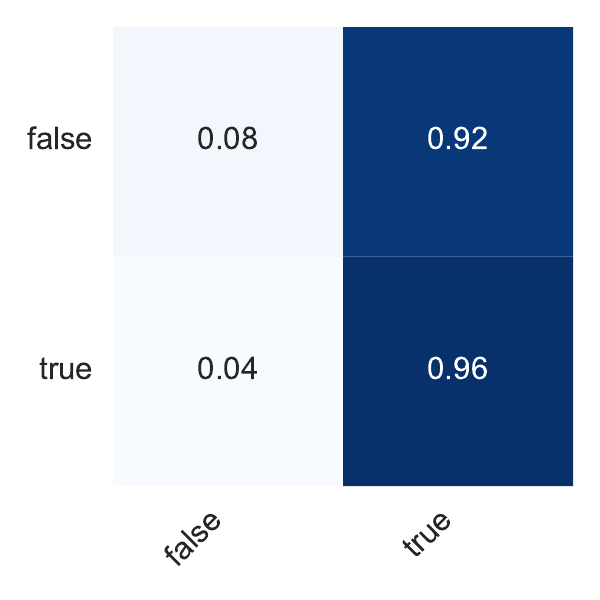}
        \subcaption{\small SD}
    \end{minipage} \\

    % ===== Row 3 =====
    \begin{minipage}[t]{0.16\linewidth}
        \centering
        \includegraphics[width=\linewidth]{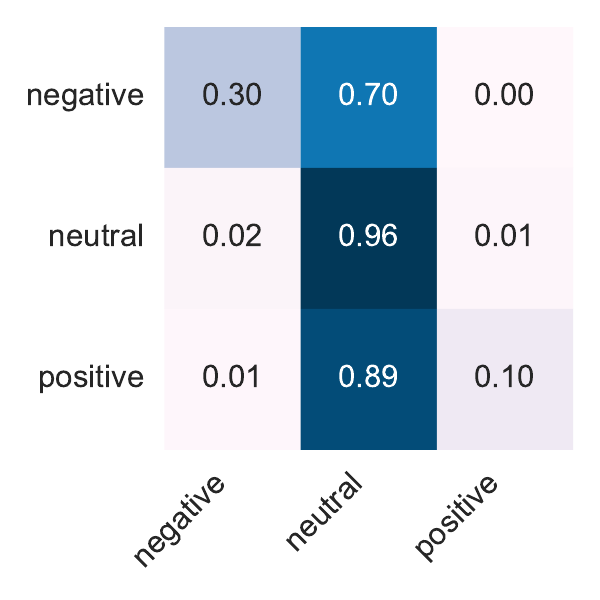}
        \subcaption{\small EIA}
    \end{minipage} \hfill
    \begin{minipage}[t]{0.20\linewidth}
        \centering
        \includegraphics[width=\linewidth]{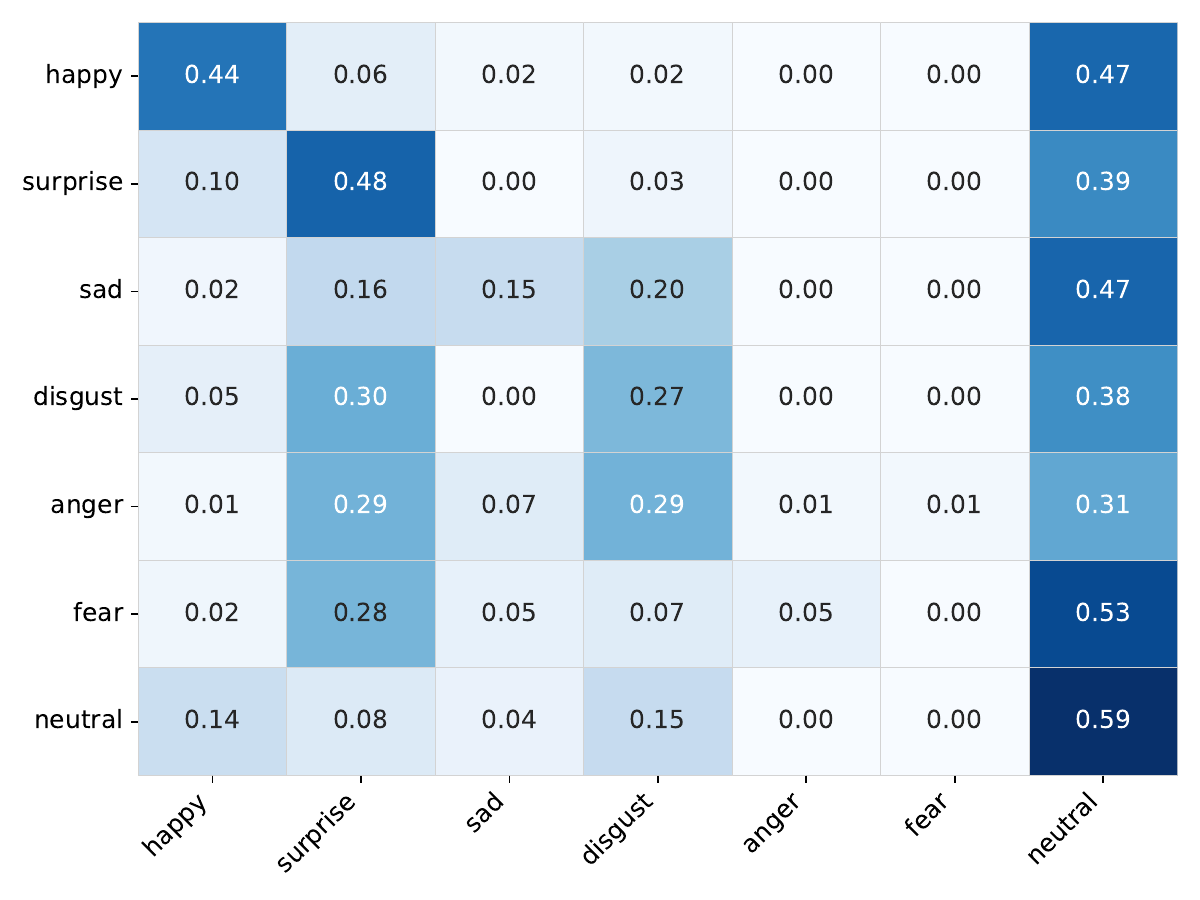}
        \subcaption{\small CEIA (emo.)}
    \end{minipage} \hfill
    \begin{minipage}[t]{0.20\linewidth}
        \centering
        \includegraphics[width=\linewidth]{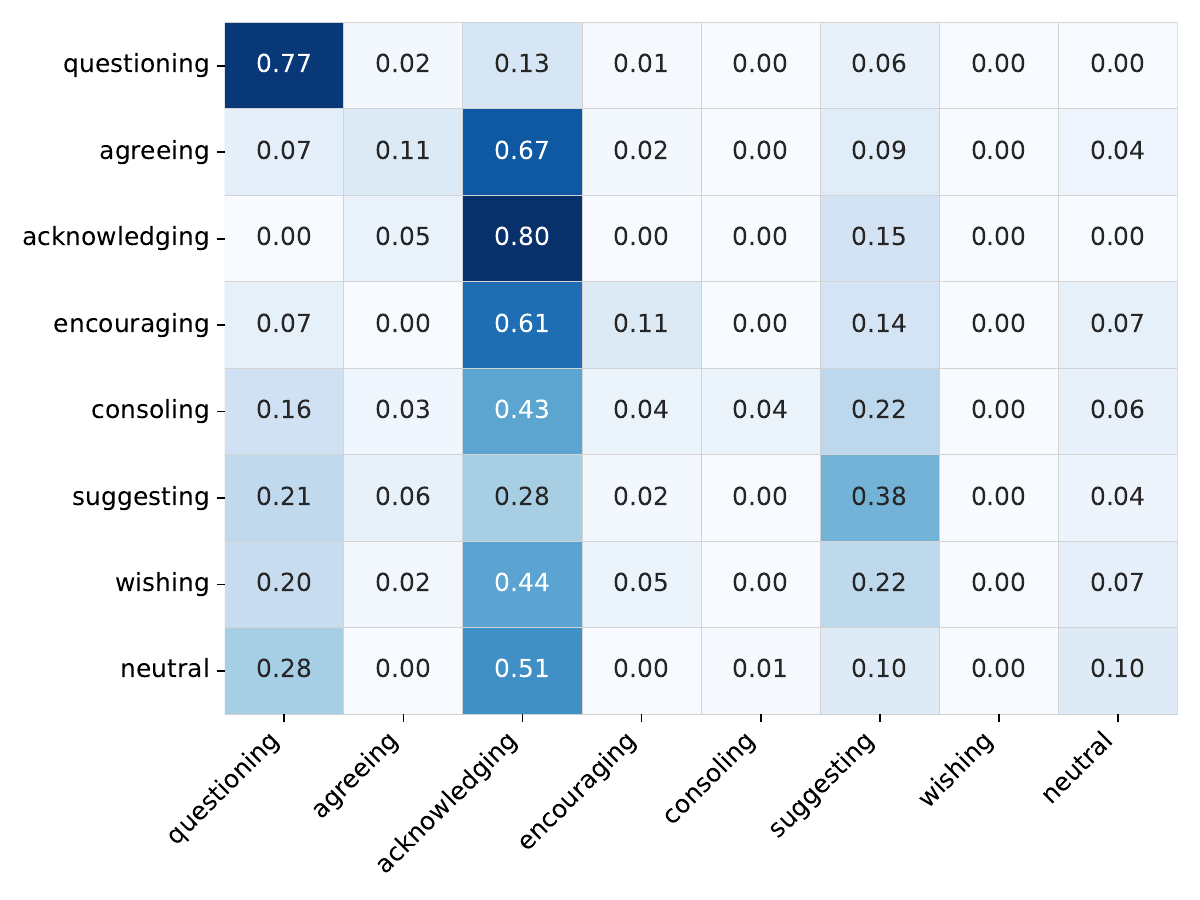}
        \subcaption{\small CEIA (int.)}
    \end{minipage} \hfill
    \begin{minipage}[t]{0.24\linewidth}
        \centering
        \includegraphics[width=\linewidth]{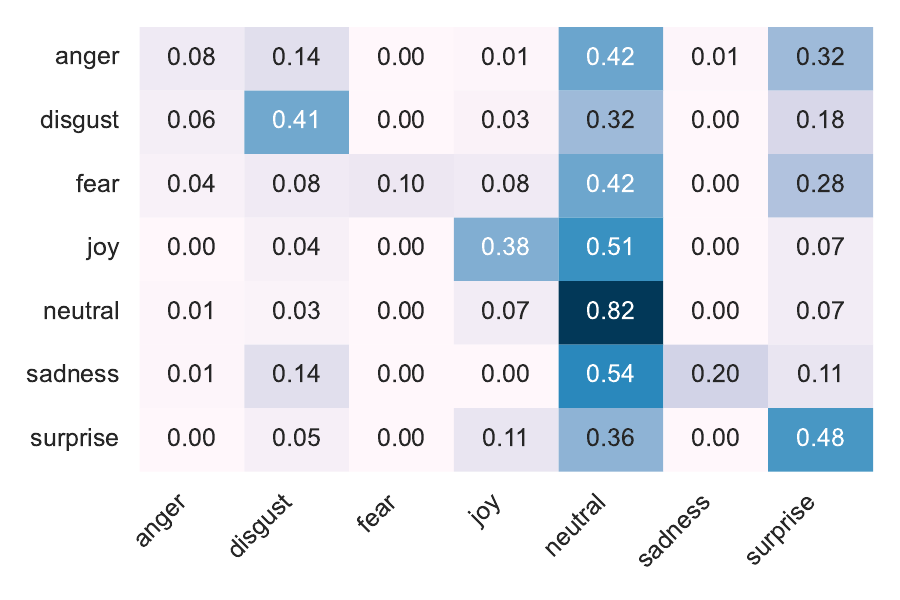}
        \subcaption{\small MPDER}
    \end{minipage} \hfill
    \begin{minipage}[t]{0.16\linewidth}
        \centering
        \includegraphics[width=\linewidth]{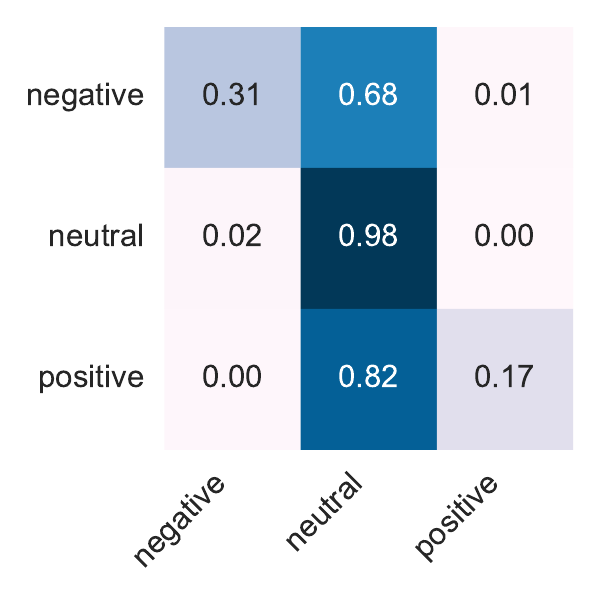}
        \subcaption{\small PEA}
    \end{minipage}

    \vspace{-2mm}
    \caption{\small Confusion matrices for LongVA-7B-DPO on each evaluation scenario of $\ours$.}
    \label{fig:confusion-LongVA-7B-DPO}
\end{figure*}

\begin{figure*}[!t]
    \centering
    
    % ===== Row 1 =====
    \begin{minipage}[t]{0.24\linewidth}
        \centering
        \includegraphics[width=\linewidth]{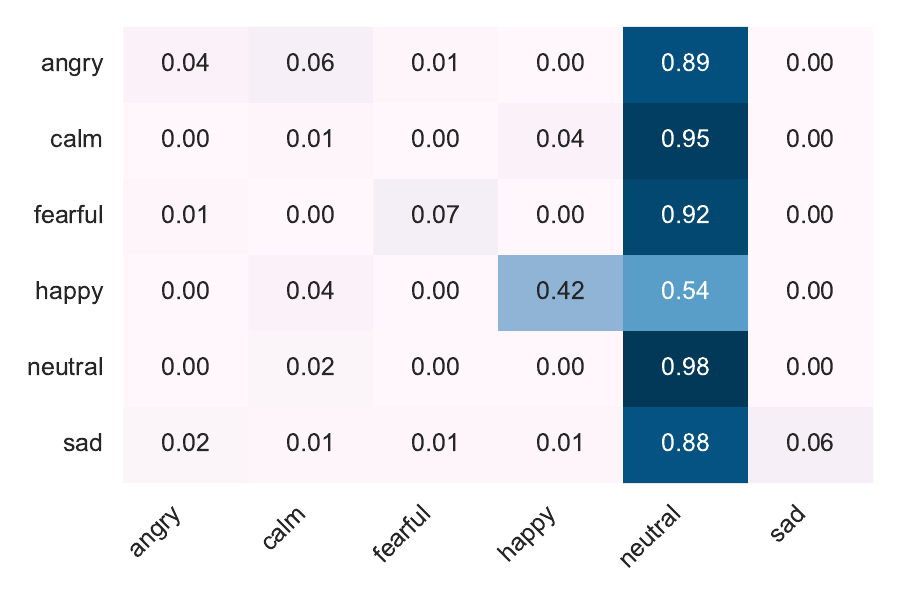}
        \subcaption{\small SOER}
    \end{minipage} \hfill
    \begin{minipage}[t]{0.24\linewidth}
        \centering
        \includegraphics[width=\linewidth]{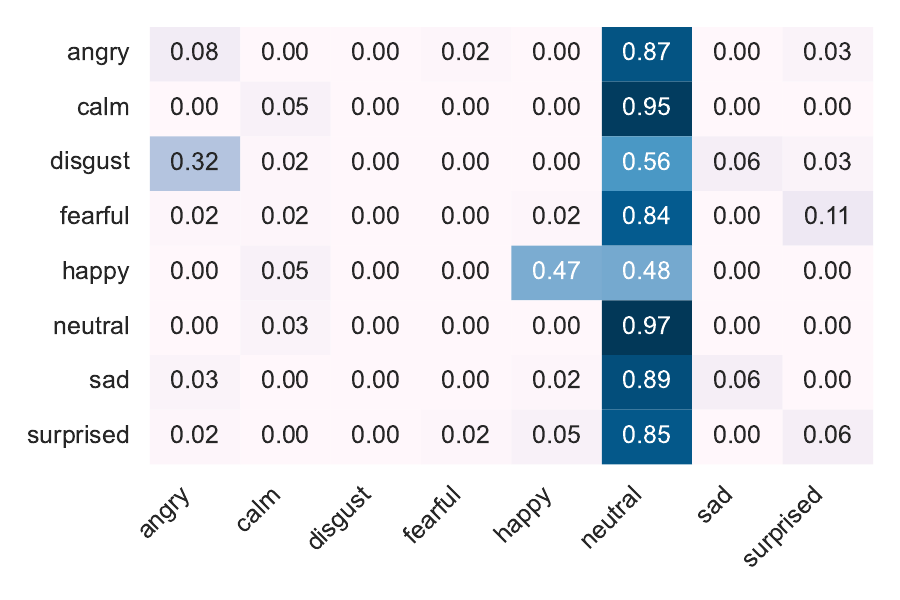}
        \subcaption{\small SPER}
    \end{minipage} \hfill
    \begin{minipage}[t]{0.16\linewidth}
        \centering
        \includegraphics[width=\linewidth]{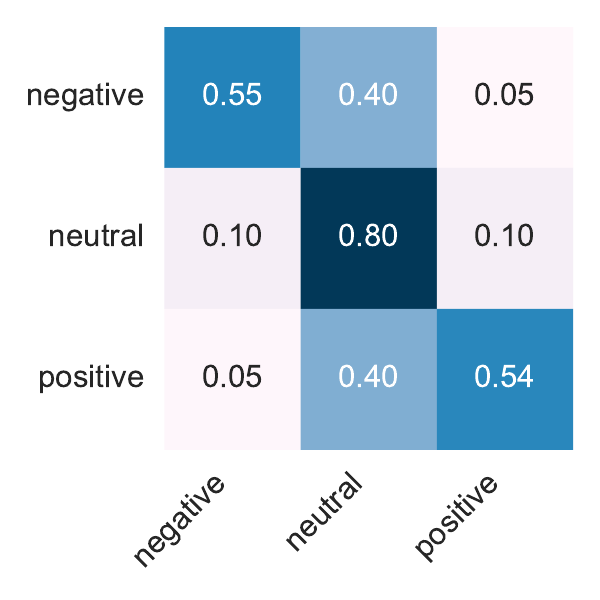}
        \subcaption{\small OSA}
    \end{minipage} \hfill
    \begin{minipage}[t]{0.16\linewidth}
        \centering
        \includegraphics[width=\linewidth]{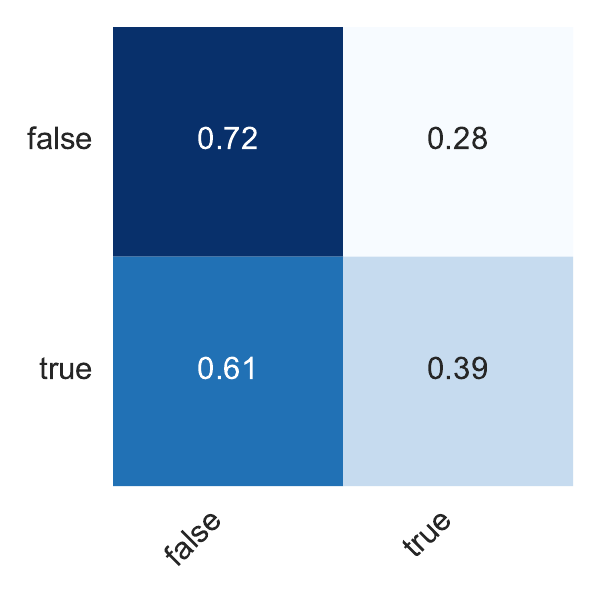}
        \subcaption{\small HU}
    \end{minipage} \\

    % ===== Row 2 =====
    \begin{minipage}[t]{0.24\linewidth}
        \centering
        \includegraphics[width=\linewidth]{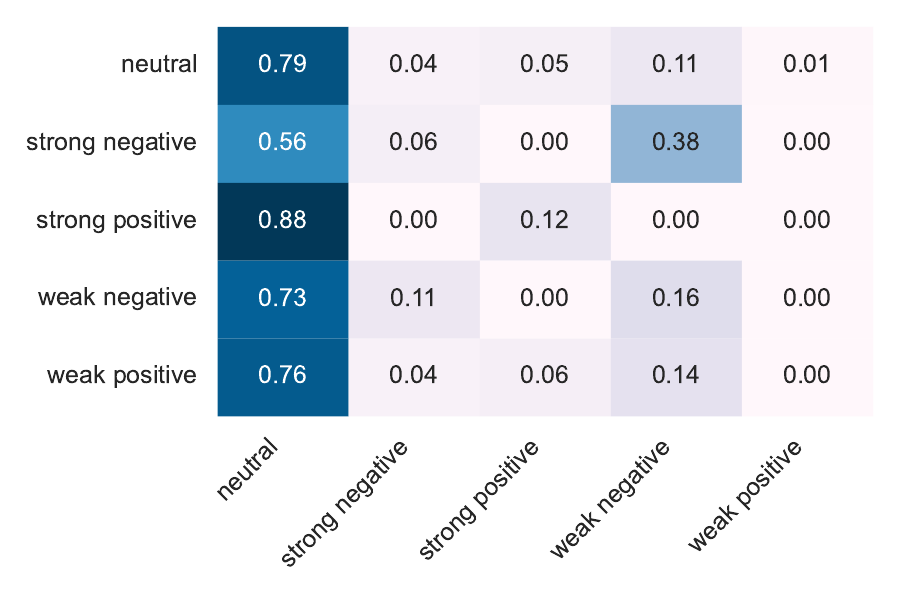}
        \subcaption{\small SCEA}
    \end{minipage} \hfill
    \begin{minipage}[t]{0.24\linewidth}
        \centering
        \includegraphics[width=\linewidth]{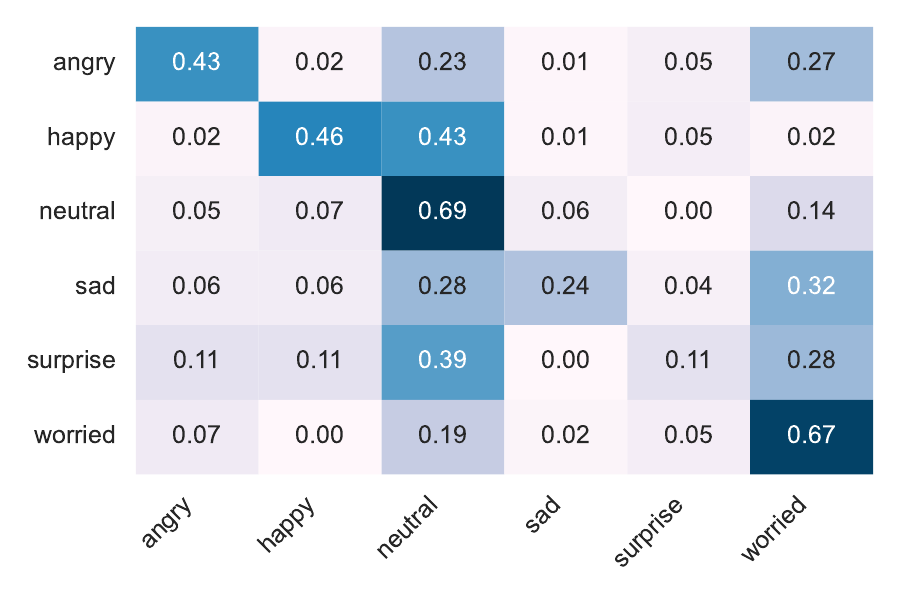}
        \subcaption{\small FGDEA}
    \end{minipage} \hfill
    \begin{minipage}[t]{0.16\linewidth}
        \centering
        \includegraphics[width=\linewidth]{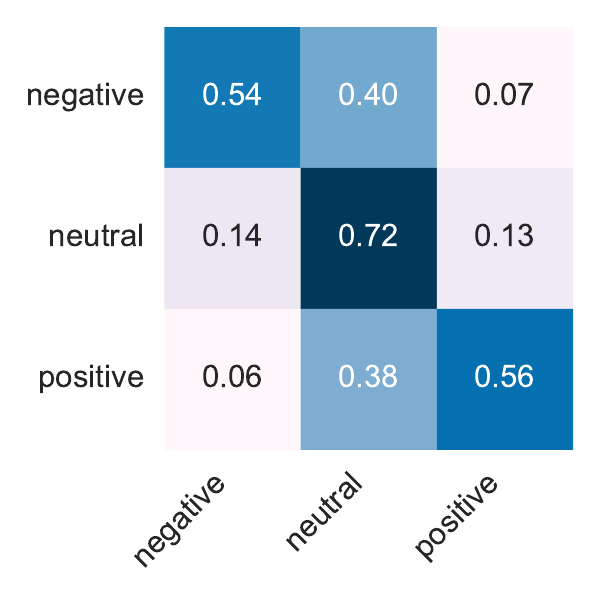}
        \subcaption{\small FCDEA}
    \end{minipage} \hfill
    \begin{minipage}[t]{0.16\linewidth}
        \centering
        \includegraphics[width=\linewidth]{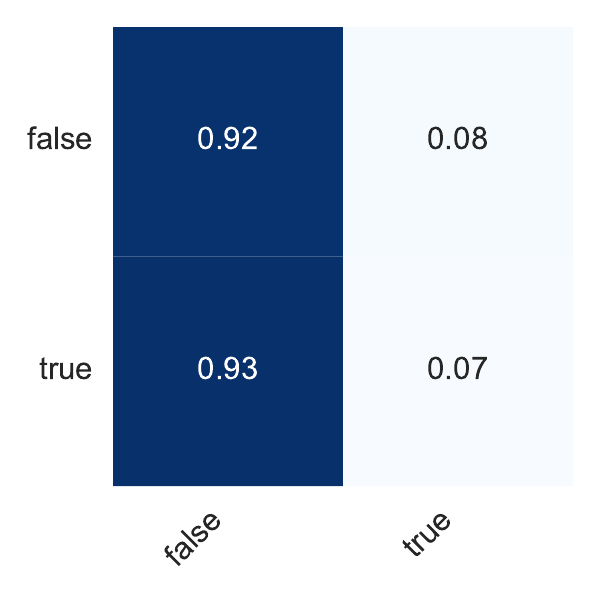}
        \subcaption{\small SD}
    \end{minipage} \\

    % ===== Row 3 =====
    \begin{minipage}[t]{0.16\linewidth}
        \centering
        \includegraphics[width=\linewidth]{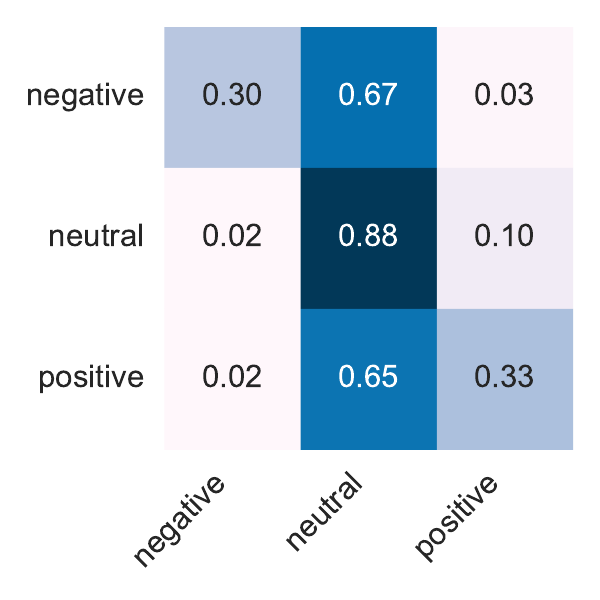}
        \subcaption{\small EIA}
    \end{minipage} \hfill
    \begin{minipage}[t]{0.20\linewidth}
        \centering
        \includegraphics[width=\linewidth]{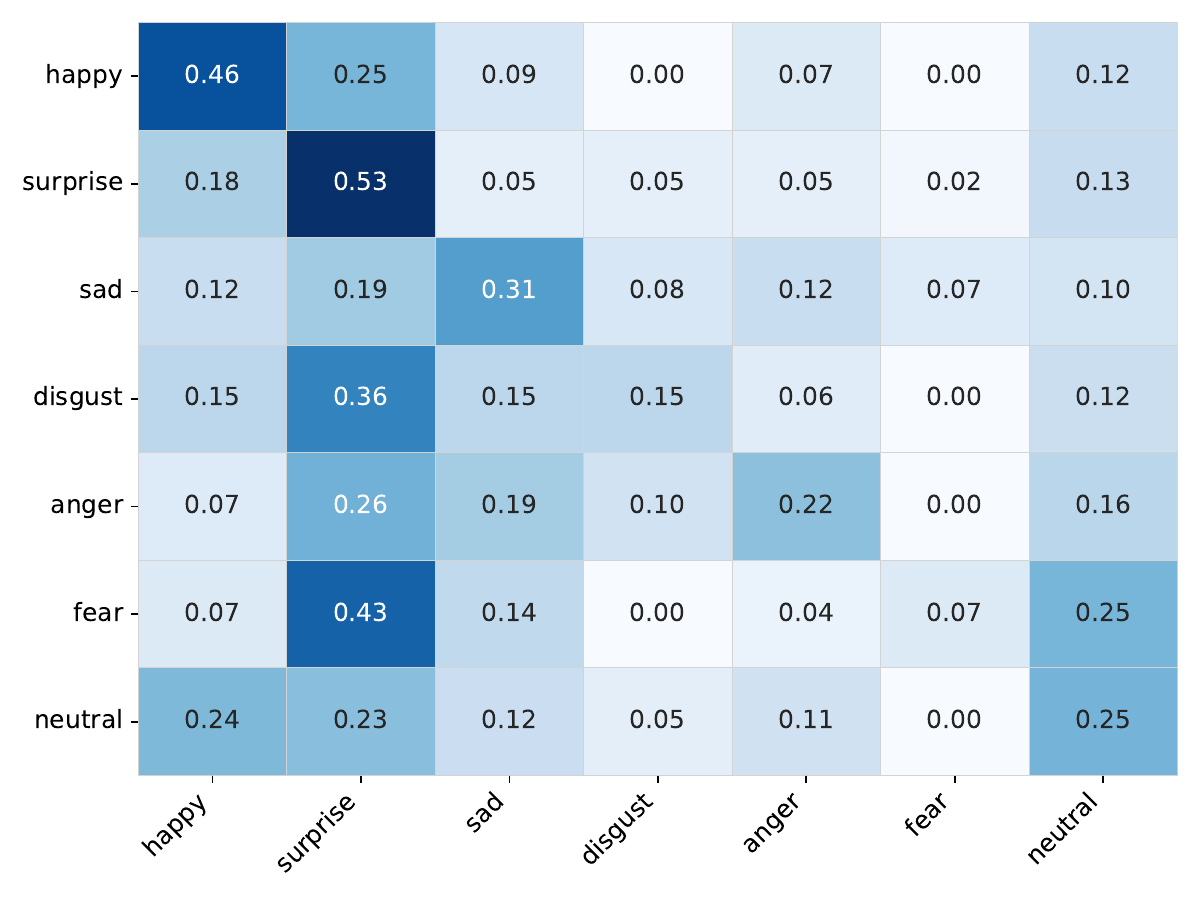}
        \subcaption{\small CEIA (emo.)}
    \end{minipage} \hfill
    \begin{minipage}[t]{0.20\linewidth}
        \centering
        \includegraphics[width=\linewidth]{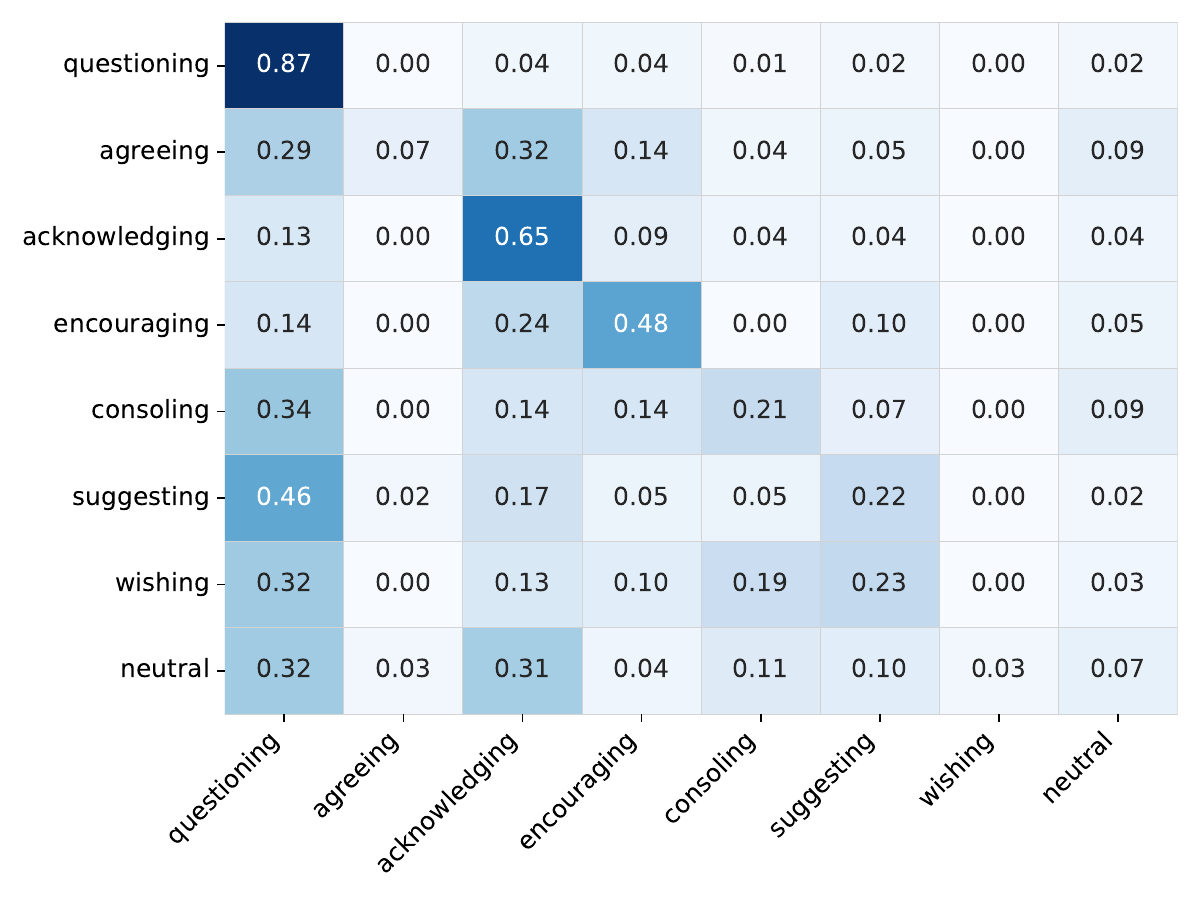}
        \subcaption{\small CEIA (int.)}
    \end{minipage} \hfill
    \begin{minipage}[t]{0.24\linewidth}
        \centering
        \includegraphics[width=\linewidth]{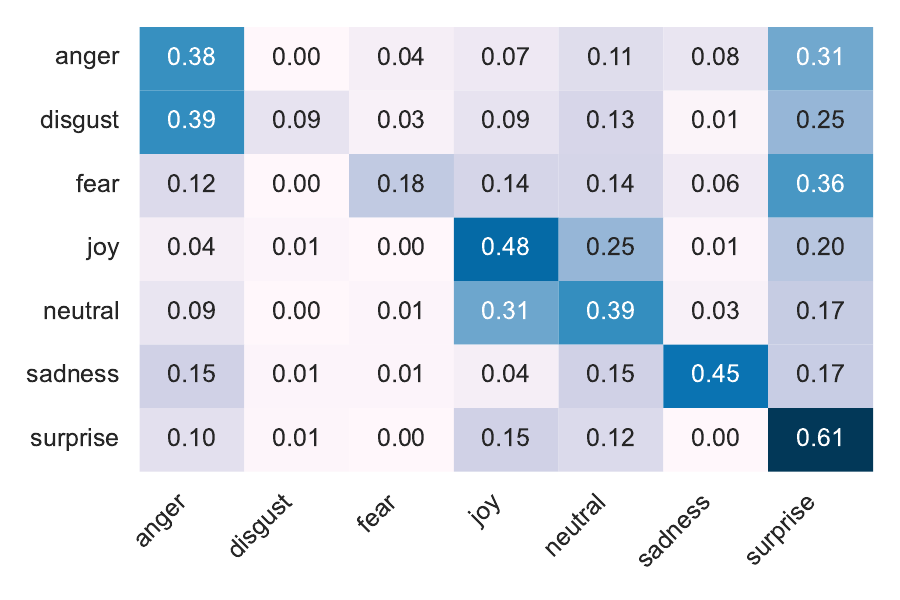}
        \subcaption{\small MPDER}
    \end{minipage} \hfill
    \begin{minipage}[t]{0.16\linewidth}
        \centering
        \includegraphics[width=\linewidth]{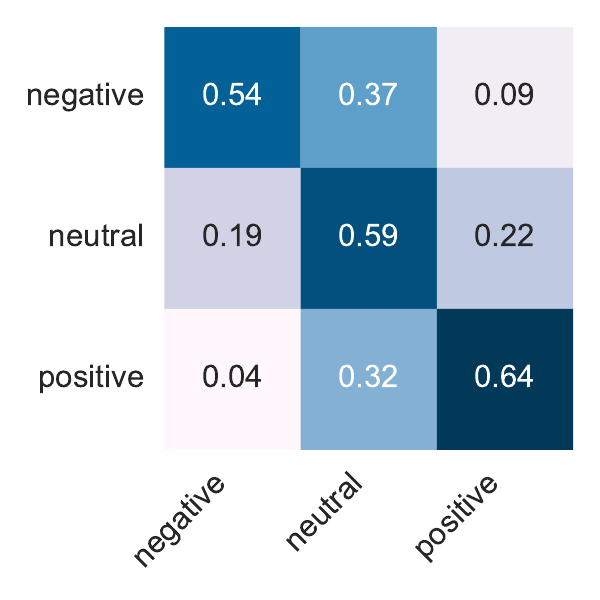}
        \subcaption{\small PEA}
    \end{minipage}

    \vspace{-2mm}
    \caption{\small Confusion matrices for MiniCPM-V-2.6-8B on each evaluation scenario of $\ours$.}
    \label{fig:confusion-MiniCPM-V-2.6-8B}
\end{figure*}

\begin{figure*}[!t]
    \centering
    
    % ===== Row 1 =====
    \begin{minipage}[t]{0.24\linewidth}
        \centering
        \includegraphics[width=\linewidth]{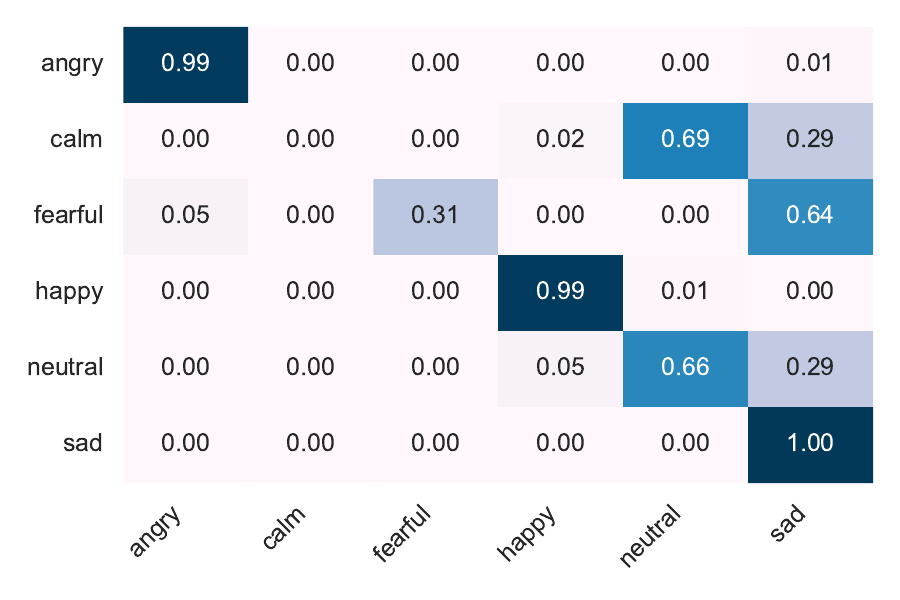}
        \subcaption{\small SOER}
    \end{minipage} \hfill
    \begin{minipage}[t]{0.24\linewidth}
        \centering
        \includegraphics[width=\linewidth]{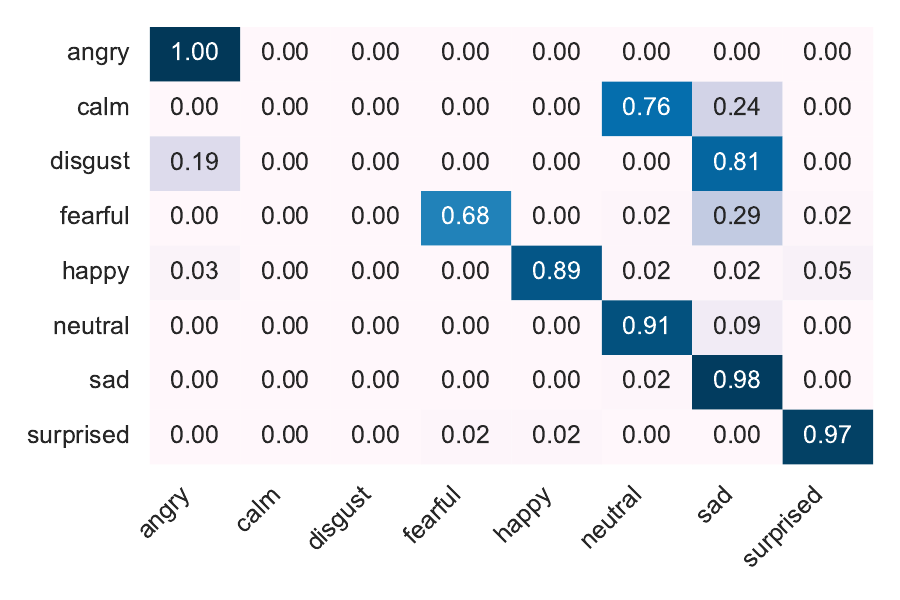}
        \subcaption{\small SPER}
    \end{minipage} \hfill
    \begin{minipage}[t]{0.16\linewidth}
        \centering
        \includegraphics[width=\linewidth]{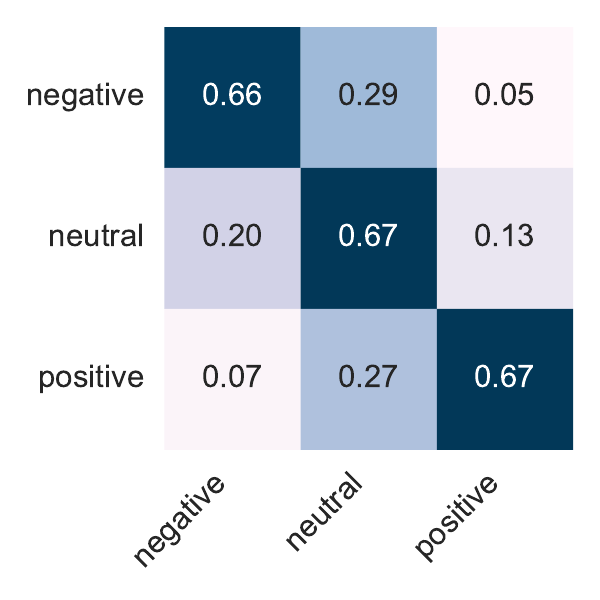}
        \subcaption{\small OSA}
    \end{minipage} \hfill
    \begin{minipage}[t]{0.16\linewidth}
        \centering
        \includegraphics[width=\linewidth]{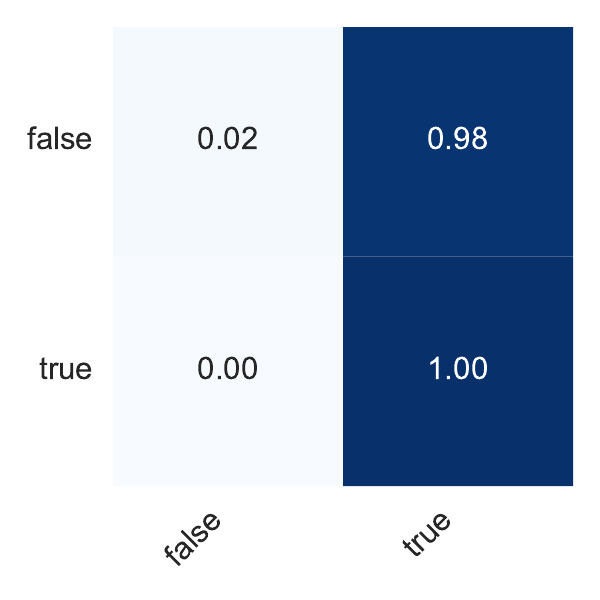}
        \subcaption{\small HU}
    \end{minipage} \\

    % ===== Row 2 =====
    \begin{minipage}[t]{0.24\linewidth}
        \centering
        \includegraphics[width=\linewidth]{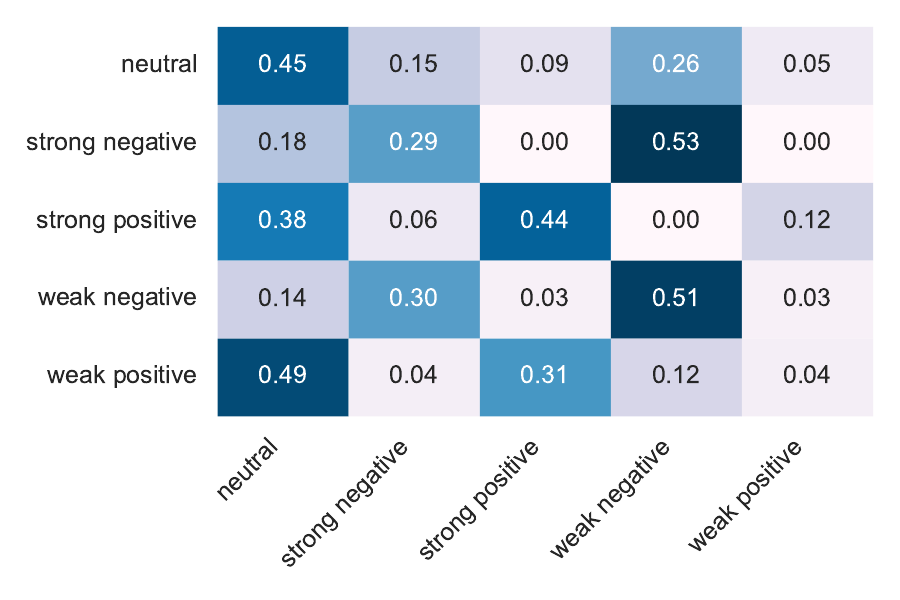}
        \subcaption{\small SCEA}
    \end{minipage} \hfill
    \begin{minipage}[t]{0.24\linewidth}
        \centering
        \includegraphics[width=\linewidth]{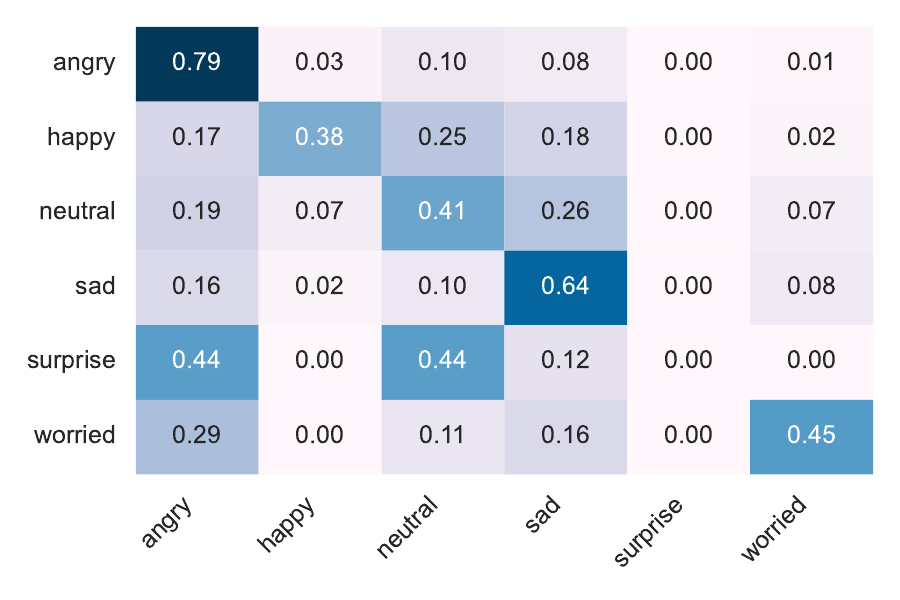}
        \subcaption{\small FGDEA}
    \end{minipage} \hfill
    \begin{minipage}[t]{0.16\linewidth}
        \centering
        \includegraphics[width=\linewidth]{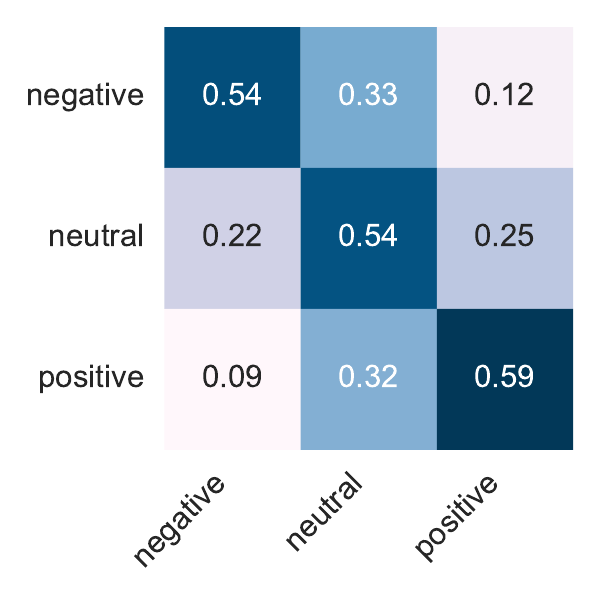}
        \subcaption{\small FCDEA}
    \end{minipage} \hfill
    \begin{minipage}[t]{0.16\linewidth}
        \centering
        \includegraphics[width=\linewidth]{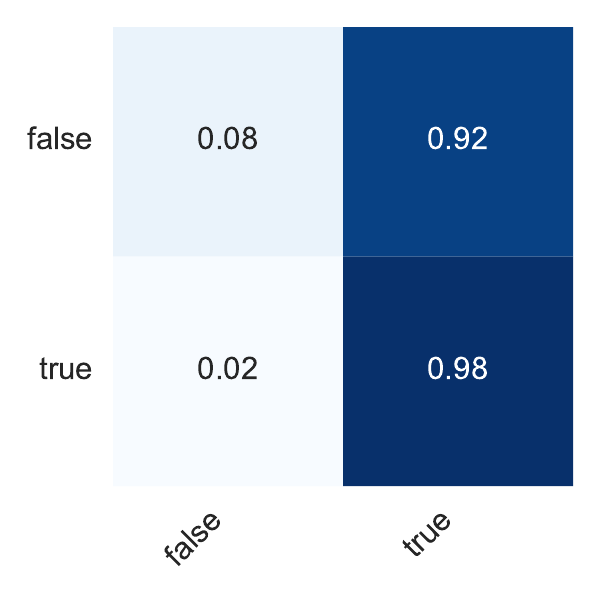}
        \subcaption{\small SD}
    \end{minipage} \\

    % ===== Row 3 =====
    \begin{minipage}[t]{0.16\linewidth}
        \centering
        \includegraphics[width=\linewidth]{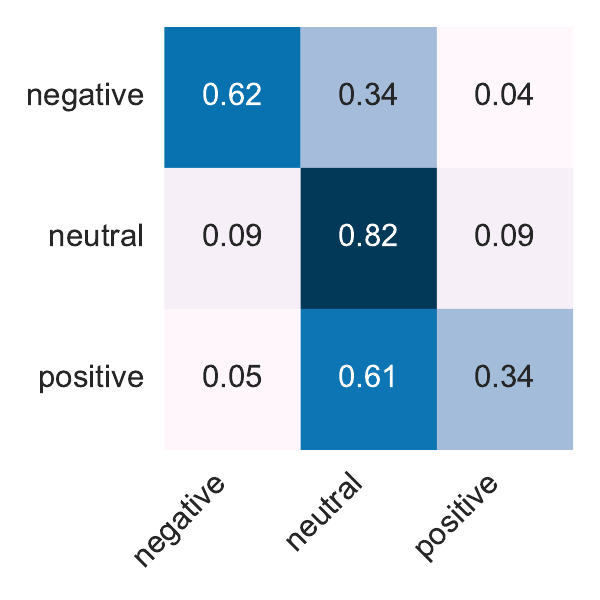}
        \subcaption{\small EIA}
    \end{minipage} \hfill
    \begin{minipage}[t]{0.20\linewidth}
        \centering
        \includegraphics[width=\linewidth]{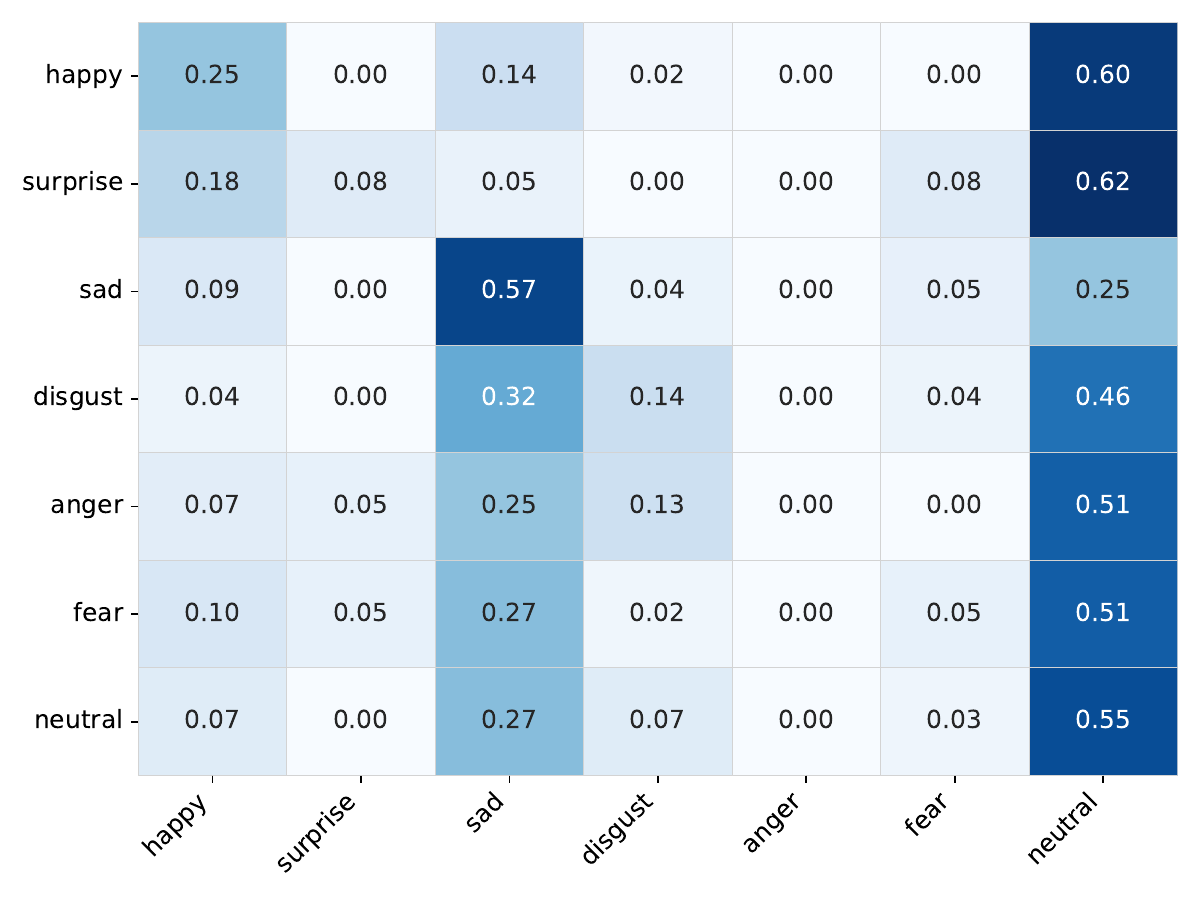}
        \subcaption{\small CEIA (emo.)}
    \end{minipage} \hfill
    \begin{minipage}[t]{0.20\linewidth}
        \centering
        \includegraphics[width=\linewidth]{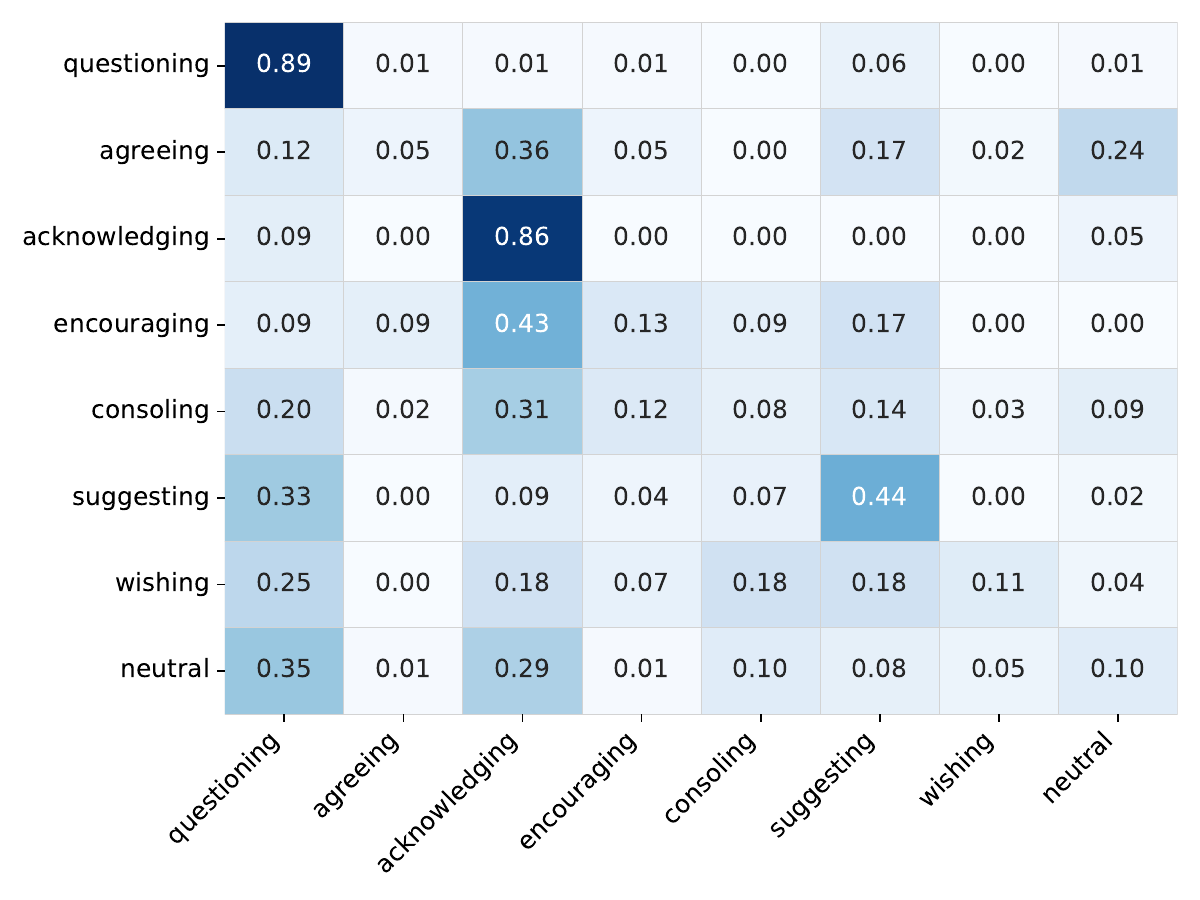}
        \subcaption{\small CEIA (int.)}
    \end{minipage} \hfill
    \begin{minipage}[t]{0.24\linewidth}
        \centering
        \includegraphics[width=\linewidth]{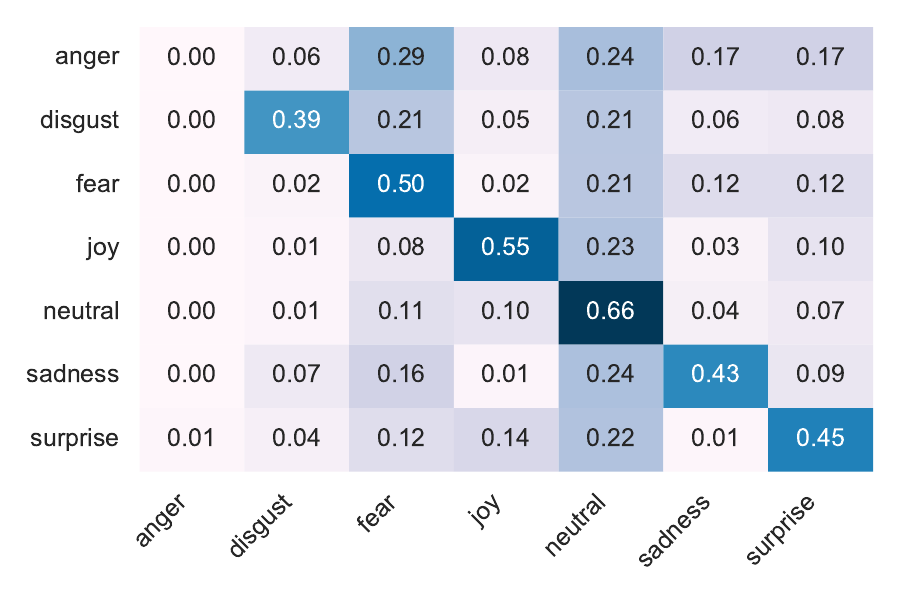}
        \subcaption{\small MPDER}
    \end{minipage} \hfill
    \begin{minipage}[t]{0.16\linewidth}
        \centering
        \includegraphics[width=\linewidth]{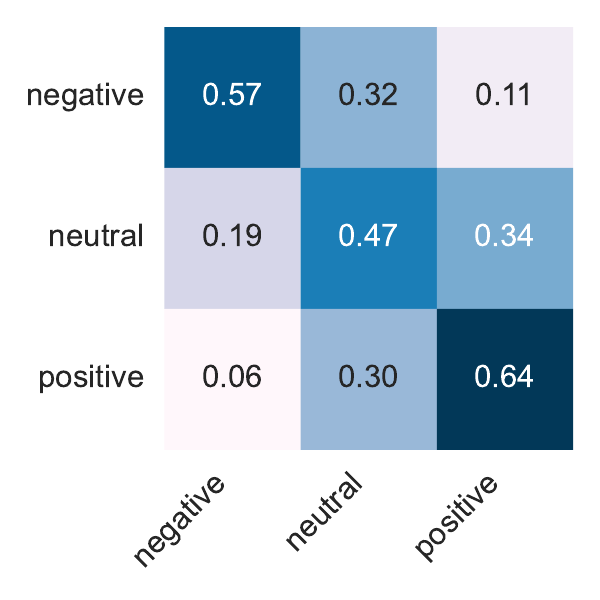}
        \subcaption{\small PEA}
    \end{minipage}

    \vspace{-3mm}
    \caption{\small Confusion matrices for Qwen2-Audio-7B-Instruct on each evaluation scenario of $\ours$.}
    \label{fig:confusion-Qwen2-Audio-7B-Instruct}
\end{figure*}

\begin{figure*}[!t]
    \centering
    
    % ===== Row 1 =====
    \begin{minipage}[t]{0.24\linewidth}
        \centering
        \includegraphics[width=\linewidth]{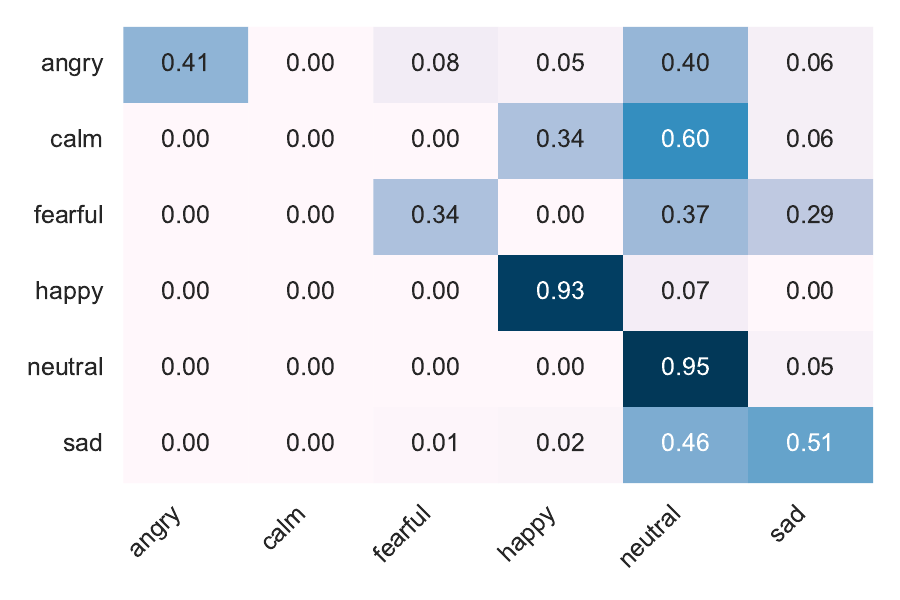}
        \subcaption{\small SOER}
    \end{minipage} \hfill
    \begin{minipage}[t]{0.24\linewidth}
        \centering
        \includegraphics[width=\linewidth]{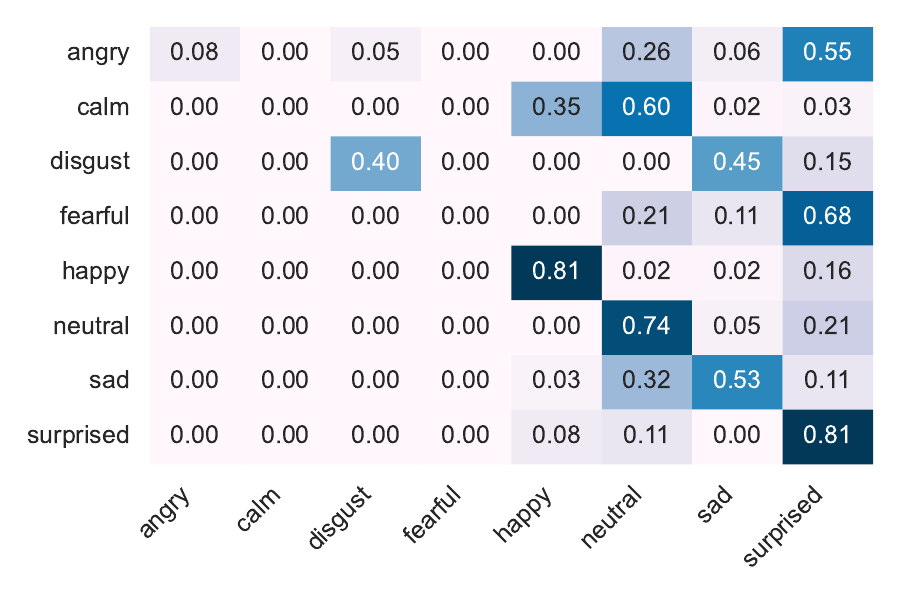}
        \subcaption{\small SPER}
    \end{minipage} \hfill
    \begin{minipage}[t]{0.16\linewidth}
        \centering
        \includegraphics[width=\linewidth]{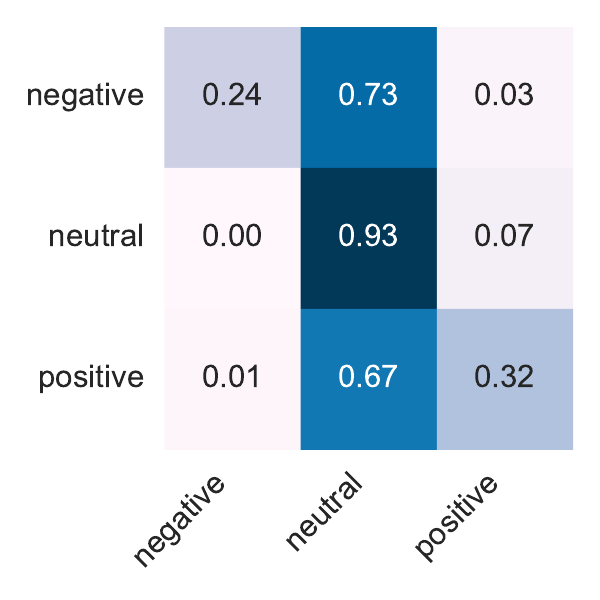}
        \subcaption{\small OSA}
    \end{minipage} \hfill
    \begin{minipage}[t]{0.16\linewidth}
        \centering
        \includegraphics[width=\linewidth]{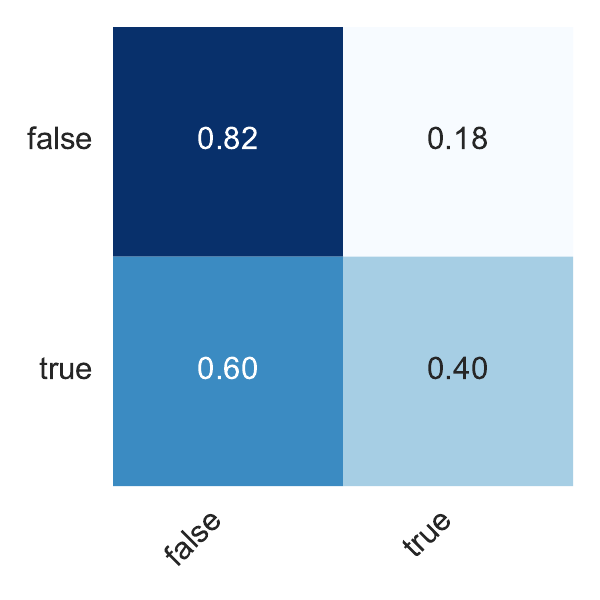}
        \subcaption{\small HU}
    \end{minipage} \\

    % ===== Row 2 =====
    \begin{minipage}[t]{0.24\linewidth}
        \centering
        \includegraphics[width=\linewidth]{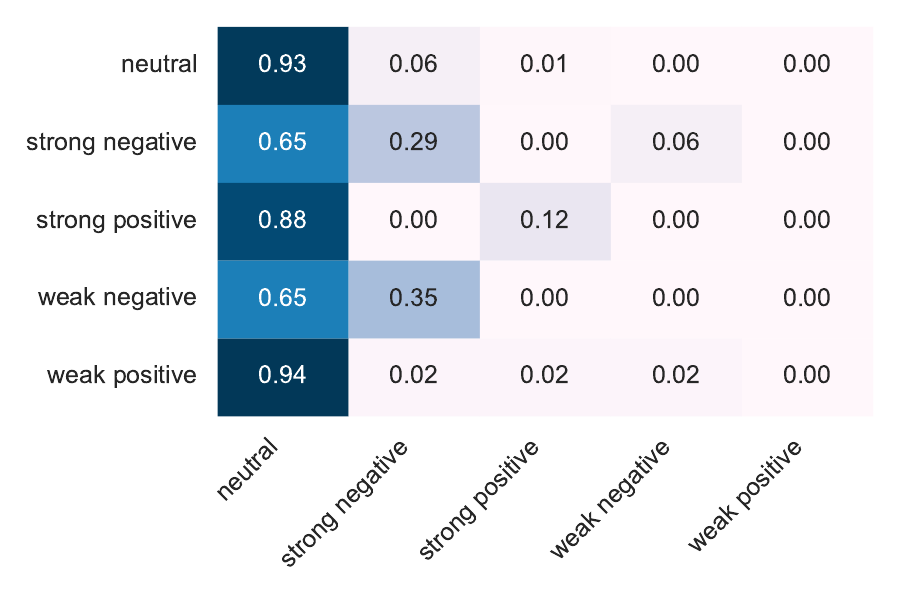}
        \subcaption{\small SCEA}
    \end{minipage} \hfill
    \begin{minipage}[t]{0.24\linewidth}
        \centering
        \includegraphics[width=\linewidth]{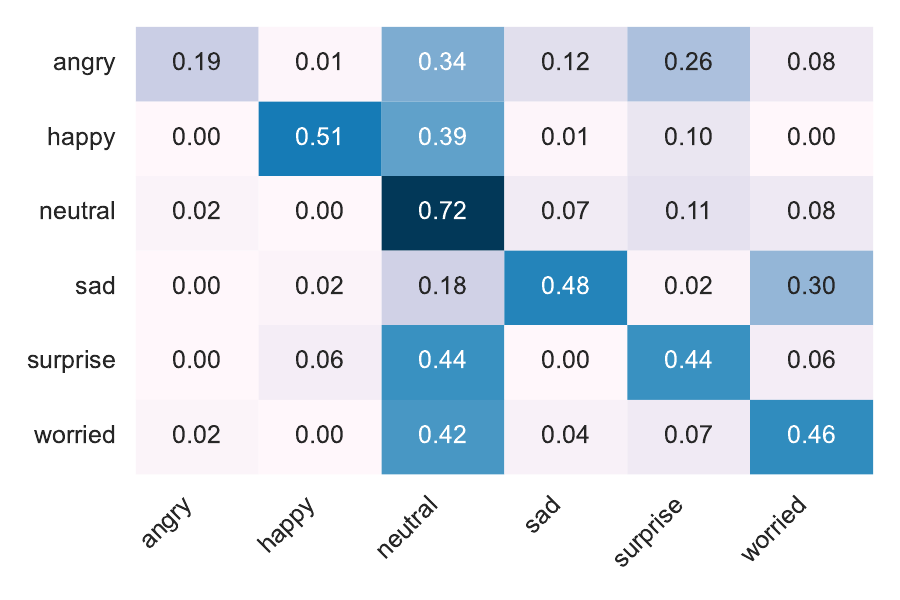}
        \subcaption{\small FGDEA}
    \end{minipage} \hfill
    \begin{minipage}[t]{0.16\linewidth}
        \centering
        \includegraphics[width=\linewidth]{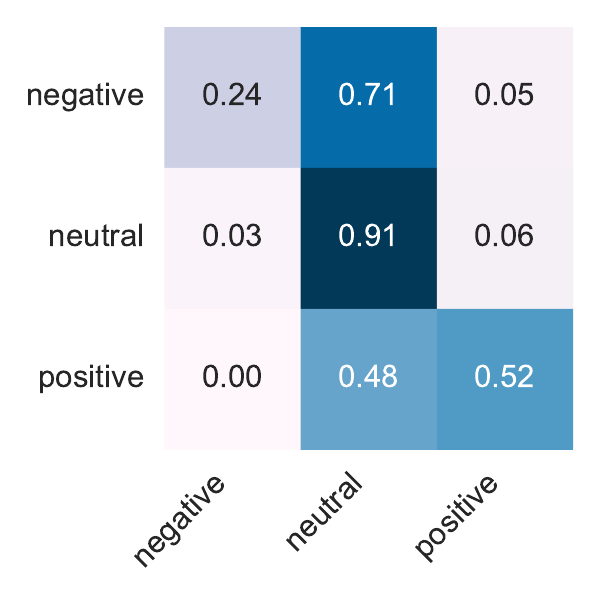}
        \subcaption{\small FCDEA}
    \end{minipage} \hfill
    \begin{minipage}[t]{0.16\linewidth}
        \centering
        \includegraphics[width=\linewidth]{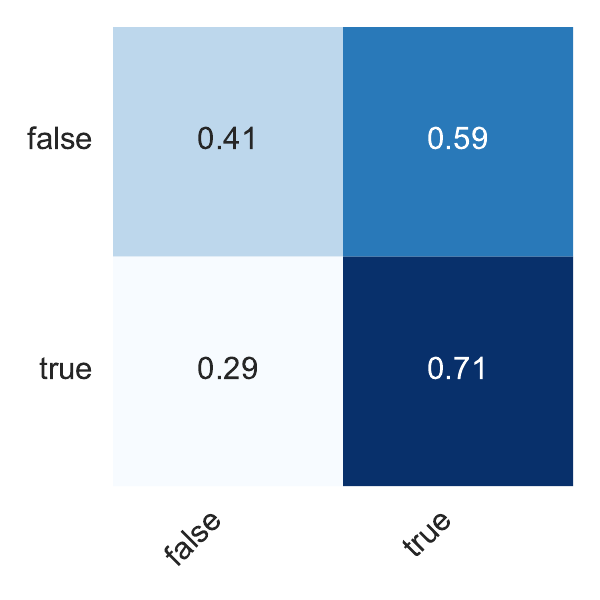}
        \subcaption{\small SD}
    \end{minipage} \\

    % ===== Row 3 =====
    \begin{minipage}[t]{0.16\linewidth}
        \centering
        \includegraphics[width=\linewidth]{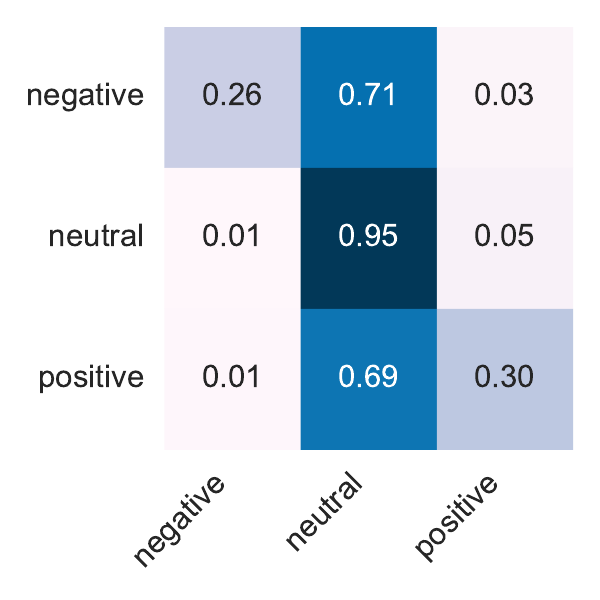}
        \subcaption{\small EIA}
    \end{minipage} \hfill
    \begin{minipage}[t]{0.20\linewidth}
        \centering
        \includegraphics[width=\linewidth]{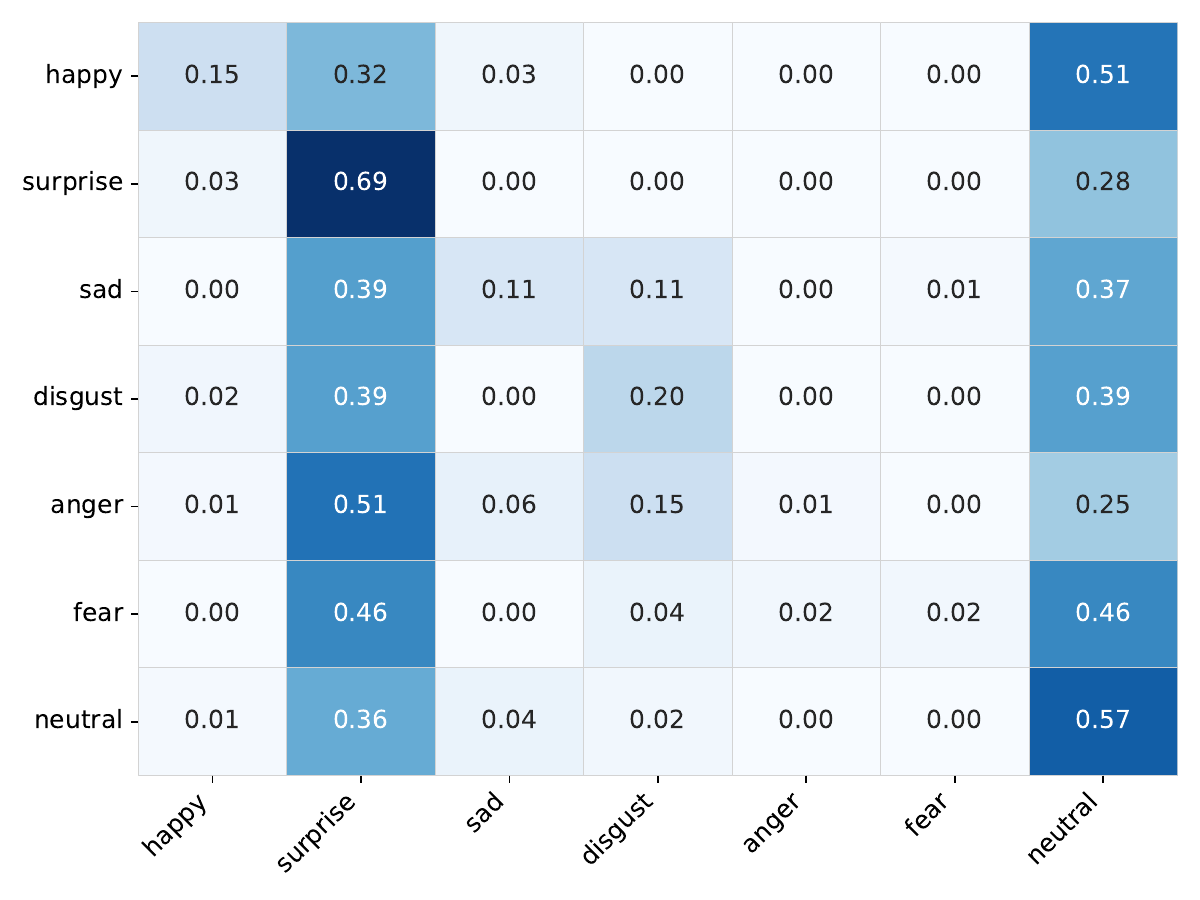}
        \subcaption{\small CEIA (emo.)}
    \end{minipage} \hfill
    \begin{minipage}[t]{0.20\linewidth}
        \centering
        \includegraphics[width=\linewidth]{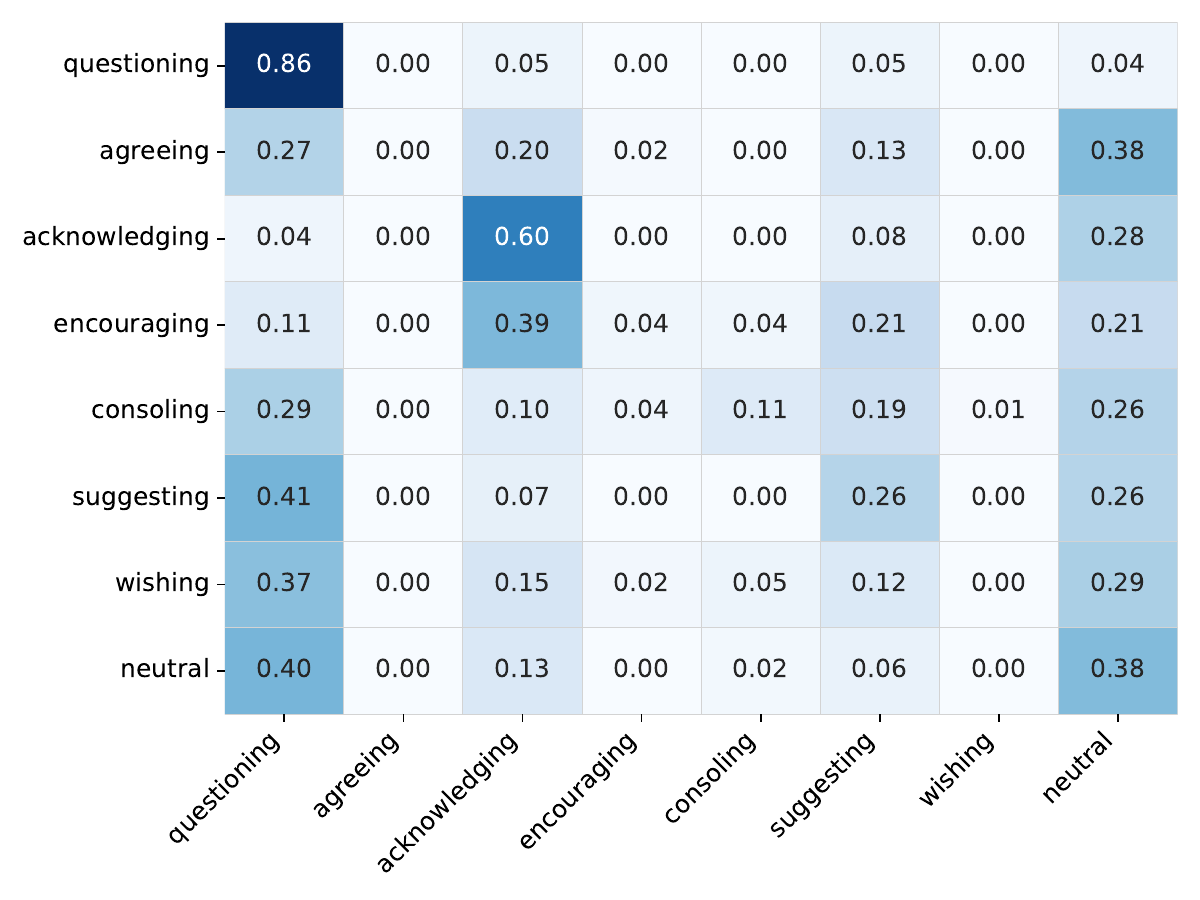}
        \subcaption{\small CEIA (int.)}
    \end{minipage} \hfill
    \begin{minipage}[t]{0.24\linewidth}
        \centering
        \includegraphics[width=\linewidth]{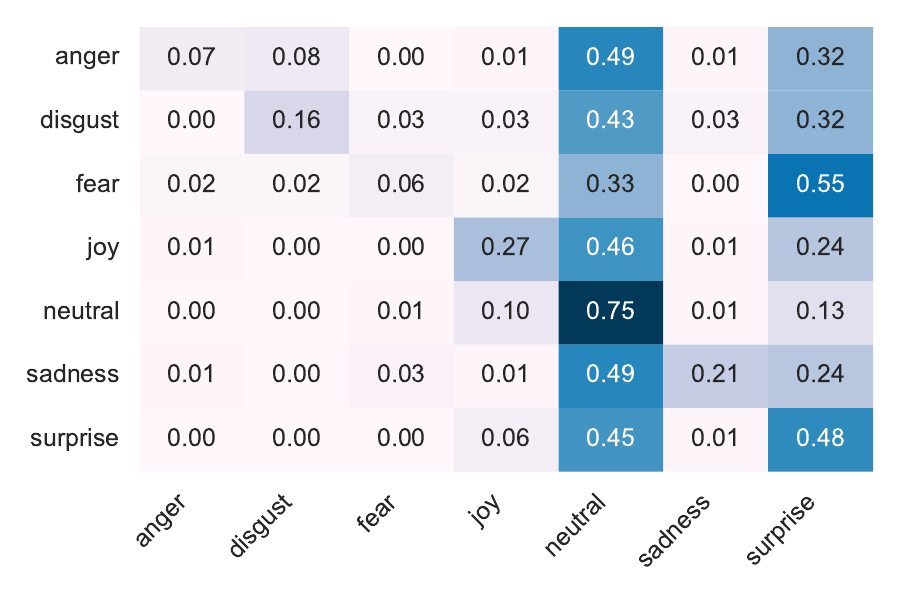}
        \subcaption{\small MPDER}
    \end{minipage} \hfill
    \begin{minipage}[t]{0.16\linewidth}
        \centering
        \includegraphics[width=\linewidth]{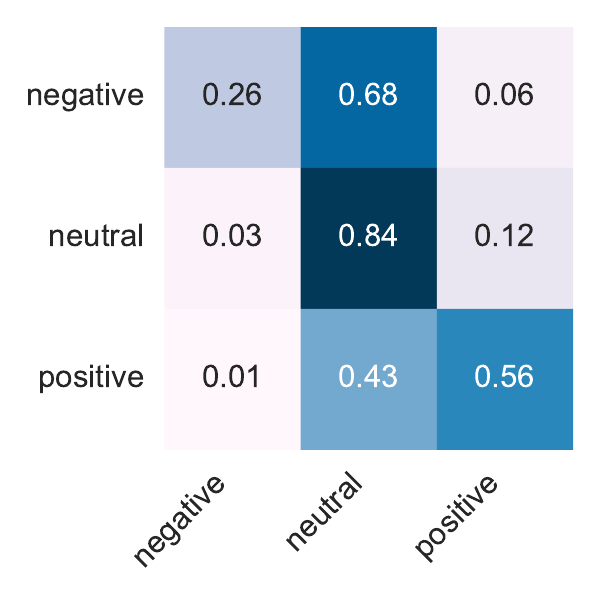}
        \subcaption{\small PEA}
    \end{minipage}

    \vspace{-3mm}
    \caption{\small Confusion matrices for VideoLLaMA2-7B on each evaluation scenario of $\ours$.}
    \label{fig:confusion-VideoLLaMA2-7B}
\end{figure*}

\begin{figure*}[!t]
    \centering
    
    % ===== Row 1 =====
    \begin{minipage}[t]{0.24\linewidth}
        \centering
        \includegraphics[width=\linewidth]{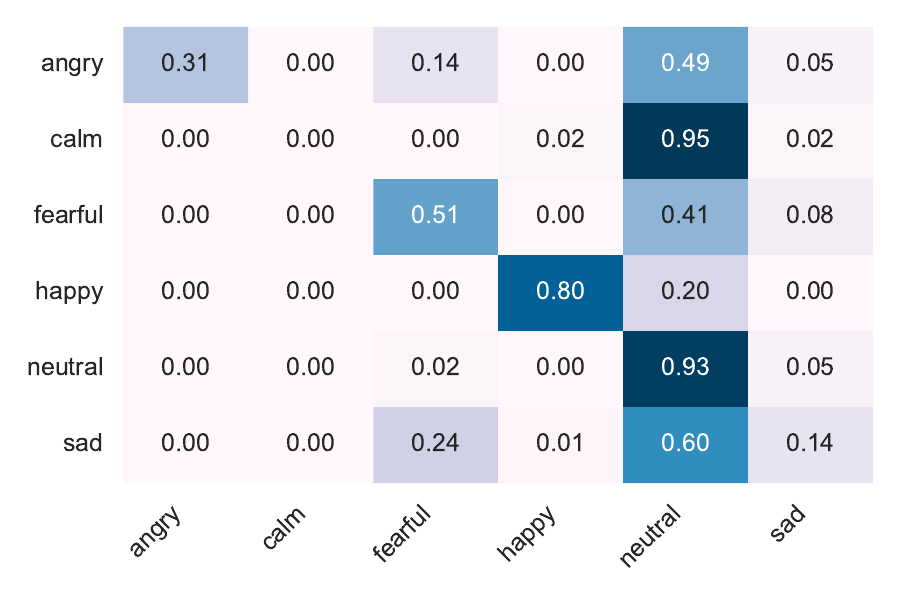}
        \subcaption{\small SOER}
    \end{minipage} \hfill
    \begin{minipage}[t]{0.24\linewidth}
        \centering
        \includegraphics[width=\linewidth]{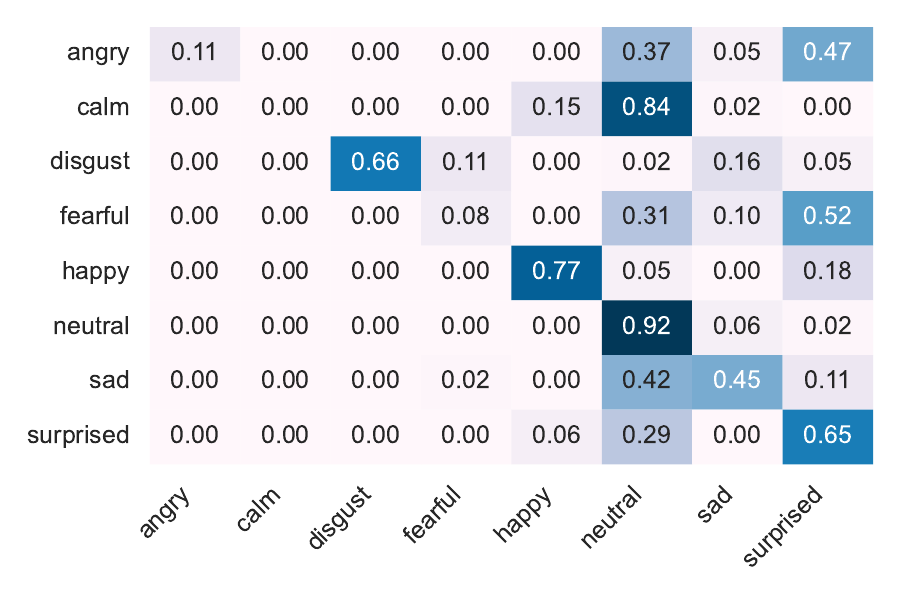}
        \subcaption{\small SPER}
    \end{minipage} \hfill
    \begin{minipage}[t]{0.16\linewidth}
        \centering
        \includegraphics[width=\linewidth]{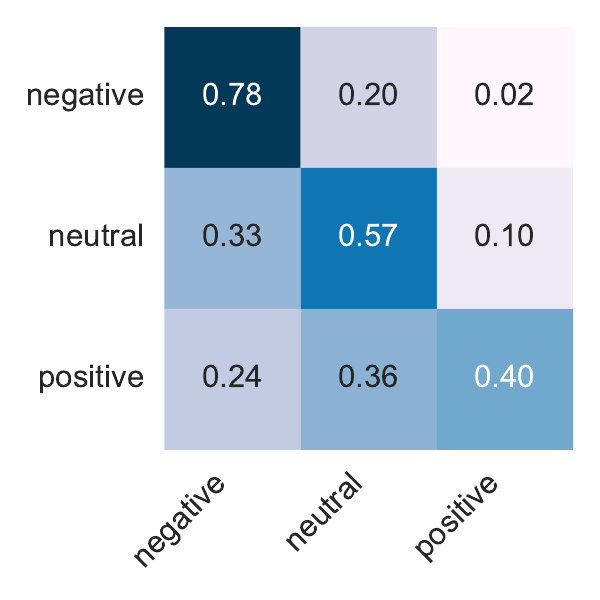}
        \subcaption{\small OSA}
    \end{minipage} \hfill
    \begin{minipage}[t]{0.16\linewidth}
        \centering
        \includegraphics[width=\linewidth]{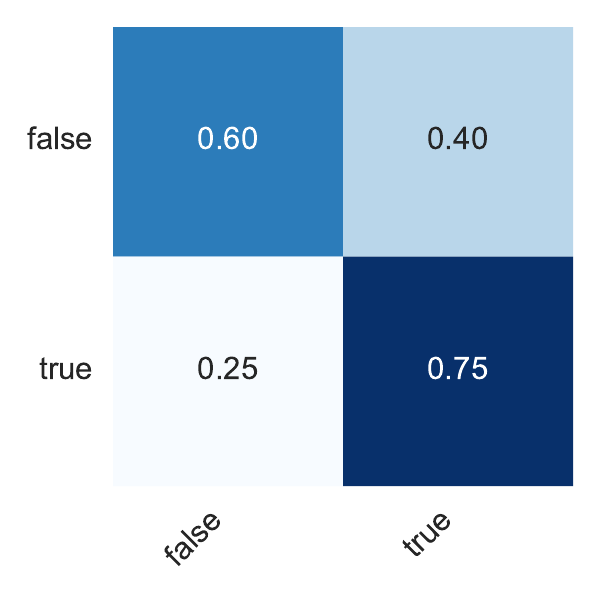}
        \subcaption{\small HU}
    \end{minipage} \\

    % ===== Row 2 =====
    \begin{minipage}[t]{0.24\linewidth}
        \centering
        \includegraphics[width=\linewidth]{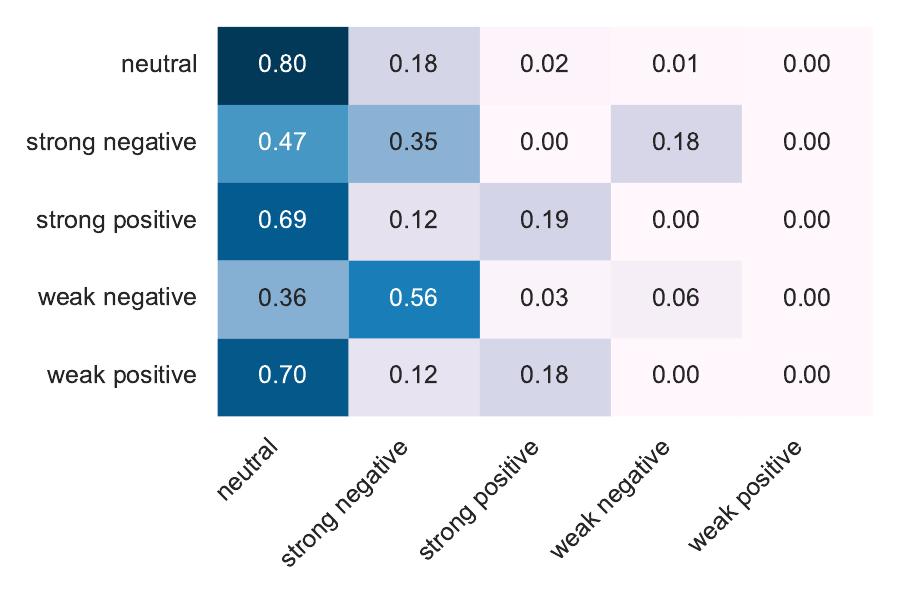}
        \subcaption{\small SCEA}
    \end{minipage} \hfill
    \begin{minipage}[t]{0.24\linewidth}
        \centering
        \includegraphics[width=\linewidth]{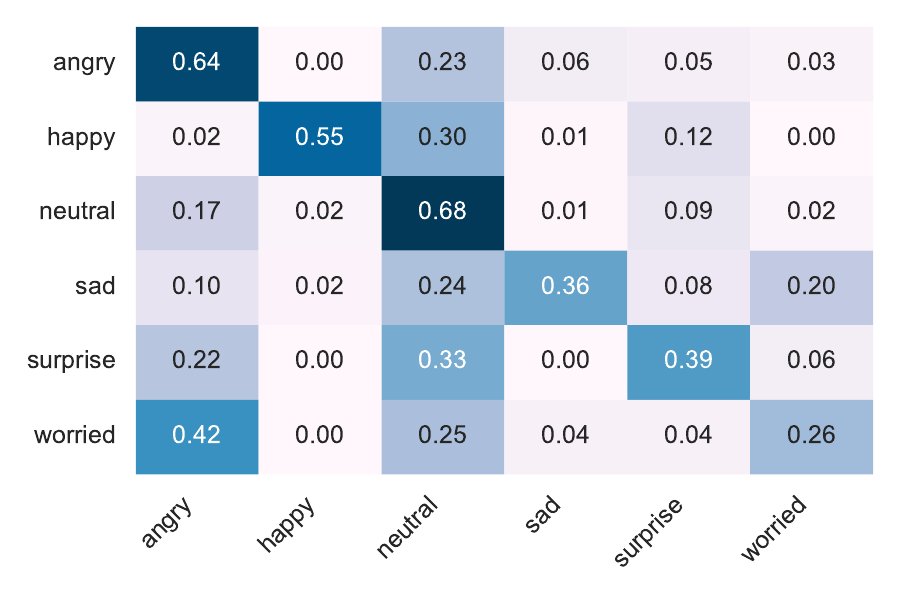}
        \subcaption{\small FGDEA}
    \end{minipage} \hfill
    \begin{minipage}[t]{0.16\linewidth}
        \centering
        \includegraphics[width=\linewidth]{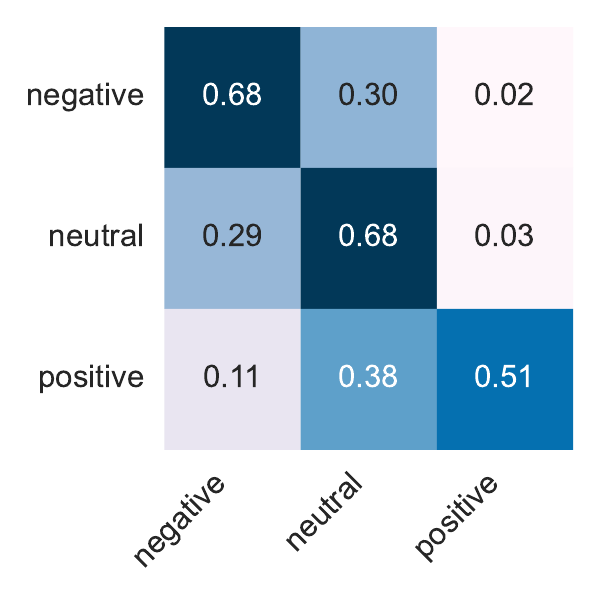}
        \subcaption{\small FCDEA}
    \end{minipage} \hfill
    \begin{minipage}[t]{0.16\linewidth}
        \centering
        \includegraphics[width=\linewidth]{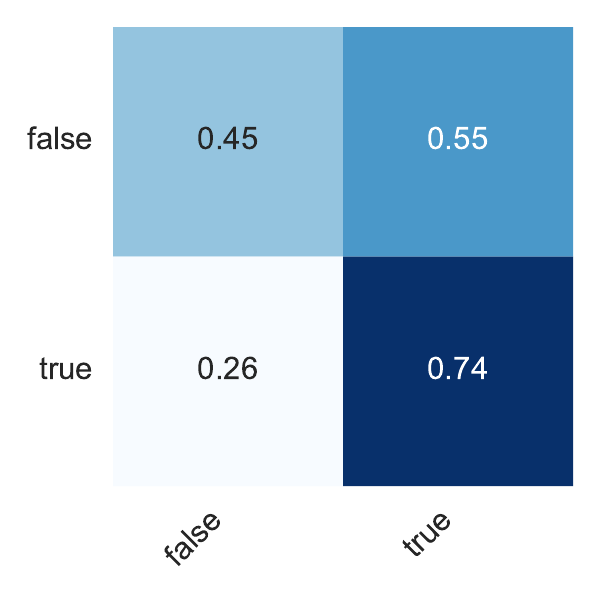}
        \subcaption{\small SD}
    \end{minipage} \\

    % ===== Row 3 =====
    \begin{minipage}[t]{0.16\linewidth}
        \centering
        \includegraphics[width=\linewidth]{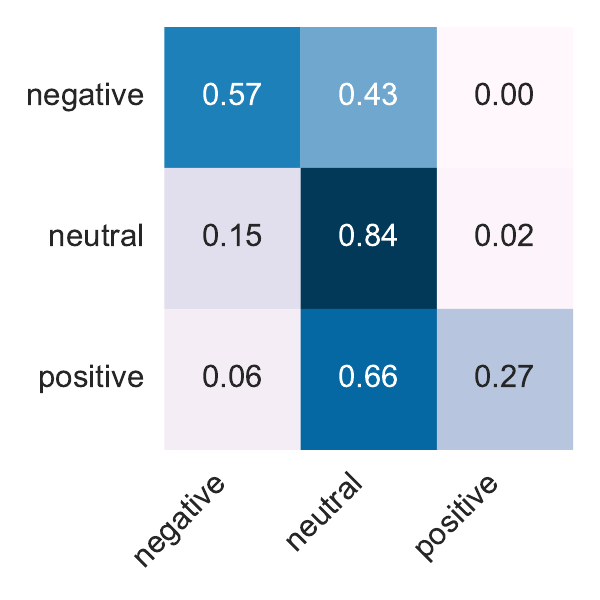}
        \subcaption{\small EIA}
    \end{minipage} \hfill
    \begin{minipage}[t]{0.20\linewidth}
        \centering
        \includegraphics[width=\linewidth]{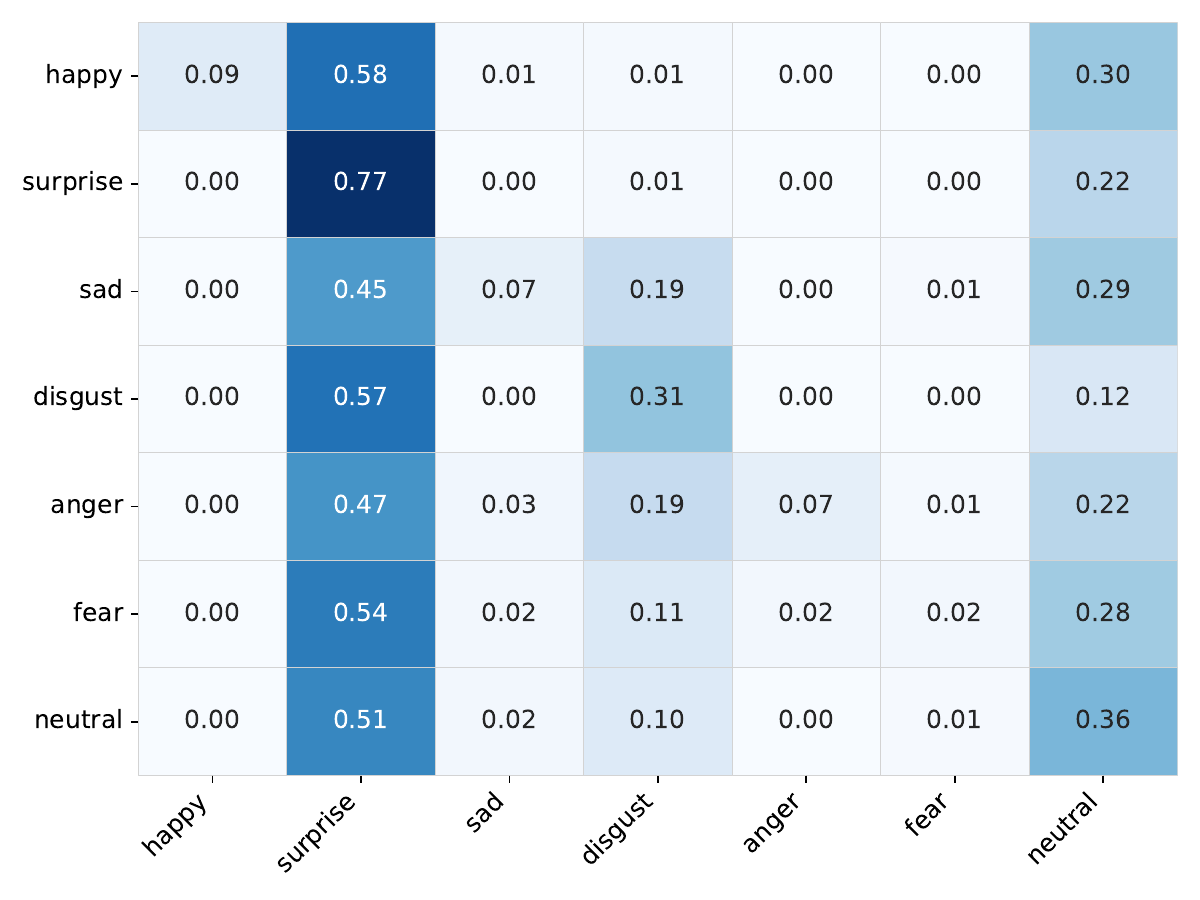}
        \subcaption{\small CEIA (emo.)}
    \end{minipage} \hfill
    \begin{minipage}[t]{0.20\linewidth}
        \centering
        \includegraphics[width=\linewidth]{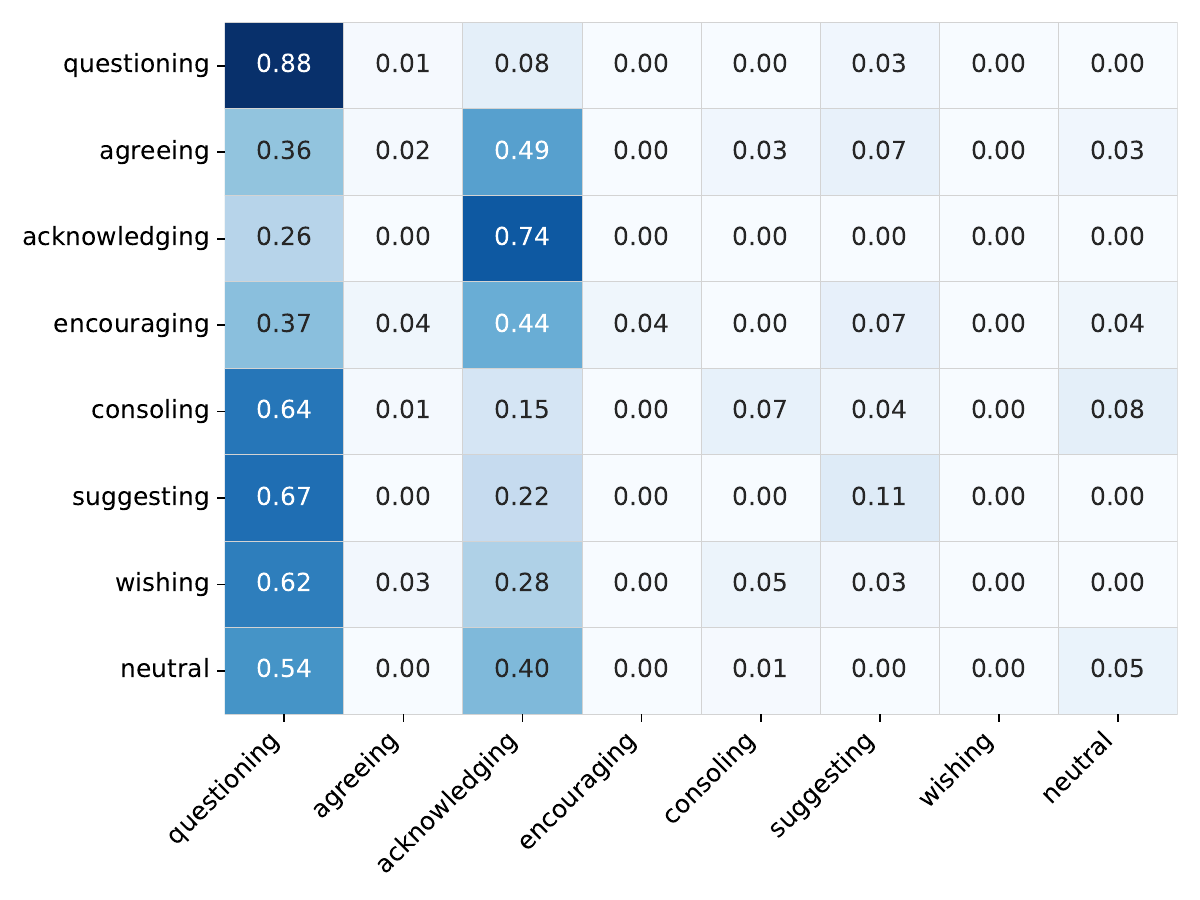}
        \subcaption{\small CEIA (int.)}
    \end{minipage} \hfill
    \begin{minipage}[t]{0.24\linewidth}
        \centering
        \includegraphics[width=\linewidth]{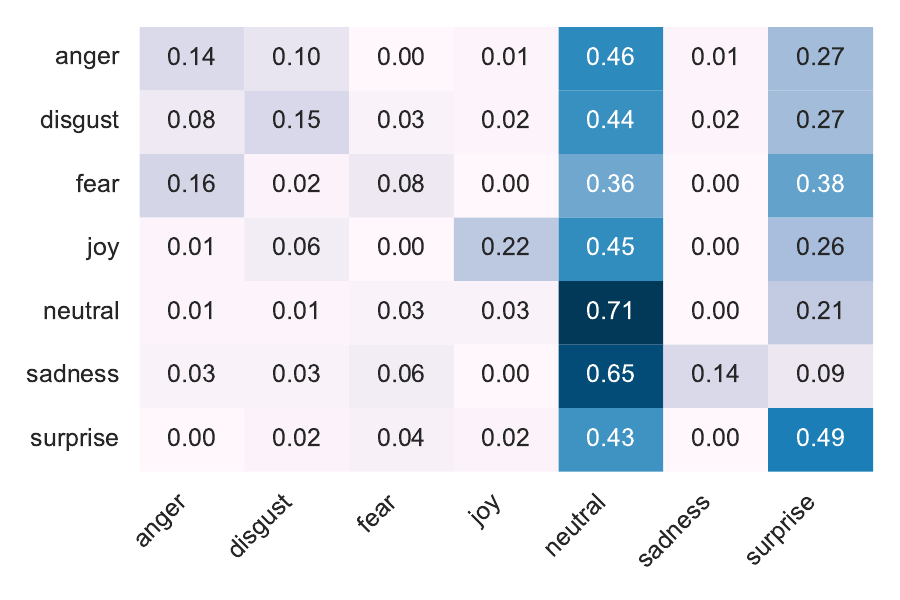}
        \subcaption{\small MPDER}
    \end{minipage} \hfill
    \begin{minipage}[t]{0.16\linewidth}
        \centering
        \includegraphics[width=\linewidth]{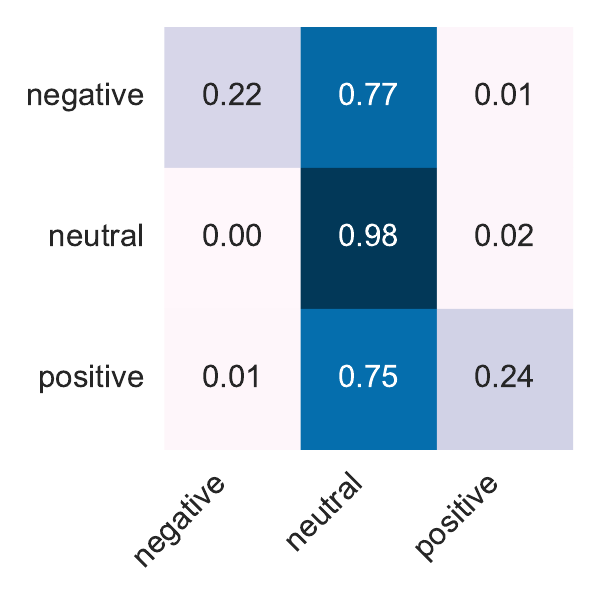}
        \subcaption{\small PEA}
    \end{minipage}

    \vspace{-2mm}
    \caption{\small Confusion matrices for VideoLLaMA2-7B-16F on each evaluation scenario of $\ours$.}
    \label{fig:confusion-VideoLLaMA2-7B-16F}
\end{figure*}

\begin{figure*}[!t]
    \centering
    
    % ===== Row 1 =====
    \begin{minipage}[t]{0.24\linewidth}
        \centering
        \includegraphics[width=\linewidth]{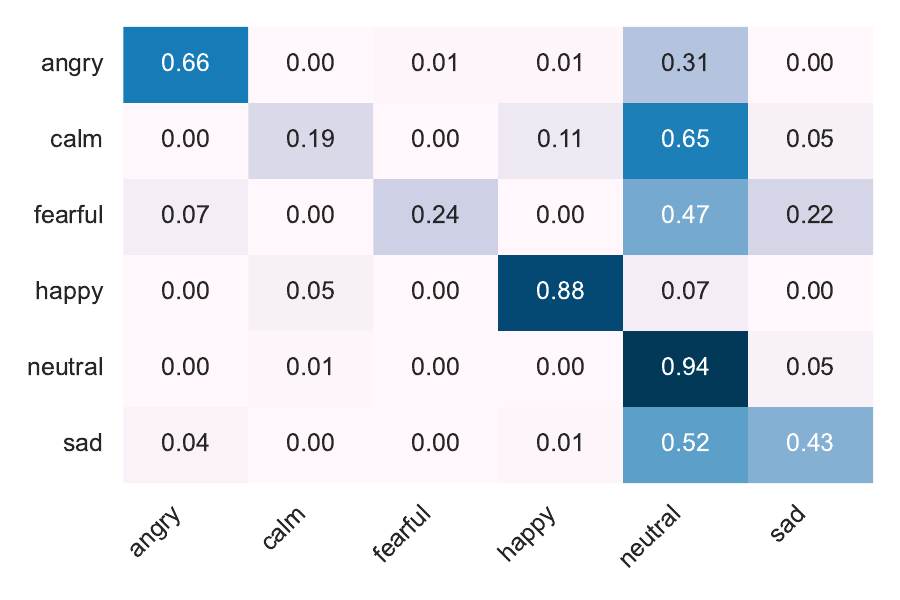}
        \subcaption{\small SOER}
    \end{minipage} \hfill
    \begin{minipage}[t]{0.24\linewidth}
        \centering
        \includegraphics[width=\linewidth]{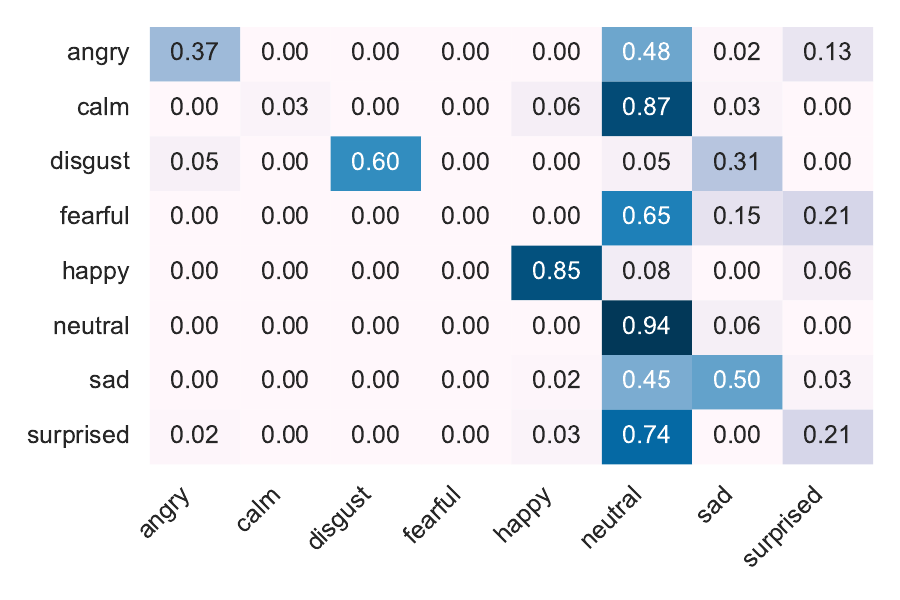}
        \subcaption{\small SPER}
    \end{minipage} \hfill
    \begin{minipage}[t]{0.16\linewidth}
        \centering
        \includegraphics[width=\linewidth]{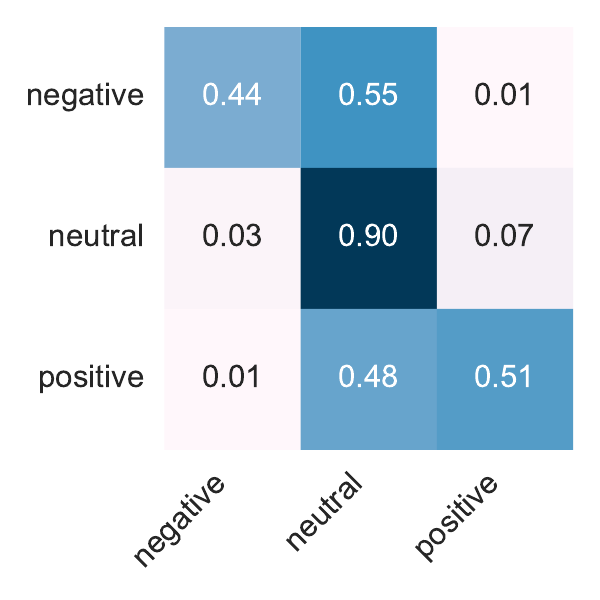}
        \subcaption{\small OSA}
    \end{minipage} \hfill
    \begin{minipage}[t]{0.16\linewidth}
        \centering
        \includegraphics[width=\linewidth]{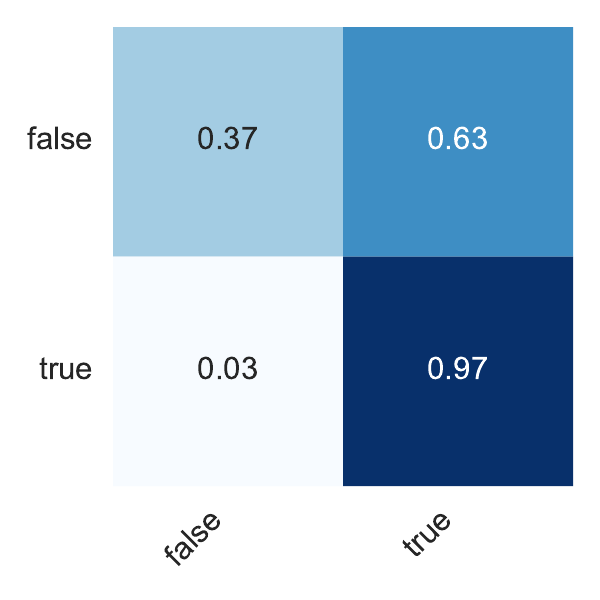}
        \subcaption{\small HU}
    \end{minipage} \\

    % ===== Row 2 =====
    \begin{minipage}[t]{0.24\linewidth}
        \centering
        \includegraphics[width=\linewidth]{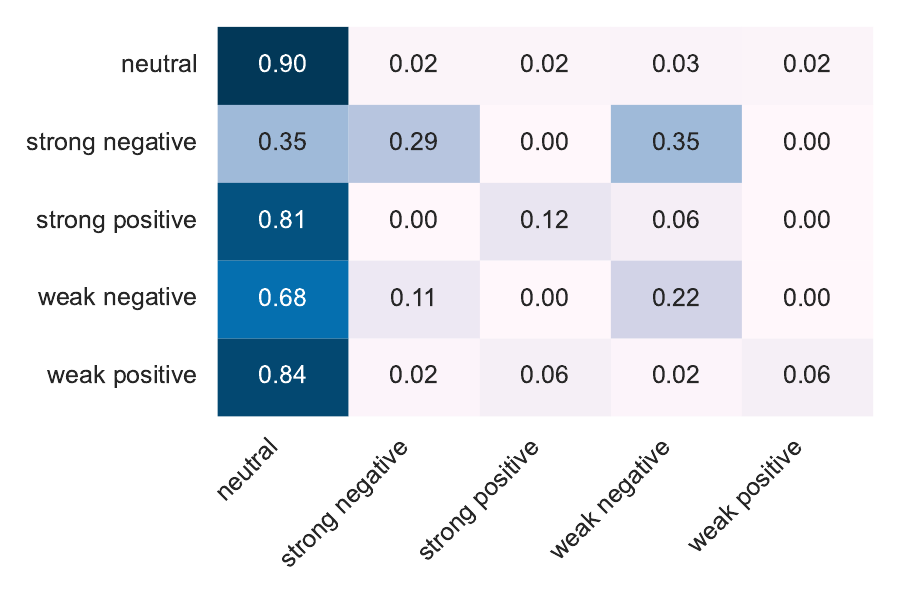}
        \subcaption{\small SCEA}
    \end{minipage} \hfill
    \begin{minipage}[t]{0.24\linewidth}
        \centering
        \includegraphics[width=\linewidth]{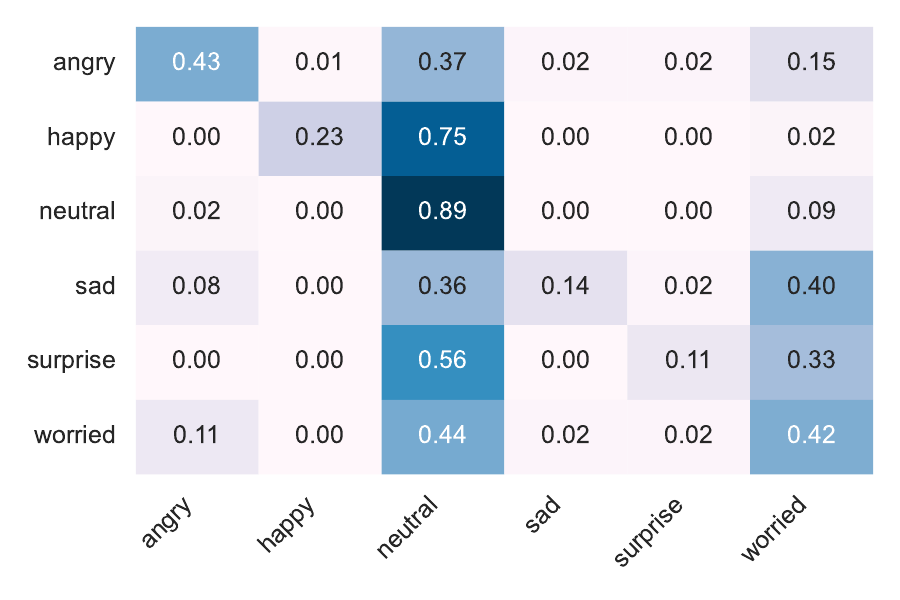}
        \subcaption{\small FGDEA}
    \end{minipage} \hfill
    \begin{minipage}[t]{0.16\linewidth}
        \centering
        \includegraphics[width=\linewidth]{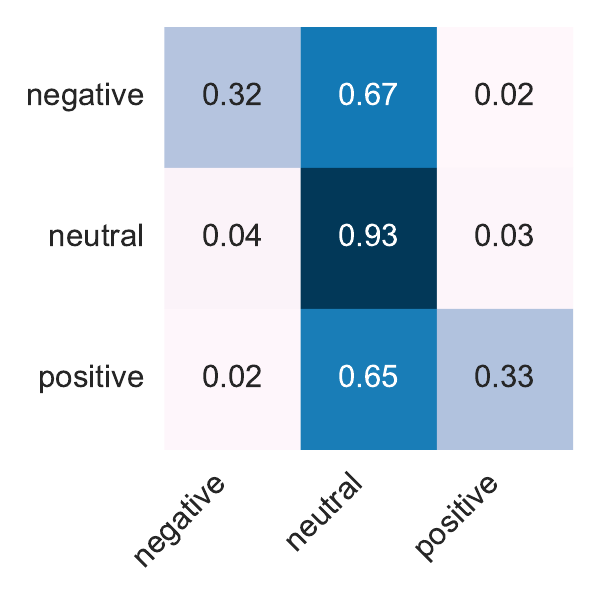}
        \subcaption{\small FCDEA}
    \end{minipage} \hfill
    \begin{minipage}[t]{0.16\linewidth}
        \centering
        \includegraphics[width=\linewidth]{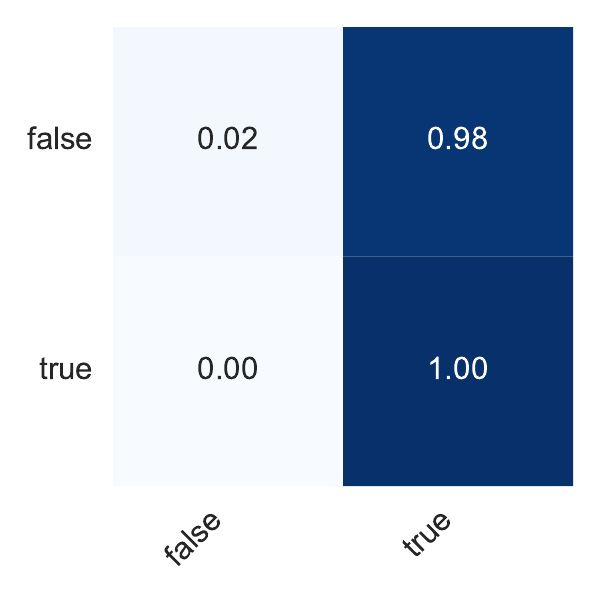}
        \subcaption{\small SD}
    \end{minipage} \\

    % ===== Row 3 =====
    \begin{minipage}[t]{0.16\linewidth}
        \centering
        \includegraphics[width=\linewidth]{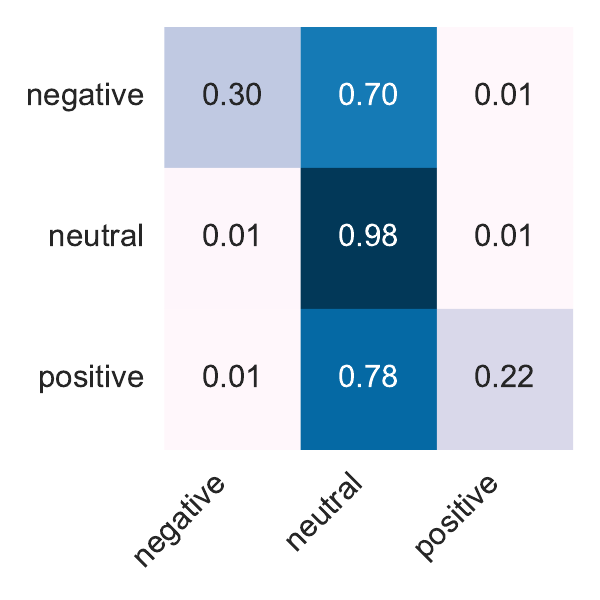}
        \subcaption{\small EIA}
    \end{minipage} \hfill
    \begin{minipage}[t]{0.20\linewidth}
        \centering
        \includegraphics[width=\linewidth]{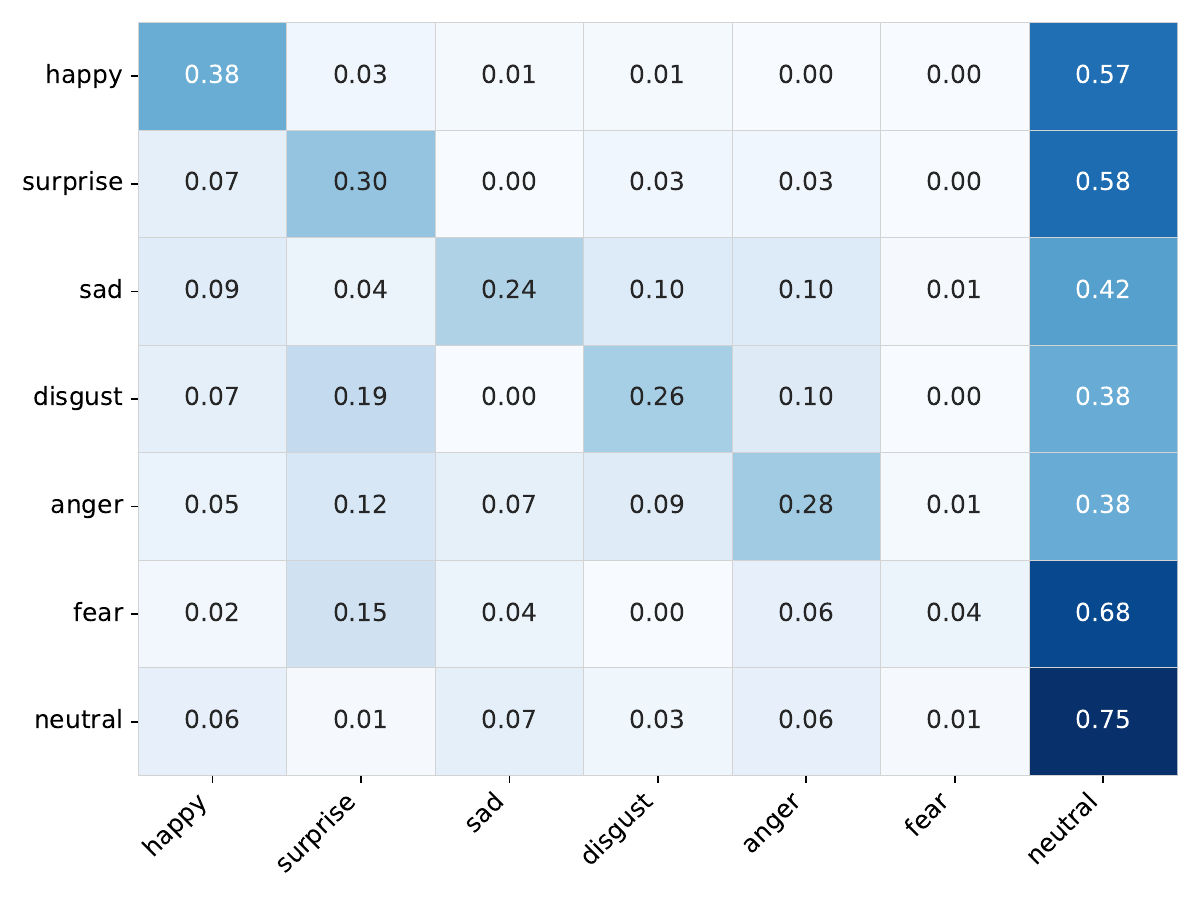}
        \subcaption{\small CEIA (emo.)}
    \end{minipage} \hfill
    \begin{minipage}[t]{0.20\linewidth}
        \centering
        \includegraphics[width=\linewidth]{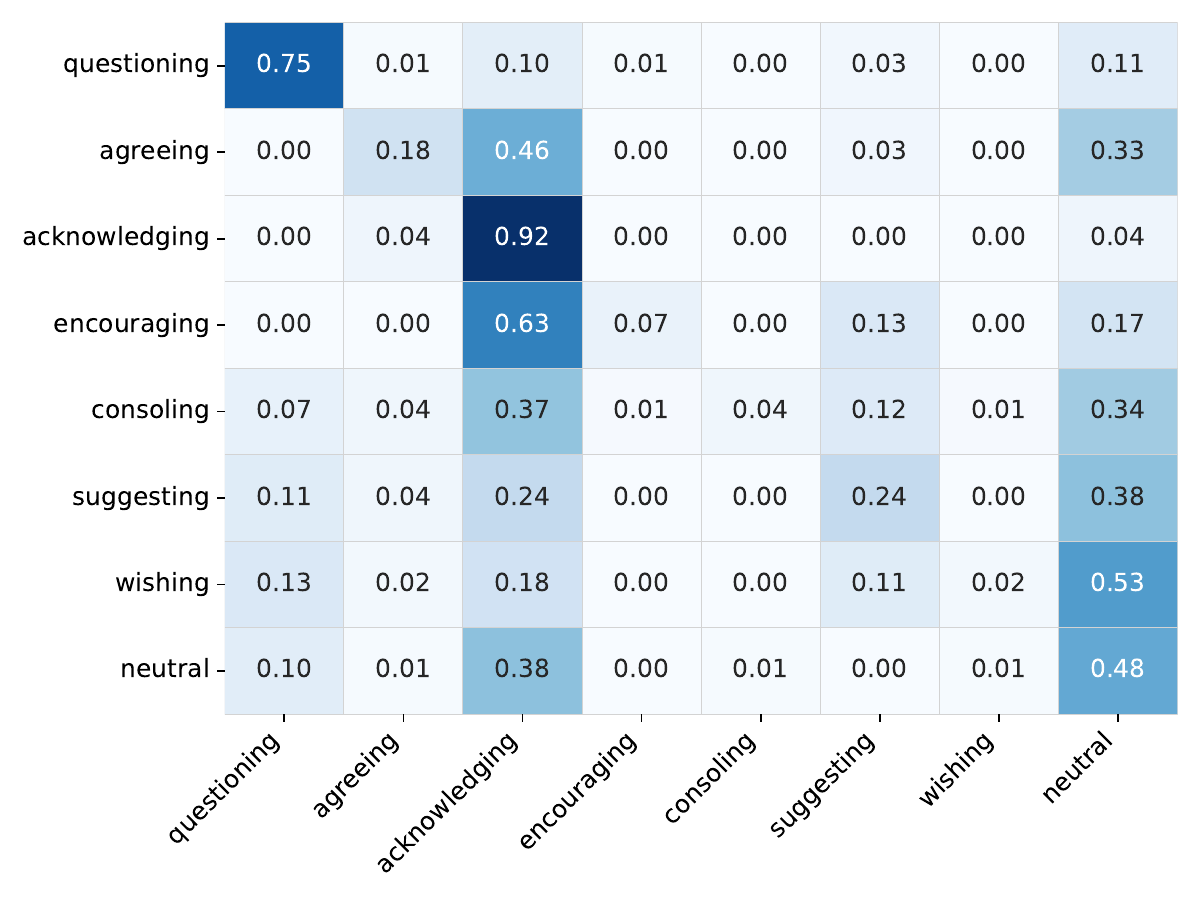}
        \subcaption{\small CEIA (int.)}
    \end{minipage} \hfill
    \begin{minipage}[t]{0.24\linewidth}
        \centering
        \includegraphics[width=\linewidth]{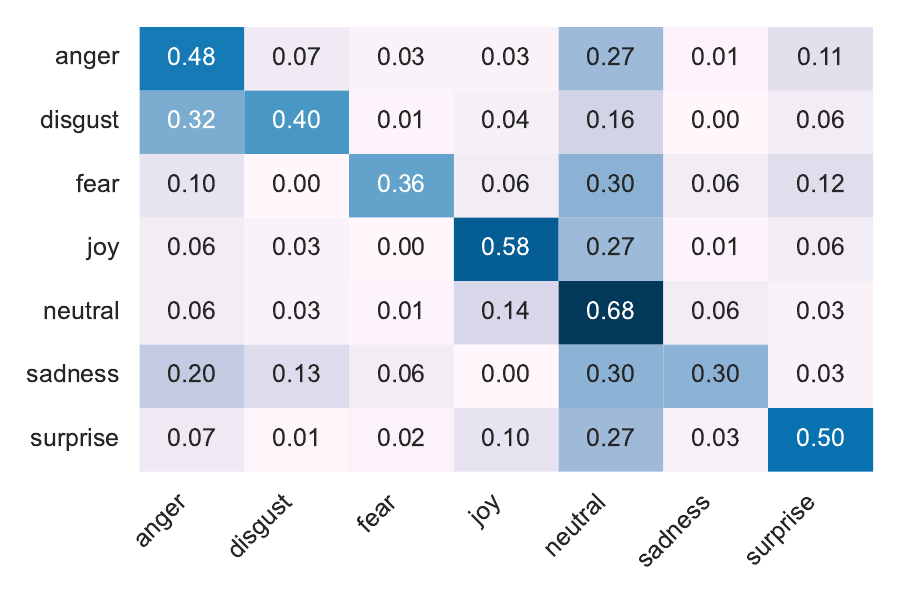}
        \subcaption{\small MPDER}
    \end{minipage} \hfill
    \begin{minipage}[t]{0.16\linewidth}
        \centering
        \includegraphics[width=\linewidth]{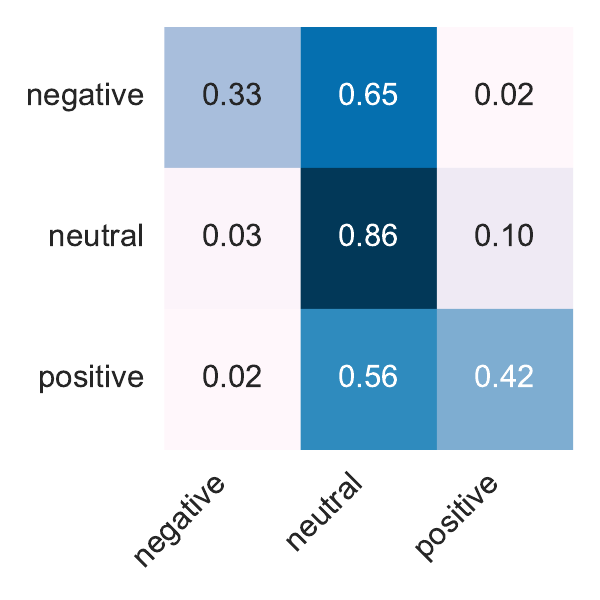}
        \subcaption{\small PEA}
    \end{minipage}

    \vspace{-2mm}
    \caption{\small Confusion matrices for VideoLLaMA2-72B on each evaluation scenario of $\ours$.}
    \label{fig:confusion-VideoLLaMA2-72B}
\end{figure*}

\begin{figure*}[!t]
    \centering
    
    % ===== Row 1 =====
    \begin{minipage}[t]{0.24\linewidth}
        \centering
        \includegraphics[width=\linewidth]{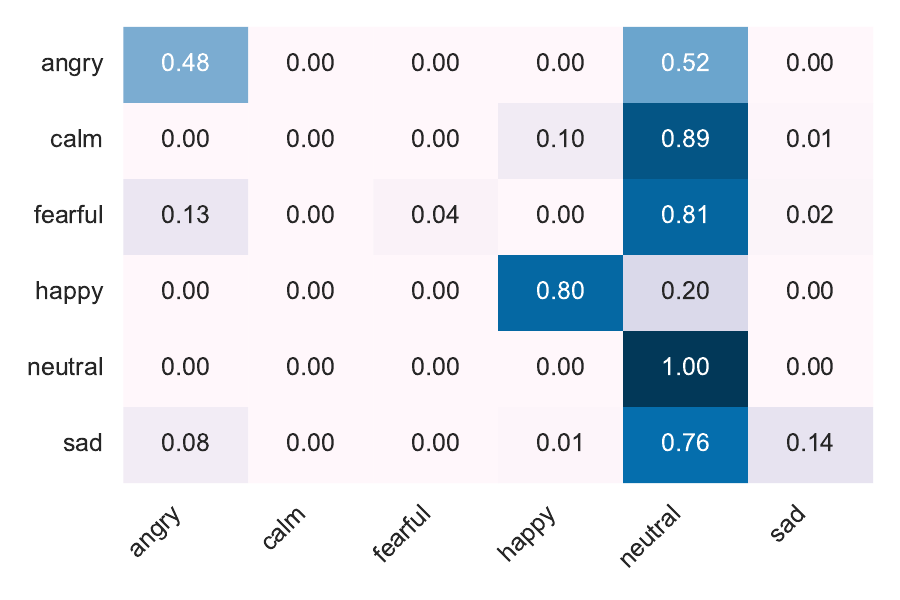}
        \subcaption{\small SOER}
    \end{minipage} \hfill
    \begin{minipage}[t]{0.24\linewidth}
        \centering
        \includegraphics[width=\linewidth]{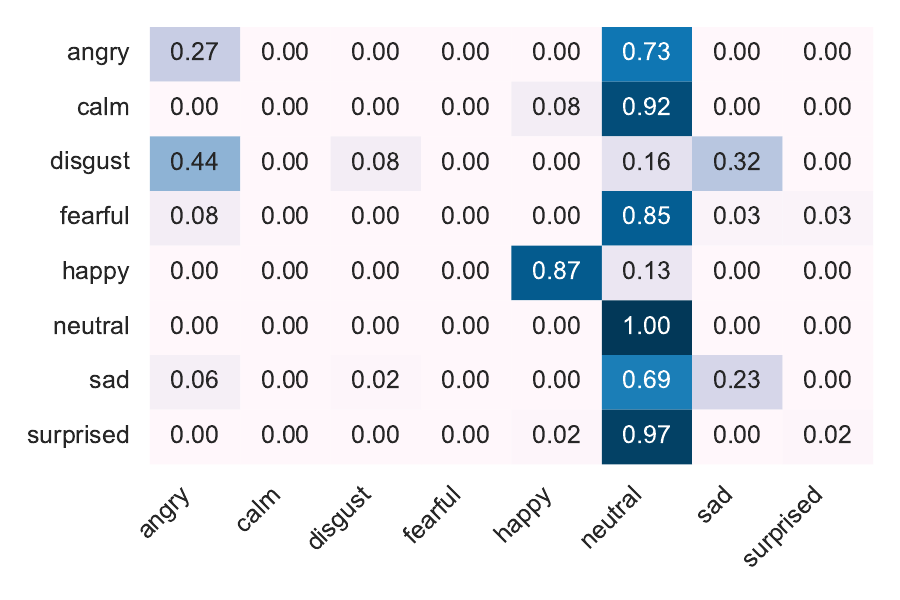}
        \subcaption{\small SPER}
    \end{minipage} \hfill
    \begin{minipage}[t]{0.16\linewidth}
        \centering
        \includegraphics[width=\linewidth]{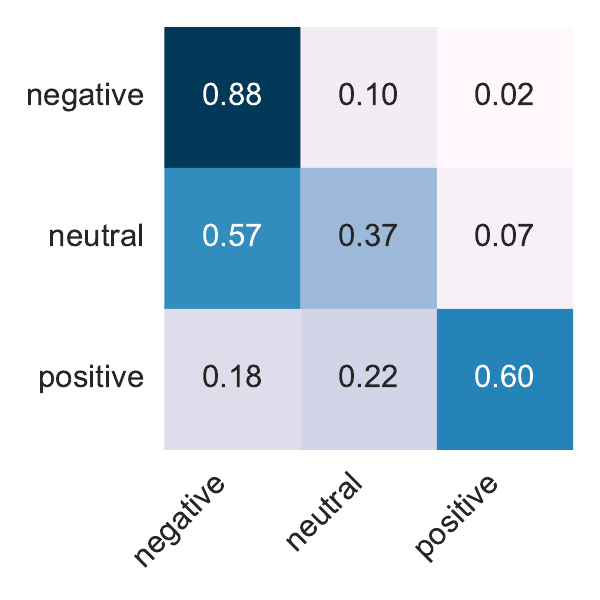}
        \subcaption{\small OSA}
    \end{minipage} \hfill
    \begin{minipage}[t]{0.16\linewidth}
        \centering
        \includegraphics[width=\linewidth]{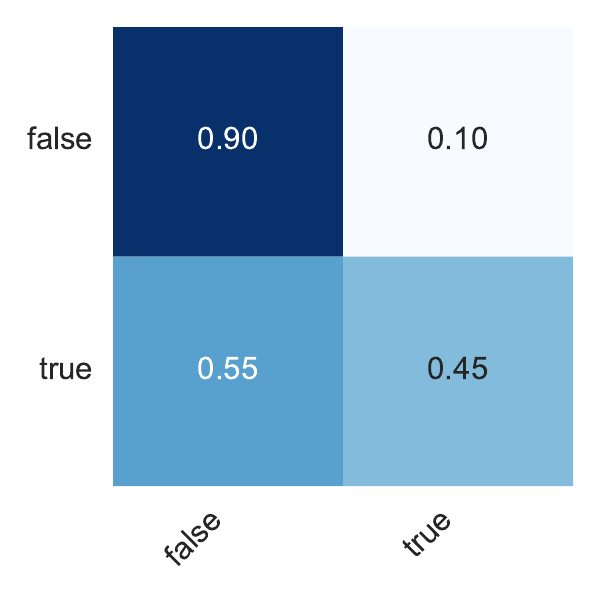}
        \subcaption{\small HU}
    \end{minipage} \\

    % ===== Row 2 =====
    \begin{minipage}[t]{0.24\linewidth}
        \centering
        \includegraphics[width=\linewidth]{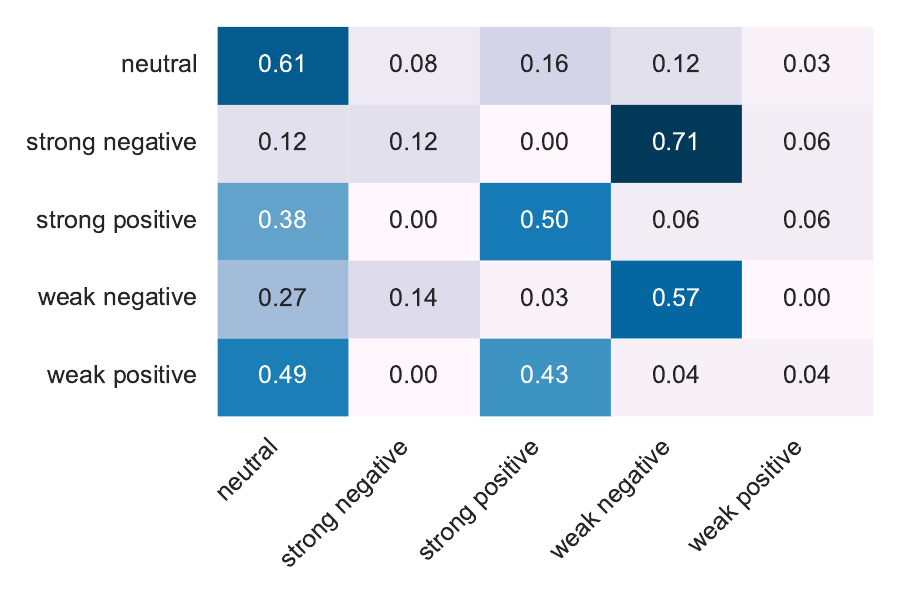}
        \subcaption{\small SCEA}
    \end{minipage} \hfill
    \begin{minipage}[t]{0.24\linewidth}
        \centering
        \includegraphics[width=\linewidth]{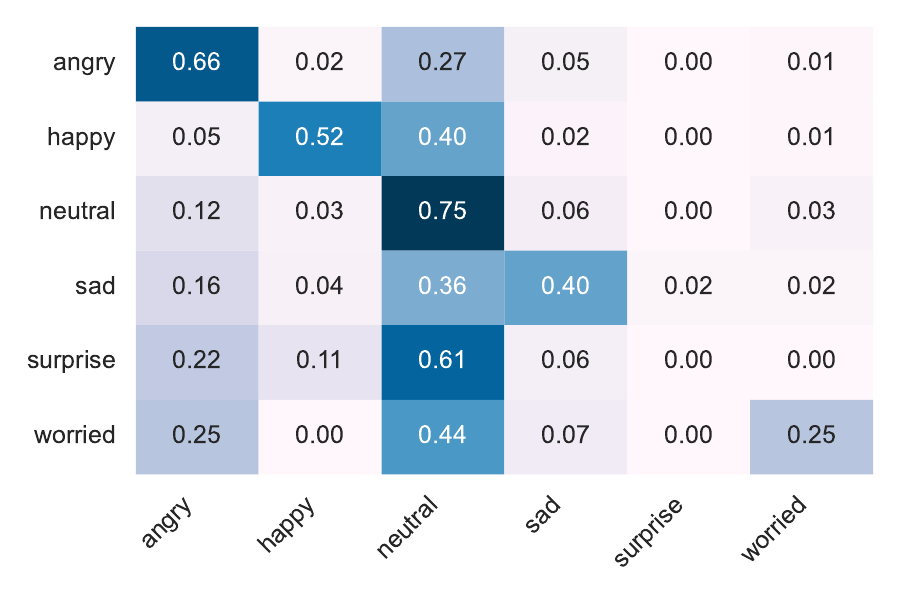}
        \subcaption{\small FGDEA}
    \end{minipage} \hfill
    \begin{minipage}[t]{0.16\linewidth}
        \centering
        \includegraphics[width=\linewidth]{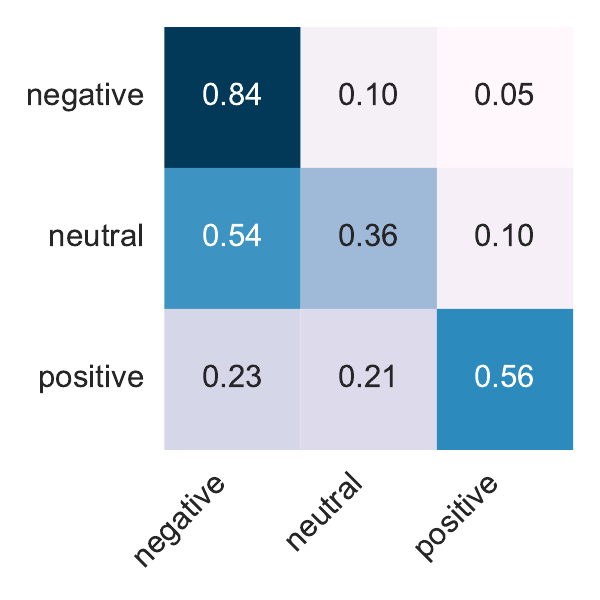}
        \subcaption{\small FCDEA}
    \end{minipage} \hfill
    \begin{minipage}[t]{0.16\linewidth}
        \centering
        \includegraphics[width=\linewidth]{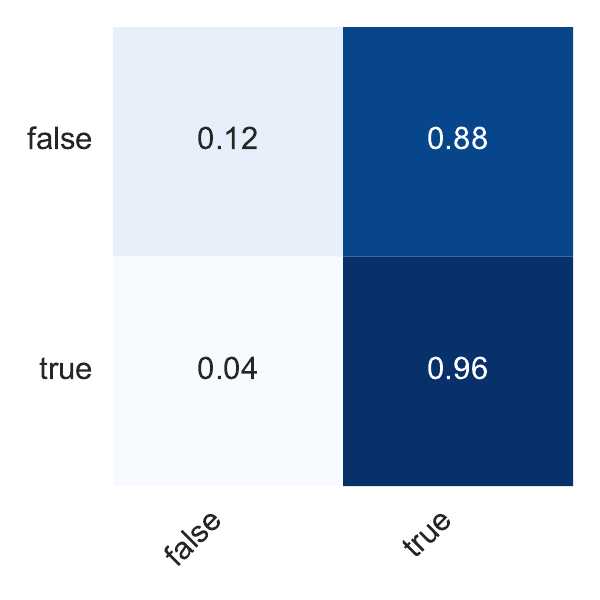}
        \subcaption{\small SD}
    \end{minipage} \\

    % ===== Row 3 =====
    \begin{minipage}[t]{0.16\linewidth}
        \centering
        \includegraphics[width=\linewidth]{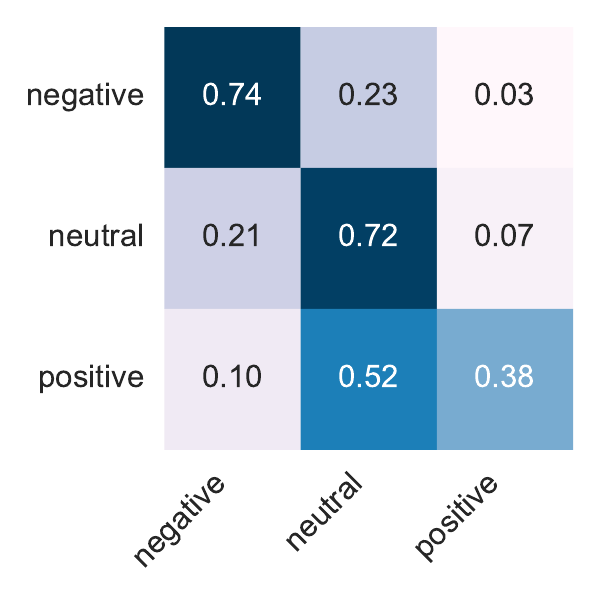}
        \subcaption{\small EIA}
    \end{minipage} \hfill
    \begin{minipage}[t]{0.20\linewidth}
        \centering
        \includegraphics[width=\linewidth]{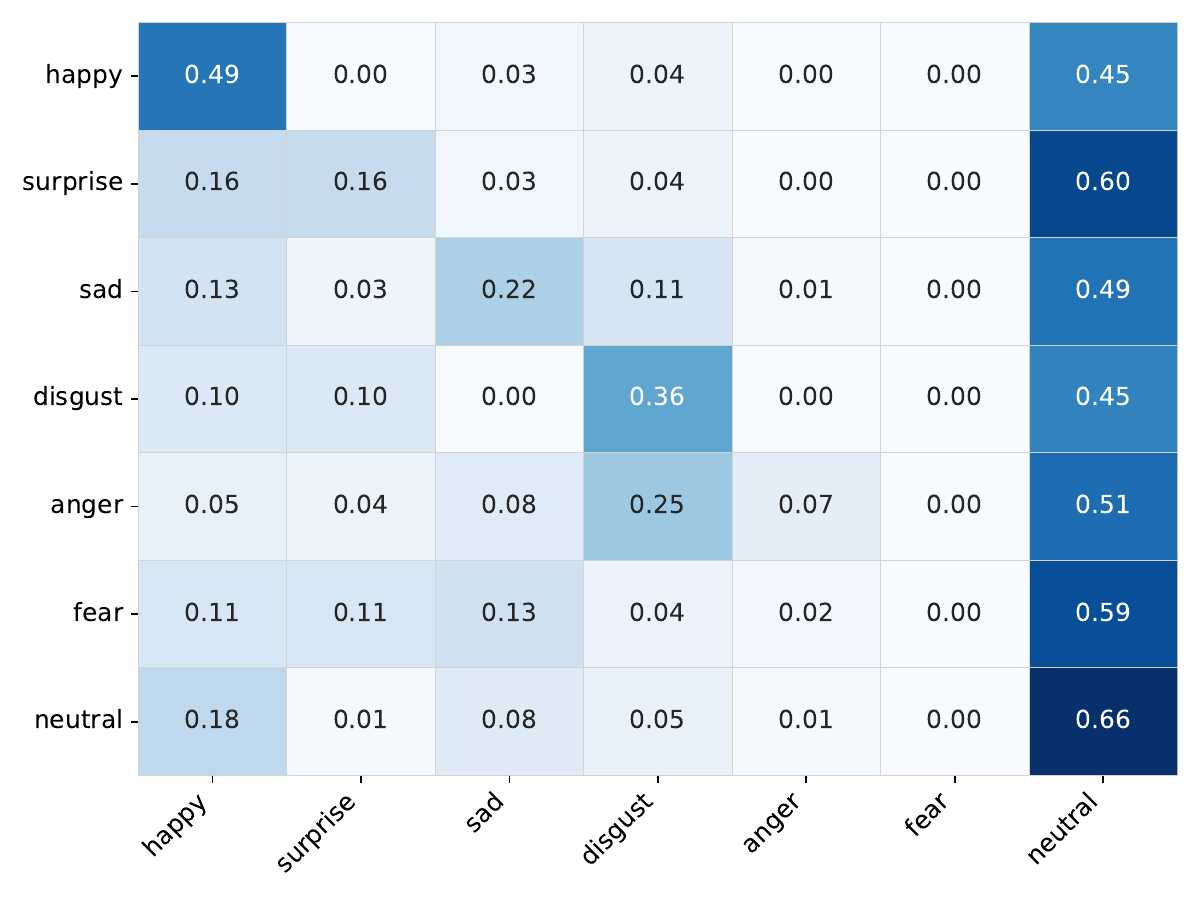}
        \subcaption{\small CEIA (emo.)}
    \end{minipage} \hfill
    \begin{minipage}[t]{0.20\linewidth}
        \centering
        \includegraphics[width=\linewidth]{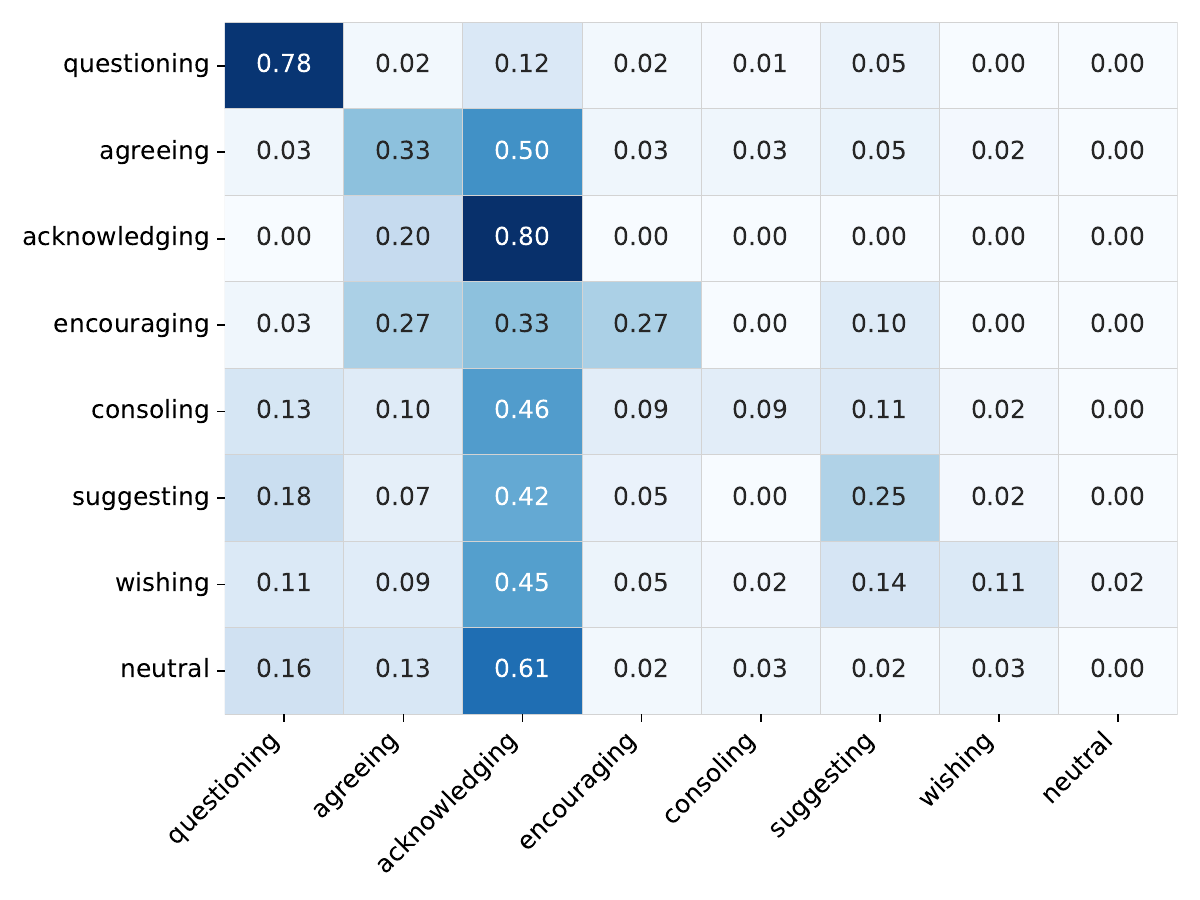}
        \subcaption{\small CEIA (int.)}
    \end{minipage} \hfill
    \begin{minipage}[t]{0.24\linewidth}
        \centering
        \includegraphics[width=\linewidth]{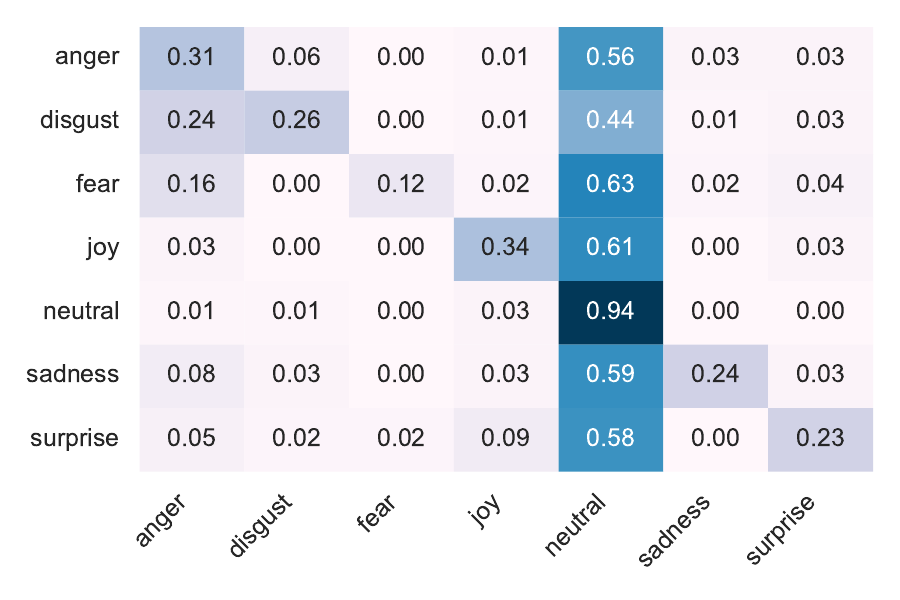}
        \subcaption{\small MPDER}
    \end{minipage} \hfill
    \begin{minipage}[t]{0.16\linewidth}
        \centering
        \includegraphics[width=\linewidth]{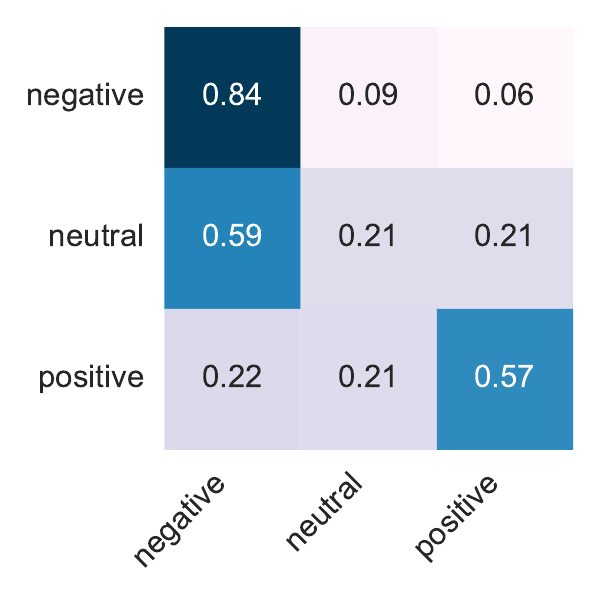}
        \subcaption{\small PEA}
    \end{minipage}

    \vspace{-3mm}
    \caption{\small Confusion matrices for VideoLLaMA2.1-7B-16F on each evaluation scenario of $\ours$.}
    \label{fig:confusion-VideoLLaMA2.1-7B-16F}
\end{figure*}

\begin{figure*}[!t]
    \centering
    
    % ===== Row 1 =====
    \begin{minipage}[t]{0.24\linewidth}
        \centering
        \includegraphics[width=\linewidth]{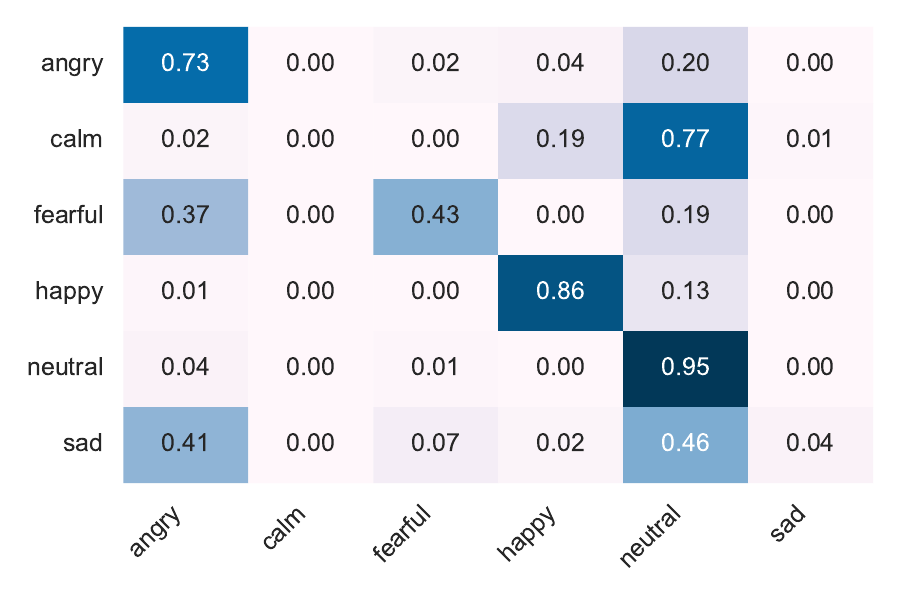}
        \subcaption{\small SOER}
    \end{minipage} \hfill
    \begin{minipage}[t]{0.24\linewidth}
        \centering
        \includegraphics[width=\linewidth]{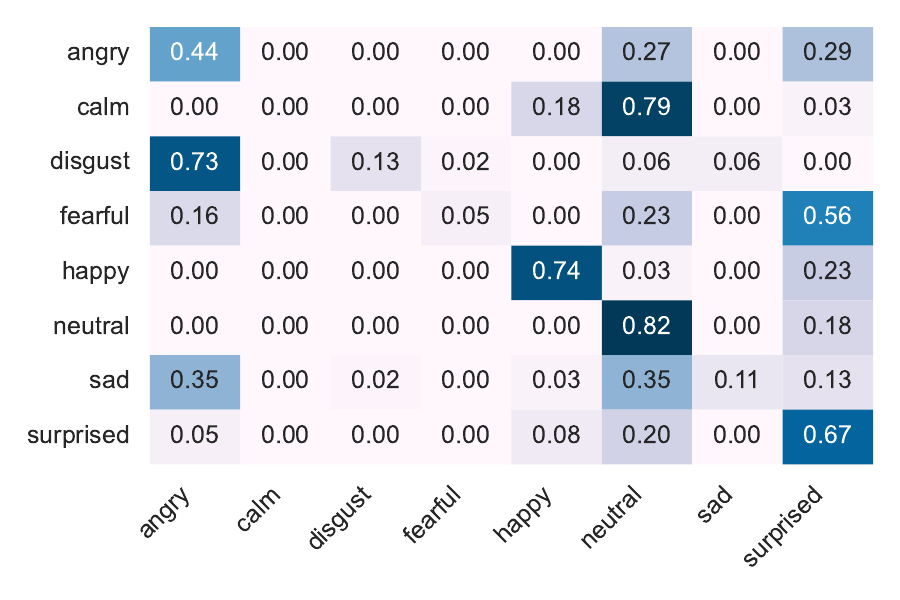}
        \subcaption{\small SPER}
    \end{minipage} \hfill
    \begin{minipage}[t]{0.16\linewidth}
        \centering
        \includegraphics[width=\linewidth]{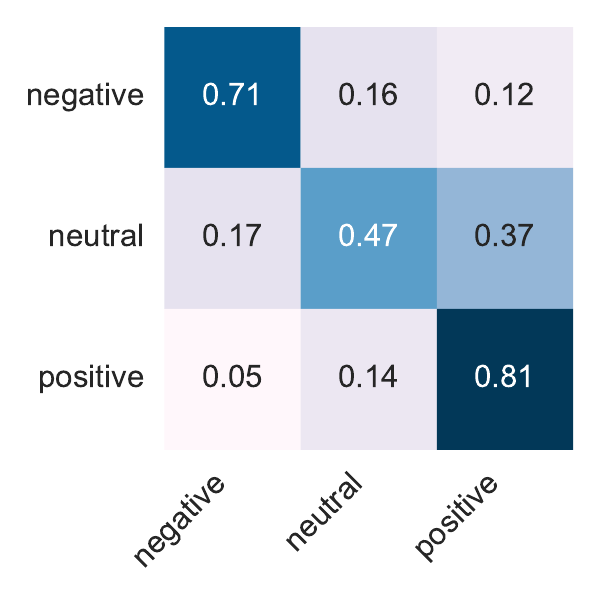}
        \subcaption{\small OSA}
    \end{minipage} \hfill
    \begin{minipage}[t]{0.16\linewidth}
        \centering
        \includegraphics[width=\linewidth]{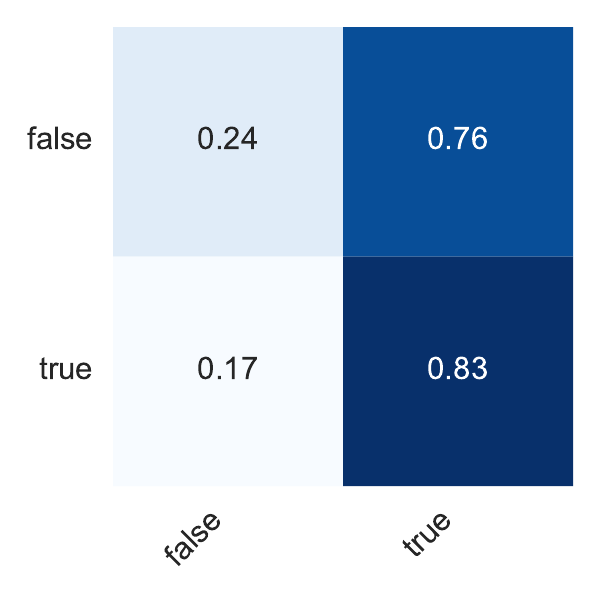}
        \subcaption{\small HU}
    \end{minipage} \\

    % ===== Row 2 =====
    \begin{minipage}[t]{0.24\linewidth}
        \centering
        \includegraphics[width=\linewidth]{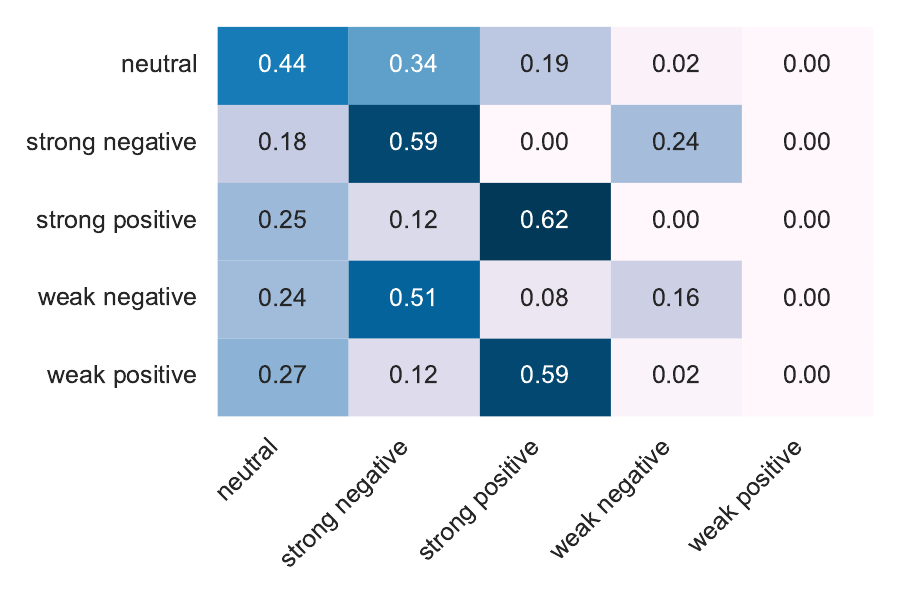}
        \subcaption{\small SCEA}
    \end{minipage} \hfill
    \begin{minipage}[t]{0.24\linewidth}
        \centering
        \includegraphics[width=\linewidth]{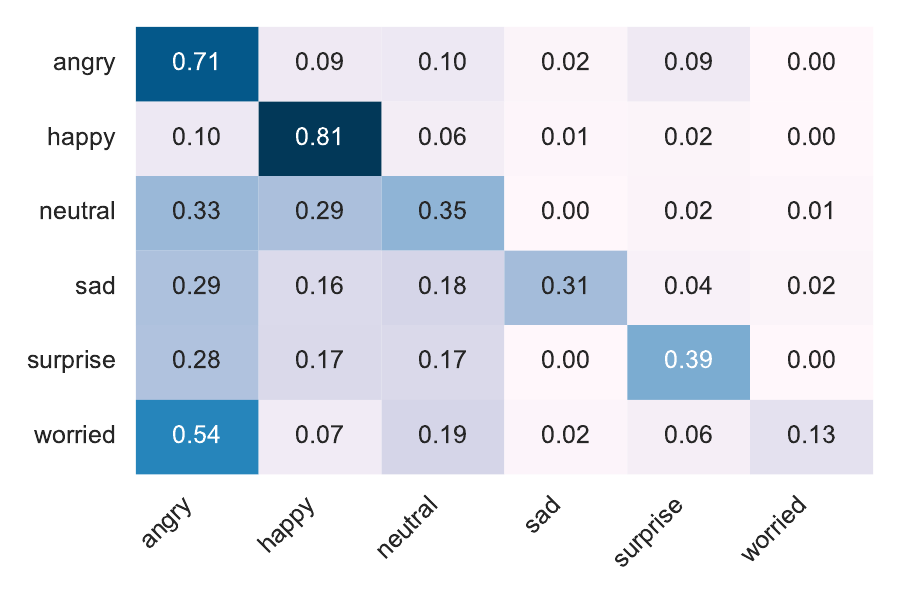}
        \subcaption{\small FGDEA}
    \end{minipage} \hfill
    \begin{minipage}[t]{0.16\linewidth}
        \centering
        \includegraphics[width=\linewidth]{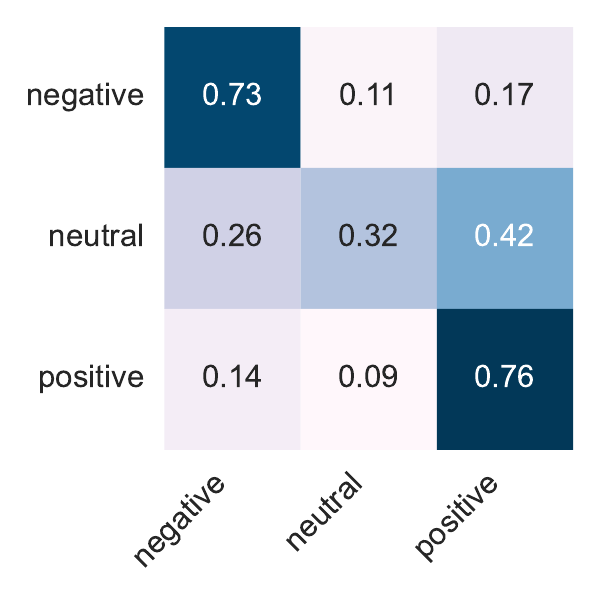}
        \subcaption{\small FCDEA}
    \end{minipage} \hfill
    \begin{minipage}[t]{0.16\linewidth}
        \centering
        \includegraphics[width=\linewidth]{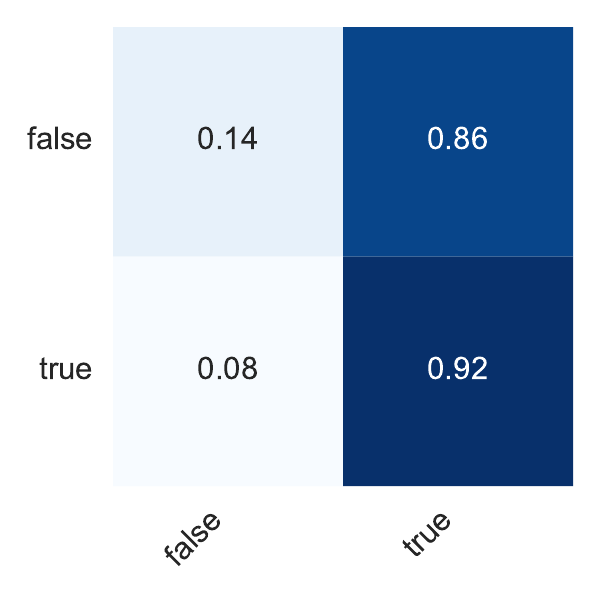}
        \subcaption{\small SD}
    \end{minipage} \\

    % ===== Row 3 =====
    \begin{minipage}[t]{0.16\linewidth}
        \centering
        \includegraphics[width=\linewidth]{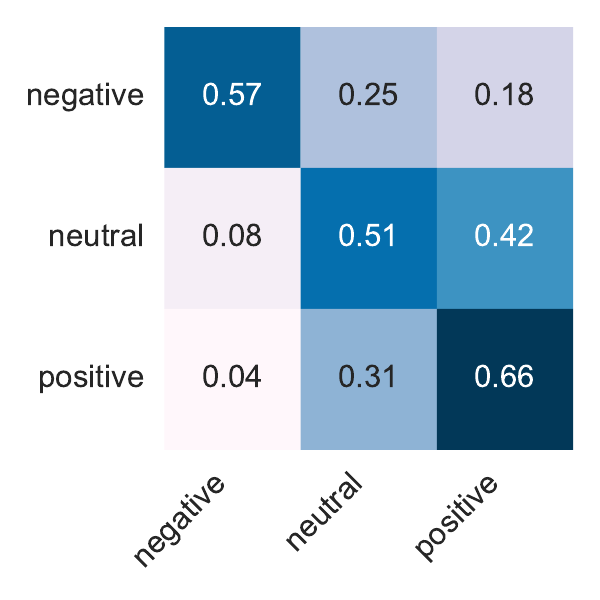}
        \subcaption{\small EIA}
    \end{minipage} \hfill
    \begin{minipage}[t]{0.20\linewidth}
        \centering
        \includegraphics[width=\linewidth]{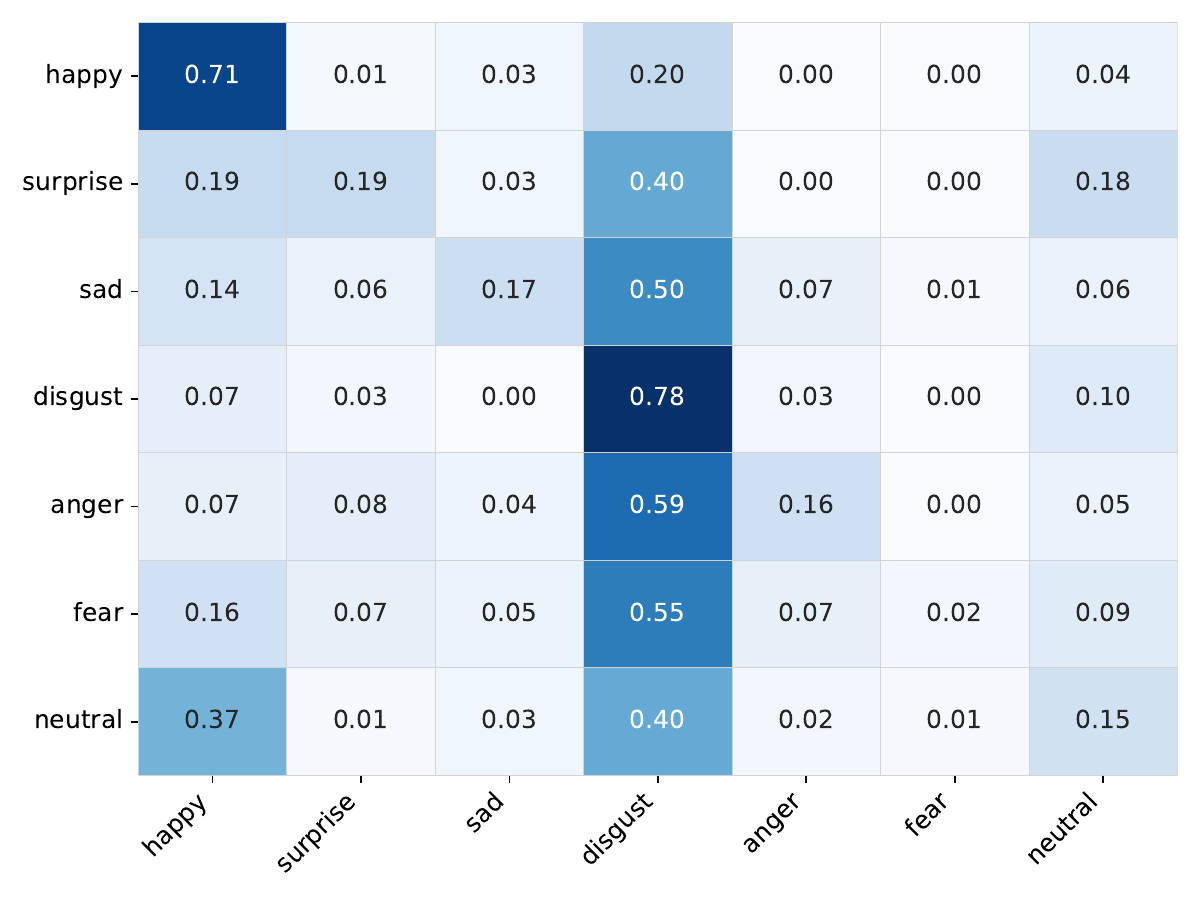}
        \subcaption{\small CEIA (emo.)}
    \end{minipage} \hfill
    \begin{minipage}[t]{0.20\linewidth}
        \centering
        \includegraphics[width=\linewidth]{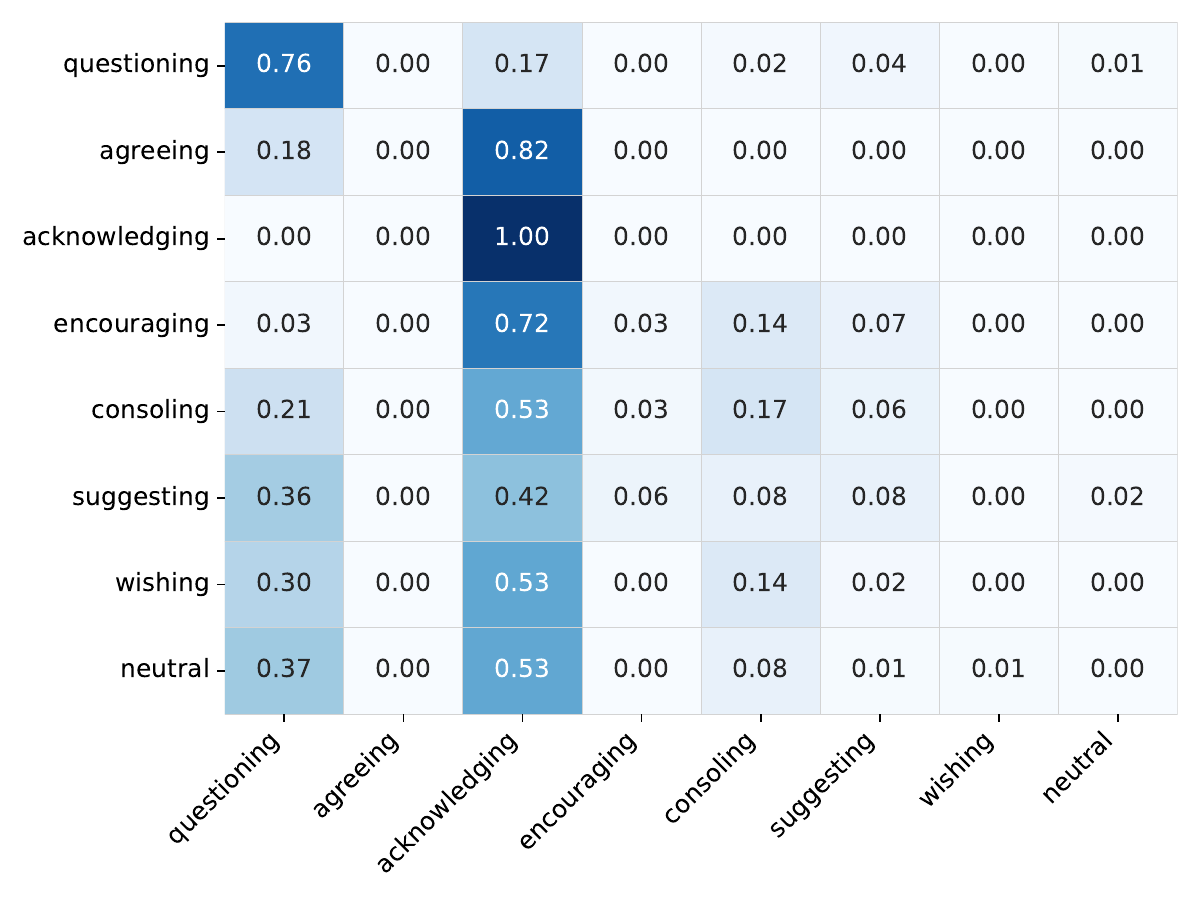}
        \subcaption{\small CEIA (int.)}
    \end{minipage} \hfill
    \begin{minipage}[t]{0.24\linewidth}
        \centering
        \includegraphics[width=\linewidth]{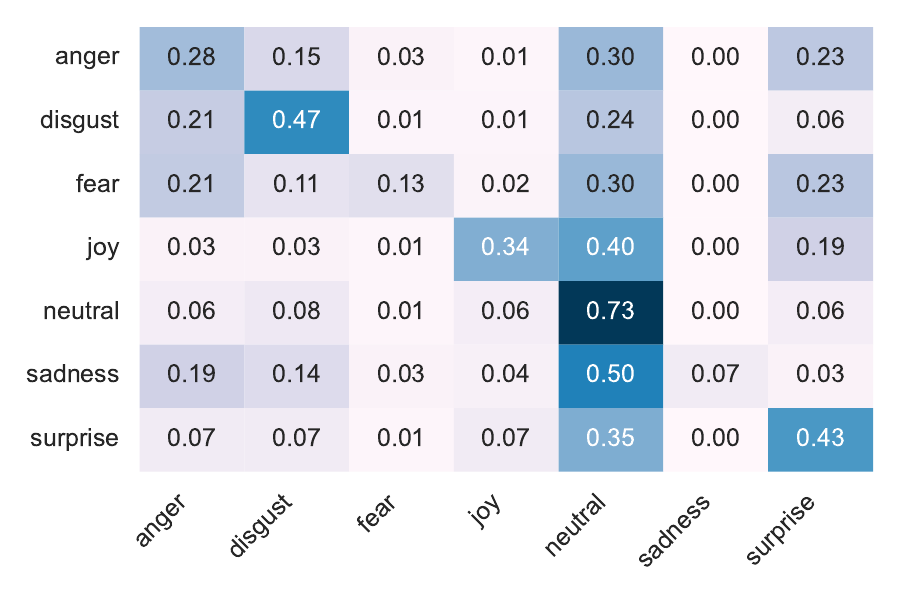}
        \subcaption{\small MPDER}
    \end{minipage} \hfill
    \begin{minipage}[t]{0.16\linewidth}
        \centering
        \includegraphics[width=\linewidth]{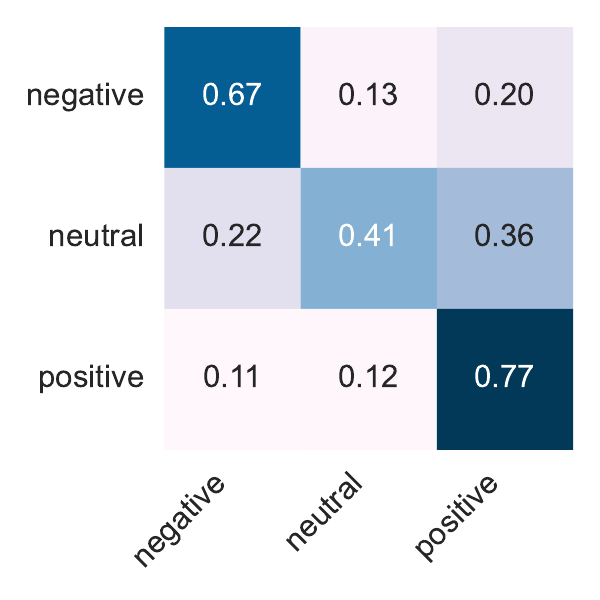}
        \subcaption{\small PEA}
    \end{minipage}

    \vspace{-2mm}
    \caption{\small Confusion matrices for VideoLLaMA2.1-7B-AV on each evaluation scenario of $\ours$.}
    \label{fig:confusion-VideoLLaMA2.1-7B-AV}
\end{figure*}

For class-wise performance, confusion matrices for more models can be found in Figure~\ref{fig:confusion-GPT5.2}-\ref{fig:confusion-VideoLLaMA2.1-7B-AV}.

\clearpage

\clearpage

\section{Prompt}\label{app:prompt}
\begin{tcolorbox}[
    colback=myblue!5!white,
    colframe=myblue!75!black,
    arc=1mm, 
    auto outer arc,
    title={Prompt From RAVDESS(song)},
    breakable
    ]\small
    
Please watch the provided video and determine the emotion it conveys. Do not provide any additional explanations or extra content. Choose one of the following labels as your final answer:  neutral, calm, happy, sad, angry, fearful. Respond in the format: \{'emotion': 'label'\}.

\end{tcolorbox}

\begin{tcolorbox}[
    colback=myblue!5!white,
    colframe=myblue!75!black,
    arc=1mm, 
    auto outer arc,
    title={Prompt From RAVDESS(speech)},
    breakable
    ]\small
    
Please watch the provided video and determine the emotion it conveys.Do not provide any additional explanations or extra content. Choose one of the following labels as your final answer: neutral, calm, happy, sad, angry, fearful, surprised, disgust. Respond in the format:\{'emotion': 'label'\}.  

\end{tcolorbox}

\begin{tcolorbox}[
    colback=myblue!5!white,
    colframe=myblue!75!black,
    arc=1mm, 
    auto outer arc,
    title={Prompt From CH-SIMSv2},
    breakable
    ]\small
    
The person in video says: \{Subtitle\}. Determine the emotion conveyed. Do not provide any additional explanations or extra content. Choose one of the following labels as your final answer: neutral, negative, positive. Respond in the format: \{'emotion': 'label'\}.  

\end{tcolorbox}

\begin{tcolorbox}[
    colback=myblue!5!white,
    colframe=myblue!75!black,
    arc=1mm, 
    auto outer arc,
    title={Prompt From FMSA-SC},
    breakable
    ]\small
    
The person in video says: \{Subtitle\}. Determine the emotion conveyed. Do not provide any additional explanations or extra content. Choose one of the following labels as your final answer: weak negative, strong negative, neutral, strong positive, weak positive. Respond in the format: \{'emotion': 'label'\}.  

\end{tcolorbox}

\begin{tcolorbox}[
    colback=myblue!5!white,
    colframe=myblue!75!black,
    arc=1mm, 
    auto outer arc,
    title={Prompt From UR-FUNNY },
    breakable
    ]\small
    
The context sentences in the video is: \{context sentences\}. The punchline sentence in the video is: \{punchline sentence\}. Choose one of the following labels as your final answer: true, false.  

\end{tcolorbox}

\begin{tcolorbox}[
    colback=myblue!5!white,
    colframe=myblue!75!black,
    arc=1mm, 
    auto outer arc,
    title={Prompt From MC-EIU },
    breakable
    ]\small
    
The person in video says:\{Subtitle\}. Analyze the emotion and intent. Choose one emotion: happy, surprise, sad, disgust, anger, fear, and neutral. Choose one intent: questioning, agreeing, acknowledging, encouraging, consoling, suggesting, wishing, and neutral. Respond in the format: \{'emotion label': 'label', 'intent label': 'label'\}. 

\end{tcolorbox}

\begin{tcolorbox}[
    colback=myblue!5!white,
    colframe=myblue!75!black,
    arc=1mm, 
    auto outer arc,
    title={Prompt From MELD },
    breakable
    ]\small
    
The person in video says: \{subtitle\}. Do not provide any additional explanations or extra content. Choose one of the following labels as your final answer: neutral, surprise, fear, sadness, joy, disgust, anger. Respond in the format: \{'emotion': 'label'\}. 

\end{tcolorbox}

\begin{tcolorbox}[
    colback=myblue!5!white,
    colframe=myblue!75!black,
    arc=1mm, 
    auto outer arc,
    title={Prompt From MER2023 },
    breakable
    ]\small
    
The person in video says: \{Subtitle\}. Do not provide any additional explanations or extra content. Choose one of the following labels as your final answer: happy, sad, neutral, angry, worried, surprise. Respond in the format: \{'emotion': 'label'\}. 

\end{tcolorbox}

\begin{tcolorbox}[
    colback=myblue!5!white,
    colframe=myblue!75!black,
    arc=1mm, 
    auto outer arc,
    title={Prompt From CMU-MOSI  },
    breakable
    ]\small
    
The person in video says: \{Subtitle\}. Do not provide any additional explanations or extra content. Choose one of the following labels as your final answer:  neutral, negative, positive. Respond in the format: \{'emotion': 'label'\}. 

\end{tcolorbox}

\begin{tcolorbox}[
    colback=myblue!5!white,
    colframe=myblue!75!black,
    arc=1mm, 
    auto outer arc,
    title={Prompt From CMU-MOSEI   },
    breakable
    ]\small
    
The person in video says: \{Subtitle\}. Do not provide any additional explanations or extra content. Choose one of the following labels as your final answer:  neutral, negative, positive. Respond in the format: \{'emotion': 'label'\}. 

\end{tcolorbox}

\begin{tcolorbox}[
    colback=myblue!5!white,
    colframe=myblue!75!black,
    arc=1mm, 
    auto outer arc,
    title={Prompt From MUStARD   },
    breakable
    ]\small
    
The person in video says: \{Subtitle\}. Does this statement express sarcasm? Do not provide any additional explanations or extra content. Choose one of the following labels as your final answer: true, false. 

\end{tcolorbox}

\begin{tcolorbox}[
    colback=myblue!5!white,
    colframe=myblue!75!black,
    arc=1mm, 
    auto outer arc,
    title={Prompt From CH-SIMS  },
    breakable
    ]\small
    
The person in video says: \{Subtitle\}. Choose one of the following labels as your final answer: neutral, negative, positive. Respond in the format: \{'emotion': 'label'\}. 

\end{tcolorbox}

\begin{tcolorbox}[
    colback=myblue!5!white,
    colframe=myblue!75!black,
    arc=1mm, 
    auto outer arc,
    title={Prompt From SMILE    },
    breakable
    ]\small
    
Reasoning task: you are to answer why the audience laughed given the video clip. The video clip from
the {Sitcom}, titled \{video title\}, with multimodal information (Utterance, Facial Action Units, Video caption,
Acoustic features(6 dimension; 1.mean of F0 contour, 2.var of F0 contour, 3. mean of energy contour, 4. var of
energy contour, 5. jitter, 6. shimmer)) is given. The audience laughing moment is marked as (audience laughing)
in certain utterances Explain why the audience laughed given the video clip, at most {40} words, starting with
'The audience laughed because '. Given video clip: \{query\}. 
\end{tcolorbox}

\clearpage
\section{Evaluation Prompt for SMILE Dataset}\label{app:evalprompt}
\begin{tcolorbox}[
    colback=myblue!5!white,
    colframe=myblue!75!black,
    arc=1mm, 
    auto outer arc,
    title={Prompt From  Logical Judgment Dimension Evaluation Criteria for
Model-Generated Reasoning   },
    breakable
    ]\small

    You need to evaluate the quality of model-generated reasoning for why a video audience laughed. You will be provided with two reasons for laughter reasoning:

1. The reason for laughter generated by the model.
2. The reference reason for laughter is annotated manually (as a benchmark).

Please score based on the following dimension, with a maximum of 5 points:

\textbf{Logical Judgment Dimension:} Based on the reference reason, evaluate the model-generated reason in terms of logical clarity, the rationality of the causal chain, and coherence with the context.

\textbf{Scoring Criteria:}
\begin{itemize}
    \item \textbf{1 Point:} The reasoning lacks logic, with unclear or missing causal relationships, and is incoherent with the context.
    \item \textbf{2 Points:} The reasoning has some logical flaws and partial causal connections but is largely incoherent with the context.
    \item \textbf{3 Points:} The reasoning is moderately logical, with clear causal links, though some minor inconsistencies with the context exist.
    \item \textbf{4 Points:} The reasoning is mostly logical, with well-defined causal relationships and strong coherence with the context.
    \item \textbf{5 Points:} The reasoning is fully logical, with clear and rational causal chains and excellent coherence with the context.
\end{itemize}

\textbf{Input:}
\begin{itemize}
    \item Reference Reason: \texttt{<reference\_reason>}
    \item Generated Reason: \texttt{<generated\_reason>}
\end{itemize}

\textbf{Output Format:}
Please strictly follow the format below to output the scoring result, and \textbf{only output the scoring result} without adding any additional explanations or text:

\texttt{Logical Judgment Dimension: <score>}
\end{tcolorbox}

\newpage
\begin{tcolorbox}[
    colback=myblue!5!white,
    colframe=myblue!75!black,
    arc=1mm, 
    auto outer arc,
    title={Prompt from Multimodal Content Association Dimension Evaluation Criteria for Model-Generated Reasoning},
    breakable
    ]
\small

You need to evaluate the quality of model-generated reasoning for why a video audience laughed. You will be provided with two reasons for laughter reasoning:

1. The reason for laughter generated by the model.
2. The reference reason for laughter is annotated manually (as a benchmark).

Please score based on the following dimension, with a maximum of 5 points:

\textbf{Multimodal Content Association Dimension:} Based on the reference reason, evaluate whether the generated text accurately reflects the interactions between language, visual, audio, and other modal contents, especially whether these contents are consistent with the triggers for laughter.

\textbf{Scoring Criteria:}
\begin{itemize}
    \item \textbf{1 Point:} The reasoning fails to associate with multimodal content, showing no consistency with language, visual, audio, or other modalities.
    \item \textbf{2 Points:} The reasoning shows minimal association with multimodal content, with limited consistency and several mismatches.
    \item \textbf{3 Points:} The reasoning moderately reflects multimodal interactions, maintaining some consistency but with noticeable gaps.
    \item \textbf{4 Points:} The reasoning strongly associates with multimodal content, showing clear consistency with most language, visual, audio, and other modalities.
    \item \textbf{5 Points:} The reasoning perfectly captures and reflects the interactions between all relevant multimodal contents, with complete consistency with the triggers for laughter.
\end{itemize}

\textbf{Input:}
\begin{itemize}
    \item Reference Reason: \texttt{<reference\_reason>}
    \item Generated Reason: \texttt{<generated\_reason>}
\end{itemize}

\textbf{Output Format:}
Please strictly follow the format below to output the scoring result, and \textbf{only output the scoring result} without adding any additional explanations or text:

\texttt{Multimodal Content Association Dimension: <score>}

\end{tcolorbox}
% bib
% \section*{Ethical Statement}

% There are no ethical issues. An optional ethics statement can be placed either in the body of the paper or in the reference pages.

% \section*{Acknowledgments}

% The preparation of these instructions and the \LaTeX{} and Bib\TeX{}
% files that implement them was supported by Schlumberger Palo Alto
% Research, AT\&T Bell Laboratories, and Morgan Kaufmann Publishers.
% Preparation of the Microsoft Word file was supported by IJCAI.  An
% early version of this document was created by Shirley Jowell and Peter
% F. Patel-Schneider.  It was subsequently modified by Jennifer
% Ballentine, Thomas Dean, Bernhard Nebel, Daniel Pagenstecher,
% Kurt Steinkraus, Toby Walsh, Carles Sierra, Marc Pujol-Gonzalez,
% Francisco Cruz-Mencia and Edith Elkind.


\end{document}